\PassOptionsToPackage{capitalize}{cleveref}
\PassOptionsToPackage{normalem}{ulem}
\AddToHook{class/article/after}{%
  \RequirePackage{amsmath,amssymb,amsthm}%
  \RequirePackage{algorithm,titlesec}%
  \RequirePackage[numbers,sort&compress]{natbib}}
\documentclass{nus_next}
\usepackage{graphicx}
\usepackage{amsmath,amssymb,amsfonts}
\usepackage{amsthm}
\usepackage{booktabs}
\usepackage{multirow}
\usepackage{colortbl}
\usepackage{xcolor}
\usepackage{array}
\usepackage{xspace}
\usepackage{microtype}
\usepackage[inline]{enumitem}
\usepackage{subcaption}
\usepackage{algorithm}
\usepackage[noend]{algpseudocode}

\newtheorem{proposition}{Proposition}

\newcommand{\our}{LEDGER\xspace}
\newcommand{\cdino}{\mathcal{C}^{\mathrm{DINO}}}
\newcommand{\csig}{\mathcal{C}^{\mathrm{SigLIP}}}
\newcommand{\himg}{h_{\mathrm{I}}}
\newcommand{\htxt}{h_{\mathrm{T}}}
\newcommand{\Lcov}{\mathcal{L}_{\mathrm{cov}}}
\newcommand{\Lrec}{\mathcal{L}_{\mathrm{rec}}}
\newcommand{\Ltm}{\mathcal{L}_{\mathrm{TM}}}
\newcommand{\Ltt}{\mathcal{L}_{\mathrm{TT}}}
\newcommand{\Lsa}{\mathcal{L}_{\mathrm{sa}}}

\newcommand{\cmon}{\mathcal{C}^{\mathrm{mon}}}

\newcommand{\supp}{the supplement\xspace}
\newcommand{\fNSys}{18}

\newcommand{\fNPixW}{twelve}
\newcommand{\fNLangW}{six}

\newcommand{\fDiscQ}{0.35}
\newcommand{\fDiscQLo}{0.17}
\newcommand{\fDiscQHi}{0.56}
\newcommand{\fDiscQPct}{35}
\newcommand{\fDiscQSceneLo}{0.33}
\newcommand{\fDiscQSceneHi}{0.39}

\newcommand{\fDiscLP}{0.10}

\newcommand{\fDiscLPMoc}{0.12}
\newcommand{\fDiscLPAcc}{0.09}
\newcommand{\fDiscH}{0.49}

\newcommand{\fDiscHPct}{49}

\newcommand{\fDiscHDet}{0.47}
\newcommand{\fDiscRefLo}{0.28}
\newcommand{\fDiscRefHi}{0.40}
\newcommand{\fDetShareLo}{67}
\newcommand{\fDetShareHi}{73}
\newcommand{\fDiscQB}{0.38}

\newcommand{\fRhoAll}{0.39}
\newcommand{\fRhoPix}{0.91}
\newcommand{\fRhoLang}{0.03}
\newcommand{\fDiscPix}{0.12}
\newcommand{\fInvPix}{8}
\newcommand{\fPairsPix}{66}

\newcommand{\fRsqQ}{0.12}
\newcommand{\fRsqQFam}{0.72}
\newcommand{\fFamOff}{0.09}
\newcommand{\fRhoPixFR}{0.66}

\newcommand{\fFamOffFR}{0.08}

\newcommand{\fRhoPixQB}{0.78}

\newcommand{\fFamOffQB}{0.06}
\newcommand{\fRhoLPC}{0.92}

\newcommand{\fRhoLangFR}{-0.60}
\newcommand{\fRhoLangQB}{-0.64}
\newcommand{\fNInvNoSeeSR}{5}
\newcommand{\fNInvNoSeeSRW}{five}
\newcommand{\fRobSeeSRWCap}{Ten}
\newcommand{\fDiscQNoSeeSR}{0.26}
\newcommand{\fRhoAllNoSeeSR}{0.65}
\newcommand{\fFamOffNoSeeSR}{0.07}
\newcommand{\fFamOffFRNoSeeSR}{0.07}
\newcommand{\fRhoLangNoSeeSR}{0.80}
\newcommand{\fRhoLangFRNoSeeSR}{-0.29}
\newcommand{\fCtrlDcLo}{0.11}
\newcommand{\fCtrlDcHi}{0.14}
\newcommand{\fCtrlDqLo}{0.002}
\newcommand{\fCtrlDqHi}{0.007}
\newcommand{\fAltRhoPix}{0.88}
\newcommand{\fAltFamOff}{0.09}

\newcommand{\fNPairs}{153}

\newcommand{\fNInvSig}{15}
\newcommand{\fNInvSigW}{fifteen}

\newcommand{\fRobSeeSR}{10}
\newcommand{\fRobCLIPPan}{2}
\newcommand{\fRobCDS}{2}
\newcommand{\fRobDACLIP}{1}
\newcommand{\fPcCMocQ}{0.97}
\newcommand{\fPcCAccQ}{0.95}
\newcommand{\fPcCDetQ}{0.92}
\newcommand{\fPcQMocC}{-0.20}
\newcommand{\fPcQAccC}{0.69}
\newcommand{\fPcQDetC}{0.34}
\newcommand{\fPcCMocLP}{0.83}
\newcommand{\fPcCAccLP}{0.55}
\newcommand{\fPcCDetLP}{0.45}
\newcommand{\fPcLPMocC}{0.07}
\newcommand{\fPcLPAccC}{0.58}
\newcommand{\fPcLPDetC}{0.59}
\newcommand{\fPcCDetDinoLP}{0.53}

\newcommand{\fPcCDetH}{0.93}
\newcommand{\fPcCMocH}{0.98}

\newcommand{\fLooPixHi}{0.47}
\newcommand{\fLooLPLo}{0.76}
\newcommand{\fLooLPHi}{0.86}
\newcommand{\fLooLPCovLo}{0.87}
\newcommand{\fLooLPCovHi}{0.93}
\newcommand{\fLooCovLo}{0.83}
\newcommand{\fLooCovHi}{0.93}
\newcommand{\fBestC}{0.75}

\newcommand{\fRefC}{0.93}

\newcommand{\fOursC}{0.87}
\newcommand{\fClosedPct}{67}
\newcommand{\fBestCFR}{0.72}
\newcommand{\fOursCFR}{0.85}

\newcommand{\fSigQ}{12.5}
\newcommand{\fSigC}{15.5}
\newcommand{\fRecLocSG}{0.58}

\newcommand{\fRecGloSG}{0.78}

\newcommand{\fWinSG}{14}
\newcommand{\fTieSG}{4}
\newcommand{\fLossSG}{2}
\newcommand{\fWinCC}{18}

\newcommand{\fWinGFE}{10}
\newcommand{\fLossGFE}{10}
\newcommand{\fPgf}{0.76}
\newcommand{\fPqb}{0.57}
\newcommand{\fFamA}{77}
\newcommand{\fFamAw}{82}
\newcommand{\fFamN}{84}
\newcommand{\fFirst}{19}

\newcommand{\fStripN}{6}

\newcommand{\fWVdE}{4.4}

\newcommand{\fWVdQ}{0.006}

\newcommand{\fWTdE}{4.4}

\newcommand{\fWTdQ}{0.010}

\newcommand{\fDetGainDino}{6.4}
\newcommand{\fDetGainCnx}{4.5}
\newcommand{\fDetGainLo}{4.3}
\newcommand{\fDetGainHi}{4.7}
\newcommand{\fDetGainDinoSG}{7.2}
\newcommand{\fDetGainLoSG}{4.8}
\newcommand{\fDetGainHiSG}{5.2}

\newcommand{\fCovOurs}{0.85}

\newcommand{\fCovBestBase}{0.72}

\newcommand{\fLabelledDetGain}{6.8}
\newcommand{\fSmallGain}{7.9}
\newcommand{\fSpeedLo}{9}
\newcommand{\fSpeedHi}{22}
\newcommand{\fInferMs}{133}
\newcommand{\fCertMs}{253}
\newcommand{\fMsCC}{22}

\newcommand{\fRetHeldLo}{7.0}
\newcommand{\fRetHeldHi}{9.4}

\newcommand{\fRetClipLoF}{0.070}
\newcommand{\fRetClipHiF}{0.094}
\newcommand{\fRetNoLangLoF}{0.076}
\newcommand{\fRetNoLangHiF}{0.077}

\newcommand{\fStrideOurs}{0.549}
\newcommand{\fStrideBase}{0.463}

\newcommand{\fFPoursThree}{0.353}
\newcommand{\fFPsgThree}{0.370}

\newcommand{\fExtAllLoPts}{2}
\newcommand{\fExtAllHiPts}{5}
\newcommand{\fSnVendorFoneStd}{2.5}
\newcommand{\fSnVendorAplsStd}{2.8}
\newcommand{\fPcSeeSR}{0.71}
\newcommand{\fExtLo}{0.02}
\newcommand{\fExtHi}{0.05}
\newcommand{\fHalOurs}{0.066}
\newcommand{\fHalSG}{0.081}
\newcommand{\fHalSeeSR}{0.214}
\newcommand{\fHalRef}{0.051}
\newcommand{\fHalCC}{0.069}
\newcommand{\fHalAR}{0.068}
\newcommand{\fHalDA}{0.088}
\newcommand{\fDotaRho}{-0.95}
\newcommand{\fPoolShareOC}{27}
\newcommand{\fPoolPrecOC}{0.82}
\newcommand{\fPoolPrecOthLo}{0.50}
\newcommand{\fPoolPrecOthHi}{0.64}
\newcommand{\fPoolRelOC}{33.3}
\newcommand{\fPoolIrrOC}{14.5}
\newcommand{\fPoolUniform}{14.3}
\newcommand{\fPoolShareHost}{32}

\newcommand{\fPoolIrrHost}{19.5}

\newcommand{\fSwapSurLo}{0.028}
\newcommand{\fSwapSurHi}{0.041}
\newcommand{\fSwapMarLo}{0.128}
\newcommand{\fSwapMarHi}{0.133}

\newcommand{\fDetSmallShare}{90}
\newcommand{\fDetRealShare}{63}
\newcommand{\fDetRealFP}{0.44}
\newcommand{\fDetRealCov}{0.74}
\newcommand{\fDetRealMap}{0.44}
\newcommand{\fPlainLcovQ}{0.916}
\newcommand{\fPlainLcovC}{0.74}
\newcommand{\fInjRecQ}{0.913}
\newcommand{\fInjRecC}{0.72}

\newcommand{\fXencVal}{0.04\text{--}0.06}

\newcommand{\fGapZero}{0.51}
\newcommand{\fNHyper}{21}

\newcommand{\fNHyperLiteW}{six}
\newcommand{\fLocalArea}{6}
\newcommand{\fLocalGradCov}{33}
\newcommand{\fLocalGradTok}{7}
\newcommand{\fGradRatio}{1.9}
\newcommand{\fMonEnc}{CLIP ViT-B/16}

\newcommand{\fLoopDc}{0.13}
\newcommand{\fInjDc}{0.17}
\newcommand{\fLoopDq}{0.011}
\newcommand{\fTotalDq}{0.019}
\newcommand{\fObjTokCLo}{0.76}
\newcommand{\fObjTokCHi}{0.79}
\newcommand{\fObjSetCLo}{0.82}
\newcommand{\fObjSetCHi}{0.83}

\newcommand{\fLcovDq}{0.008}
\newcommand{\fMgCovDq}{0.006}
\newcommand{\fMgCovDw}{0.012}

\newcommand{\fParams}{0.40}
\newcommand{\fTotalParams}{428.4}
\newcommand{\fGpuHours}{10{,}056}
\newcommand{\fGenGF}{936}
\newcommand{\fCertGF}{1850}
\newcommand{\fLightGF}{1.6}
\newcommand{\fEncGF}{912}
\newcommand{\fTeaserBaseQ}{0.958}
\newcommand{\fTeaserOursQ}{0.958}

\newcommand{\fTeaserNW}{six}
\newcommand{\fTeaserBaseW}{three}

\newcommand{\fTeaserScene}{11}

\newcommand{\fNReassign}{11}
\newcommand{\fNLabelled}{61}
\newcommand{\fNLabelledDraw}{20}

\newcommand{\fSnOursFone}{0.691}
\newcommand{\fSnOursSmall}{0.56}
\newcommand{\fSnOursApls}{0.656}
\newcommand{\fSnVendFone}{0.681}
\newcommand{\fSnVendSmall}{0.52}
\newcommand{\fSnVendApls}{0.642}
\newcommand{\fSnTiles}{1{,}360}
\newcommand{\fSnBuildings}{61{,}240}
\newcommand{\fSnRoadTiles}{346}
\newcommand{\fSnRoadKm}{1{,}412}

\newcommand{\fPlugQLo}{0.004}
\newcommand{\fPlugQHi}{0.006}
\newcommand{\fPlugCovLo}{0.09}
\newcommand{\fPlugCovHi}{0.19}
\newcommand{\fPlugDetLo}{4.2}
\newcommand{\fPlugDetHi}{4.5}

\newcommand{\fRaSG}{0.036}
\newcommand{\fRaGap}{0.103}

\newcommand{\fHumPref}{68}
\newcommand{\fHumScenes}{19}
\newcommand{\fHumScenesTwo}{11}
\newcommand{\fHumNScenes}{20}

\newcommand{\fHumP}{4.0\times10^{-5}}
\newcommand{\fHumPrefTwo}{52}
\newcommand{\fHumPTwo}{0.82}
\newcommand{\fHumRaters}{5}
\newcommand{\fHumPairs}{300}
\newcommand{\fCountOurs}{1.8}
\newcommand{\fCountBase}{2.6}
\newcommand{\fCountP}{0.002}
\newcommand{\fFacAG}{+0.003}
\newcommand{\fFacFG}{+0.001}
\newcommand{\fFacAFG}{+0.011}
\newcommand{\fFacInt}{+0.006}
\newcommand{\fFacGateH}{+0.022}

\newcommand{\fFacGmainH}{+0.015}

\newcommand{\maxsizebox}[2]{\resizebox{\ifdim\width>#1 #1\else\width\fi}{!}{#2}}
\definecolor{ourrow}{RGB}{255,243,235}
\definecolor{ourhead}{RGB}{199,115,26}
\newcommand{\best}[1]{\textbf{#1}}
\newcommand{\second}[1]{\underline{#1}}
\newcommand{\sig}{$^{\star}$}
\newcommand{\sigw}{$^{\dagger}$}
\providecommand{\etal}{\emph{et al.}}

\newsavebox{\ReportTableBox}

\usepackage{omni-compat}
\usepackage{wide-tables}
\renewcommand{\floatpagefraction}{.93}
\makeatletter
\AtBeginEnvironment{ReportFlowTable}{%
  \let\ReportSavedTableLabel\label
  \def\label#1{%
    \edef\@currentHref{report.longtable.#1}%
    \llap{\hbox to\linewidth{\Hy@raisedlink{\hyper@anchorstart{\@currentHref}\hyper@anchorend}\hss}}%
    \ReportSavedTableLabel{#1}}}
\makeatother
\makeatletter
\newcommand{\ReportMainTablePlacement}{%
  \let\ReportMakeFloat\@xfloat
  \def\@xfloat##1[##2]{\ReportMakeFloat{##1}[!htbp]}%
  \renewcommand{\floatpagefraction}{.72}}
\makeatother

\DeclareCaptionFont{ReportTableCaption}{\fontsize{9}{10.4}\selectfont}
\usepackage{xurl}
\hypersetup{pdfdisplaydoctitle=true,pdftitle={Closing the Loop on the Retrievability Gap},
 pdfauthor={Guangyuan Dong, Wenhao Li, Zhen Tian, Jiaqi Zhang, Haoyu Wang, Junhao Xu, Mingzhe Liu, Pan Wang, Fei Shen, Tat-Seng Chua},
 pdfsubject={Technical Report}}
\crefname{section}{Sec.}{Secs.}
\Crefname{section}{Section}{Sections}
\crefname{table}{Tab.}{Tabs.}
\Crefname{table}{Table}{Tables}
\crefname{proposition}{Proposition}{Propositions}
\newcommand{\MFigTeaser}{\ref{fig:teaser}}
\newcommand{\MFigGap}{\ref{fig:gap}}
\newcommand{\MFigMethod}{\ref{fig:framework}}
\newcommand{\MFigQual}{\ref{fig:qualitative}}
\newcommand{\MTabMain}{\ref{tab:main}}
\newcommand{\MTabDown}{\ref{tab:downstream}}
\newcommand{\MTabIndep}{\ref{tab:independence}}
\newcommand{\MTabGains}{\ref{tab:gains}}
\newcommand{\MSecGap}{\ref{sec:gap}}
\newcommand{\MSecMethod}{\ref{sec:method}}
\newcommand{\MSecGeneral}{\ref{sec:general}}
\newcommand{\MProp}{\ref{prop:area}}
\newcommand{\MEqAttn}{\ref{eq:attn}}
\newcommand{\MEqPresence}{\ref{eq:presence}}
\newcommand{\MEqLcov}{\ref{eq:lcov}}
\newcommand{\SRefSecTeaser}{\ref{sec:teaser_supp}}
\newcommand{\SRefSecProof}{\ref{sec:proof}}
\newcommand{\SRefSecMatched}{\ref{sec:matched}}
\newcommand{\SRefSecLabelled}{\ref{sec:labelled}}

\newcommand{\SRefSecDomains}{\ref{sec:domains}}
\newcommand{\SRefSecRecipe}{\ref{sec:recipe}}
\newcommand{\SRefTabAblation}{\ref{tab:ablation_full}}
\newcommand{\SRefAlgTrain}{\ref{alg:train}}
\newcommand{\SRefTabMapping}{\ref{tab:mapping}}
\newcommand{\SRefTabQueries}{\ref{tab:queries}}
\newcommand{\SRefTabProbeVal}{\ref{tab:probe_val}}
\newcommand{\SRefTabFmow}{\ref{tab:fmow}}
\newcommand{\SRefTabArchive}{\ref{tab:archive}}

\newcommand{\SRefTabStrip}{\ref{tab:strip}}
\newcommand{\SRefTabFamilies}{\ref{tab:families}}
\newcommand{\SRefTabInversions}{\ref{tab:inversions}}
\newcommand{\SRefTabPartial}{\ref{tab:partial}}
\newcommand{\SRefTabVocabSplit}{\ref{tab:vocab_split}}
\newcommand{\SRefTabSwap}{\ref{tab:swap}}
\newcommand{\SRefTabCross}{\ref{tab:cross}}
\newcommand{\SRefTabRetadapt}{\ref{tab:retadapt}}

\newcommand{\SRefTabObjective}{\ref{tab:objective}}
\newcommand{\SRefTabLadder}{\ref{tab:ladder}}
\newcommand{\SRefTabControls}{\ref{tab:controls}}
\newcommand{\SRefTabAreabins}{\ref{tab:areabins}}

\newcommand{\SRefTabLedgernoise}{\ref{tab:ledgernoise}}
\newcommand{\SRefTabAdapter}{\ref{tab:adapter}}

\newcommand{\SRefTabGaming}{\ref{tab:gaming}}
\newcommand{\SRefTabPostproc}{\ref{tab:postproc}}
\newcommand{\SRefTabConcurrent}{\ref{tab:concurrent}}

\newcommand{\SRefTabDownstreamRR}{\ref{tab:downstream_rr}}
\newcommand{\SRefTabSpacenet}{\ref{tab:spacenet_aoi}}
\newcommand{\SRefTabPlugin}{\ref{tab:plugin}}
\newcommand{\SRefTabPerscene}{\ref{tab:perscene_wv3_rr}}
\providecommand{\ReportContentsSection}[1]{\par\addvspace{.6em}\noindent
  \hyperref[#1]{\textbf{\ref*{#1}\quad\nameref*{#1}}\nobreak\hfill\textbf{\pageref*{#1}}}\par}
\providecommand{\ReportContentsSubsection}[1]{\par\noindent
  \hyperref[#1]{\ref*{#1}\quad\nameref*{#1}\nobreak\dotfill\pageref*{#1}}\par}
\newcommand{\ReportSupplementContentsLeft}{%
  \ReportContentsSection{sec:theory}%
  \ReportContentsSubsection{toc:supp:1:1}%
  \ReportContentsSubsection{sec:proof}%
  \ReportContentsSubsection{toc:supp:1:3}%
  \ReportContentsSubsection{sec:monitors}%
  \ReportContentsSubsection{sec:matched}%
  \ReportContentsSubsection{sec:coupling}%
  \ReportContentsSection{sec:algos}%
  \ReportContentsSection{supp:sec:protocol}%
  \ReportContentsSubsection{sec:data}%
  \ReportContentsSubsection{sec:labelled}%
  \ReportContentsSubsection{sec:annot}%
  \ReportContentsSubsection{sec:vocab}%
  \ReportContentsSubsection{sec:encoders}%
  \ReportContentsSubsection{sec:evaluators}%
  \ReportContentsSubsection{sec:retrievalprotocol}%
  \ReportContentsSubsection{sec:spacenet}%
  \ReportContentsSubsection{sec:retadapt}%
  \ReportContentsSubsection{sec:human}%
  \ReportContentsSubsection{sec:stats}%
  \ReportContentsSection{sec:impl}%
  \ReportContentsSubsection{toc:supp:4:1}%
  \ReportContentsSubsection{toc:supp:4:2}%
  \ReportContentsSubsection{sec:recipe}%
  \ReportContentsSubsection{toc:supp:4:4}%
}
\newcommand{\ReportSupplementContentsRight}{%
  \ReportContentsSection{supp:sec:results}%
  \ReportContentsSubsection{toc:supp:5:1}%
  \ReportContentsSubsection{sec:gap_supp}%
  \ReportContentsSubsection{toc:supp:5:3}%
  \ReportContentsSubsection{toc:supp:5:4}%
  \ReportContentsSubsection{sec:objectives}%
  \ReportContentsSubsection{toc:supp:5:6}%
  \ReportContentsSubsection{sec:controls}%
  \ReportContentsSubsection{toc:supp:5:8}%
  \ReportContentsSubsection{toc:supp:5:9}%
  \ReportContentsSubsection{toc:supp:5:10}%
  \ReportContentsSubsection{sec:teaser_supp}%
  \ReportContentsSection{sec:domains}%
  \ReportContentsSubsection{toc:supp:6:1}%
  \ReportContentsSubsection{toc:supp:6:2}%
  \ReportContentsSection{sec:limits}%
}

\newcommand{\ReportTitle}{Closing the Loop on the Retrievability Gap}
\title{\textcolor[HTML]{B07900}{Closing} \textcolor[HTML]{C54F1D}{the Loop}
  \textcolor[HTML]{377FAE}{on the} \textcolor[HTML]{7962A9}{Retrievability Gap}}
\author{\parbox{.997\textwidth}{\centering
  Guangyuan Dong, Wenhao Li, Zhen Tian, Jiaqi Zhang, Haoyu Wang,\\
  Junhao Xu, Mingzhe Liu, Pan Wang, Fei Shen$^{\dagger}$, Tat-Seng Chua$^{\dagger}$}}
\affiliation{NUS-Tsinghua NExT Research Centre, National University of Singapore}
\date{November 2026}
\abstract{%
  \begingroup
  \def\abstract{}
  \begin{abstract}
A frozen vision-language encoder now guides image restoration and later indexes what it produced. Over a fixed set of eighteen pansharpening systems on one WorldView-III test set, the fidelity index that ranks them, Q2$^n$ with a reference and HQNR without one, picks the less retrievable of two systems one time in three and one in two; the disagreement lies between the pixel-supervised and the language-conditioned family, and even the best system falls short of the reference, small objects first. We call this the retrievability gap and trace it to the area weighting of position-averaged losses. \our closes the loop around the encoder: it indexes the input into a set of named concepts, generates, re-encodes the output alone with the same frozen encoder, and minimises a set-level soft-Jaccard loss on the coverage of that set, a label-free target available at test time, so that every present concept of a scale receives a gradient share independent of its area. Measured outside the loop by label-grounded probes, held-out retrievers, language-free evaluators, detectors and human raters, \our matches or exceeds the strongest baselines in fidelity on four sensors, raises retrieval by seven to nine points within its vocabulary and two to five outside it, closes two thirds of the reduced-resolution gap to the reference, lifts extraction on raw SpaceNet imagery above the vendor product in each of four cities, and transfers to existing generators as a training-only loss at no inference cost.
\end{abstract}
  \endgroup}
\pretocmd{\beginabstract}{\pdfbookmark[1]{Abstract}{report.abstract}}{}{}

\makeatletter
\newcommand*{\ps@nextreport}{%
  \ps@plain
  \def\@oddhead{\raisebox{0pt}[\headheight][0pt]{%
    \ReportOriginalImage[width=16.5mm]{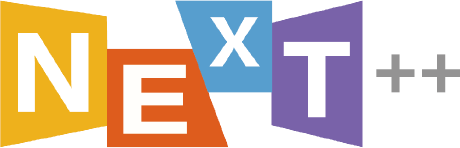}}\hfil}%
  \let\@evenhead\@oddhead}
\makeatother

\newcommand{\OmniCoverLogo}[4]{%
  \href{#1}{\raisebox{-.5\height}{%
    \ReportOriginalImage[width=#2,height=#3,keepaspectratio]{logos/#4}}}}
\fancypagestyle{firststyle}{%
  \fancyhf{}%
  \fancyhead[L]{\makebox[\headwidth][c]{%
    \OmniCoverLogo{https://www.nextcenter.org/}{23mm}{8mm}{next.png}\hspace{4mm}%
    \OmniCoverLogo{https://nus.edu.sg/}{22mm}{8mm}{nus.pdf}\hspace{4mm}%
    \OmniCoverLogo{https://www.tsinghua.edu.cn/}{26mm}{8mm}{tsinghua.pdf}\hspace{4mm}%
    \OmniCoverLogo{https://www.antresearch.com/}{23mm}{8mm}{ant-research.png}\hspace{4mm}%
    \OmniCoverLogo{https://www.pku.edu.cn/}{27mm}{8mm}{pku.pdf}}}%
  \fancyfoot[C]{\normalfont\small\thepage}}
\newcommand{\OmniPartnerSeparator}{\hspace{.17em}\textperiodcentered\hspace{.17em}}
\newcommand{\OmniCollaborationNote}{%
  \noindent\begin{minipage}{\linewidth}%
  \normalfont\rmfamily\fontsize{7.7}{10}\selectfont
  \noindent{\color[HTML]{A9A0B8}\rule{.22\linewidth}{.4pt}}\par\vspace{1.2mm}%
  \noindent$^{\dagger}$\,Corresponding authors\par
  \noindent\mbox{%
    \color[HTML]{888888}\fontfamily{fontmedium}\fontseries{m}\upshape\fontsize{6.7}{9.4}\selectfont
    \spaceskip=.22em\relax
    {Collaborating institutions}\hspace{.55em}%
    {Tsinghua University}\OmniPartnerSeparator
    {Ant Research}\OmniPartnerSeparator
    {Peking University}\OmniPartnerSeparator
    {University of Glasgow}\OmniPartnerSeparator
    {Shandong University}\OmniPartnerSeparator
    {Nankai University}}\par\end{minipage}\par}

\newcommand{\ReportTOCRow}[3]{%
  \par\noindent\hyperref[#1]{%
    \begin{minipage}[t]{.110\linewidth}\raggedright{#2\color{nextblue}\ref*{#1}}\end{minipage}\hfill
    \begin{minipage}[t]{.785\linewidth}\raggedright{#2\color{black}\nameref*{#1}}\end{minipage}\hfill
    \begin{minipage}[t]{.068\linewidth}\raggedleft{#2\color{nextblue}\pageref*{#1}}\end{minipage}}%
  \par\vspace{#3}}
\renewcommand{\ReportContentsSection}[1]{\par\addvspace{3.8pt}\ReportTOCRow{#1}{\sffamily}{2pt}}
\renewcommand{\ReportContentsSubsection}[1]{\ReportTOCRow{#1}{}{1.1pt}}
\newcommand{\ReportContentsPage}{%
  \noindent\begin{minipage}{\linewidth}
  \pdfbookmark[0]{Supplementary Material}{report.supplement}%
  \pdfbookmark[1]{Contents}{report.suppcontents}%
  \noindent{\titlefont\sffamily Supplementary Material}\hfill
  {\sffamily\color{nextblue}\fontsize{11}{13}\selectfont Contents}\par
  \vspace{4pt}{\color{nextblue}\hrule height .7pt}\vspace{6pt}
  \begingroup\parskip=0pt\parindent=0pt\normalfont\fontsize{9.2}{10.8}\selectfont
  \noindent\begin{minipage}[t]{.51\textwidth}\ReportSupplementContentsLeft\end{minipage}\hfill
  \begin{minipage}[t]{.466\textwidth}\ReportSupplementContentsRight\end{minipage}\par
  \endgroup\end{minipage}\par}

\begin{document}
  \newgeometry{left=25mm,right=25mm,top=18mm,bottom=14mm,headheight=23pt,headsep=2mm,footskip=6.8mm}%
  \pdfbookmark[0]{\ReportTitle}{report.title}%
  \begingroup\setlength{\parskip}{0pt}\setlength{\parindent}{0pt}%
  \patchcmd{\mymaketitle}{\tcbset{top=6mm}}{\tcbset{top=3mm}}{}{}
  \patchcmd{\mymaketitle}{\tcbset{bottom=6mm}}{\tcbset{bottom=3mm}}{}{}
  \maketitle\vspace{-6mm}
  \noindent\begin{minipage}{\linewidth}
    \centering
    \captionsetup{type=figure,singlelinecheck=false,skip=4pt}%
    \renewcommand{\includegraphics}[2][]{%
      \ReportOriginalImage[width=\linewidth]{##2}}%
    \includegraphics[width=\columnwidth]{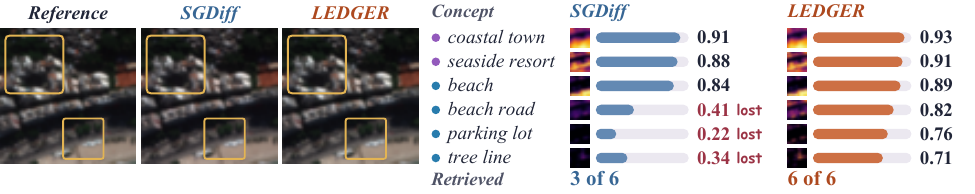}
\caption{\textbf{Looks right, retrieves wrong.} A WorldView-III test tile (scene~$\fTeaserScene$, chosen because both scene-level Q2$^n$ scores round to $\fTeaserBaseQ$) restored by SGDiff, the strongest pixel-supervised network, and by \our. In the input ledger, purple dots mark global concepts and blue dots mark meso-scale concepts. The frozen index that found \fTeaserNW{} concepts in the input reads each output alone: \fTeaserBaseW{} remain in SGDiff's (the beach road, the parking lot and the tree line are lost), all \fTeaserNW{} in ours. The index's readings explain the method; outside the loop, the held-out OpenCLIP retriever scores the beach-road query $0.31$ on ours against $0.22$ on SGDiff's and the probe finds the parking lot in ours only (\supp, Sec.~\SRefSecTeaser); every result is measured outside the loop (\cref{fig:gap}).}
\label{fig:teaser}
  \end{minipage}\par
  \vfill\vspace{.5mm}\OmniCollaborationNote
  \endgroup
  \restoregeometry
\pagestyle{nextreport}
\begingroup
\ReportMainFloatPlacement
\ReportMainTablePlacement
\section{Introduction}\label{sec:intro}

\noindent\textbf{Problem.} Image restoration has acquired a second reader: a frozen vision-language model (VLM) conditions diffusion and restoration networks~\cite{rombach2022ldm,wu2024seesr,yu2024supir,rao2022denseclip,cheng2024clipdenoising}, and the same family of encoders then indexes the restored images for search. Pansharpening, in which a sharp panchromatic band guides the reconstruction of coarse multispectral imagery, is where the two readers meet: its pixel-fidelity baselines are strong~\cite{duan2024canconv,bandara2022hypertransformer}, its products are archived and queried by the tile, and perceptual metrics~\cite{zhang2018lpips}, detectors~\cite{xia2018dota} and held-out retrievers~\cite{cherti2023openclip} measure what a pixel metric misses.

\noindent\textbf{Observation.} We call an image \emph{retrievable} when the concepts of its scene can still be found in it by an index (in information retrieval~\cite{azzopardi2008retrievability} the term measures how readily a system retrieves a document; here it is per image) and measure it by \emph{coverage}, the recall of a tile's annotated concepts under a frozen evaluator, an evaluation measure, not an archive index. \cref{sec:gap} measures it for a fixed set of $\fNSys$ trained systems and finds a \emph{retrievability gap} with two sides (\cref{fig:teaser,fig:gap}). Its first side is measurement: Q2$^n$, SAM and ERGAS with a reference and HQNR~\cite{arienzo2022hqnr} without one order systems differently from coverage in $\fDiscQPct$--$\fDiscHPct\%$ of the pairs; within the pixel-supervised family fidelity and coverage agree, whereas language-conditioned systems recover more of the scene at equal fidelity. Its second side is recovery: the best of the $\fNSys$ systems reaches $\fBestC$ coverage against $\fRefC$ for the reference tiles, and the smallest concepts go first.

\noindent\textbf{Cause and method.} That distortion and task performance can diverge is known~\cite{blau2018perception,shermeyer2019sr}; our claim is specific: a pixel loss, and equally a feature loss averaged over positions, rewards each concept in proportion to its area, so the few tokens of a boat are outweighed by the land cover around them (\cref{prop:area}). \our (\cref{sec:method}) changes what is rewarded: one frozen SigLIP-2 encoder~\cite{tschannen2025siglip2} indexes the input into a set of named concepts with confidences, the input's \emph{ledger}; the generator produces an output; the same frozen encoder re-encodes the output alone; and a soft-Jaccard loss on the overlap between the recovered set and the ledger is minimised through the encoder, which gives every present concept of a scale a gradient share that no longer grows with its area. The target is the input's own ledger, label-free and available at test time; no prior restoration method verifies its output against the concepts of its own input. Its core adds \fNHyperLiteW{} hyper-parameters of its own to a reconstruction recipe and joins existing generators during training.

\noindent\textbf{Evidence outside the loop.} A generator trained against one encoder could learn to satisfy that encoder alone, so no in-loop quantity is used as evidence: every gain is reported under evaluators at a stated distance from the training signal, from a probe grounded in external labels to benchmarks outside our vocabulary, segmenters on raw SpaceNet imagery~\cite{vanetten2018spacenet} and human raters. Our contributions:
\begin{itemize}[leftmargin=*,nosep]
\item a rank-based measurement of the retrievability gap over $\fNSys$ systems, two sensors and two protocols: fidelity indices and coverage disagree on $\fDiscQPct$--$\fDiscHPct\%$ of system pairs, all $\fNInvSig$ seed-robust rank inversions lie between the two families, and coverage stays predictive of retrieval and recognition given Q2$^n$ or LPIPS and of detection given Q2$^n$;
\item \our, a closed loop around one frozen encoder in which the input's ledger is injected into the generator and, after the output is re-encoded alone, serves as the target of a set-level coverage loss; an area-invariance analysis that carries the per-class rebalancing of Dice-type losses to open-vocabulary concepts of one scale; and a training-only form for existing generators. With architecture and encoder fixed, the loss adds $+0.04$ coverage and $+0.03$ mAP over the strongest position-averaged feature loss and $+0.02$--$0.03$ over area-invariant per-concept losses, concentrated on the smallest concepts ($+0.13$ recall on concepts of at most four tokens);
\item first or within noise of the leader in every fidelity cell of four sensors; $+\fRetHeldLo$ to $+\fRetHeldHi$ mAP on vocabulary queries under six held-out evaluators, $+\fExtAllLoPts$ to $+\fExtAllHiPts$ on external benchmarks, $+\fDetGainLo$ to $+\fDetGainDino$ detection mAP under four heads, extraction on raw sensor data above the vendor product, and controls tying the gain to concept size and anchors.
\end{itemize}

\section{Related Work}\label{sec:related}

\noindent\textbf{Semantics in restoration and fusion.}
Text conditions latent diffusion and instruction-tuned editors~\cite{rombach2022ldm,brooks2023ip2p}; SeeSR~\cite{wu2024seesr}, SUPIR~\cite{yu2024supir}, PASD~\cite{yang2024pasd} and CoSeR~\cite{sun2024coser} read prompts or embeddings from the degraded input to steer a diffusion prior; SeD~\cite{li2024sed} conditions the discriminator on semantics; DenseCLIP~\cite{rao2022denseclip}, MaskCLIP~\cite{zhou2022maskclip}, CLIPDenoising~\cite{cheng2024clipdenoising}, DA-CLIP~\cite{luo2024daclip} and VLMIR~\cite{yang2025vlmir} use frozen CLIP features or embeddings as priors, CLIP-LIT~\cite{liang2023cliplit} and CLIP-IQA~\cite{wang2023clipiqa} as a loss or a quality judge; text-aware restoration~\cite{min2025tair} keeps legible text through a jointly trained text spotter, an entity-level counterpart of our concern. In pansharpening, SGDiff~\cite{xing2025sgdiff} routes diffusion experts with semantics read from the input, semantic prototypes order the tokens of a linear-attention backbone~\cite{li2026msps}, and CLIPPan~\cite{jian2025clippan} scores the fused output against protocol-level text. None verifies that each input concept survives in the output for the index that will serve it.

\noindent\textbf{Task-driven objectives and consistency.}
Task-driven enhancement, super-resolution and coding~\cite{liu2018denoisinghl,son2020urie,shermeyer2019sr,haris2021taskdriven,kim2024sr4ir,zhang2023superyolo,bai2018sodmtgan,duan2020vcm} and fusion trained through task heads~\cite{tang2022seafusion,liu2022tardal,liu2023segmif} put a downstream model into the loop and transfer part of one task's gain, tied to that model (recognition degrades with image quality in model-specific ways~\cite{dodge2016quality}). Perceptual and modulation losses~\cite{johnson2016perceptual,zhang2018lpips,wang2018sftgan,perez2018film,park2019spade} sharpen content without saying which content must be present~\cite{blau2018perception}. Our loop is the open-vocabulary, set-level instance of CyCADA's semantic consistency~\cite{hoffman2018cycada}, a frozen model's reading of the input held fixed as the target for the output. Reward fine-tuning~\cite{clark2024draft,prabhudesai2023alignprop} and concept neglect in text-to-image generation~\cite{chefer2023attendexcite,hu2023tifa,cho2024dsg} share the concern that a global score is met while entities drop out; features tuned against one CLIP transfer to others~\cite{zhao2023advvlm}, hence our evaluators without language.

\noindent\textbf{Set-level losses, retrievability and pansharpening.}
Dice~\cite{milletari2016vnet} and Generalised Dice~\cite{sudre2017gdl} allocate gradient per class rather than per pixel, and soft Jaccard~\cite{rahman2016jaccard}, the Lov\'asz hinge~\cite{berman2018lovasz} and Jaccard metric losses~\cite{wang2023jaccard} optimise set overlap directly; \cref{prop:area} carries this rebalancing to concepts read through a frozen encoder, with token area as the quantity that must not dominate. Pansharpening spans component substitution~\cite{aiazzi2007gsa,vivone2019bdsdpc}, detail-injecting CNNs~\cite{yang2017pannet,deng2021fusionnet,jin2022lagconv}, transformers~\cite{bandara2022hypertransformer,zhou2022panformer,cao2026scaleformer}, diffusion~\cite{xing2024crossdiff,meng2024pandiff,zhong2024ssdiff}, state-space~\cite{he2025panmamba} and adaptive-convolution models~\cite{duan2024canconv,wang2025arconv}, evaluated on PanCollection~\cite{deng2022pancollection} under the Wald protocol~\cite{wald1997fusion} and full-resolution indices~\cite{vivone2021new,arienzo2022hqnr}; remote-sensing image-text models~\cite{liu2024remoteclip,zhang2024georsclip,wang2024skyscript,yuan2022rsitmd,lu2018rsicd} and Scale-MAE~\cite{reed2023scalemae}, a multi-scale extension of SatMAE~\cite{cong2022satmae}, are held-out evaluators (\cref{sec:protocol}).

\section{The Retrievability Gap}\label{sec:gap}

\noindent\textbf{Setting and measures.} Let $\mathbf{X}$ be a degraded observation, $\mathbf{A}$ a registered guide (the panchromatic band), $\mathbf{Y}$ the reference and $\hat{\mathbf{Y}}{=}\mathcal{G}_\theta(\mathbf{X},\mathbf{A})$ a generator's output. For the concept set $\mathcal{Q}$ of a tile and the set $\hat{\mathcal{Q}}$ an evaluator finds in $\hat{\mathbf{Y}}$, the \emph{coverage} of the output is the concept recall $\mathcal{C}=|\mathcal{Q}\cap\hat{\mathcal{Q}}|/|\mathcal{Q}|$, and its precision $\mathcal{P}_{\mathrm{c}}{=}|\mathcal{Q}\cap\hat{\mathcal{Q}}|/|\hat{\mathcal{Q}}|$ counts introduced concepts. Here coverage is \emph{label-grounded}, written $\cdino$: $\mathcal{Q}$ holds the concepts that the tile's human annotations entail, synonyms merged and nothing finer added (\supp, Table~\SRefTabMapping), and $\hat{\mathcal{Q}}$ comes from a linear probe on a frozen DINOv2~\cite{oquab2024dinov2} fitted on external labels and read at its maximum over the tile, so the probe shares a concept set and a max-type read-out with the training loss of \cref{sec:method} but not backbone or data.

\noindent\textbf{Discordance.}\label{def:discordance} For systems $\mathcal{M}$ and measures $f,g$, larger being better, the \emph{discordance} $\delta(f,g)$ is the fraction of pairs of $\mathcal{M}$ that $f$ and $g$ order oppositely, ties counting one half: Kendall's proportion of discordant pairs, $(1{-}\tau_a)/2$, the probability that the better of two systems by $f$ is the worse by $g$ ($0$ when they agree, $0.5$ when unrelated).

\begin{figure}[t]
\centering
\includegraphics[width=\columnwidth]{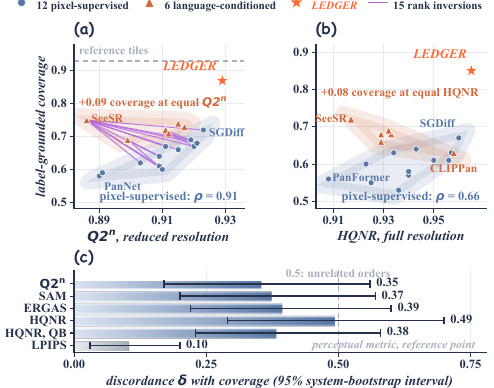}
\caption{\textbf{The retrievability gap on WorldView-III}, $\fNSys$ systems that do not close the loop (\our for reference, outside every statistic). (a)~Q2$^n$ on the $20$ reduced-resolution scenes against label-grounded coverage on their $180$ annotated tiles; lines: the $\fNInvSig$ seed-robust rank inversions, all cross-family. (b)~HQNR on the $20$ full-resolution scenes against coverage on their $980$ tiles, local concepts included. (c)~Discordance $\delta$ of each index with coverage, $95\%$ system-bootstrap intervals; QB: QuickBird.}
\label{fig:gap}
\end{figure}

\noindent\textbf{Comparison set.} The set is fixed in advance: every system we trained that neither re-indexes its output nor trains on coverage, $\fNSys$ in all (\cref{sec:protocol}); \fNPixW{} are \emph{pixel-supervised}, eleven published networks and a plain U-Net, and in the other \fNLangW{}, the \emph{language-conditioned} systems, a frozen VLM conditions or supervises the generator: CLIPPan (native to pansharpening), the adapted DA-CLIP and SeeSR, and three open-loop conditioning designs on our backbone and encoder, included as controls. SGDiff, pixel-supervised but routed by a description of the input, is counted with the former; the other assignment changes no conclusion (\supp, Table~\SRefTabFamilies). Two WorldView-III test sets are never pooled: with a reference, the annotated tiles carry meso and global concepts from their scene label ($|\mathcal{Q}|$ one or two, a scene-level concept recall); at full resolution (HQNR) their boxes add local concepts ($|\mathcal{Q}|$ two to nine).

\noindent\textbf{Fidelity indices capture little of retrievability.} Over the $\fNSys$ systems, $\delta(\mathrm{Q2}^n,\cdino){=}\fDiscQ$ (\cref{fig:gap}c): Q2$^n$ selects the less retrievable of two systems in one pair out of three. The $95\%$ interval is $[\fDiscQSceneLo,\fDiscQSceneHi]$ over scenes and $[\fDiscQLo,\fDiscQHi]$ over systems, and the value describes this comparison set: $\fDiscQNoSeeSR$ without SeeSR, $\fDiscPix$ inside the pixel-supervised family. HQNR gives $\fDiscH$ at full resolution, local concepts included ($\fDiscQB$ on a QuickBird subset). LPIPS comes closest ($\fDiscLP$): much of the disagreement is the perception--distortion trade-off~\cite{blau2018perception} read by a semantic probe; coverage adds the name of what is lost, which is what lets \cref{sec:area,sec:general} resolve the gap by concept and by size and turn it into a loss.

\noindent\textbf{The disagreement lies between two families.} Inside the pixel-supervised family fidelity and coverage agree (Spearman $\rho{=}\fRhoPix$, \cref{fig:gap}a): $\fInvPix$ of $\fPairsPix$ pairs are inverted, none beyond seed noise; language-conditioned systems lie above this trend. A family indicator added to a linear model of coverage on Q2$^n$ raises $R^2$ from $\fRsqQ$ to $\fRsqQFam$, with a coefficient of $+\fFamOff$ coverage ($+\fFamOffFR$ for HQNR at full resolution, $+\fFamOffQB$ on QuickBird). Of the $\fNPairs$ pairs, $\fNInvSig$ are inverted beyond seed noise and Holm-corrected paired tests (\supp, Table~\SRefTabInversions); in each the more retrievable system is language-conditioned and the one with the higher Q2$^n$ pixel-supervised. \fRobSeeSRWCap{} of the \fNInvSigW{} involve SeeSR, whose diffusion prior also introduces concepts (\cref{tab:downstream}), so its recall overstates what it preserves; without it $\delta$ is $\fDiscQNoSeeSR$, the family coefficient $+\fFamOffNoSeeSR$ and the \fNInvNoSeeSRW{} remaining robust inversions are again cross-family. The language-conditioned family is small and half of it was built by us, so the offset is a statement about these six systems; the cleanest comparison is the three conditioning designs that share architecture, data and recipe with the plain U-Net and gain $\fCtrlDcLo$--$\fCtrlDcHi$ coverage over it at a Q2$^n$ within $\fCtrlDqHi$. Inside that family fidelity and coverage are unrelated with a reference ($\rho{=}\fRhoLang$) and opposed without one ($\rho{=}{\fRhoLangFR}$), so the indices order systems reliably only among pixel-supervised networks, more closely with a reference ($\rho{=}\fRhoPix$) than without ($\rho{=}\fRhoPixFR$).

\noindent\textbf{Coverage predicts downstream use.} Given Q2$^n$, the partial Spearman correlation of coverage with OpenCLIP retrieval mAP, ConvNeXt-B scene accuracy and detection mAP under a ConvNeXt-T head (no backbone shared with the probe) is $\fPcCMocQ$, $\fPcCAccQ$ and $\fPcCDetQ$; given LPIPS, the stronger control, $\fPcCMocLP$, $\fPcCAccLP$ and $\fPcCDetLP$, while LPIPS given coverage retains $\fPcLPMocC$, $\fPcLPAccC$ and $\fPcLPDetC$: coverage subsumes LPIPS for retrieval, complements it for recognition and matches it for detection (\supp, Table~\SRefTabPartial).

\noindent\textbf{A shortfall remains for every system.} Language conditioning narrows the gap without closing it: the most retrievable baseline reaches $\cdino{=}\fBestC$ against $\fRefC$ for the reference tiles under the same probe, and the deficit grows as concepts shrink (the recall of the strongest pixel-supervised network falls from $\fRecGloSG$ on global to $\fRecLocSG$ on local concepts). All $\fNSys$ systems train with position-averaged objectives, which \cref{sec:area} shows to under-serve small concepts and \cref{tab:gains} puts to the test.

\section{Method}\label{sec:method}

\begin{figure*}[t]
\centering
\includegraphics[width=\textwidth]{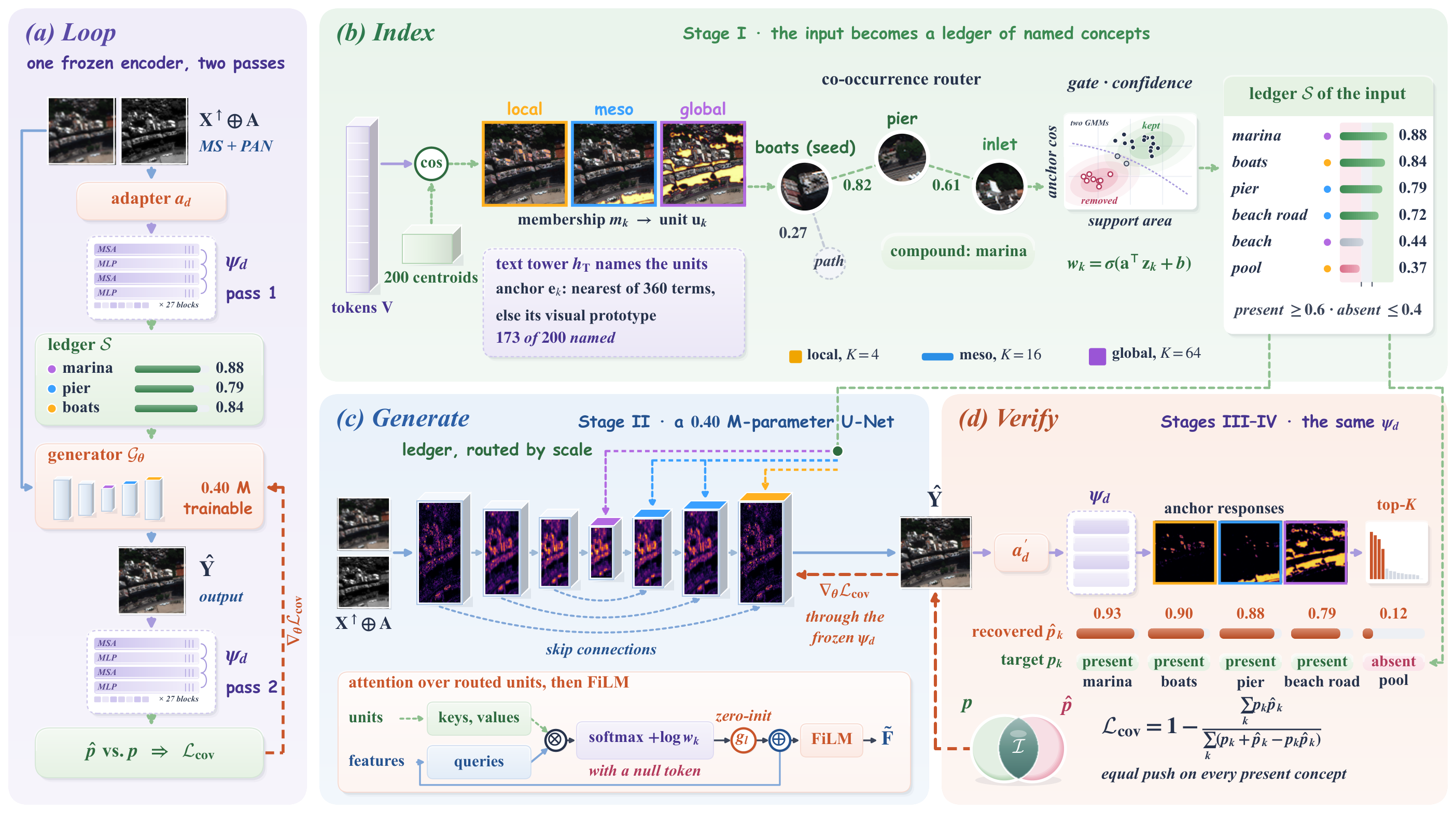}
\caption{\textbf{\our: one frozen encoder, used twice.} (a)~The loop. (b)~Stage~I names the encoder's tokens as units at three scales, composes and gates them and fixes a confidence for each: the ledger. (c)~Stage~II injects the ledger by scale into a U-Net. (d)~Stages~III--IV read a presence per anchor from the output alone and compare it with the detached target by a soft Jaccard. Green: the ledger; orange: the only route of the coverage gradient.}
\label{fig:framework}
\end{figure*}

\subsection{The loop}\label{sec:loop}
\noindent\textbf{Terms.} The \emph{host} is the frozen VLM $\psi{=}(\psi_d,\himg,\htxt)$: dense encoder, per-token head into the joint image-text space, text encoder. A \emph{unit} is a soft group of tokens around one centroid of the token space, its \emph{anchor} the joint-space vector naming it, the \emph{ledger} of an input its list of units with confidences and scale tags, and the \emph{presence} of an anchor in an image how strongly the image's tokens respond to it; host quantities are \emph{in-loop} and never evidence. The host is SigLIP-2 So400m/16 NaFlex~\cite{tschannen2025siglip2} ($428$\,M parameters), fed at twice the output grid so that one token covers $8{\times}8$ output pixels. \cref{fig:framework} shows the four stages of the full model, compound units and gates included (\cref{sec:extensions}); \our-lite is the same four stages without them (pseudocode: \supp, Alg.~\SRefAlgTrain).

\noindent\textbf{Stage I: index the input.} The indexed input is $\mathbf{X}^{\uparrow}\oplus\mathbf{A}$, so that panchromatic-only structure enters the ledger, mapped to the host's three channels by a bounded $1{\times}1$ adapter $a_d$ initialised as a natural-colour composite. Dense tokens $\mathbf{V}$ are softly assigned to $K_c{=}200$ centroids $\{c_k\}$ from $k$-means on training tokens, $m_k{=}\mathrm{softmax}_k(\cos(\mathbf{V},c_k)/\tau)$, and pooled into units $\mathbf{u}_k{=}\sum m_k\mathbf{V}/\sum m_k$. The anchor $\mathbf{e}_k$ of a centroid is the text embedding of its nearest term in a $360$-term vocabulary when their cosine exceeds $0.20$ ($173$ of $200$), else its visual prototype. A logistic calibrator, fitted once on training patches and frozen, turns the unit-anchor cosine, the support area $a_k$ and the scale tag $s_k$ (local, meso or global, by footprint) into a confidence $w_k$; the ledger is $\mathcal{S}{=}\{(\mathbf{u}_k,w_k,s_k)\}$.

\noindent\textbf{Stage II: generate.} $\mathcal{G}_\theta$ is a four-level U-Net~\cite{ronneberger2015unet} with $\fParams$\,M trainable parameters predicting the detail residual. The ledger is injected by scale (local units at the full-resolution decoder level, meso at the two intermediate ones, global at the bottleneck). At level $l$ the routed units, plus a learned null token, are attended with one query per position,
\begin{equation}
\tilde{\mathbf{F}}^{(l)}=\mathbf{F}^{(l)}+g_l\,\mathrm{softmax}_k\!\Bigl(\tfrac{\mathbf{Q}_l\mathbf{K}_l^{\top}}{\sqrt{c_l}}+\log w_k\Bigr)\mathbf{Z}_l,
\label{eq:attn}
\end{equation}
followed by FiLM~\cite{perez2018film}. The bias $\log w_k$ lets a weak unit act only where its content matches, the null token lets a position receive nothing, and the zero-initialised gate $g_l$ starts from the plain U-Net; $\Lrec{=}\|\hat{\mathbf{Y}}{-}\mathbf{Y}\|_1{+}0.1(1{-}\mathrm{SSIM})$~\cite{wang2004ssim}.

\noindent\textbf{Stage III: re-index the output alone.} The output is re-encoded through its own adapter, $\bar{\mathbf{V}}{=}\psi_d(a'_d(\hat{\mathbf{Y}}))$, without the guide, which the archived product lacks. Each anchor yields a response map $\tilde{\mathbf{M}}_k{=}\sigma((\himg(\bar{\mathbf{V}})^{\top}\mathbf{e}_k{-}\mu_k)/\tau_k)$ and the recovered presence is a smooth maximum of its $K$ strongest responses,
\begin{equation}
\hat p_k=\tau_p\log\Bigl(\tfrac{1}{K_{s_k}}\textstyle\sum_{(i,j)\in\mathrm{top}\text{-}K_{s_k}}\exp\bigl(\tilde{\mathbf{M}}_k(i,j)/\tau_p\bigr)\Bigr),
\label{eq:presence}
\end{equation}
with $\tau_p{=}0.1$ and $K_s{=}(4,16,64)$ for the three scales.

\noindent\textbf{Stage IV: set-level coverage.} The target is read from the frozen confidences without gradient, in three zones: present ($p_k{=}1$) for $w_k\ge0.6$, absent ($p_k{=}0$) for $w_k\le0.4$, ignored between. Over the remaining concepts,
\begin{equation}
\Lcov=1-\frac{\sum_k p_k\hat p_k}{\sum_k\bigl(p_k+\hat p_k-p_k\hat p_k\bigr)}=1-\frac{\mathcal{I}}{\mathcal{U}},
\label{eq:lcov}
\end{equation}
the soft Jaccard~\cite{rahman2016jaccard} of the concept set. Its derivative is $-1/\mathcal{U}$ on every present concept and $+\mathcal{I}/\mathcal{U}^2$ on every absent one, so missing and introduced concepts are both penalised, and all present concepts alike. The binary zones fix the sign of the derivative~\cite{wang2023jaccard}. The gradient reaches $\theta$ through the frozen host. The reconstruction and semantic gradients are accumulated in two backward passes; the semantic gradient is rescaled whenever its norm at the output exceeds $5\%$ of the reconstruction gradient's (the ratio settles at $1.9\%$) and is discarded on the Stage-I adapter, so that confidences, target and the input units are constants for $\Lcov$ and the adapter is trained by $\Lrec$ alone (\supp, Table~\SRefTabAdapter).

\subsection{Why a set-level loss helps small concepts}\label{sec:area}
Let the output have $T$ tokens and present concept $k$ a support of $a_k$ tokens. Assume, as an idealisation, that the per-token output gradients, of the loss terms in (i) and of the anchor responses in (ii), are mutually orthogonal with equal norm ($g$ for the responses), and an equal sigmoid slope $\sigma'$ on every support.
\begin{proposition}[Area invariance]\label{prop:area}
(i) Under a token-averaged feature loss $\mathcal{L}_{\mathrm{tok}}{=}\frac1T\sum_t\ell(\bar{\mathbf{V}}_t,\mathbf{V}^{\ast}_t)$ with equal per-token residuals, the share of the squared output-space gradient norm attributable to concept $k$ is $a_k/\sum_j a_j$, its area share. (ii) Under $\Lcov$ with the presence of \cref{eq:presence}, every present concept of scale $s$ whose support fills its top-$K$ window ($a_k\ge K_s$) receives the same gradient norm $\sigma'g/(\mathcal{U}\sqrt{K_s})$ whatever its area, hence the share $1/|\mathcal{P}_s|$ of the squared norm within its scale class; one with $a_k<K_s$ receives at least this norm once its response margin $m_k$ over the background tokens of its window exceeds $\tau_p\ln(1+\sqrt{K_s/a_k})$ ($0.11$ for one token in a window of four), and up to $\sigma'g/(\mathcal{U}\sqrt{a_k})$ as the margin grows (exact form in \supp, Sec.~\SRefSecProof); across classes only $K_s$ differs.
\end{proposition}
\noindent The proof (\supp, Sec.~\SRefSecProof) uses the constant coefficient $-1/\mathcal{U}$ of \cref{eq:lcov} on every present concept and the softmax weights of \cref{eq:presence} over the $K_s$ strongest tokens. Below the margin a sub-window concept receives less; after the warm-up $94\%$ of the present local concepts of the validation tiles exceed a margin of $0.3$, one reason the coverage term starts only then. The statement concerns the distribution of gradient, not its amount or convergence, and its idealisation does not hold for transformer tokens; on validation tiles, concepts of at most four tokens occupy $\fLocalArea\%$ of the area and receive $\fLocalGradTok\%$ of the squared gradient norm under a token-averaged loss and $\fLocalGradCov\%$ under $\Lcov$. Area invariance holds for every loss on the top-$K$ presence, BCE included; the set-level form adds the coupling through $\mathcal{U}$, pushing each present concept by $1/\mathcal{U}$ whatever its presence and each introduced one by $\mathcal{I}/\mathcal{U}^2$.


\begin{table*}[t]
\centering
\caption{\textbf{Pixel fidelity on four sensors.} Reduced (ERGAS, SAM, Q2$^n$) and full resolution ($D_\lambda$, $D_s$, HQNR); mean of three seeds $\pm$ std over $20$ scenes; WV-II unseen. \sig\,/\,\sigw: \our first and separated from the second best at $p{<}0.01$\,/\,$0.05$ (Holm-corrected paired Wilcoxon over scenes; strip-level checks, weaker networks and per-scene values: \supp, Tables~\SRefTabPerscene--\SRefTabStrip). Bold: best; underline: second. DA-CLIP, SeeSR: adapted official code. Inference per $256^2$ scene: \our $\fInferMs$\,ms, CANConv $\fMsCC$\,ms, diffusion networks $1.2$--$2.9$\,s.}
\label{tab:main}
\scriptsize
\setlength{\tabcolsep}{2.0pt}
\renewcommand{\arraystretch}{0.73}
\ReportMethodMatrix{%
\toprule
\ReportMethodRow{Data & Metric}{LAGConv & CrossDiff & SSDiff & Pan-Mamba & CANConv & ARConv}{SGDiff & CLIPPan & DA-CLIP & SeeSR & \our}
\midrule
\ReportMethodRow{\multirow{6}{*}{GF-2} & ERGAS$\downarrow$}{0.692$\pm$0.12 & 0.861$\pm$0.13 & \best{0.603$\pm$0.13} & 0.641$\pm$0.13 & 0.634$\pm$0.13 & 0.629$\pm$0.12}{0.618$\pm$0.12 & 0.742$\pm$0.12 & 0.731$\pm$0.12 & 0.864$\pm$0.12 & \second{0.606$\pm$0.13}}
\ReportMethodRow{ & SAM$\downarrow$}{0.792$\pm$0.14 & 0.945$\pm$0.14 & \second{0.673$\pm$0.15} & 0.712$\pm$0.14 & 0.712$\pm$0.14 & 0.703$\pm$0.14}{0.692$\pm$0.14 & 0.801$\pm$0.14 & 0.788$\pm$0.14 & 0.902$\pm$0.15 & \best{0.648$\pm$0.15}\sig}
\ReportMethodRow{ & Q2$^n\uparrow$}{0.980$\pm$0.007 & 0.972$\pm$0.007 & \second{0.984$\pm$0.007} & 0.982$\pm$0.006 & 0.983$\pm$0.007 & 0.983$\pm$0.007}{0.984$\pm$0.007 & 0.978$\pm$0.007 & 0.979$\pm$0.007 & 0.971$\pm$0.007 & \best{0.987$\pm$0.007}\sig}
\ReportMetricRule
\ReportMethodRow{ & $D_\lambda\downarrow$}{0.029$\pm$0.008 & 0.021$\pm$0.009 & 0.017$\pm$0.007 & 0.022$\pm$0.007 & 0.020$\pm$0.008 & 0.019$\pm$0.008}{0.017$\pm$0.008 & \second{0.016$\pm$0.007} & 0.023$\pm$0.008 & 0.027$\pm$0.007 & \best{0.013$\pm$0.008}\sig}
\ReportMethodRow{ & $D_s\downarrow$}{0.078$\pm$0.009 & 0.056$\pm$0.010 & 0.027$\pm$0.008 & 0.043$\pm$0.010 & 0.062$\pm$0.010 & 0.051$\pm$0.009}{\second{0.024$\pm$0.010} & 0.025$\pm$0.010 & 0.047$\pm$0.010 & 0.057$\pm$0.010 & \best{0.021$\pm$0.010}\sigw}
\ReportMethodRow{ & HQNR$\uparrow$}{0.895$\pm$0.016 & 0.924$\pm$0.017 & 0.957$\pm$0.014 & 0.935$\pm$0.016 & 0.919$\pm$0.017 & 0.931$\pm$0.016}{0.959$\pm$0.017 & \second{0.960$\pm$0.016} & 0.931$\pm$0.016 & 0.918$\pm$0.016 & \best{0.966$\pm$0.017}\sig}
\midrule
\ReportMethodRow{\multirow{6}{*}{QB} & ERGAS$\downarrow$}{3.838$\pm$0.32 & 4.436$\pm$0.33 & 3.641$\pm$0.33 & 3.702$\pm$0.32 & 3.660$\pm$0.34 & 3.642$\pm$0.35}{\second{3.586$\pm$0.33} & 4.118$\pm$0.33 & 3.934$\pm$0.32 & 4.372$\pm$0.32 & \best{3.392$\pm$0.33}\sig}
\ReportMethodRow{ & SAM$\downarrow$}{4.561$\pm$0.80 & 5.389$\pm$0.87 & 4.478$\pm$0.84 & 4.598$\pm$0.83 & 4.519$\pm$0.83 & 4.441$\pm$0.87}{\second{4.364$\pm$0.82} & 4.862$\pm$0.86 & 4.731$\pm$0.80 & 5.104$\pm$0.84 & \best{4.148$\pm$0.88}\sig}
\ReportMethodRow{ & Q2$^n\uparrow$}{0.933$\pm$0.061 & 0.923$\pm$0.061 & 0.934$\pm$0.062 & 0.936$\pm$0.062 & 0.936$\pm$0.063 & \second{0.938$\pm$0.065}}{0.938$\pm$0.063 & 0.918$\pm$0.064 & 0.927$\pm$0.062 & 0.904$\pm$0.064 & \best{0.946$\pm$0.063}\sig}
\ReportMetricRule
\ReportMethodRow{ & $D_\lambda\downarrow$}{0.085$\pm$0.013 & 0.060$\pm$0.014 & 0.032$\pm$0.012 & 0.040$\pm$0.011 & 0.037$\pm$0.014 & 0.039$\pm$0.013}{\best{0.024$\pm$0.013} & 0.026$\pm$0.011 & 0.048$\pm$0.014 & 0.054$\pm$0.013 & \second{0.025$\pm$0.011}}
\ReportMethodRow{ & $D_s\downarrow$}{0.068$\pm$0.011 & 0.077$\pm$0.011 & \second{0.037$\pm$0.010} & 0.046$\pm$0.011 & 0.051$\pm$0.012 & 0.040$\pm$0.013}{0.044$\pm$0.010 & 0.039$\pm$0.012 & 0.054$\pm$0.011 & 0.066$\pm$0.011 & \best{0.033$\pm$0.013}\sigw}
\ReportMethodRow{ & HQNR$\uparrow$}{0.853$\pm$0.021 & 0.868$\pm$0.022 & 0.933$\pm$0.020 & 0.916$\pm$0.020 & 0.914$\pm$0.024 & 0.923$\pm$0.023}{0.933$\pm$0.022 & \second{0.936$\pm$0.022} & 0.901$\pm$0.022 & 0.884$\pm$0.022 & \best{0.944$\pm$0.022}\sig}
\midrule
\ReportMethodRow{\multirow{6}{*}{WV-III} & ERGAS$\downarrow$}{2.311$\pm$0.53 & 2.731$\pm$0.51 & 2.117$\pm$0.51 & 2.191$\pm$0.52 & 2.166$\pm$0.52 & 2.147$\pm$0.52}{\second{2.052$\pm$0.49} & 2.412$\pm$0.53 & 2.286$\pm$0.50 & 2.634$\pm$0.53 & \best{1.962$\pm$0.51}\sig}
\ReportMethodRow{ & SAM$\downarrow$}{3.119$\pm$0.58 & 3.637$\pm$0.60 & 2.855$\pm$0.57 & 2.924$\pm$0.59 & 2.939$\pm$0.56 & 2.896$\pm$0.59}{\second{2.780$\pm$0.57} & 3.206$\pm$0.59 & 3.071$\pm$0.57 & 3.412$\pm$0.61 & \best{2.680$\pm$0.59}\sig}
\ReportMethodRow{ & Q2$^n\uparrow$}{0.909$\pm$0.067 & 0.891$\pm$0.065 & 0.915$\pm$0.065 & 0.919$\pm$0.066 & 0.920$\pm$0.067 & 0.921$\pm$0.063}{\second{0.923$\pm$0.065} & 0.899$\pm$0.068 & 0.911$\pm$0.064 & 0.886$\pm$0.064 & \best{0.929$\pm$0.065}\sig}
\ReportMetricRule
\ReportMethodRow{ & $D_\lambda\downarrow$}{0.036$\pm$0.007 & 0.031$\pm$0.007 & 0.014$\pm$0.006 & 0.018$\pm$0.006 & 0.020$\pm$0.007 & 0.015$\pm$0.005}{\second{0.013$\pm$0.006} & 0.013$\pm$0.006 & 0.028$\pm$0.006 & 0.033$\pm$0.007 & \best{0.011$\pm$0.007}\sig}
\ReportMethodRow{ & $D_s\downarrow$}{0.043$\pm$0.007 & 0.030$\pm$0.007 & 0.031$\pm$0.007 & 0.039$\pm$0.007 & 0.031$\pm$0.008 & 0.028$\pm$0.008}{\second{0.028$\pm$0.007} & 0.029$\pm$0.007 & 0.044$\pm$0.007 & 0.052$\pm$0.008 & \best{0.024$\pm$0.007}\sig}
\ReportMethodRow{ & HQNR$\uparrow$}{0.923$\pm$0.012 & 0.940$\pm$0.013 & 0.956$\pm$0.012 & 0.943$\pm$0.011 & 0.950$\pm$0.013 & 0.957$\pm$0.012}{\second{0.960$\pm$0.012} & 0.958$\pm$0.013 & 0.929$\pm$0.012 & 0.917$\pm$0.013 & \best{0.965$\pm$0.013}\sig}
\midrule
\ReportMethodRow{\multirow{3}{*}{\shortstack{WV-II\\(unseen)}} & ERGAS$\downarrow$}{4.132$\pm$0.37 & 5.231$\pm$0.34 & 3.961$\pm$0.36 & 4.032$\pm$0.36 & 4.010$\pm$0.36 & 3.939$\pm$0.34}{\second{3.874$\pm$0.34} & 4.112$\pm$0.34 & 4.104$\pm$0.36 & 4.427$\pm$0.33 & \best{3.705$\pm$0.35}\sig}
\ReportMethodRow{ & SAM$\downarrow$}{5.236$\pm$0.47 & 6.301$\pm$0.47 & 5.118$\pm$0.45 & 5.183$\pm$0.47 & 5.154$\pm$0.49 & 5.077$\pm$0.46}{\second{4.975$\pm$0.46} & 5.222$\pm$0.48 & 5.232$\pm$0.49 & 5.586$\pm$0.49 & \best{4.800$\pm$0.49}\sig}
\ReportMethodRow{ & Q2$^n\uparrow$}{0.841$\pm$0.047 & 0.775$\pm$0.047 & 0.849$\pm$0.048 & 0.848$\pm$0.046 & 0.852$\pm$0.045 & 0.854$\pm$0.047}{\second{0.858$\pm$0.047} & 0.844$\pm$0.044 & 0.843$\pm$0.047 & 0.824$\pm$0.047 & \best{0.868$\pm$0.047}\sig}
\bottomrule
}
\end{table*}

\subsection{The full model}\label{sec:extensions}
The full model adds two refinements to \our-lite. \emph{Compound units.} Concepts such as \emph{marina} are compositions: a router appends to a seed unit its most co-occurring units (at most three steps) under a relational anchor from $45$ compositions or the normalised sum of the members' anchors; overlapping units are merged and two Gaussian-mixture gates remove fragmentary and weakly named units. \emph{Auxiliary terms.} A presence-gated text-mask term $\Ltm$ constrains where a concept returns, a text-image term $\Ltt$ ties scale-pooled output features to the input's scene text, and a unit-alignment term $\Lsa$ compares input and output units; all are phased in after a reconstruction warm-up. In the \emph{training-only form} nothing is injected and the host only computes the loss, so $\Lcov$ joins another generator's objective with inference unchanged (\cref{sec:general}). The same $\fNHyper$ hyper-parameters serve every sensor and domain; \our-lite keeps \fNHyperLiteW{} of its own and inherits the rest unchanged (templates, $K_{\mathrm{meso}}$, $K_{\mathrm{global}}$, zones, cap, schedule); it keeps $0.80$ of the full model's $0.85$ coverage and $0.923$ of its $0.929$ Q2$^n$, so the refinements are an optional gain (\supp, Table~\SRefTabAblation).

\section{Benchmarks and Evaluation Protocols}\label{sec:protocol}

\noindent\textbf{Data.} We use four PanCollection sensors~\cite{deng2022pancollection}: GaoFen-2 and QuickBird (four bands), WorldView-III and WorldView-II (eight bands; WorldView-II held out), with the official training and validation patches and $20$ reduced-resolution (Wald protocol~\cite{wald1997fusion}) and $20$ full-resolution test scenes per sensor; fidelity is ERGAS, SAM, Q2$^n$~\cite{garzelli2009q2n}, $D_\lambda$, $D_s$ and HQNR~\cite{arienzo2022hqnr}. Design choices used the validation split alone.

\noindent\textbf{Annotated scenes and labelled data.} The $40$ WorldView-III test scenes were annotated before any evaluation: a scene label per $128$-px tile ($\kappa{=}0.89$), $2{,}314$ oriented boxes of $11$ DOTA classes on the full-resolution scenes, and per-tile relevance for $60$ concept queries ($\kappa{=}0.84$): $980$ full-resolution tiles with concepts of all three scales and $180$ reduced-resolution tiles with a reference, carrying meso and global concepts. A \emph{labelled aerial benchmark} of $20{,}000$ DOTA tiles with AID and NWPU-RESISC45~\cite{xia2018dota,xia2017aid,cheng2017nwpu}, processed under the WorldView-III reduced-resolution protocol applied to RGB tiles (panchromatic band derived from the tile under the sensor's modulation transfer function; \supp, Sec.~\SRefSecLabelled), trains and tests the evaluators and, because its guide derives from the reference tile, is reported as a labelled benchmark and never as a sensor result. Three external benchmarks, labelled and captioned by others, follow the same construction: $30$ queries on the fMoW categories~\cite{christie2018fmow} that a synonym and hypernym check places outside the vocabulary (lists and rule in \supp, Table~\SRefTabFmow), and the full RSITMD and RSICD caption sets~\cite{yuan2022rsitmd,lu2018rsicd}. SpaceNet~2 and~3~\cite{vanetten2018spacenet} provide raw WorldView-3 imagery of four cities, scored by building F1 and road APLS~\cite{vanetten2020roads} with vendor-trained segmenters (\cref{tab:downstream}d).

\noindent\textbf{Compared systems.} The $\fNSys$ systems of \cref{sec:gap} are eleven published networks from PanNet~\cite{yang2017pannet} to SGDiff~\cite{xing2025sgdiff}, retrained with their official code and recipes under one data and budget protocol; three VLM-guided methods, CLIPPan~\cite{jian2025clippan}, which is native to pansharpening, and DA-CLIP~\cite{luo2024daclip} and SeeSR~\cite{wu2024seesr}, run from their official implementations with the input layer adapted to multispectral data (adaptation points in \supp, Sec.~\SRefSecRecipe); and four open-loop controls on our backbone and encoder: a plain U-Net, counted with the pixel-supervised family, and three conditioning designs (DenseCLIP-, CLIPDenoising- and DA-CLIP-style), counted with the language-conditioned one. Variants of our method are reported separately, and two concurrent pansharpeners~\cite{li2026msps,cao2026scaleformer} that appeared after the set was frozen are evaluated in \supp (Table~\SRefTabConcurrent) outside every statistic. \our trains with AdamW~\cite{loshchilov2019adamw} for $1{,}000$ epochs.

\noindent\textbf{Evaluators.} Evaluators are ordered by distance from the training signal along two axes. \emph{Backbone.} The host is in the loop. Four CLIP-family retrievers never enter training, OpenCLIP ViT-L/14~\cite{cherti2023openclip}, RemoteCLIP~\cite{liu2024remoteclip}, GeoRSCLIP~\cite{zhang2024georsclip} and the BLIP-2 matching head~\cite{li2023blip2}; members of the host's family, they are held out, not independent. Two evaluators have no language in their backbone: multi-label heads on an ImageNet-supervised ConvNeXt-B~\cite{liu2022convnext} and on Scale-MAE~\cite{reed2023scalemae}. \emph{Concept space.} The $60$ queries, the coverage probe and the two multi-label heads use concepts of the training vocabulary; the fMoW, RSITMD and RSICD benchmarks and the held-out anchors of \cref{sec:independence} do not. Detection heads~\cite{xie2021orcnn} are trained on the labelled benchmark with DINOv2, ConvNeXt-T and ResNet-50 backbones and end to end; hallucination is measured as in CHAIR~\cite{rohrbach2018chair} over the exhaustively annotated box classes; a monitor encoder (\fMonEnc{}) monitors training and scores no result.

\noindent\textbf{Statistics.} Every method is the mean of three seeds, and $\pm$ is the std over test units. Significance marks are Holm-corrected paired Wilcoxon tests over scenes; because the $20$ scenes of a sensor are crops of $4$--$6$ acquisition strips, every marked cell is also checked at strip level against its runner-up: all or all but one strip favour \our, and a mixed-effects model with a random strip intercept gives $p{<}0.05$ in $18$ of the $19$ marked cells ($p{<}0.01$ in $12$) and $0.054$ in the other (\supp, Table~\SRefTabStrip). $\Delta$ is against the strongest baseline of its row.

\section{Results}\label{sec:results}

\subsection{Pixel fidelity}\label{sec:fidelity}
\begin{figure*}[t]
\centering
\includegraphics[width=0.92\textwidth]{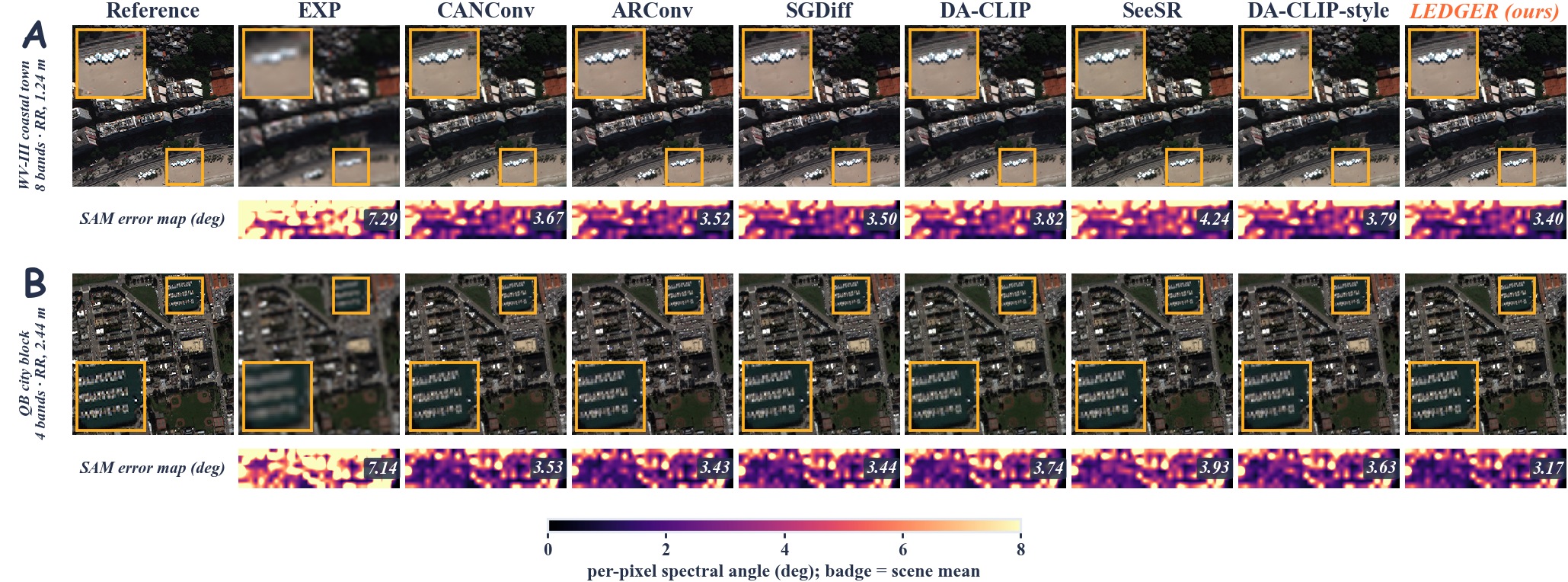}
\caption{\textbf{Reduced-resolution results} on two annotated test scenes (A: WorldView-III coastal town, output GSD $1.24$\,m; B: QuickBird city block, $2.44$\,m). Insets enlarge the gold ROI; strips: per-pixel spectral-angle error on a shared scale; badge: scene mean.}
\label{fig:qualitative}
\end{figure*}

\our ranks first in $\fFirst$ of the $21$ cells of \cref{tab:main} (\cref{fig:qualitative}) and second within noise in the other two, GF-2 ERGAS behind SSDiff and QB $D_\lambda$ behind SGDiff ($p{=}\fPgf$ and $\fPqb$ unadjusted). Against SGDiff, the strongest pixel-supervised network, the margin on WorldView-III is $\fWVdE\%$ in ERGAS and $\fWVdQ$ in Q2$^n$, present on all $\fStripN$ acquisition strips and preserved on the unseen WorldView-II sensor ($\fWTdE\%$, $\fWTdQ$). \cref{tab:gains}b attributes $\fLoopDq$ of the $\fTotalDq$ Q2$^n$ gain over the plain U-Net to the loop, and the controls locate it: at a matched gradient neither a semantics-free high-frequency loss nor the set loss on shuffled anchors moves Q2$^n$ from the plain U-Net's level, and the gain persists on the unseen sensor with both adapters fixed; it is tied to the anchor--concept correspondence, not to the added gradient or to spectral information read by the encoder (\supp, Sec.~\SRefSecMatched). The encoder runs once per scene ($\fInferMs$\,ms; CANConv $\fMsCC$\,ms, diffusion networks $1.2$--$2.9$\,s; \cref{tab:main}); where that cost is not acceptable, the training-only form adds $\fPlugQLo$--$\fPlugQHi$ Q2$^n$ to the generator it joins at unchanged cost (\cref{sec:general}).

\subsection{Retrievability and downstream perception}\label{sec:retrieval}
\cref{tab:downstream}a reports what the fidelity indices miss. On the annotated full-resolution scenes, label-grounded coverage rises from $\fCovBestBase$ (strongest baseline) to $\fCovOurs$, and detection mAP by $\fDetGainDino$ under the DINOv2 head and $\fDetGainCnx$ under the ConvNeXt-T head, which shares no backbone with the probe. On the labelled aerial benchmark (guide derived from the reference tile) \our adds $\fLabelledDetGain$ detection mAP and $\fSmallGain$ small-object AP (\supp, Table~\SRefTabDownstreamRR). The gain does not come from introduced content: over the exhaustively annotated box classes the hallucinated-concept rate equals that of the pixel-supervised networks ($\fHalOurs$ against $\fHalAR$ for ARConv); SeeSR's prior introduces them three times as often.

\begin{table}[t]
\centering
\caption{\textbf{Downstream perception, evaluators by distance from the training signal, and raw sensor data.} (a)~Annotated WorldView-III scenes, $980$ full-resolution tiles; seed std. (b)~$60$ vocabulary queries, mAP (host and BLIP-2 rows in \supp; the two language-free heads score the $39$ queries whose concepts carry labels, Table~\SRefTabQueries); query-bootstrap std of \our. (c)~External benchmarks labelled and captioned by others; $95\%$ interval of the gain. (d)~Raw WorldView-3 SpaceNet imagery, no fine-tuning, segmenters trained on the vendor product; last column: vendor product. $\Delta$: against the strongest of all compared baselines (all methods in \supp).}
\label{tab:downstream}\label{tab:independence}
\scriptsize
\setlength{\tabcolsep}{3pt}
\renewcommand{\arraystretch}{0.71}
\maxsizebox{\columnwidth}{%
\begin{tabular}{lcc>{\columncolor{ourrow}}ccc}
\toprule
Measure / evaluator & SGDiff & SeeSR & \our & $\Delta$ & std\,/\,CI\,/\,ref. \\
\midrule
\multicolumn{6}{l}{\emph{(a) Annotated scenes}} \\
\midrule
Det.\ mAP$_{50}$, ConvNeXt-T head & 33.2 & 33.9 & \best{38.4} & $+$4.5 & $\pm$0.8 \\
Coverage $\cdino$ & 0.67 & 0.72 & \best{0.85} & $+$0.13 & $\pm$0.007 \\
\midrule
\multicolumn{6}{l}{\emph{(b) Vocabulary queries, mAP; evaluator (relation to training)}} \\
\midrule
OpenCLIP (CLIP, held out) & 0.417 & 0.433 & \best{0.503} & $+$0.070 & $\pm$0.026 \\
RemoteCLIP (CLIP, held out) & 0.430 & 0.453 & \best{0.540} & $+$0.087 & $\pm$0.025 \\
GeoRSCLIP (CLIP, held out) & 0.450 & 0.453 & \best{0.547} & $+$0.094 & $\pm$0.026 \\
ConvNeXt-B head (no language) & 0.380 & 0.397 & \best{0.473} & $+$0.076 & $\pm$0.028 \\
Scale-MAE head (no language) & 0.393 & 0.410 & \best{0.487} & $+$0.077 & $\pm$0.028 \\
\midrule
\multicolumn{6}{l}{\emph{(c) External benchmarks, labelled and captioned by others}} \\
\midrule
fMoW-OOV mAP, OpenCLIP & 0.37 & 0.39 & \best{0.43} & $+$0.04 & [0.02, 0.06]  \\
fMoW-OOV mAP, ConvNeXt-B head & 0.34 & 0.36 & \best{0.40} & $+$0.04 & [0.01, 0.06]  \\
RSITMD R@5, OpenCLIP & 0.41 & 0.43 & \best{0.48} & $+$0.05 & [0.04, 0.06] \\
\midrule
\multicolumn{6}{l}{\emph{(d) SpaceNet~2/3, raw sensor data, no fine-tuning; small: below $60$\,m$^2$}} \\
\midrule
Building F1 & 0.673 & 0.662 & \best{0.691} & $+$0.018 & 0.681 \\
Small-building F1 & 0.50 & 0.51 & \best{0.56} & $+$0.05 & 0.52 \\
Road APLS & 0.633 & 0.618 & \best{0.656} & $+$0.023 & 0.642 \\
\bottomrule
\end{tabular}%
}
\end{table}

\begin{table}[t]
\centering
\caption{\textbf{Where the gain comes from} (WV-III; seed std $0.002$ Q2$^n$, $0.01$ coverage, $0.005$ mAP). (a)~One term added to $\Lrec$ on the injected architecture, same schedule, no auxiliary term, at the largest weight keeping validation ERGAS within $1\%$; GDL: inverse-area re-weighting. (b)~Components added in turn; mAP under OpenCLIP.}
\label{tab:gains}
\scriptsize
\setlength{\tabcolsep}{3pt}
\renewcommand{\arraystretch}{0.72}
\maxsizebox{\columnwidth}{%
\begin{tabular}{lccc}
\toprule
Setting & Q2$^n\uparrow$ & $\cdino\uparrow$ & mAP$\uparrow$ \\
\midrule
\multicolumn{4}{l}{\emph{(a) Same architecture and frozen encoder; one term added to $\Lrec$}} \\
\midrule
none & 0.913 & 0.72 & 0.42 \\
token features $\to$ degraded input & 0.921 & 0.76 & 0.44 \\
token features $\to$ reference & 0.922 & 0.78 & 0.45 \\
anchor responses $\to$ reference & 0.923 & 0.79 & 0.46 \\
token features $\to$ reference, GDL-weighted & 0.924 & 0.81 & 0.47 \\
per-concept BCE on top-$K$ presences & 0.925 & 0.80 & 0.48 \\
detector loss (task-driven)~\cite{haris2021taskdriven,kim2024sr4ir} & 0.915 & 0.74 & 0.44 \\
set-level $\Lcov$, reference set & 0.923 & 0.82 & 0.48 \\
set-level $\Lcov$, input set (\our's term) & \best{0.924} & \best{0.83} & \best{0.49} \\
\midrule
\multicolumn{4}{l}{\emph{(b) Components, added in turn}} \\
\midrule
Plain U-Net & 0.910 & 0.55 & 0.37 \\
+ ledger injection, no loop & 0.913 & 0.72 & 0.42 \\
+ $\Ltt$, $\Ltm$ (open-loop \our) & 0.918 & 0.72 & 0.43 \\
\rowcolor{ourrow}+ $\Lcov$, $\Lsa$: closed loop (\our) & \best{0.929} & \best{0.85} & \best{0.50} \\
\midrule
anchors shuffled / paraphrased & 0.914 / 0.928 & 0.60 / 0.84 & 0.36 / 0.50 \\
\bottomrule
\end{tabular}%
}
\end{table}

\subsection{How far from the loop the gain holds}\label{sec:independence}
\cref{tab:downstream}b,c order the evaluators by their distance from the training signal. On queries of the training vocabulary, the gain over the strongest baseline is $+\fRetClipLoF$ to $+\fRetClipHiF$ mAP under the four held-out CLIP-family retrievers and $+\fRetNoLangLoF$ and $+\fRetNoLangHiF$ under the two evaluators without language: the recovered content is readable by models outside training, still as vocabulary concepts. Two measurements leave that space: on anchors held out of the vocabulary before training ($25\%$ of the terms in synonym groups, three draws) the coverage gain over SGDiff is $+0.07$ with the centroids kept as visual prototypes and $+0.05$ with them removed (\supp, Table~\SRefTabSwap); on external benchmarks labelled and captioned by others it is $+\fExtLo$ to $+\fExtHi$ under every evaluator (mAP and R@5 $+0.04$ to $+0.05$, R@1 $+0.02$ to $+0.03$; \supp, Table~\SRefTabVocabSplit), so the gain keeps its sign beyond the vocabulary under every backbone. At archive scale, among $39{,}020$ raw WorldView-3 distractor tiles with pooled relevance judgments of every method's top ranks, the OpenCLIP margin is $+0.06$ mAP ($0.38$ against $0.32$; \supp, Table~\SRefTabArchive). A LoRA~\cite{hu2022lora} fitted to SGDiff's outputs recovers only $\fRaSG$ of its $\fRaGap$ host-mAP deficit (\supp, Table~\SRefTabRetadapt); under the budgets tested, the shortfall remains in the pixels. The recovered content survives archive post-processing (quantisation, JPEG~2000 and resampling move the gains by at most $0.01$) and is spectrally real inside the masks of the smallest concepts (\supp, Table~\SRefTabPostproc). The regime in which the host is satisfied at the expense of the pixels is excluded and monitored: pixel optimisation with $\Lcov$ alone drives in-loop coverage to $0.97$ while PSNR falls to $21.3$\,dB and the probe to $0.41$; the reconstruction weight and the gradient cap keep training out of it, and the in-loop--monitor gap ($\fXencVal$ in our runs, $\fGapZero$ there) would register a drift (\supp, Table~\SRefTabGaming).

\subsection{Generalisation and analysis}\label{sec:general}
\noindent\textbf{Raw sensor data and a training-only loss.} Applied without fine-tuning to raw WorldView-3 SpaceNet imagery (\cref{tab:downstream}d), \our is the only learned method above the vendor product on all three measures (building F1 $\fSnOursFone$ against $\fSnVendFone$, small-building F1 $\fSnOursSmall$ against $\fSnVendSmall$, road APLS $\fSnOursApls$ against $\fSnVendApls$; $\fSnVendorFoneStd$ and $\fSnVendorAplsStd$ seed std with one segmenter per task; positive in all four cities; every method and city in \supp, Table~\SRefTabSpacenet). Added to CANConv, ARConv and the plain U-Net during training only (\supp, Table~\SRefTabPlugin), $\Lcov$ raises coverage by $\fPlugCovLo$--$\fPlugCovHi$ and detection mAP by $\fPlugDetLo$--$\fPlugDetHi$ at unchanged inference, and Q2$^n$ by $\fPlugQLo$--$\fPlugQHi$; the distance from the U-Net's $0.74$ to the full model's $0.85$ is injection at inference ($+0.09$, the one-term row of \cref{tab:gains}a) and the unit-alignment term ($+0.02$; \supp, Table~\SRefTabLadder).

\noindent\textbf{Source of the gain.} \cref{tab:gains}a holds the injected architecture, the frozen encoder and the schedule fixed and changes only the term added to $\Lrec$: feature losses averaged over positions reach $\fObjTokCLo$--$\fObjTokCHi$ coverage; re-weighting the token loss by inverse concept area with the ledger's membership masks, the Generalised-Dice counterpart of \cref{prop:area}, reaches $0.81$, per-concept BCE on the top-$K$ presences $0.80$ and the set-level loss $\fObjSetCLo$--$\fObjSetCHi$ with either ledger as target ($p{<}0.01$ against each alternative, scene-paired test on seed-averaged values); swapped inside the full model the order is the same ($0.82$, $0.82$, $0.85$; \supp, Table~\SRefTabObjective): area invariance alone, by re-weighting or by a per-concept presence loss, recovers part of the margin over position-averaged losses and the set-level coupling the rest. Block~b separates architecture from objective: the ledger lifts coverage by $+\fInjDc$ when injected and by a further $+\fLoopDc$ when verified, and the loop carries most of the fidelity ($\fLoopDq$ of $\fTotalDq$ Q2$^n$), retrieval ($+0.07$ of $+0.13$ mAP) and, by \cref{prop:area}, small-concept gains, which no per-concept or position-averaged objective matches at equal architecture. A task-driven detector loss~\cite{haris2021taskdriven,kim2024sr4ir} on the same architecture recovers $\fDetSmallShare\%$ of our small-object gain on its own benchmark, $\fDetRealShare\%$ on the real scenes and $0.02$ of the $0.07$ retrieval gain (\cref{tab:gains}a).

\noindent\textbf{The gain follows concept size; the design is robust.} Binned by token support (\supp, Table~\SRefTabAreabins), the recall gain over the strongest position-averaged loss is $+0.13$ on concepts of at most four tokens, $+0.08$ on $5$--$16$, $+0.03$ on $17$--$64$ and none on tile-filling ones, which that loss already recovers at $0.90$, as \cref{prop:area} predicts. Changing one design choice at a time moves coverage by at most $0.05$ (\supp, Table~\SRefTabControls). At test time a null ledger still gives $0.73$ coverage, above the open-loop model (part of what the loop teaches lives in the weights), and $50\%$ spurious units raise the hallucination rate by $0.007$ (\supp, Table~\SRefTabLedgernoise).

\noindent\textbf{Other domains and raters.} The recipe transfers to histopathology~\cite{bejnordi2017camelyon} and natural photographs~\cite{lin2014coco} (\supp, Sec.~\SRefSecDomains); $\fHumRaters$ raters prefer \our to SGDiff for small objects in $\fHumPref\%$ of $\fHumPairs$ blind pairs ($\fHumScenes$ of $\fHumNScenes$ scenes, all six strips; $p{=}\fHumP$), do not separate them on naturalness ($\fHumPrefTwo\%$, $p{=}\fHumPTwo$; low agreement, inconclusive) and prefer \our to SeeSR in a non-significant $59\%$.

\noindent\textbf{Limitations.} Like the compared networks, \our trains one model per sensor; transfer across sensors without fine-tuning is reported in \supp (Table~\SRefTabCross), and a single multi-sensor model is left to future work. The vocabulary, compositions and templates are written per domain. The full model runs the frozen encoder once per scene ($\fInferMs$\,ms); where an inference budget rules this out, the training-only form applies (\cref{sec:general}). Reduced-resolution coverage is a scene-level recall, so the small-concept findings rest on the full-resolution protocol, where human raters validate the probe (\supp, Table~\SRefTabProbeVal).

\FloatBarrier
\section{Conclusion}\label{sec:conclusion}
Guided restoration is now read by an index it did not optimise for: fidelity indices disagree with retrievability on one system pair in three to one in two, rank systems reliably only within the pixel-supervised family, and every system loses small concepts first, as the area weighting of position-averaged objectives predicts. A set-level coverage loss through one frozen encoder removes that weighting, closes two thirds of the reduced-resolution gap to the reference under outside evaluators and keeps its sign beyond its vocabulary.

\endgroup
\par\FloatBarrier\addvspace{12pt}
\ReportContentsPage
\par\addvspace{12pt}
\appendix
\setcounter{section}{0}
\setcounter{subsection}{0}
\setcounter{figure}{0}
\setcounter{table}{0}
\setcounter{equation}{0}
\setcounter{algorithm}{0}
\setcounter{proposition}{0}
\renewcommand{\thefigure}{S\arabic{figure}}
\renewcommand{\thetable}{S\arabic{table}}
\renewcommand{\theequation}{S\arabic{equation}}
\renewcommand{\thealgorithm}{S\arabic{algorithm}}
\renewcommand{\theHsection}{supp.\Alph{section}}
\renewcommand{\theHsubsection}{\theHsection.\arabic{subsection}}
\renewcommand{\theHfigure}{supp.\arabic{figure}}
\renewcommand{\theHtable}{supp.\arabic{table}}
\renewcommand{\theHequation}{supp.\arabic{equation}}
\providecommand{\theHalgorithm}{}
\renewcommand{\theHalgorithm}{supp.\arabic{algorithm}}
\providecommand{\theHproposition}{}
\renewcommand{\theHproposition}{supp.\arabic{proposition}}
\makeatletter
\providecommand{\theHALG@line}{}
\renewcommand{\theHALG@line}{\theHalgorithm.\arabic{ALG@line}}
\makeatother

\noindent This document contains the proof of the area-invariance proposition with its measured counterpart, its relation to class-rebalanced overlap losses, the gradient analysis and a matched-gradient control (\cref{sec:theory}); pseudocode for index fitting, the full training step, inference and the training-only form, and for every evaluation protocol (\cref{sec:algos}); the protocol details of the data splits and their uses, the labelled aerial benchmark, the annotated scenes and the concepts each protocol covers, the query bank, the vocabulary and the eleven manual centroid re-assignments, the evaluators and their two axes of distance from the training signal, the SpaceNet protocol, the retriever-adaptation alternative, the human study and the statistics (\cref{supp:sec:protocol}); the architecture, index fitting, training recipe, hyper-parameters and compute (\cref{sec:impl}); the complete results behind every number of the main paper, including every per-scene value behind the paired tests at the precision the tests use, the full analysis of the retrievability gap on two sensors and two protocols, and additional visualisations (\cref{supp:sec:results}); two further domains (\cref{sec:domains}); and scope and broader impact (\cref{sec:limits}). References of the form ``Main Table~\MTabMain'' point to the main paper. All numbers follow the seed and statistics protocol of \cref{sec:stats}.

\section{Theory: area invariance and the gradient through the encoder}\label{sec:theory}

\subsection{Setting and notation}\label{toc:supp:1:1}
Write $q_k{=}\hat p_k$ for the recovered presence of concept $k$ (the top-$K$ smooth maximum of Main Eq.~\MEqPresence), $\mathcal{P}{=}\{k:p_k{=}1\}$ for the present set of the detached three-zone target, $\mathcal{A}$ for the absent set, and exclude the ignored zone from every sum. The soft intersection and union are $\mathcal{I}{=}\sum_{k\in\mathcal{P}}q_k$ and $\mathcal{U}{=}|\mathcal{P}|{+}\sum_{k\in\mathcal{A}}q_k$, so that $\Lcov{=}1{-}\mathcal{I}/\mathcal{U}$. The output $\hat{\mathbf{Y}}$ is encoded into $T$ tokens; $S_k(t)$ is the anchor response at token $t$, $\tilde{\mathbf{M}}_k(t){=}\sigma((S_k(t){-}\mu_k)/\tau_k)$, and $\mathbf{g}_{k,t}{=}\nabla_{\hat{\mathbf{Y}}}S_k(t)$ the output-space gradient of one token response. Present concept $k$ has an active support $\Omega_k$ of $a_k$ tokens.

\subsection{Proposition 1 and its proof}\label{sec:proof}
\begin{proposition}[Area invariance; Main Proposition~\MProp]\label{supp:prop:area}
Assume that the per-token gradients at the output are mutually orthogonal across tokens and of equal norm: in (i) the gradients $\nabla_{\hat{\mathbf{Y}}}\ell_t$ of the per-token loss terms, of norm $g'$; in (ii) the response gradients $\{\mathbf{g}_{k,t}\}$, of norm $g$; and assume that within a support all responses are equal. (i)~Under a token-averaged feature loss $\mathcal{L}_{\mathrm{tok}}{=}\frac1T\sum_t\ell(\bar{\mathbf{V}}_t,\mathbf{V}^{\ast}_t)$, the share of $\|\nabla_{\hat{\mathbf{Y}}}\mathcal{L}_{\mathrm{tok}}\|^2$ attributable to the tokens of concept $k$ is $a_k/\sum_j a_j$. (ii)~Under $\Lcov$ with the top-$K$ presence, where $K{=}K_s$ depends on the scale tag $s$ of the concept, every present concept of scale $s$ with $a_k\ge K_s$ receives the output-space gradient norm $\sigma'_k\,g/(\mathcal{U}\sqrt{K_s})$, where $\sigma'_k{=}\tilde{\mathbf{M}}_k(1{-}\tilde{\mathbf{M}}_k)/\tau_k$ is the common sigmoid slope on its support; a present concept with $a_k<K_s$, whose top-$K$ window also holds $K_s{-}a_k$ background tokens of response $r_-$ below the support response $r_+$, receives $\sigma'_k\,g\,\sqrt{a_k}\,\omega_+/\mathcal{U}$ with $\omega_+{=}1/\bigl(a_k+(K_s{-}a_k)\varepsilon_k\bigr)$ and $\varepsilon_k{=}e^{-(r_+-r_-)/\tau_p}$ (the background slopes being negligible), a norm that increases from $\sigma'_k g\sqrt{a_k}/(\mathcal{U}K_s)$ at $\varepsilon_k{=}1$ to $\sigma'_k g/(\mathcal{U}\sqrt{a_k})$ as $\varepsilon_k\to0$ and equals the window-filling norm $\sigma'_k g/(\mathcal{U}\sqrt{K_s})$ at $\varepsilon_k{=}\sqrt{a_k}/(\sqrt{K_s}+\sqrt{a_k})$, that is, at the response margin $r_+{-}r_-{=}\tau_p\ln(1+\sqrt{K_s/a_k})$; above this margin a sub-window concept receives at least the window-filling norm, below it less; within a scale class the share of a present concept in $\sum_{k\in\mathcal{P}_s}\|\nabla_k\|^2$ is $1/|\mathcal{P}_s|$ when all its concepts have $a_k\ge K_s$ and equal slopes, whatever their areas, and across classes the norms differ through $K_s$ alone, by a factor of $\sqrt{K_g/K_l}{=}4$ between global and local concepts. Part (i) is a statement about squared norms and part (ii) about norms; the main text uses the squared norm in (i) and the norm in (ii) accordingly.
\end{proposition}

\noindent\textbf{Proof.} (i) The gradient of the token-averaged loss is $\nabla\mathcal{L}_{\mathrm{tok}}{=}\frac1T\sum_t\nabla_{\hat{\mathbf{Y}}}\ell_t$, a sum of $T$ token terms. Under the assumption each term $\frac1T\nabla_{\hat{\mathbf{Y}}}\ell_t$ has norm $g'/T$ and the terms are orthogonal (the assumption is placed on the loss gradients themselves, since orthogonal response gradients of equal norm would not by themselves make the Jacobian-vector products $\nabla_{\hat{\mathbf{Y}}}\ell_t$ orthogonal or equal in norm), so $\|\nabla\mathcal{L}_{\mathrm{tok}}\|^2{=}T\,g'^2/T^2$ and the part of the sum over $\Omega_k$ has squared norm $a_k g'^2/T^2$. Its share is $a_k/\sum_j a_j$ when the supports partition the tokens that carry residual. \\
(ii) By the quotient rule, $\partial\Lcov/\partial q_k{=}-(\mathcal{U}\partial_k\mathcal{I}-\mathcal{I}\partial_k\mathcal{U})/\mathcal{U}^2$; for $k\in\mathcal{P}$, $\partial_k\mathcal{I}{=}1$ and $\partial_k\mathcal{U}{=}0$, giving $-1/\mathcal{U}$ independently of $k$; for $k\in\mathcal{A}$, $\partial_k\mathcal{I}{=}0$ and $\partial_k\mathcal{U}{=}1$, giving $+\mathcal{I}/\mathcal{U}^2$. The top-$K$ mellowmax $q_k{=}\tau_p\log\frac1K\sum_{t\in\mathrm{top}\text{-}K}e^{\tilde{\mathbf{M}}_k(t)/\tau_p}$ has $\partial q_k/\partial\tilde{\mathbf{M}}_k(t){=}\omega_{k,t}{=}e^{\tilde{\mathbf{M}}_k(t)/\tau_p}/\sum_{t'\in\mathrm{top}\text{-}K}e^{\tilde{\mathbf{M}}_k(t')/\tau_p}$ for $t$ in the top-$K$ and zero otherwise; the weights are a convex combination. With equal responses on the support and $a_k\ge K$, the top-$K$ set is any $K$ tokens of $\Omega_k$ and $\omega_{k,t}{=}1/K$ on them. By the chain rule
\begin{equation}
\begin{aligned}
\nabla_{\hat{\mathbf{Y}}}q_k&=\sum_{t\in\mathrm{top}\text{-}K}\omega_{k,t}\,\sigma'_k\,\mathbf{g}_{k,t},\\
\|\nabla_{\hat{\mathbf{Y}}}q_k\|^2&=\sigma'^2_k g^2\sum_t\omega_{k,t}^2=\frac{\sigma'^2_k g^2}{K},
\end{aligned}
\end{equation}
using orthogonality and equal norms; multiplying by $1/\mathcal{U}$ gives the stated norm, and tokens outside the top-$K$ have $\omega_{k,t}{=}0$, so enlarging the background does not change $\nabla q_k$. When $a_k<K$ the window is completed by $K{-}a_k$ background tokens; with responses $r_+$ on the support and $r_-$ on the background the softmax weights are $\omega_+{=}e^{r_+/\tau_p}/(a_ke^{r_+/\tau_p}+(K{-}a_k)e^{r_-/\tau_p}){=}1/(a_k+(K{-}a_k)\varepsilon_k)$ on the support and $\omega_-{=}\varepsilon_k\omega_+$ on the background, $\varepsilon_k{=}e^{-(r_+-r_-)/\tau_p}$. The squared norm is $g^2\bigl(a_k\sigma'^2_k\omega_+^2+(K{-}a_k)\sigma'^2_{-}\omega_-^2\bigr)$; background tokens have responses near $0$ or $1$ and therefore slopes $\sigma'_-$ near zero, so the second term is negligible and $\|\nabla q_k\|{=}\sigma'_kg\sqrt{a_k}\,\omega_+$. At $\varepsilon_k{=}1$ (no margin: the window is one uniform average) this is $\sigma'_kg\sqrt{a_k}/K$, below the window-filling value $\sigma'_kg/\sqrt{K}$ by the factor $\sqrt{a_k/K}$; it increases monotonically as $\varepsilon_k$ falls, equals $\sigma'_kg/\sqrt{K}$ when $\sqrt{a_k}/(a_k+(K{-}a_k)\varepsilon_k){=}1/\sqrt{K}$, that is, at $\varepsilon_k{=}\sqrt{a_k}/(\sqrt{K}+\sqrt{a_k})$ or a margin of $\tau_p\ln(1+\sqrt{K/a_k})$, and tends to $\sigma'_kg/\sqrt{a_k}$ as $\varepsilon_k\to0$. With $\tau_p{=}0.1$ the margin needed is $0.11$ for one token in a window of four, $0.16$ for one token in a window of sixteen and $0.22$ for one in a window of sixty-four; a margin of $0.3$ gives $\varepsilon_k{=}e^{-3}{\approx}0.05$, so a one-token concept in a window of four has $\omega_+{=}0.87$ and a norm within $13\%$ of the limit, a two-token concept within $5\%$. After the reconstruction warm-up $94\%$ of the present local concepts of the validation tiles have a margin of at least $0.3$; below the margin the norm of a sub-window concept grows with its area toward the window-filling value (for $K{=}4$ and a margin of $0.05$ the norms of one- to four-token concepts are in the ratio $0.71:0.88:0.96:1$), so for the smallest concepts the statement is one about trained models, in which the margin condition holds, and not about initialisation, which is one reason the coverage term is switched on after the warm-up (\cref{sec:monitors}). When all present concepts have $a_k\ge K$ and equal slopes, every present concept has the same squared norm $\sigma'^2g^2/(K\mathcal{U}^2)$ and the share is $1/|\mathcal{P}|$. \hfill$\square$

\begin{figure}[!htbp]
\centering
\includegraphics[width=\columnwidth]{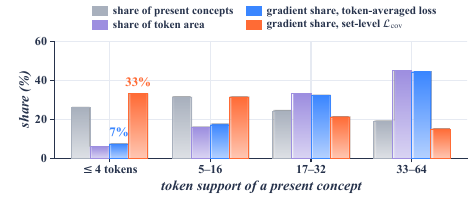}
\caption{Measured gradient shares on the WorldView-III validation patches ($64$ tokens each) at the end of training, present concepts binned by token support (the data of \cref{tab:area}). A token-averaged feature loss distributes gradient in proportion to area; under $\Lcov$ the share per present concept falls with the support, which is the allocation that \cref{supp:prop:area} describes.}
\label{fig:area}
\end{figure}

\noindent\textbf{What the proposition says and does not say.} It compares how two losses \emph{distribute} gradient across the concepts of one tile, under an idealisation (per-token loss and response gradients of equal norm that are mutually orthogonal, and equal responses within a support) which the tokens of a transformer do not satisfy and which serves to make the comparison exact. It does not claim that $\Lcov$ is larger than $\Lrec$ (it is capped at $5\%$ of it), nor that training converges, nor that coverage must improve; it says that whatever gradient the coverage term carries is split by concept count, not by area, which is what a retrieval index needs; concepts smaller than their top-$K$ window receive at least an equal share once their response margin exceeds $\tau_p\ln(1+\sqrt{K_s/a_k})$, which holds for $94\%$ of the present local concepts after the warm-up, so in the trained model the share of a present concept does not grow with its area. The measured counterpart (\cref{tab:area}, \cref{fig:area}) shows that the idealisation is close to the real case: token gradients on real tiles are correlated and unequal, yet concepts of at most four tokens, holding $\fLocalArea\%$ of the area, receive $\fLocalGradTok\%$ of the squared output-space gradient norm of a token-averaged loss and $\fLocalGradCov\%$ of that of $\Lcov$ (the per-concept quantity of the proposition, computed as stated in \cref{tab:area}).

\noindent\textbf{Relation to Dice-type losses.} Dice~\cite{milletari2016vnet} and Generalised Dice~\cite{sudre2017gdl} losses address class imbalance in segmentation by normalising a mask overlap per class, so that gradient is allocated per class rather than per pixel; Generalised Dice additionally weights each class by the inverse of its squared volume. \cref{supp:prop:area} states the same principle for a different object. No per-pixel target is assumed: the classes are open-vocabulary concepts that exist only as responses of a frozen encoder, each concept is first reduced to one scalar by the top-$K$ presence, and the overlap is then computed once per image over the set of concepts. What the proposition adds is not the principle but the object it is applied to and the aggregate under which the share of a concept is independent of both its area and the token grid; \cref{tab:area} measures that the allocation takes effect on real tiles.

\noindent\textbf{What distinguishes the set-level form from a per-concept loss.} Area invariance is a property of the top-$K$ presence and holds for every loss placed on it: per-concept binary cross-entropy on $q_k$, focal cross-entropy, soft Dice and the Lov\'asz hinge all inherit it, and \cref{supp:prop:area} accordingly explains the margin of presence-based losses over position-averaged ones in \cref{tab:objectives_full} ($0.80$--$0.83$ against $0.76$--$0.79$), not the margin among them. The set-level form differs from a per-concept loss in how the concepts of one tile are coupled. Under $\Lcov$ the coefficient on a present concept is $-1/\mathcal{U}$, the same for every present concept whatever its own presence $q_k$, so the last concept still missing is pushed as hard as the first, and the coefficient on an introduced concept, $+\mathcal{I}/\mathcal{U}^2$, grows with the recovered mass $\mathcal{I}$, so suppression of introduced content is deferred until the present set is largely recovered. Per-concept BCE has coefficient $-1/q_k$ on a present concept, which concentrates the push on the concept with the lowest presence and lets it fade as $q_k\to1$, and a coefficient on introduced concepts that does not depend on the rest of the set. \cref{tab:objective} separates the per-concept losses (BCE, focal BCE) from the set-level overlap losses by $0.02$--$0.03$ coverage ($p{<}0.01$); the three overlap losses lie within $0.02$ of one another, inside seed noise, as expected of losses whose per-sample gradients differ by a positive scalar (soft Dice and soft Jaccard satisfy $D{=}2J/(1{+}J)$ and $\nabla L_D{=}2\nabla L_J/(1{+}J)^2$ on the same presence vector), so the table supports the set-level coupling and not an ordering among these losses. The margin of $\Lcov$ over per-concept BCE on the same presences is $0.03$ coverage in both settings of \cref{sec:objectives} ($p{<}0.01$ by the scene-paired test on seed-averaged values, \cref{sec:stats}), concentrated on the smallest concepts (\cref{tab:areabins}).

\begin{table}[!htbp]
\centering
\caption{Measured gradient shares behind \cref{supp:prop:area} (the $1{,}080$ WorldView-III validation patches, $64$ tokens each on the $2\times$ grid, the validation tiles of Main Sec.~\MSecMethod{} and \cref{sec:monitors}; epoch $1{,}000$; present concepts binned by the size of their token support, which on these patches cannot exceed $64$). Columns: share of present concepts in the bin, share of the token area they occupy, share of the squared output-space gradient norm they receive under a token-averaged feature loss (SigLIP tokens $\to$ reference, the strongest feature baseline of \cref{tab:objectives_full}), and under $\Lcov$ with top-$K$ presence. The share of a bin is $\sum_{k\in\mathrm{bin}}\|\nabla_k\|^2/\sum_k\|\nabla_k\|^2$, where $\nabla_k$ is the output-space gradient of the terms that belong to concept $k$ alone ($c_k\nabla q_k$ under $\Lcov$; the sum of its tokens' terms under the token loss), so cross terms between concepts do not enter; a token shared by two supports ($3\%$ of tokens) is counted with each. This is the quantity of \cref{supp:prop:area}. The token-averaged loss distributes gradient in proportion to area; under the set-level loss the share per present concept decreases with the support ($1.27$, $1.00$, $0.88$ and $0.79$ times that of an average present concept), as the top-$K$ windows of \cref{supp:prop:area} predict, less steeply than in its idealisation because the token gradients of real patches are correlated and unequal and because sub-window concepts below their margin receive less.}
\label{tab:area}
\scriptsize
\setlength{\tabcolsep}{4pt}
\maxsizebox{\columnwidth}{%
\begin{tabular}{lcccc}
\toprule
Support size & concepts (\%) & token area (\%) & gradient share, token loss (\%) & gradient share, $\Lcov$ (\%) \\
\midrule
$\le4$ tokens & 26 & 6 & 7 & 33 \\
$5$--$16$ & 31 & 16 & 17 & 31 \\
$17$--$32$ & 24 & 33 & 32 & 21 \\
$33$--$64$ & 19 & 45 & 44 & 15 \\
\bottomrule
\end{tabular}%
}
\end{table}

\subsection{Why the target is binarised}\label{toc:supp:1:3}
With a soft target $p_k\in(0,1)$ the same computation gives $\partial\Lcov/\partial q_k{=}-\bigl(p_k-(1{-}p_k)J\bigr)/\mathcal{U}$ with $J{=}\mathcal{I}/\mathcal{U}$, which vanishes when $p_k/(1{-}p_k){=}J$ and changes sign below it. The instance $p{=}(\tfrac15,\tfrac45)$, $q{=}(\tfrac12,\tfrac13)$ gives $J{=}0.25$ and $\partial\Lcov/\partial q_1\propto-(0.2-0.8\cdot0.25){=}0$: the half-recovered first concept receives no push. Wang~\etal~\cite{wang2023jaccard} analyse this soft-label behaviour of Jaccard losses in general; the detached three-zone target (present $\ge0.6$, absent $\le0.4$, ignored between) is our design choice to stay in the regime where the sign is fixed, and the ignore zone is where the two annotators of \cref{sec:annot} disagreed most.

\subsection{Remarks on the frozen-encoder gradient}\label{sec:monitors}
The coverage gradient reaches $\theta$ as $\nabla_\theta\Lcov{=}\mathbf{J}_{\mathcal{G}}^{\top}\sum_k c_k\nabla_{\hat{\mathbf{Y}}}q_k$ with $c_k{=}-1/\mathcal{U}$ on present and $+\mathcal{I}/\mathcal{U}^2$ on absent concepts. It can vanish in three ways, which are monitored on a fixed batch of $32$ \emph{validation} patches every $10$ epochs and logged in \cref{tab:trainlog} and \cref{fig:training}: (C1)~saturation, when every token of a present concept has $\tilde{\mathbf{M}}_k\in\{0,1\}$ up to $0.05$, so $\sigma'_k{=}0$ ($59\%$ of present concepts at initialisation, $11.8\%$ at epoch $100$, $4.6\%$ at epoch $200$, below $4\%$ afterwards; this is why $\Lcov$ starts at epoch $200$); (C2)~cancellation, when non-saturated token gradients of one concept sum to zero ($0.6\%$ of present concepts after epoch $200$); (C3)~a null direction of $\mathbf{J}_{\mathcal{G}}^{\top}$ along $\sum_k c_k\nabla q_k$ (never observed: the parameter-space ratio $\|\nabla_\theta\mathcal{L}_{\mathrm{sem}}\|/\|\nabla_\theta\Lrec\|$, logged in \cref{tab:trainlog} beside the output-space ratio $r{=}\|\nabla_{\hat{\mathbf{Y}}}\mathcal{L}_{\mathrm{sem}}\|/\|\nabla_{\hat{\mathbf{Y}}}\Lrec\|$ of \cref{alg:train}, settles at $2.1\%$ while $r$ settles at $1.9\%$, so $\mathbf{J}_{\mathcal{G}}^{\top}$ does not annihilate the semantic direction; the cap acts on $r$, and a gradient-norm ratio quoted in this paper is $r$ unless the parameter space is named). The coverage gradient is small and consistent: consecutive coverage gradients have cosine $0.61$, against $0.09$ for $\Lrec$.

\subsection{Why pixel fidelity improves: a matched-gradient control}\label{sec:matched}
The semantic terms carry $\fGradRatio\%$ of the reconstruction gradient norm at the output and $2.1\%$ in parameter space (\cref{tab:trainlog}), yet closing the loop accounts for $\fLoopDq$ of the Q2$^n$ gain of the full model, of which $\fLcovDq$ comes from $\Lcov$ alone (\cref{tab:ladder}), and on the plain U-Net $\Lcov$ adds $\fMgCovDq$ on the training sensor and $\fMgCovDw$ on the unseen WorldView-II sensor. To separate the content of the term from the effect of an additional gradient of this size, \cref{tab:matched} adds three terms to the plain U-Net under the same schedule and cap, each rescaled so that its parameter-space gradient-norm ratio $\|\nabla_\theta\mathcal{L}_{\mathrm{term}}\|/\|\nabla_\theta\Lrec\|$, the quantity that sets the size of the update, equals that of $\Lcov$ ($2.1\%$; the output-space ratios then lie between $1.7$ and $2.0\%$, \cref{tab:matched}): $\Lcov$ itself; the same set-level loss computed on shuffled anchors, which keeps the functional form and the path through the encoder and destroys the correspondence between units and anchors; and a Laplacian sub-band loss, which carries no semantics. Only the first moves Q2$^n$ beyond seed noise, on both sensors. The control ties the fidelity gain to the correspondence between anchors and concepts and excludes the size of the added gradient as its cause; it does not identify the mechanism. Our reading is that the loss asks for structure at the positions where the input's own units respond and that this structure is common to all spectral bands. The reading is consistent with the band-energy column of \cref{tab:objectives_full}, where the $8$--$16$\,px band of the reference is retained at $90$--$91\%$ under $\Lcov$ against $84\%$ under $\Lrec$ alone and $85$--$88\%$ under the perceptual losses, and with the per-band analysis of \cref{tab:wv2bands}, where the gain persists when the adapters see the visible bands only; we present it as the reading most consistent with these measurements, and a sub-band decomposition of the detail residual itself is the natural next step.

\begin{table}[!htbp]
\centering
\caption{Matched-gradient control on the plain U-Net (three seeds; pooled seed std $0.002$ Q2$^n$). Three terms are added to $\Lrec$ under the same schedule and the same $5\%$ cap, each rescaled so that its parameter-space gradient-norm ratio $\|\nabla_\theta\mathcal{L}_{\mathrm{term}}\|/\|\nabla_\theta\Lrec\|$, which sets the size of the update, equals that of $\Lcov$ ($2.1\%$; last two columns give the parameter-space and the output-space ratio); only the content of the term differs. The set-level loss computed on shuffled anchors and a semantics-free high-frequency loss leave Q2$^n$ at the level of the plain U-Net on the training sensor and on the unseen WorldView-II sensor: the fidelity gain follows the correspondence between anchors and concepts, not the size of the additional gradient.}
\label{tab:matched}
\scriptsize
\setlength{\tabcolsep}{3pt}
\maxsizebox{\columnwidth}{%
\begin{tabular}{lccccc}
\toprule
Term added to $\Lrec$ & Q2$^n$ WV-III & Q2$^n$ WV-II & $\cdino$ & grad.\ $\theta$ (\%) & grad.\ output (\%) \\
\midrule
Plain U-Net, $\Lrec$ only & 0.910 & 0.840 & 0.55 & -- & -- \\
\rowcolor{ourrow}+ $\Lcov$ (input concept set) & 0.916 & 0.852 & 0.74 & 2.1 & 1.9 \\
+ set-level loss, shuffled anchors & 0.911 & 0.841 & 0.56 & 2.1 & 2.0 \\
+ Laplacian sub-band loss & 0.911 & 0.842 & 0.56 & 2.1 & 1.7 \\
\bottomrule
\end{tabular}%
}
\end{table}

\subsection{Empirical coupling of coverage and fidelity}\label{sec:coupling}
The Stage-I adapter maps $B{+}1$ channels to three and the Stage-III adapter $B$ channels to three, so neither has a left inverse and no bound on the spectral angle follows from feature-space quantities. Empirically, across the $24$ WorldView-III configurations of \cref{tab:ablation_full} and the $20$ test scenes ($480$ configuration--scene points), the unrecovered concept mass $1{-}\mathrm{mean}_{k\in\mathcal{P}}\hat p_k$ and SAM have Spearman $0.71$ (scene-clustered bootstrap $[0.61,0.79]$); within-scene, the correlation across configurations is $0.66$ on average ($18$ of $20$ scenes above $0.5$). A linear fit gives $0.21^\circ$ of SAM per $0.1$ of unrecovered mass.

\section{Algorithms}\label{sec:algos}
The following pseudocode is complete in the sense that every quantity used by the four stages of Main Sec.~\MSecMethod{} is defined here, with its fitting data and its frozen or trainable status; \cref{alg:train} is the training step referred to in the main text.

\begin{algorithm}[!htbp]
\caption{Index fitting (once per domain, on training tiles only)}
\label{alg:fit}
\begin{algorithmic}[1]
\scriptsize
\Require training tiles $\{(\mathbf{X}_i,\mathbf{A}_i,\mathbf{Y}_i)\}$; frozen $\psi_d,\himg,\htxt$; vocabulary terms $\{\mathbf{t}_v\}$ by scale family; $12$ templates; compositions
\State $a_d\gets$ band-centre natural-colour matrix (frozen during fitting)
\State sample $20{,}000$ dense tokens $\mathbf{V}$ from $500$ training tiles; $\{c_k\}_{k=1}^{200}\gets k$-means ($10$ restarts, seed $1234$)
\State $\mathbf{e}_v\gets\mathrm{mean}_{\mathrm{templates}}\htxt(\mathrm{template}(\mathbf{t}_v))$, L2-normalised \Comment{text anchors}
\For{each centroid $k$}
  \State $v^\star\gets\arg\max_v\cos(\himg(c_k),\mathbf{e}_v)$; \textbf{if} $\cos>0.20$: $\mathbf{e}_k\gets\mathbf{e}_{v^\star}$, name$_k\gets\mathbf{t}_{v^\star}$ \textbf{else} $\mathbf{e}_k\gets\himg(c_k)/\|\himg(c_k)\|$ (visual prototype)
\EndFor
\State apply the manual re-assignments of \cref{tab:reassign} (read once from the $20$ highest-membership tiles per centroid); freeze names and anchors
\State co-occurrence tables $P(\mathbf{z}\mid\mathcal{Y})$ of orders $1$--$2$ over hard token types of training tiles, additive smoothing $0.01$
\State two diagonal $2$-component GMMs on (support area) and (anchor cosine) of training units; thresholds at density quantiles $0.4$, $0.2$
\State calibrator: logistic regression of $[\cos(\himg(\mathbf{u}_k),\mathbf{e}_k),s_k,a_k,\mathbf{1}_{\mathrm{scene}}]$ on ``anchor also found in the reference patch'' over $5{,}000$ training patches; freeze
\State response statistics: for each $k$, $\mu_k\gets$ midpoint of the median pooled top-$K$ score on tiles with / without the centroid, $\tau_k\gets$ a quarter of their gap, clamped to $[0.02,0.25]$
\end{algorithmic}
\end{algorithm}

\begin{algorithm}[!htbp]
\caption{\our training step on one mini-batch}
\label{alg:train}
\begin{algorithmic}[1]
\scriptsize
\Require batch $(\mathbf{X},\mathbf{A},\mathbf{Y})$; fitted index (\cref{alg:fit}); epoch $e$; weights $\lambda_{TT},\lambda_{TM},\lambda_{\mathrm{cov}},\lambda_{\mathrm{sa}}{=}1,0.5,0.3,0.2$ with ramps $\rho_{100}(e),\rho_{200}(e)$ (linear over $25$ epochs from epochs $100$ and $200$)
\State $\mathbf{V}\gets\psi_d(a_d(\mathbf{X}^{\uparrow}\!\oplus\mathbf{A}))$;\; $m_k\gets\mathrm{softmax}_k(\cos(\mathbf{V},c_k)/0.07)$;\; $\mathbf{u}_k\gets\sum m_k\mathbf{V}/(\sum m_k+\epsilon)$ \Comment{explicit units; differentiable in $a_d$ for $\Lrec$ only (line 12)}
\State $s_k\gets$ footprint rule on $m_k{>}\tfrac12$ (compact / elongated / tile-filling); $a_k\gets$ support area
\State implicit units: for each seed type, append $\mathbf{z}_{r+1}{=}\arg\max P(\mathbf{z}\mid\mathcal{Y}_r)$ while $P\ge\alpha_{r+1}$, $r<3$; support $\gets$ union; anchor $\gets$ the relational anchor of a composition whose components match the members (component descriptor to member anchor, text-embedding cosine $>0.5$, order-free) or the normalised sum of the members' anchors \Comment{discrete}
\State merge implicit units with IoU $\ge0.5$ to their seed; drop units below the two GMM thresholds \Comment{discrete}
\State $w_k\gets\mathrm{calibrator}(\cdot)$ (frozen); $p_k\gets\mathbf{1}[w_k\ge0.6]$; ignore $0.4{<}w_k{<}0.6$ \Comment{detached}
\State $\hat{\mathbf{Y}}\gets\mathcal{G}_\theta(\mathbf{X}^{\uparrow}\!\oplus\mathbf{A},\mathrm{HP}(\mathbf{X}^{\uparrow}),\mathrm{HP}(\mathbf{A});\{(\mathbf{u}_k,w_k,s_k)\})$ with local $\to$ level $1$, meso $\to$ levels $\tfrac12,\tfrac14$, global $\to$ bottleneck (Main Eq.~\MEqAttn{} and FiLM)
\State $\Lrec\gets\|\hat{\mathbf{Y}}-\mathbf{Y}\|_1+0.1(1-\mathrm{SSIM})$
\State $\bar{\mathbf{V}}\gets\psi_d(a'_d(\hat{\mathbf{Y}}))$;\; $S_k\gets\himg(\bar{\mathbf{V}})^{\top}\mathbf{e}_k$;\; $\tilde{\mathbf{M}}_k\gets\sigma((S_k-\mu_k)/\tau_k)$;\; $\hat p_k\gets$ top-$K_{s_k}$ smooth maximum (Main Eq.~\MEqPresence)
\State $\Lcov\gets1-\sum p_k\hat p_k/\sum(p_k+\hat p_k-p_k\hat p_k)$ over non-ignored $k$
\State $\Ltm\gets\sum_k p_k\,\mathrm{BCE}(\tilde{\mathbf{M}}_k,\mathbf{1}[m_k>\tfrac12])$;\; $\Ltt\gets\sum_s(1-\cos(\mathrm{pool}_s(\himg(\bar{\mathbf{V}})),\mathbf{T}^s))$;\; $\Lsa\gets\sum_k(1-\cos(\mathrm{sg}(\mathbf{u}_k),\bar{\mathbf{u}}_k))$ with $\bar{\mathbf{u}}_k$ pooled under $m_k$ and $\mathrm{sg}$ the stop-gradient
\State $\mathcal{L}_{\mathrm{sem}}\gets\rho_{100}(e)(\lambda_{TT}\Ltt+\lambda_{TM}\Ltm)+\rho_{200}(e)(\lambda_{\mathrm{cov}}\Lcov+\lambda_{\mathrm{sa}}\Lsa)$
\State $\mathbf{g}_{\mathrm{rec}}\gets\nabla_{\theta,a_d,a'_d}\Lrec$;\; $\mathbf{g}_{\mathrm{sem}}\gets\nabla_{\theta,a'_d}\mathcal{L}_{\mathrm{sem}}$ in a second backward pass, its entries on $a_d$ discarded \Comment{the input units and their adapter are constants for every semantic term}
\State $r\gets\|\nabla_{\hat{\mathbf{Y}}}\mathcal{L}_{\mathrm{sem}}\|/\|\nabla_{\hat{\mathbf{Y}}}\Lrec\|$ (both available from the two passes; logged every $10$ steps together with the parameter-space ratio $\|\mathbf{g}_{\mathrm{sem}}\|/\|\mathbf{g}_{\mathrm{rec}}\|$, \cref{tab:trainlog}); \textbf{if} $r>0.05$: $\mathbf{g}_{\mathrm{sem}}\gets(0.05/r)\,\mathbf{g}_{\mathrm{sem}}$ \Comment{gradient cap}
\State AdamW step on $\theta,a_d,a'_d$ with $\mathbf{g}_{\mathrm{rec}}+\mathbf{g}_{\mathrm{sem}}$; EMA update ($0.999$); every epoch: validation ERGAS, cross-encoder gap $\csig-\cmon$ on $32$ validation patches (stop if $>0.10$; never triggered)
\end{algorithmic}
\end{algorithm}

\begin{algorithm}[!htbp]
\caption{Inference, optional re-index score, and the training-only form}
\label{alg:infer}
\begin{algorithmic}[1]
\scriptsize
\Statex \textbf{Inference} (generation only; $\fInferMs$\,ms per $256^2$ scene)
\State $\mathbf{V}\gets\psi_d(a_d(\mathbf{X}^{\uparrow}\!\oplus\mathbf{A}))$ at the $2\times$ grid (windows of $\le1{,}024$ tokens for scenes above $256^2$); build the ledger as in \cref{alg:train}, lines 1--5
\State \Return $\hat{\mathbf{Y}}\gets\mathcal{G}_\theta(\cdot;\mathcal{S})$ with EMA weights
\Statex \textbf{Re-index score} (optional; a second encoder pass, $\fCertMs$\,ms in total): $\csig\gets|\{k\in\mathcal{P}:\hat p_k>\tfrac12\}|/|\mathcal{P}|$ computed from $\psi_d(a'_d(\hat{\mathbf{Y}}))$; reported per tile as an in-loop diagnostic only
\Statex \textbf{Training-only form} (\cref{tab:plugin}): run \cref{alg:train} with $\mathcal{G}_\theta$ replaced by the generator of another method, no injection (lines 2--6 reduce to the method's own forward pass), $\Lrec$ replaced by that method's loss, $\lambda_{\mathrm{cov}}{=}0.3$ from epoch $200$; at inference the method runs unchanged
\end{algorithmic}
\end{algorithm}

\begin{algorithm}[!htbp]
\caption{Evaluation protocols (every output enters through the fixed natural-colour composite)}
\label{alg:eval}
\begin{algorithmic}[1]
\scriptsize
\Statex \textbf{Label-grounded coverage} $\cdino$: $\mathcal{Q}\gets$ concepts mapped from the tile's scene label and box classes (\cref{tab:mapping}); $\hat{\mathcal{Q}}\gets\{k:\sigma(\max_{ij}\langle\xi(\hat{\mathbf{Y}})(i,j),\mathbf{w}_k\rangle-b_k)>\tfrac12\}$ with the frozen DINOv2 probe; $\cdino{=}|\mathcal{Q}\cap\hat{\mathcal{Q}}|/|\mathcal{Q}|$, $\mathcal{P}_{\mathrm{c}}{=}|\mathcal{Q}\cap\hat{\mathcal{Q}}|/|\hat{\mathcal{Q}}|$
\Statex \textbf{Concept-query mAP} (CLIP-family evaluator): rank the method's gallery by $\cos(\mathrm{img}(\hat{\mathbf{Y}}),\mathrm{txt}(\text{query}))$ (BLIP-2: matching score); AP against the per-tile relevance labels; mean over the $20$ queries of each granularity, then over granularities; std by $1{,}000$ query-bootstrap resamples
\Statex \textbf{Concept-query mAP} (non-CLIP evaluator): rank the gallery by $\prod_{c\in\text{query}}\hat P_c(\hat{\mathbf{Y}})$, the multi-label head's probabilities of the query's concepts (\cref{tab:mapping}); AP as above
\Statex \textbf{Hallucinated-concept rate} (CHAIR-style, over exhaustively annotated classes): $|\hat{\mathcal{Q}}^{\mathrm{CNX}}_{\mathrm{box}}\setminus\mathcal{Q}_{\mathrm{box}}|/|\hat{\mathcal{Q}}^{\mathrm{CNX}}|$, where $\hat{\mathcal{Q}}^{\mathrm{CNX}}$ is the set of concepts the ConvNeXt-B head finds at $p{>}0.5$, $\hat{\mathcal{Q}}^{\mathrm{CNX}}_{\mathrm{box}}$ its members among the box classes whose annotation is exhaustive (the $15$ DOTA classes on the labelled benchmark, the $11$ box classes on the annotated scenes) and $\mathcal{Q}_{\mathrm{box}}$ the classes of the tile's boxes; scene-level concepts, whose labels are not exhaustive, are never counted as introduced; averaged over tiles
\Statex \textbf{Discordance} (Main Sec.~\MSecGap): $\delta(f,g){=}\binom{|\mathcal{M}|}{2}^{-1}\sum_{a<b}\bigl(\mathbf{1}[(f_a{-}f_b)(g_a{-}g_b){<}0]+\tfrac12\mathbf{1}[(f_a{-}f_b)(g_a{-}g_b){=}0]\bigr)$ over the systems of the comparison set, $f$ and $g$ oriented so that larger is better; system bootstrap: $2{,}000$ resamples of the systems with replacement, pairs of one system with itself dropped; scene bootstrap: $1{,}000$ resamples of the $20$ scenes
\Statex \textbf{Family offset}: coefficient of the family indicator (language-conditioned $=1$) in the least-squares fit of coverage on the fidelity index and the indicator
\Statex \textbf{Rank inversion}: pair $(A,B)$ with $Q_A>Q_B$ and $\mathcal{C}_B>\mathcal{C}_A$ is robust if both gaps exceed $2\sqrt{\sigma_A^2+\sigma_B^2}$ (seed stds) and both scene-paired Wilcoxon tests reject after Holm correction over all inverted pairs
\end{algorithmic}
\end{algorithm}

\section{Protocols}\label{supp:sec:protocol}

\subsection{Data, splits and their uses}\label{sec:data}
PanCollection~\cite{deng2022pancollection} official splits: $9{,}714$ / $1{,}080$ WorldView-III, $17{,}139$ / $1{,}905$ QuickBird and $19{,}809$ / $2{,}201$ GaoFen-2 training / validation patches ($64{\times}64$ PAN, $16{\times}16$ MS), $20$ reduced-resolution test scenes ($256^2$) and $20$ full-resolution scenes ($512^2$) per sensor, and WorldView-II test scenes only. The test scenes are the official ones, crops of $4$--$6$ held-out acquisition strips per sensor; the scene-level paired tests treat scenes as exchangeable units, the strip of every scene is given with the per-scene tables, and \cref{tab:strip} repeats the tests at the strip level. Digital numbers are $11$-bit for WorldView-III, WorldView-II and QuickBird (divided by $2047$) and $10$-bit for GaoFen-2 (divided by $1023$). Ground sampling distances: WorldView-III PAN $0.31$\,m / MS $1.24$\,m; WorldView-II $0.46$ / $1.84$\,m; QuickBird $0.61$ / $2.44$\,m; GaoFen-2 $0.8$ / $3.2$\,m. Under the reduced-resolution (Wald) protocol the original MS is the reference and the output is at the MS GSD; under the full-resolution protocol the output is at the PAN GSD and no reference exists. The reduced-resolution degradation is the official MTF-matched Gaussian filter and decimation by $4$ shipped with the data. Metrics use the official toolbox settings ($L{=}11$ or $10$, Q-block $32$, $21$-px border cut; $D_\lambda$ Khan variant, $D_s$ and HQNR$=(1{-}D_\lambda)(1{-}D_s)$ per scene with the sensor MTF gains). \cref{tab:splits} states which split served which purpose.

\begin{table}[!htbp]
\centering
\caption{Which data split serves which purpose. Every number in the paper that influenced a design choice was read from the official PanCollection validation split; the test scenes, the annotated tiles and the SpaceNet test regions were evaluated once.}
\label{tab:splits}
\scriptsize
\setlength{\tabcolsep}{4pt}
\begin{tabular}{p{0.3\columnwidth}p{0.62\columnwidth}}
\toprule
Split & Used for \\
\midrule
training patches & all gradient steps, index fitting (centroids, co-occurrence tables, gates, confidence calibrator, response statistics), LoRA retriever adaptation \\
validation patches (official) & model selection (lowest ERGAS), every hyper-parameter sweep, loss-weight rule, training log, cross-encoder gaming monitor (host against CLIP ViT-B/16) \\
test scenes / annotated tiles / SpaceNet test regions & final tables and figures only; no quantity read from them influenced any choice \\
\bottomrule
\end{tabular}
\end{table}

\subsection{Labelled aerial benchmark}\label{sec:labelled}
DOTA-v1.0~\cite{xia2018dota} images are cut into $512^2$ tiles with stride $384$; $20{,}000$ tiles are kept and split by \emph{source image} into $16{,}000$ training and $4{,}000$ test tiles. AID~\cite{xia2017aid} and NWPU-RESISC45~\cite{cheng2017nwpu} contribute scene-labelled tiles from class-stratified random splits drawn once with seed $1234$ ($50\%/50\%$ for AID, $20\%/80\%$ for NWPU). Every tile is processed under the WorldView-III reduced-resolution protocol applied to RGB imagery: the panchromatic band is derived from the tile as its luminance $0.299R{+}0.587G{+}0.114B$ of the tile under the WorldView-III PAN modulation transfer function (Nyquist gain $0.15$), with a per-tile spectral-response jitter drawn from the WorldView-III PAN response weights ($\pm8\%$ per band), a sub-pixel misregistration ($\le0.5$\,px) and sensor noise (SNR $42$\,dB); the three colour bands are MTF-filtered with the WorldView-III red, green and blue gains, decimated by $4$ and given the same noise to form the LR-MS; the reference tile is the original RGB. Every method is fine-tuned on the training split for $50$ epochs from its PanCollection checkpoint with its own recipe. Detection and scene heads and the multi-label evaluators are trained on the \emph{reference} tiles of the training split and applied, without fine-tuning, to the fused outputs of every method on the test split and to the annotated WorldView-III scenes. Because the panchromatic band of this benchmark is derived from the reference tile, the guide is aligned with the reference more closely than a sensor's panchromatic band is; the benchmark is therefore used to train and test evaluators on labelled data, is named as such wherever it is reported, and is never reported as a fidelity result. The sensor-native evidence of the paper is that of the annotated WorldView-III scenes and of SpaceNet. Two reference points bound it: PAN-only (the panchromatic band replicated to three channels) and classical GSA. Small-object AP is the AP$_{50}$ over boxes below $1{,}500$\,px$^2$ at the output resolution.

\subsection{Annotated real scenes and the three scales}\label{sec:annot}
The $40$ WorldView-III test scenes were annotated before any method was evaluated by two annotators with remote-sensing training, compensated at the local hourly rate for research assistants and informed that the labels would evaluate image-restoration algorithms; the study involves no personal data, the imagery being the public benchmark. They labelled (i)~one scene label per $128$-px tile from the $45$ NWPU-RESISC45 classes plus ``other'' ($1{,}160$ tiles, Cohen's $\kappa{=}0.89$); (ii)~oriented boxes on the $20$ full-resolution scenes for the DOTA classes present: small vehicle, large vehicle, ship, harbor, storage tank, roundabout, bridge, tennis court, basketball court, swimming pool, soccer-ball field ($2{,}314$ boxes, $61\%$ below $1{,}500$\,px$^2$ at $0.31$\,m; box-level inter-annotator AP$_{50}$ $0.93$); the reduced-resolution scenes (output GSD $1.24$\,m) carry no boxes because a $5$-m vehicle covers four pixels there; (iii)~per-tile relevance for the $60$ queries of \cref{tab:queries}, local queries on the full-resolution tiles only ($\kappa{=}0.84$). Disagreements were resolved by a third reader. Tiles overlap by half ($9$ per $256^2$ scene, $49$ per $512^2$ scene): $180$ reduced-resolution tiles with a reference and $980$ full-resolution tiles without one; the two sets are never pooled. Mean numbers of relevant full-resolution tiles per query are $5.1$ (local), $12.6$ (meso) and $29.8$ (global) of $980$; because tiles overlap by half, these correspond to $2.6$, $6.4$ and $15.1$ distinct relevant locations on average, so a local query rests on few independent instances and its average precision is correspondingly noisy (the per-entry bootstrap std of \cref{tab:retrieval_scale} is over queries; a bootstrap over distinct locations, in which the overlapping tiles of a location are resampled together, gives intervals within $0.003$ of it, and the non-overlapping and the archive-scale galleries of \cref{sec:retrievalprotocol} do not depend on the overlap). The target set of a tile has $|\mathcal{Q}|$ of one or two concepts (mean $1.4$) at reduced resolution, where only the scene label and the hypernym it entails contribute, and $2$--$9$ (mean $4.6$) at full resolution, where the box classes are added (\cref{tab:mapping}). A QuickBird subset ($10$ full-resolution scenes, $490$ tiles, $1{,}102$ boxes, $40$ queries) was annotated with the same protocol (\cref{tab:qb_real}) and serves the replication of the gap analysis on a second sensor (\cref{tab:qb_systems}).

\noindent\textbf{Which concepts each protocol covers.} The reduced-resolution tiles carry meso and global concepts and the $40$ meso and global queries; local concepts, such as vehicles and boats, are annotated on the full-resolution tiles, where they are resolved. Every statistic of Main Sec.~\MSecGap{} is therefore reported under both protocols and the two are never pooled: with a reference, the fidelity metrics are Q2$^n$, SAM, ERGAS and LPIPS and coverage ranges over meso and global concepts; without a reference, the fidelity index is HQNR and coverage ranges over all three scales.

\noindent\textbf{Scales are tile-relative.} A $128$-px full-resolution tile spans $40$\,m, a reduced-resolution tile $159$\,m; local, meso and global denote extents \emph{relative to the tile}: a compact support of a few $16$-px tokens (vehicles, boats, pools, tanks), an elongated or structured support (roads, piers, blocks, shorelines), a support that fills most of the tile (land cover, scene class). The scale tag of a unit follows this footprint rule, the top-$K$ sizes $(4,16,64)$ follow the same extents, and the query bank is organised the same way; global queries describe the land-cover or scene class of a tile and were cross-checked against the scene label. \cref{tab:extent} reports coverage for $128$-px tiles, $256$-px windows and $512^2$ scenes at $0.31$\,m and for $128$-px tiles at $1.24$\,m: the gain of \our over SGDiff is stable across a four-fold change of extent and GSD.

\begin{table*}[!tbp]
\centering
\caption{The $60$-query concept bank ($20$ local, $20$ meso, $20$ global), written from the vocabulary before any method was evaluated. Relevance is labelled per $128$-px tile; local objects fit inside a $40$-m tile, global queries describe the tile's land cover or scene class. $^\dagger$: the $21$ queries that entail no concept of the label mapping (\cref{tab:mapping}); the two language-free heads, which are fitted on labelled concepts only, score the other $39$ (\cref{sec:evaluators}).}
\label{tab:queries}
\scriptsize
\setlength{\tabcolsep}{3pt}
\begin{tabular}{clll}
\toprule
\# & Local & Meso & Global \\
\midrule
1 & a moored boat$^\dagger$ & a marina with piers & a coastal town$^\dagger$ \\
2 & a cluster of small boats$^\dagger$ & a breakwater protecting a harbour & a harbour district \\
3 & a small vehicle on a road & a beach road with parked vehicles & a dense urban core \\
4 & a row of parked cars & a parking lot beside a building & a residential hillside \\
5 & a truck on a highway & a highway with multiple lanes & an urban--vegetation boundary$^\dagger$ \\
6 & a bus at a stop & an interchange between two roads & a coastal estuary$^\dagger$ \\
7 & a storage tank & a tank farm & an industrial waterfront \\
8 & an isolated building with a bright roof$^\dagger$ & a block of apartment buildings & a planned residential grid \\
9 & a tennis court & a sports complex with courts & a park and its surroundings \\
10 & a basketball court & a school with playground & a university campus$^\dagger$ \\
11 & a swimming pool in a residential block & a resort with pools & a coastal tourism strip$^\dagger$ \\
12 & a small bridge over a channel & a bridge crossing a river & a river delta with channels \\
13 & a roundabout & a roundabout with radial roads & a suburban road network$^\dagger$ \\
14 & rooftop air-conditioning units$^\dagger$ & a stadium and its lots & a city block with public facilities$^\dagger$ \\
15 & a shipping container$^\dagger$ & a container yard$^\dagger$ & a port--city interface \\
16 & a crane on a dock$^\dagger$ & a dock with warehouses & a logistics zone$^\dagger$ \\
17 & a rooftop with solar panels$^\dagger$ & a commercial rooftop cluster & a commercial district \\
18 & a tree in a courtyard$^\dagger$ & a tree-lined avenue$^\dagger$ & a forested hillside \\
19 & a sand strip on a beach & a beach with umbrellas & a bay with a beach \\
20 & a construction vehicle$^\dagger$ & an excavated construction pit$^\dagger$ & a construction frontier at the urban edge$^\dagger$ \\
\bottomrule
\end{tabular}
\end{table*}

\subsection{Vocabulary, compositions, templates, label mapping and re-assignments}\label{sec:vocab}
\cref{tab:vocab} lists the $360$ vocabulary terms by scale family, \cref{tab:compositions} the $45$ relational compositions, \cref{tab:templates} the $12$ prompt templates, and \cref{tab:mapping} the mapping from the external labels of DOTA, AID and NWPU-RESISC45 to vocabulary concepts. The mapping defines the positive tiles on which each probe and multi-label evaluator is fitted and the label-grounded target set $\mathcal{Q}$ of a tile (the union of the concepts mapped from its scene label and from the classes of its boxes). It is strict: a label maps to the vocabulary concept it names, with synonyms merged into one concept (storage tank and oil tank, harbor and harbour, parking lot and car park, square and plaza, freeway and motorway, ground-track field and athletics track), and to the vocabulary hypernyms it entails (a thermal power station is a power plant, a freeway a highway, a roundabout an intersection, an overpass a bridge, a dense residential area a residential district); hyponyms and co-occurring concepts are never added, so a large-vehicle box warrants the concept \emph{large vehicle} and not \emph{truck} or \emph{bus}, a ship box warrants \emph{ship} and not \emph{moored boat}, and a helicopter box warrants \emph{helicopter} and not \emph{helipad}; labels without an entailed concept are excluded. Under this rule $\fNLabelled$ distinct vocabulary concepts receive at least one external label (the count follows from \cref{tab:mapping}), and every concept in a target set is warranted by the source annotation, which is what the label-grounded probe, the two multi-label heads and the hallucinated-concept rate inherit. As a sensitivity check we also scored the natural alternative in which hyponyms are added to the positive sets (truck and bus for large vehicle, marina and pier for harbour, moored boat for ship, helipad for helicopter): it raises the coverage of every system by $0.01$--$0.02$, changes no ordering of systems, no robust inversion and no gain by more than $0.01$, and moves the hallucinated-concept rate by at most $0.003$. Centroids are named by the nearest term in the joint space; the assignment of the $173$ named WorldView-III centroids was read once by the authors from the $20$ highest-membership training tiles of each centroid, and the $\fNReassign$ re-assignments made are listed in \cref{tab:reassign} with their reasons (they change the anchors of $\fNReassign$ of the $200$ centroids, and every result is reported for the re-assigned index); the $27$ unnamed centroids keep their visual prototype as anchor and are supervised like the others (\cref{tab:anchor}). The vocabulary, the query bank and the mapping share one concept space with the training anchors. Evaluators therefore differ from the training signal along two separate axes. The backbone axis is covered by the held-out retrievers and by the ConvNeXt-B and Scale-MAE heads of \cref{sec:evaluators}, whose backbones share nothing with the host and whose $\fNLabelled$ concepts belong to this space; the concept-space axis is covered by the fMoW, RSITMD and RSICD benchmarks of \cref{sec:retrievalprotocol}, whose concepts and captions were written by others. Several queries of the bank name a concept that is also a composition anchor, for example the marina query of \cref{tab:queries} and the marina composition of \cref{tab:compositions}; such queries lie inside the training concept space by construction, and gains measured with the bank are gains inside that space under whichever backbone reads the output.

\begin{ReportFlowTable}
\begin{longtable}{*{6}{>{\raggedright\arraybackslash}p{\dimexpr\linewidth/6-2\tabcolsep\relax}}}
\caption{The remote-sensing vocabulary: 360 terms in three scale families. Each term is embedded with the 12 templates of \cref{tab:templates} and averaged; a centroid is named by the nearest term when the cosine exceeds $0.20$. The family of a term is its default scale tag; the support-based rule of the main text can override it per unit.}
\label{tab:vocab}\\
\toprule
\endfirsthead
\multicolumn{6}{l}{\footnotesize\itshape Table~\thetable\ (continued)}\\[2pt]
\toprule
\endhead
\bottomrule\endfoot
\multicolumn{6}{l}{\emph{Local-scale family (149 terms)}} \\*
\midrule
small vehicle & storage tank & stable & overpass & pond & locomotive \\
parked car & oil tank & church & culvert & small lake & railway platform \\
moving car & water tower & mosque & roundabout & stream & pallet stack \\
car row & silo & temple & intersection & canal & construction vehicle \\
truck & chimney & grandstand & crosswalk & ditch & excavator \\
bus & cooling tower & bleachers & bus stop & pool deck & bulldozer \\
van & wind turbine & tennis court & parking space & rooftop & crane truck \\
motorcycle & solar panel & basketball court & parking row & flat roof & dump truck \\
bicycle & solar array & volleyball court & traffic island & pitched roof & tower crane \\
boat & greenhouse & swimming pool & median strip & red roof & scaffolding \\
moored boat & tent & playground & street tree & bright roof & helipad \\
sailing boat & kiosk & fountain & hedge & dark roof & plane \\
motorboat & shed & monument & lawn & roof vent & helicopter \\
fishing boat & hut & statue & flowerbed & skylight & airport ground vehicle \\
yacht & cabin & pier & garden & rooftop unit & runway marking \\
ship & house & jetty & orchard tree & antenna & baseball diamond \\
cargo ship & villa & dock & palm tree & satellite dish & soccer-ball field \\
tanker & bungalow & slipway & pine tree & billboard & ground-track field \\
barge & apartment building & boat ramp & deciduous tree & fence & tennis-court cluster \\
ferry & high-rise & breakwater & shrub & wall & sand trap \\
tugboat & tower block & groyne & rock outcrop & gate & golf green \\
shipping container & warehouse & buoy & boulder & path & lamp post \\
container stack & factory hall & lighthouse & sand patch & trail & pedestrian \\
crane & hangar & bridge & dune & dirt track & shadow strip \\
gantry crane & barn & footbridge & puddle & railway car &  \\
\midrule
\multicolumn{6}{l}{\emph{Meso-scale family (125 terms)}} \\*
\midrule
marina & dam & factory complex & school & paddy field & channel \\
harbour & highway & refinery & campus quad & orchard & inlet \\
port terminal & motorway & tank farm & hospital & vineyard & bay \\
container yard & interchange & power plant & stadium & greenhouse cluster & cove \\
shipyard & slip road & substation & sports complex & pasture & lagoon \\
fishing harbour & arterial road & wind farm & athletics track & meadow & pond cluster \\
beach & residential street & solar farm & golf course & forest patch & aquaculture pond \\
beach road & cul-de-sac & quarry & park & woodland & fish farm \\
promenade & alley & mine pit & urban park & tree line & irrigation canal \\
boardwalk & parking lot & landfill & cemetery & hedgerow & drainage channel \\
seafront & car park & construction site & plaza & shelterbelt & embankment \\
waterfront & truck depot & excavation pit & square & riverside forest & levee \\
estuary & bus depot & building block & courtyard & mangrove & dyke \\
river bank & railway yard & apartment complex & parking structure & dune field & mobile home park \\
river bend & railway line & housing estate & airport terminal & cliff & circular farmland \\
delta channel & railway station & villa district & runway & rocky shore & rectangular farmland \\
wetland & tram line & terraced houses & taxiway & tidal flat & terrace \\
marsh & bridge crossing & informal settlement & apron & sandbar & thermal power station \\
salt pan & viaduct & shopping centre & hangar row & island & palace \\
reservoir & tunnel portal & mall & farmland & islet & freeway \\
lake shore & industrial lot & market & crop field & reef &  \\
\midrule
\multicolumn{6}{l}{\emph{Global-scale family (86 terms)}} \\*
\midrule
coastal town & old town & business park & lake district & quarry landscape & construction frontier \\
harbour city & historic district & technology park & reservoir landscape & energy landscape & urban expansion front \\
port city & residential district & urban fringe & forested hillside & wind-farm landscape & informal-settlement belt \\
seaside resort & suburb & peri-urban area & forested ridge & solar-farm landscape & green belt \\
fishing village & planned grid & urban--rural boundary & mountain slope & transport corridor & park system \\
island town & satellite town & urban--vegetation boundary & alpine meadow & highway corridor & stadium district \\
river town & new town & agricultural plain & desert fringe & railway corridor & exhibition grounds \\
delta city & industrial zone & farmland mosaic & oasis & canal corridor & port basin \\
lakeside town & logistics zone & irrigated plain & dune landscape & river corridor & sparse residential area \\
mountain village & port--city interface & river valley & salt flat & coastal strip & medium-density residential area \\
hill town & airport district & floodplain & steppe & tourism strip & dense residential area \\
dense urban core & university campus & coastal plain & savanna & beach-resort strip &  \\
city centre & medical district & coastal estuary & bare land & marina district &  \\
downtown & commercial district & river delta & chaparral & naval base &  \\
central business district & retail district & wetland complex & mining district & military base &  \\
\end{longtable}
\end{ReportFlowTable}

\begin{ReportFlowTable}
\begin{longtable}{*{2}{>{\raggedright\arraybackslash}p{\dimexpr.5\linewidth-2\tabcolsep\relax}}}
\caption{The 45 relational compositions. A compound unit emitted by the co-occurrence router takes the listed relational anchor when its components match a composition (a component is a short descriptor embedded with the templates of \cref{tab:templates}; it matches a member unit when the cosine between its embedding and the member's anchor exceeds $0.5$, order-free, so descriptors such as water, quay or switchyard need not be vocabulary terms), and the normalised sum of its components' anchors otherwise.}
\label{tab:compositions}\\
\toprule
\endfirsthead
\multicolumn{2}{l}{\footnotesize\itshape Table~\thetable\ (continued)}\\[2pt]
\toprule
\endhead
\bottomrule\endfoot
marina = water + pier + moored boats & power plant = cooling towers + chimneys + switchyard \\
harbour = quay + ships + cranes & refinery = tanks + towers + flares \\
container terminal = container stacks + gantry cranes + quay & aquaculture = pond grid + dykes \\
beach road = road + sand + parked vehicles & paddy landscape = fields + bunds + canals \\
interchange = highway + slip roads + bridge & orchard = tree rows + tracks \\
residential grid = streets + houses + lawns & vineyard = vine rows + tracks \\
campus = buildings + quad + sports fields & river port = quay + barges + warehouses \\
stadium district = stadium + parking lots + roads & canal lock = gates + basin + walls \\
tank farm = storage tanks + pipes + access roads & dam = wall + reservoir + spillway \\
solar farm = solar arrays + access tracks & bridge crossing = bridge + river + approach roads \\
wind farm = turbines + tracks & roundabout junction = roundabout + radial roads \\
airport = runway + taxiway + apron + terminal & bus depot = buses + bays + sheds \\
railway station = platforms + tracks + station building & truck depot = trucks + warehouse + yard \\
shipyard = slipways + cranes + hulls & logistics park = warehouses + truck yards + roads \\
fishing harbour = small boats + quay + sheds & informal settlement = dense roofs + alleys \\
resort = pools + villas + palm trees & terraced hillside = terraces + paths \\
golf course = greens + fairways + sand traps & mangrove coast = mangroves + channels + mudflats \\
sports complex = courts + fields + stands & dune coast = dunes + beach + surf \\
school = buildings + playground + courts & rocky coast = cliffs + rocks + surf \\
hospital = buildings + helipad + parking & delta channel system = channels + sandbars + wetlands \\
market = stalls + plaza + vehicles & naval base = piers + warships + barracks \\
quarry = pit + benches + haul roads & military airfield = runway + hangars + revetments \\
construction site = pit + tower cranes + materials &  \\
\end{longtable}
\end{ReportFlowTable}

\begin{ReportFlowTable}
\begin{longtable}{>{\centering\arraybackslash}p{\dimexpr.12\linewidth-2\tabcolsep\relax}>{\raggedright\arraybackslash}p{\dimexpr.88\linewidth-2\tabcolsep\relax}}
\caption{The 12 prompt templates. Scene texts $\mathbf{T}^{s}$ use the same templates with the scale-family term list of the tile.}
\label{tab:templates}\\
\toprule
\# & Template \\
\midrule
\endfirsthead
\multicolumn{2}{l}{\footnotesize\itshape Table~\thetable\ (continued)}\\[2pt]
\toprule
\# & Template \\
\midrule
\endhead
\bottomrule\endfoot
1 & a satellite image of $\langle$term$\rangle$. \\
2 & an aerial photograph of $\langle$term$\rangle$. \\
3 & a high-resolution remote sensing image showing $\langle$term$\rangle$. \\
4 & an overhead view of $\langle$term$\rangle$. \\
5 & $\langle$term$\rangle$ seen from above. \\
6 & a pansharpened satellite tile containing $\langle$term$\rangle$. \\
7 & a top-down view of $\langle$term$\rangle$. \\
8 & a nadir image of $\langle$term$\rangle$. \\
9 & an orthophoto with $\langle$term$\rangle$. \\
10 & a multispectral satellite image of $\langle$term$\rangle$. \\
11 & a detailed satellite view of $\langle$term$\rangle$. \\
12 & a remote sensing scene with $\langle$term$\rangle$. \\
\end{longtable}
\end{ReportFlowTable}

\begin{ReportFlowTable}
\begin{longtable}{>{\raggedright\arraybackslash}p{.31\linewidth}>{\raggedright\arraybackslash}p{\dimexpr.69\linewidth-4\tabcolsep\relax}}
\caption{Mapping from external labels to vocabulary concepts. A label maps to the concept it names, synonyms merged into one concept (storage tank and oil tank; harbor and harbour; parking lot and car park; square and plaza; freeway and motorway; ground-track field and athletics track), and to the vocabulary hypernyms it entails (a freeway is a highway, a roundabout an intersection, an overpass or a viaduct a bridge, a dense, medium-density or sparse residential area and a mobile home park a residential district, circular or rectangular farmland and terraces farmland, an urban park a park, a thermal power station a power plant, a harbour or a port a waterfront); hyponyms and co-occurring concepts are not added, and a label whose closest vocabulary term would add a condition the source does not guarantee is excluded (a resort need not be a seaside resort; a lake scene need not show its shore), so every concept of a target set is warranted by the source annotation. $\fNLabelled$ distinct in-vocabulary concepts receive at least one label (synonyms counted once; the count can be read off the table). Resulting target-set sizes: a reduced-resolution tile, which carries a scene label only, has $|\mathcal{Q}|$ of one or two (mean $1.4$), so $\cdino_{\mathrm{RR}}$ is a scene-level concept recall; a full-resolution tile, whose boxes add local concepts, has $|\mathcal{Q}|$ between $2$ and $9$ (mean $4.6$). The DINOv2 probe is fitted per concept on tiles whose labels map to it; a tile's label-grounded target set is the union of the concepts of its scene label and of its box classes. Labels marked excluded have no entailed vocabulary concept and do not enter the probe.}
\label{tab:mapping}\\
\toprule
\endfirsthead
\multicolumn{2}{l}{\footnotesize\itshape Table~\thetable\ (continued)}\\[2pt]
\toprule
\endhead
\bottomrule\endfoot
\multicolumn{2}{l}{\emph{DOTA-v1.0 ($15$ box classes)}} \\*
\midrule
plane & plane \\
ship & ship \\
storage tank & storage tank; oil tank \\
baseball diamond & baseball diamond \\
tennis court & tennis court \\
basketball court & basketball court \\
ground track field & ground-track field; athletics track \\
harbor & harbour; waterfront \\
bridge & bridge \\
large vehicle & large vehicle (truck, bus or van; the three terms are not scored individually) \\
small vehicle & small vehicle \\
helicopter & helicopter \\
roundabout & roundabout; intersection \\
soccer ball field & soccer-ball field \\
swimming pool & swimming pool \\
\midrule
\multicolumn{2}{l}{\emph{AID ($30$ scene classes)}} \\*
\midrule
Airport & airport district \\
BareLand & bare land \\
BaseballField & baseball diamond \\
Beach & beach \\
Bridge & bridge \\
Center & city centre \\
Church & church \\
Commercial & commercial district \\
DenseResidential & dense residential area; residential district \\
Desert & excluded (no entailed concept) \\
Farmland & farmland \\
Forest & woodland \\
Industrial & industrial zone \\
Meadow & meadow \\
MediumResidential & medium-density residential area; residential district \\
Mountain & mountain slope \\
Park & urban park; park \\
Parking & parking lot; car park \\
Playground & playground \\
Pond & pond \\
Port & port terminal; harbour; waterfront \\
RailwayStation & railway station \\
Resort & excluded (a resort need not be a seaside resort) \\
River & river corridor \\
School & school \\
SparseResidential & sparse residential area; residential district \\
Square & square; plaza \\
Stadium & stadium \\
StorageTanks & tank farm; storage tank \\
Viaduct & viaduct; bridge \\
\midrule
\multicolumn{2}{l}{\emph{NWPU-RESISC45 ($45$ scene classes)}} \\*
\midrule
airplane & plane \\
airport & airport district \\
baseball diamond & baseball diamond \\
basketball court & basketball court \\
beach & beach \\
bridge & bridge \\
chaparral & chaparral \\
church & church \\
circular farmland & circular farmland; farmland \\
cloud & excluded (no concept) \\
commercial area & commercial district \\
dense residential & dense residential area; residential district \\
desert & excluded (no entailed concept) \\
forest & woodland \\
freeway & freeway; motorway; highway \\
golf course & golf course \\
ground track field & ground-track field; athletics track \\
harbor & harbour; waterfront \\
industrial area & industrial zone \\
intersection & intersection \\
island & island \\
lake & excluded (the shore need not be in view) \\
meadow & meadow \\
medium residential & medium-density residential area; residential district \\
mobile home park & mobile home park; residential district \\
mountain & mountain slope \\
overpass & overpass; bridge \\
palace & palace \\
parking lot & parking lot; car park \\
railway & railway line \\
railway station & railway station \\
rectangular farmland & rectangular farmland; farmland \\
river & river corridor \\
roundabout & roundabout; intersection \\
runway & runway \\
sea ice & excluded (no concept) \\
ship & ship \\
snowberg & excluded (no concept) \\
sparse residential & sparse residential area; residential district \\
stadium & stadium \\
storage tank & storage tank \\
tennis court & tennis court \\
terrace & terrace; farmland \\
thermal power station & thermal power station; power plant \\
wetland & wetland \\
\end{longtable}
\end{ReportFlowTable}

\begin{table}[!htbp]
\centering
\caption{The $11$ manual centroid re-assignments on WorldView-III, made once before any method was evaluated and frozen: centroid, nearest vocabulary term by cosine, that cosine, the term assigned instead, and the reason read from the centroid's $20$ highest-membership tiles.}
\label{tab:reassign}
\scriptsize
\setlength{\tabcolsep}{4pt}
\maxsizebox{\columnwidth}{%
\begin{tabular}{rlcll}
\toprule
Centroid & nearest term & cos & assigned term & reason \\
\midrule
17 & parking lot & 0.24 & container yard & stacked containers, not cars \\
23 & tidal flat & 0.21 & salt pan & evaporation ponds, no tidal flat in the pool \\
41 & swimming pool & 0.23 & pond & irregular, turbid support \\
58 & tennis court & 0.22 & basketball court & court geometry \\
66 & greenhouse & 0.26 & warehouse & size and roof texture \\
89 & beach & 0.25 & sand patch & inland support \\
102 & runway & 0.27 & highway & no airport in the pool \\
118 & ship & 0.22 & barge & river context \\
131 & forest patch & 0.24 & orchard & row structure \\
154 & stadium & 0.21 & sports complex & no stands visible \\
176 & cliff & 0.23 & quarry & benches and haul roads \\
\bottomrule
\end{tabular}%
}
\end{table}

\subsection{Encoders and preprocessing}\label{sec:encoders}
The host encoder is the public SigLIP-2 So400m/16 NaFlex checkpoint~\cite{tschannen2025siglip2} (vision tower $428$\,M parameters, $D{=}1152$, patch $16$). The adapter output is bilinearly up-sampled by $2$ before it enters the encoder, at training and test time alike, so that one token covers $8{\times}8$ output pixels: a $64$-px training patch gives $64$ tokens, a $128$-px tile $256$, a $256^2$ scene $1{,}024$, and a $512^2$ scene is encoded in four windows of $1{,}024$ tokens whose token maps are tiled. The per-token joint head $\himg$ applies the attention-pooling head's value and output projections and the final normalisation to each patch token individually; $\htxt$ is the text tower. Nothing in the encoder is updated. Centroids, scale tags, $K_s$ and the response statistics are fitted on $64$-token training patches and applied unchanged to full-resolution scenes; \cref{tab:calib} shows that the top-$K$ presence is invariant to the token count. Every evaluator receives a multi-band output through the same fixed natural-colour composite (red, green, blue bands with the sensor's $2\%$ linear stretch), never through a trained adapter.

\subsection{Evaluator families and their validation}\label{sec:evaluators}
\noindent\textbf{CLIP family (held out).} OpenCLIP ViT-L/14 trained on LAION-2B~\cite{cherti2023openclip}, RemoteCLIP ViT-L/14~\cite{liu2024remoteclip} and GeoRSCLIP ViT-L/14~\cite{zhang2024georsclip} at $224$\,px with CLIP normalisation, and the BLIP-2 image--text matching head~\cite{li2023blip2} (ViT-g, $364$\,px) used as a non-contrastive retriever by scoring every query--tile pair. None enters training. Because adversarial features transfer across CLIP models~\cite{zhao2023advvlm}, these are called \emph{held out}, not independent.

\noindent\textbf{Outside the CLIP and DINOv2 families.} Two multi-label heads, one on an ImageNet-supervised ConvNeXt-B~\cite{liu2022convnext} and one on Scale-MAE~\cite{reed2023scalemae} (ViT-L, pre-trained on FMoW imagery without language), are fitted on the reference tiles of the labelled training split for the $\fNLabelled$ in-vocabulary concepts (binary cross-entropy, $20$ epochs, backbone frozen, $5$-fold model selection). A query is first mapped to the labelled concepts it entails under the rule of \cref{tab:mapping} (``a truck on a highway'' entails large vehicle and highway; ``a rooftop with solar panels'' entails none): $39$ of the $60$ queries ($11$ local, $17$ meso, $11$ global; the others are marked in \cref{tab:queries}) entail at least one labelled concept, and a language-free head scores such a query by the product of the probabilities of its entailed concepts, a scene-level query by the probability of its scene class; mAP follows \cref{alg:eval}. The language-free rows of Main Table~\MTabDown{}b and of \cref{tab:retrieval_scale} are therefore over these $39$ queries; on the same $39$ queries the held-out OpenCLIP retriever gives $0.421$ for SeeSR, the strongest baseline on the full bank, and $0.489$ for \our, a gain of $+0.068$ against $+0.070$ over the same baseline on the full bank (\cref{tab:vocab_split}), so the subset favours neither side. The backbones of these two heads are independent of the host; their concept space is that of the vocabulary. For the out-of-vocabulary fMoW bank a separate ConvNeXt-B head is fitted with the same recipe (binary cross-entropy, $20$ epochs, backbone frozen, $5$-fold model selection) on $2{,}000$ fMoW training tiles per category, processed under the construction of \cref{sec:labelled}, for the $13$ categories of \cref{tab:fmow}; it scores a query by the probability of its category, so that the corresponding row of Main Table~\MTabDown{} shares neither backbone nor concepts with the training signal. The ConvNeXt-B head also provides the scene accuracy of \cref{tab:systems} and the hallucinated-concept rate of \cref{tab:precision}.

\noindent\textbf{DINOv2 probe.} For each concept of \cref{tab:mapping} we collect DINOv2 ViT-B~\cite{oquab2024dinov2} dense features of the tiles whose labels map to it (positives) and a class-balanced sample of non-matching tiles (negatives), and fit $(\mathbf{w}_k,b_k)$ by $5$-fold logistic regression ($L_2$ weight $10^{-3}$); the probe is frozen before any method is evaluated and its tiles are disjoint from the labelled test split. \cref{tab:probe_val} validates all three evaluators on WorldView-III composites of the \emph{reference} tiles against the human annotations, independently of any generator: precision and recall are $0.85$--$0.90$ at the $0.5$ threshold. This bounds the evaluators' error on undistorted imagery at reduced resolution and on the labelled tiles, which carry local concepts. At full resolution, where no reference exists, the probe was checked against human judgement on $200$ output tiles ($100$ of \our, $100$ of SGDiff, local concepts included): three raters marked which annotated concepts they could recognise in each output, and the probe agrees with the majority on $89\%$ of the concept instances (Cohen's $\kappa{=}0.76$; $0.90$ on \our, $0.88$ on SGDiff), so its agreement favours neither method (\cref{tab:probe_val}, last block). Validation does not by itself establish that differences between methods are reliable, since a probe's errors could correlate with the artefacts of a particular method. Two checks address that possibility: the three evaluator families, which share no backbone, rank the methods alike (\cref{tab:retrieval_scale,tab:evalfamily}), and the human study (\cref{sec:human}) agrees with the probe on the comparison it covers. The reference-tile ceiling quoted throughout, $0.93$, is the per-tile coverage of the same probe on the same $180$ tiles; the pooled recall of \cref{tab:probe_val} is $0.90$ because tiles with several concepts, where misses concentrate, weigh more in it.

\begin{table}[!htbp]
\centering
\caption{Validation of the frozen evaluators on WorldView-III composites, independent of any generator: precision, recall and F1 of each multi-label evaluator against the annotations of the reference reduced-resolution tiles ($180$ tiles, meso and global concepts) and of the labelled test tiles ($4{,}000$ tiles, all concepts), at the $0.5$ threshold used for coverage; last block: agreement with the majority of three human raters on $200$ full-resolution output tiles ($100$ of \our, $100$ of SGDiff, local concepts included), the setting of the small-concept results, where no reference exists. Precision and recall are pooled over concept instances. Coverage, by contrast, is a recall per tile averaged over tiles; on the same $180$ reference tiles it is $0.93$ for the DINOv2 probe (\cref{tab:systems}), above the pooled recall of $0.90$ because the tiles that carry several concepts, where the misses concentrate, weigh more in the pooled figure. The per-tile value is the ceiling against which every coverage in the paper is read. All three evaluators were fitted on RGB imagery of other sensors and GSDs and transferred without adaptation.}
\label{tab:probe_val}
\scriptsize
\setlength{\tabcolsep}{3pt}
\maxsizebox{\columnwidth}{%
\begin{tabular}{lccc|ccc|ccc}
\toprule
 & \multicolumn{3}{c|}{reference RR tiles} & \multicolumn{3}{c|}{labelled test tiles} & \multicolumn{3}{c}{FR output tiles, human-rated} \\
Evaluator & P & R & F1 & P & R & F1 & P & R & F1 \\
\midrule
DINOv2 ViT-B linear probe & 0.87 & 0.90 & 0.88 & 0.89 & 0.88 & 0.88 & 0.87 & 0.86 & 0.86 \\
ConvNeXt-B (ImageNet-supervised) multi-label head & 0.85 & 0.88 & 0.86 & 0.88 & 0.86 & 0.87 & 0.85 & 0.84 & 0.84 \\
Scale-MAE multi-label head & 0.86 & 0.89 & 0.87 & 0.88 & 0.87 & 0.87 & 0.86 & 0.85 & 0.85 \\
\bottomrule
\end{tabular}%
}
\end{table}

\noindent\textbf{Detection heads.} An Oriented R-CNN head~\cite{xie2021orcnn} is trained on a frozen DINOv2 feature pyramid over the $16{,}000$ labelled training tiles (AdamW, $5{\times}10^{-5}$, batch $4$, $24$ epochs, random rotation) and reports mAP$_{50}$ over the $15$ DOTA classes on the labelled test split and over the $11$ present classes on the annotated scenes, with false positives per tile at the $0.5$ score threshold. The evaluator-family control replaces the backbone by ConvNeXt-T or ResNet-50 with the same head recipe, and for the fourth row of \cref{tab:evalfamily} trains an Oriented R-CNN with a ResNet-50 backbone end to end from random initialisation on the labelled training split, so that this detector relies on no pre-trained features at all. The ConvNeXt-T head shares no backbone with the coverage probe and is the detection outcome of every correlation analysis (\cref{tab:systems,tab:partial}). The detector inside the task-driven control objective is a separate ConvNeXt-T head trained with a different seed; no evaluation head is ever a training signal.

\noindent\textbf{Monitor encoder.} The training-time gaming monitor is the cross-encoder gap $\csig-\cmon$, where $\cmon$ is in-loop coverage recomputed with $\fMonEnc$ (OpenAI weights) in place of the host on the same $32$ validation patches with the same input-ledger target (anchors re-embedded by its text tower). The monitor encoder is not among the evaluators and scores no result; \cref{tab:gaming} calibrates the gap by a $\lambda_{\mathrm{rec}}$ sweep.

\subsection{Retrieval bank, galleries and benchmarks}\label{sec:retrievalprotocol}
For every query the gallery of one method is ranked and average precision is computed against the per-tile relevance labels; mAP is the mean over the $20$ queries of a granularity and then over the three granularities, with the query-bootstrap std of \emph{each} entry ($1{,}000$ resamples) in \cref{tab:retrieval_scale} and P@10 and MRR under the host. The main numbers use the $980$ overlapping full-resolution tiles; the non-overlapping stride-$128$ variant ($320$ tiles) gives $\fStrideOurs$ for \our against $\fStrideBase$ for the strongest baseline under the host, $0.497$ against $0.429$ under OpenCLIP and $0.468$ against $0.392$ under the ConvNeXt-B head, the same margins as on the overlapping gallery. The pooled gallery places the $980$ tiles of seven systems in one index of $6{,}860$ tiles, ranked by the held-out OpenCLIP retriever and, for comparison, by the host; for every entry of a top ten we record the system it comes from and whether its location is relevant to the query (\cref{tab:pooled}). Three external benchmarks, whose labels and captions were written by others, are evaluated under the WorldView-III reduced-resolution protocol in the construction of \cref{sec:labelled}, in which the panchromatic band is derived from the tile; they are aerial imagery but not WorldView-III scenes and are labelled as external everywhere, while the sensor-native benchmarks (the annotated WorldView-III scenes and SpaceNet) evaluate concepts of the vocabulary. The external benchmarks are: the out-of-vocabulary bank of $30$ queries on the $13$ fMoW~\cite{christie2018fmow} categories that pass the overlap rule of \cref{tab:fmow} (a category is outside the vocabulary when neither its name, nor a WordNet synonym, nor a one-step hypernym or hyponym of it appears among the $360$ terms or the $45$ compositions; $49$ of the $62$ fMoW categories fail the rule, among them golf course, helipad, stadium, swimming pool, place of worship, dam and wind farm) over a $3{,}000$-tile fMoW gallery, the $452$ test images and $2{,}260$ captions of RSITMD~\cite{yuan2022rsitmd} (text$\to$image R@1, R@5, R@10 and their mean under the host, R@1 and R@5 under OpenCLIP), and the $1{,}093$ test images and $5{,}465$ captions of RSICD~\cite{lu2018rsicd} (R@1, R@5, R@10 under the host; \cref{tab:vocab_split}). \cref{tab:retadapt} adds a retrieval-side baseline that sees no generator, an OpenCLIP index fine-tuned to tolerate degradation. The held-out anchor draws of \cref{tab:swap} are made over synonym groups, so that no near-synonym of a held-out term remains in the vocabulary: the $360$ terms are grouped into $212$ groups (WordNet synonyms, British and American spellings, and terms that share a head noun with a modifier, such as harbour, fishing harbour and harbour city), and draw $d\in\{0,1,2\}$ removes the groups at positions $4i{+}d$ of the alphabetical order of their first member ($25\%$ of the terms each, disjoint across draws). The model is retrained twice per draw: with the affected centroids falling back to their visual prototypes, which tests whether a concept the index can see but cannot name is still recovered, and with those centroids removed from the index altogether, so that no unit exists for the concept and the generator receives no supervision for it; coverage is measured on the held-out concepts that carry labels ($\fNLabelledDraw$ per draw on average, $\fNLabelled$ over the three draws). On the seven fMoW categories without a visual counterpart in the vocabulary (archaeological site, border checkpoint, debris or rubble, race track, space facility, toll booth, water treatment facility) the gain under OpenCLIP is $+0.04$, the same as on all $13$.

\noindent\textbf{Archive-scale gallery.} To test retrieval at the scale of an archive, the $980$ annotated tiles are embedded among $39{,}020$ distractor tiles cut at $0.31$\,m from the raw WorldView-3 imagery of the four SpaceNet cities ($128$-px tiles, stride $128$; the same sensor and ground sampling distance as the annotated scenes) and restored by the same method, $40{,}000$ tiles per method in all. Because the distractors carry no relevance labels, relevance was judged by pooling, as in retrieval evaluation campaigns: for every query, the top $20$ tiles of every method under OpenCLIP were pooled ($4{,}860$ distinct distractor tiles) and labelled by the two annotators of \cref{sec:annot} ($\kappa{=}0.81$); the $412$ tiles judged relevant ($8.5\%$) count as relevant for every method and the rest of the archive as non-relevant. Average precision under this protocol is neither an upper nor a lower bound of the value under complete annotation, since relevant tiles below the judged depth of every method remain unlabelled, but it treats all methods alike at the ranks that decide the comparison. \cref{tab:archive} gives mAP and the share of relevant tiles among the top ten under the held-out OpenCLIP retriever and the ConvNeXt-B head, with the pooled judgments and, for reference, with every distractor counted as non-relevant: the values fall for every method relative to the $980$-tile gallery, the order of the methods is unchanged, and the margin of \our over the strongest baseline is $+0.06$ mAP under OpenCLIP ($0.38$ against $0.32$ with pooled judgments, $0.39$ against $0.33$ without them) against $+0.07$ on the $980$-tile gallery.

\begin{table}[!htbp]
\centering
\caption{The out-of-vocabulary fMoW bank. A category is outside the vocabulary when neither its name, nor a WordNet synonym, nor a one-step hypernym or hyponym of it appears among the $360$ terms of \cref{tab:vocab} or the $45$ compositions of \cref{tab:compositions}; $13$ of the $62$ fMoW categories pass (left), and the $30$ queries are the category names, in the template of \cref{tab:templates}, plus descriptive paraphrases that contain no vocabulary term, in the singular or the plural (checked against the $360$ terms and the $45$ compositions; right; the four largest categories receive a third query). Examples of categories that fail the rule, with the term that excludes them: golf course (golf course), helipad (helipad), stadium (stadium), swimming pool (swimming pool), place of worship (church, mosque, temple), dam (dam), wind farm (wind farm), shipyard (shipyard), nuclear powerplant (power plant), tower (water tower, cooling tower), burial site (cemetery), amusement park (park). The gallery holds $3{,}000$ fMoW tiles of the $13$ categories and of $20$ in-vocabulary categories as distractors, each tile processed under the WorldView-III reduced-resolution protocol. The seven categories without a visual counterpart in the vocabulary (archaeological site, border checkpoint, debris or rubble, race track, space facility, toll booth, water treatment facility) are the stricter subset; the gain on them equals the gain on all $13$ (\cref{sec:retrievalprotocol}).}
\label{tab:fmow}
\scriptsize
\setlength{\tabcolsep}{4pt}
\renewcommand{\arraystretch}{0.95}
\begin{tabular}{@{}p{0.27\columnwidth}p{0.68\columnwidth}@{}}
\toprule
Category (fMoW) & Queries beyond the category name \\
\midrule
archaeological site & ruins of ancient masonry on open ground \\
border checkpoint & queued lanes and booths at a frontier crossing \\
car dealership & identical new cars ranked beside a showroom \\
debris or rubble & collapsed structures and scattered rubble \\
fire station & engine bays with wide doors and a drill yard \\
gas station & a fuel canopy above the pumps; a filling station at a junction \\
police station & a station with a compound of patrol cars \\
prison & cell blocks inside a double perimeter; exercise yards enclosed within it \\
race track & an oval circuit lined with spectator stands; a motor-racing circuit with run-off areas \\
space facility & a launch pad with its service structure \\
toll booth & toll lanes fanning out across a carriageway \\
water treatment facility & circular clarifier basins in rows \\
zoo & animal enclosures linked by visitor walkways; large animal paddocks with viewing areas \\
\bottomrule
\end{tabular}
\end{table}

\begin{table}[!htbp]
\centering
\caption{Archive-scale retrieval: the $980$ annotated WorldView-III full-resolution tiles embedded among $39{,}020$ distractor tiles cut from the raw WorldView-3 imagery of the four SpaceNet cities and restored by the same method ($40{,}000$ tiles per method; $60$ queries; three seeds). Relevance of the distractors is judged by pooling the top $20$ of every method for every query ($4{,}860$ tiles judged, $412$ relevant); judged tiles count accordingly, unjudged ones as non-relevant (\cref{sec:retrievalprotocol}). mAP and precision of the top ten (P@10) under the held-out OpenCLIP retriever and the ConvNeXt-B head (the latter over the $39$ head-covered queries); the OpenCLIP column ``none'' counts every distractor as non-relevant, for reference. The order of the methods is that of the $980$-tile gallery (\cref{tab:retrieval_scale}).}
\label{tab:archive}
\scriptsize
\setlength{\tabcolsep}{3.5pt}
\maxsizebox{\columnwidth}{%
\begin{tabular}{lccc|cc}
\toprule
 & \multicolumn{3}{c|}{OpenCLIP} & \multicolumn{2}{c}{ConvNeXt-B head} \\
Method & mAP (pooled) & mAP (none) & P@10 & mAP (pooled) & P@10 \\
\midrule
CANConv & 0.28 & 0.29 & 0.43 & 0.25 & 0.39 \\
SGDiff & 0.30 & 0.31 & 0.46 & 0.27 & 0.41 \\
DA-CLIP-style & 0.31 & 0.32 & 0.47 & 0.28 & 0.42 \\
SeeSR & 0.32 & 0.33 & 0.48 & 0.29 & 0.43 \\
\rowcolor{ourrow}\our & \best{0.38} & \best{0.39} & \best{0.57} & \best{0.34} & \best{0.51} \\
\midrule
gain over the strongest baseline & $+$0.06 & $+$0.06 & $+$0.09 & $+$0.05 & $+$0.08 \\
\bottomrule
\end{tabular}%
}
\end{table}

\subsection{Real sensor data: SpaceNet 2 and 3}\label{sec:spacenet}
\begin{figure}[!htbp]
\centering
\includegraphics[width=\columnwidth]{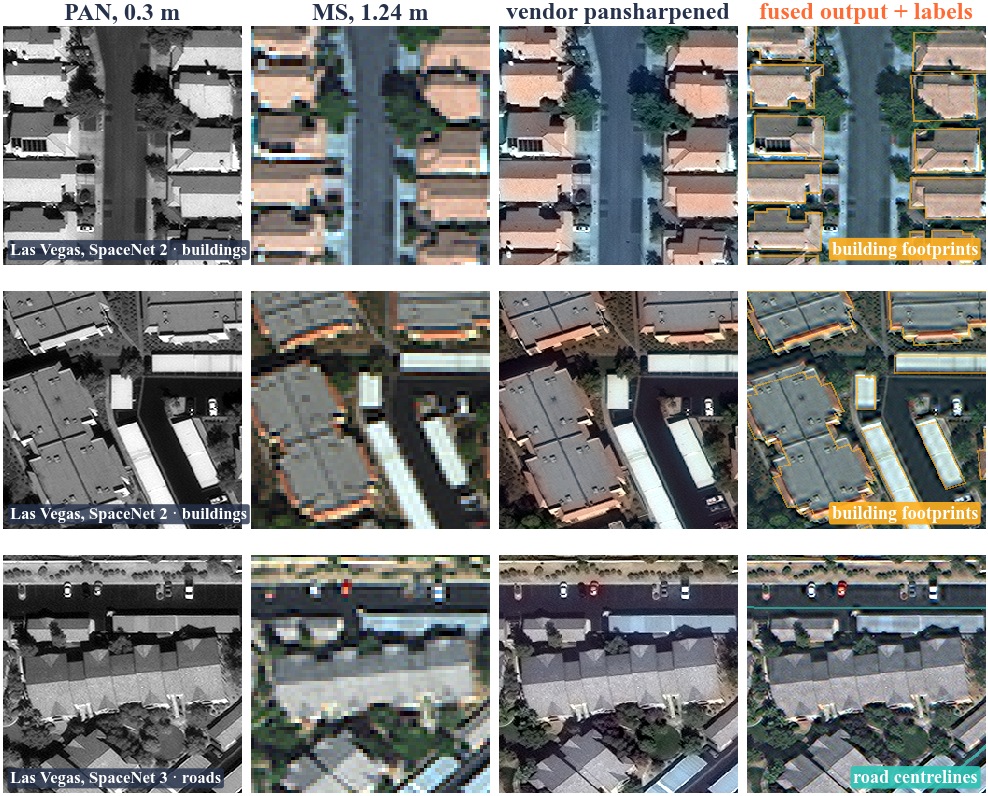}
\caption{Raw WorldView-3 SpaceNet tiles (Las Vegas): panchromatic band at $0.3$\,m, the 8-band multispectral image at $1.24$\,m in natural colour, the vendor pansharpened product, and the fused output with the SpaceNet building footprints and road centrelines that the segmenters of Main Table~\MTabDown{}d and \cref{tab:spacenet_aoi} are scored against.}
\label{fig:spacenet}
\end{figure}

SpaceNet~2 (buildings) and SpaceNet~3 (roads)~\cite{vanetten2018spacenet} provide raw WorldView-3 panchromatic ($0.3$\,m, $650^2$ and $1300^2$ tiles) and 8-band multispectral ($1.24$\,m) imagery of Las Vegas, Paris, Shanghai and Khartoum with building footprints and road centrelines, together with the vendor pansharpened product (PS-MS / PS-RGB). Every pansharpening model is the WorldView-III PanCollection checkpoint applied \emph{without fine-tuning} to the raw PAN + MS pairs after the same $11$-bit scaling. The official training / test split of each city is used; the segmenters are trained on the vendor pansharpened product of the training regions (a U-Net for buildings, F1 at IoU $\ge0.5$ after polygonisation; a U-Net plus skeletonisation and graph extraction for roads, scored by APLS~\cite{vanetten2020roads}) and applied to every method's output on the test regions ($\fSnTiles$ building tiles with $\fSnBuildings$ footprints; $\fSnRoadTiles$ road tiles with $\fSnRoadKm$\,km of centrelines); the vendor product itself is scored with the same segmenters. Small buildings are footprints below $60$\,m$^2$. \cref{tab:spacenet_aoi} gives every method, the per-city breakdown and reduced-resolution fidelity on $400$ SpaceNet crops; \cref{fig:spacenet} shows real tiles with their labels.

\subsection{Adapting the retriever, not the generator}\label{sec:retadapt}
A rank-$16$ LoRA~\cite{hu2022lora} on the query, key and value projections of the host image tower is fitted on the training pairs so that the embedding of a generator's output matches the embedding of the reference tile (symmetric InfoNCE over the batch, temperature $0.07$, $5$ epochs, learning rate $10^{-4}$, text tower frozen). Three retrievers result: the frozen host, the host adapted to SGDiff outputs, and the host adapted to \our outputs. \cref{tab:retadapt} scores SGDiff and \our outputs under each, together with the reference tiles under the same retriever (an adapted index must still serve the reference-quality imagery already in an archive), the frozen OpenCLIP index, and the label-grounded measures that no retriever change can move.

\subsection{Human study}\label{sec:human}
$\fHumRaters$ raters with remote-sensing experience, blind to the method and compensated at the local hourly rate for research assistants, saw $\fHumPairs$ pairs of $128$-px full-resolution tiles ($15$ per scene; \our against SGDiff at the same location, left--right order randomised) and answered two questions per pair: Q1, aligned with the training objective, ``in which tile are small objects and their arrangement easier to recognise?''; Q2, objective-neutral, ``which tile looks more like a natural satellite photograph?''. A pair counts for \our when the majority of raters chose its tile. Because the $15$ pairs of a scene are not independent, the unit of the test is the scene: a scene counts for \our when the majority of its pairs does, which cannot tie with $15$ pairs, and the reported $p$ is that of the exact two-sided binomial sign test over the $\fHumNScenes$ scenes ($\fHumScenes$ of $\fHumNScenes$ scenes for Q1, $p{=}\fHumP$; $\fHumScenesTwo$ of $\fHumNScenes$ for Q2, $p{=}\fHumPTwo$). Because the scenes are crops of six acquisition strips, the test is repeated on strip means: all six strips favour \our on Q1 (exact sign-flip $p{=}0.031$, the smallest value six strips allow) and three of six on Q2. In a counting task on $80$ tiles with annotated vehicle counts the raters missed $\fCountOurs$ vehicles per tile on \our tiles against $\fCountBase$ on SGDiff tiles (two-sided Wilcoxon signed-rank test on the per-scene means, $p{=}\fCountP$) and counted $0.3$ against $0.4$ spurious vehicles per tile ($p{=}0.21$), so the gain is in omissions, not in additions. A second block of $200$ pairs compares \our with SeeSR, the most retrievable baseline, under the same protocol ($10$ pairs per scene): \our is preferred for small objects in $59\%$ of the pairs ($14$ of $20$ scenes, $p{=}0.12$) and SeeSR for naturalness in $56\%$ ($12$ of $20$, $p{=}0.50$), neither significant; the diffusion prior renders plausible texture, which raters read as natural, and the counting task was not repeated for it. Agreement among raters is moderate on Q1 (Krippendorff's $\alpha$ of $0.66$ and $0.58$) and low on Q2 ($0.41$ and $0.47$), so Q2 is reported as inconclusive rather than as a tie. The $p$-values belong to the scene counts; the pair shares are descriptive. \cref{tab:human} gives the results.

\begin{table}[!htbp]
\centering
\caption{Human study: $5$ raters, two questions per blind pair, left--right randomised, $128$-px full-resolution tiles at the same location. Block~1: $300$ pairs against SGDiff ($15$ per scene); block~2: $200$ pairs against SeeSR ($10$ per scene), the most retrievable baseline. Preference: share of pairs in which the majority of raters chose \our. Test: exact two-sided binomial sign test over the $20$ scenes; a scene counts for \our when the majority of its pairs does (ties are impossible with an odd number of pairs per scene, and are split for block~2), and scenes rather than pairs are the units because the pairs of one scene are not independent; the $p$-values belong to the scene counts, not to the pair shares. Q1 concerns what the training objective targets, and its wording was fixed before the study; Q2 does not. Agreement among raters (Krippendorff's $\alpha$) is moderate on Q1 and low on Q2. Counting task: $80$ tiles with annotated vehicle counts, scored for omissions and spurious counts.}
\label{tab:human}
\scriptsize
\setlength{\tabcolsep}{4pt}
\maxsizebox{\columnwidth}{%
\begin{tabular}{>{\raggedright\arraybackslash}p{.40\linewidth}cccc}
\toprule
Question & \shortstack{\our\\preferred} & \shortstack{scenes\\won} & \shortstack{Krippendorff\\$\alpha$} & $p$ \\
\midrule
Q1: in which tile are small objects and their arrangement easier to recognise? & 68\% & 19 / 20 & 0.66 & $4.0\times10^{-5}$ \\
Q2: which tile looks more like a natural satellite photograph? & 52\% & 11 / 20 & 0.41 & $0.82$ \\
\midrule
Q1, \our against SeeSR ($200$ pairs) & 59\% & 14 / 20 & 0.58 & $0.12$ \\
Q2, \our against SeeSR ($200$ pairs) & 44\% & 8 / 20 & 0.47 & $0.50$ \\
\midrule
counting task ($80$ tiles): vehicles missed per tile, \our / SGDiff & \multicolumn{4}{l}{$1.8$ / $2.6$; two-sided Wilcoxon signed-rank test on the $20$ per-scene means, $p{=}0.002$} \\
counting task: spurious vehicles counted per tile, \our / SGDiff & \multicolumn{4}{l}{$0.3$ / $0.4$; same test, $p{=}0.21$} \\
\bottomrule
\end{tabular}%
}
\end{table}

\subsection{Seeds and statistics}\label{sec:stats}
Every method, \our included, is trained with seeds $0,1,2$ and every table reports the mean over these three seeds; \our was additionally trained with seeds $3,4$ on WorldView-III; these two seeds enter only \cref{tab:fiveseeds} and the row of \our in \cref{tab:seeds}. Every fixed random draw uses seed $1234$. Metrics are averaged over seeds per test unit and tables report the mean over units with the std over units; \cref{tab:seeds} gives the seed std of Q2$^n$ for every method and \cref{tab:seeds_down} the seed std of every downstream measure for every method of \cref{tab:downstream_rr}. Family~A: paired Wilcoxon signed-rank tests of \our against SGDiff, ARConv, SSDiff and CANConv on the $21$ cells of Main Table~\MTabMain, Holm-corrected over the $\fFamN$ tests: $\fFamA$ adjusted $p$-values are below $0.01$ and $\fFamAw$ below $0.05$; the remaining two are the cells in which \our is second, GF-2 ERGAS against SSDiff and QB $D_\lambda$ against SGDiff, with unadjusted $p{=}\fPgf$ and $\fPqb$ (adjusted $1.00$), and the main text calls these margins ``within noise'', which is a statement about the test, not an equivalence claim. Family~B: best against second-best in every cell over all compared methods (the up-sampled input, a reference point, excluded), Holm-corrected over the $21$ tests; this family produces the marks, which are therefore identical in Main Table~\MTabMain{} and in \cref{tab:full_rr_a,tab:full_rr_b,tab:full_fr}. The per-scene values behind both families are printed in \cref{tab:perscene_wv3_rr,tab:perscene_qb_rr,tab:perscene_gf2_rr,tab:perscene_wv2_rr,tab:perscene_wv3_fr,tab:perscene_qb_fr,tab:perscene_gf2_fr} at exactly the precision the tests use (three decimals for ERGAS and SAM, four for Q2$^n$, $D_\lambda$, $D_s$ and HQNR), so every $p$ can be recomputed from the printed numbers. Because the $20$ scenes of a sensor are crops of $4$--$6$ acquisition strips, \cref{tab:strip} repeats every marked comparison at the strip level in two ways: the exact sign-flip permutation test on strip means, which reports the direction of the effect and whose smallest attainable $p$ is $0.031$, $0.062$ and $0.125$ with six, five and four strips, and a linear mixed-effects model of the per-scene paired differences with a random intercept per strip (REML; $t$-test of the mean difference with strips$-1$ degrees of freedom, the conservative choice for four to six clusters), which uses the within-strip replication and is the strip-level significance quoted in the main text. The marks of Main Table~\MTabMain{} are those of the scene-level Family~B test; \cref{tab:strip} tests every cell against its actual runner-up, and the full-resolution per-scene values are printed for all three sensors with CLIPPan, the runner-up of three full-resolution cells (\cref{tab:perscene_wv3_fr,tab:perscene_qb_fr,tab:perscene_gf2_fr}). Every marked cell has all or all but one strip in favour of \our; the mixed model gives $p{<}0.05$ in $18$ of the $19$ marked cells, $p{<}0.01$ in $12$, and $0.054$ in the remaining one (WorldView-III $D_\lambda$, five of six strips in favour). Retrieval std is the per-entry query bootstrap; downstream std is over seeds; the discordance of Main Sec.~\MSecGap{} carries a system bootstrap and, for its headline value, a scene bootstrap; rank inversions, family offsets and partial correlations follow \cref{alg:eval}. Unless a counterpart is named, every gain quoted in the paper is taken against the strongest baseline of its row and is the difference of the two printed values. The comparison set of Main Sec.~\MSecGap{} was fixed before the analysis: every trained system that does not close the loop, i.e.\ that neither re-indexes its output nor trains on coverage, $\fNSys$ systems in all, three of which share our backbone and encoder by design as open-loop controls; \our-lite, the open-loop model and the ablation configurations are variants of our method and are not part of it. KID~\cite{binkowski2018kid} replaces FID, which is biased at our sample sizes. Differences between ablation configurations smaller than about twice the pooled seed std of \cref{tab:ablation_full} are described as within noise. Where a comparison between configurations carries a $p$-value (the one-term and replacement settings of \cref{sec:objectives}), it is the scene-paired Wilcoxon test on the per-scene values averaged over the three seeds ($20$ pairs; seeds are averaged, never pooled as independent observations), Holm-corrected over the alternatives named in the same sentence. The gains on the external benchmarks carry $95\%$ bootstrap intervals of the paired gain, over queries for the banks and over captions for RSITMD and RSICD (\cref{tab:vocab_split}).

\section{Implementation}\label{sec:impl}

\subsection{Architecture}\label{toc:supp:4:1}
\begin{figure*}[!tbp]
\centering
\includegraphics[width=\textwidth]{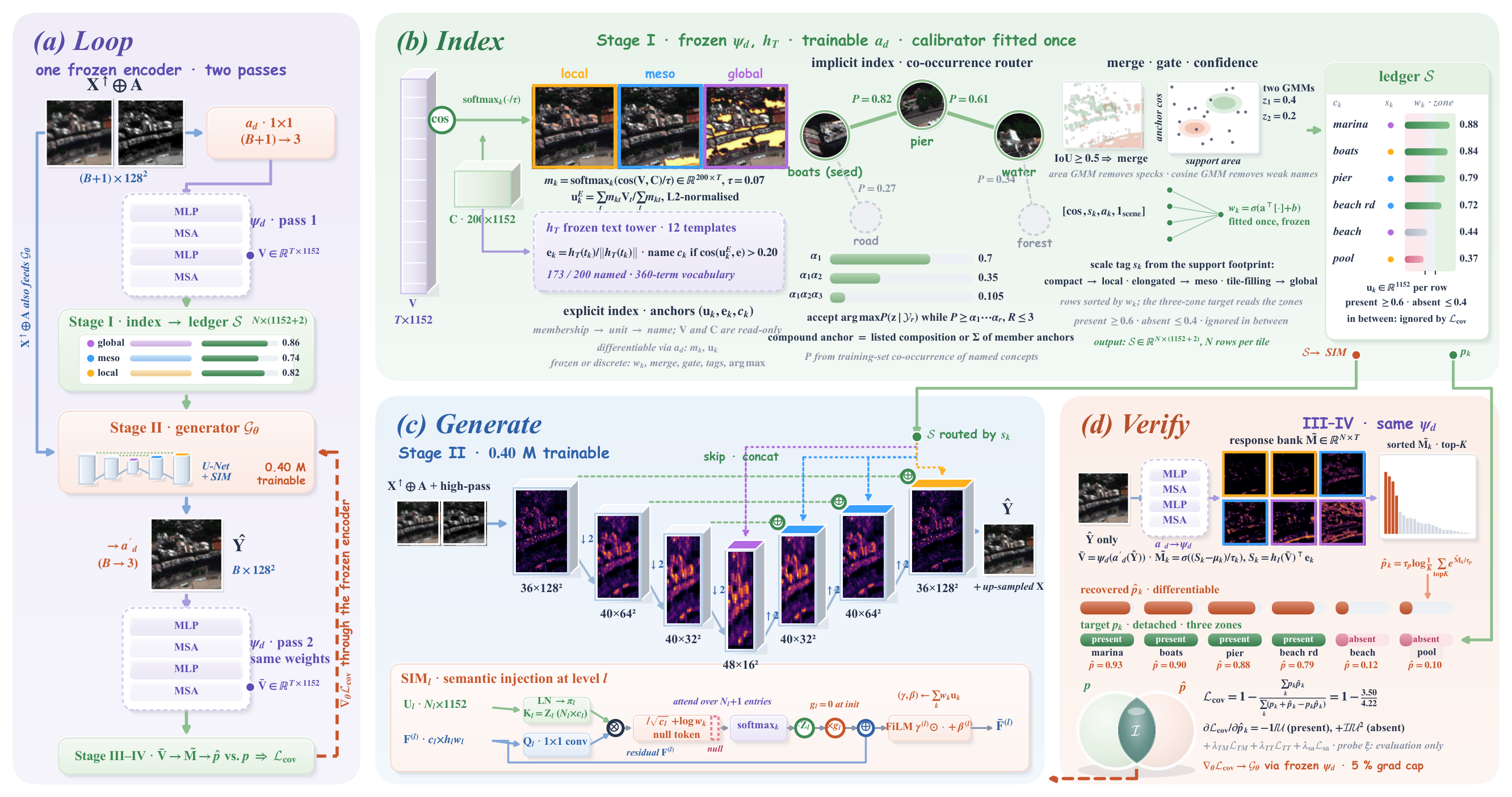}
\caption{Module-level diagram of the full model on a real WorldView-III tile, extending Main Fig.~\MFigMethod{} with every tensor shape, formula and hyper-parameter (best viewed enlarged). (a)~Loop: the two passes of the frozen encoder and the single dashed return path. (b)~Index: membership of tokens over centroids, naming by the text tower, the co-occurrence router with its probabilities and thresholds, IoU merge, two-GMM gate and the frozen confidence calibrator, and the ledger as a table. (c)~Generate: U-Net feature volumes, skip concatenations, the ledger routed by scale to each decoder SIM, and the SIM block diagram. (d)~Verify: response bank of the re-encoded output, sorted top-$K$ responses, detached three-zone target and the soft-Jaccard loss.}
\label{fig:framework_supp}
\end{figure*}

\begin{figure*}[!tbp]
\centering
\includegraphics[width=\textwidth]{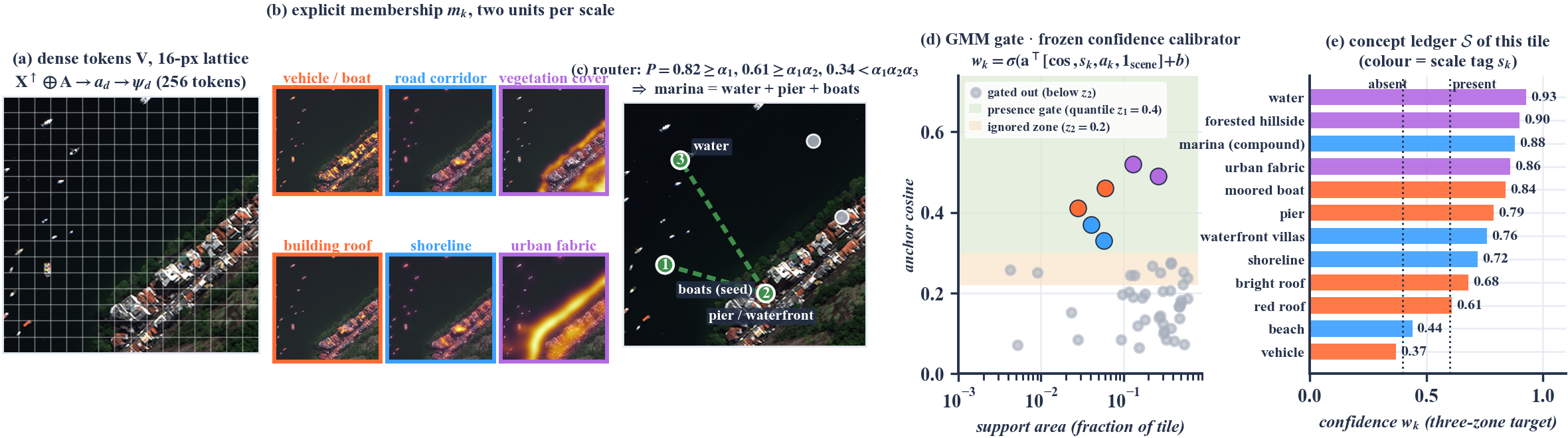}
\caption{Stage I anatomy on a WorldView-III marina tile: dense tokens on the lattice, explicit membership at three tile-relative scales, the co-occurrence router with its acceptance probabilities, the gate in the (support area, anchor cosine) plane, and the resulting ledger with confidences and scale tags.}
\label{fig:anatomy_index}
\end{figure*}

\begin{figure*}[!tbp]
\centering
\includegraphics[width=\textwidth]{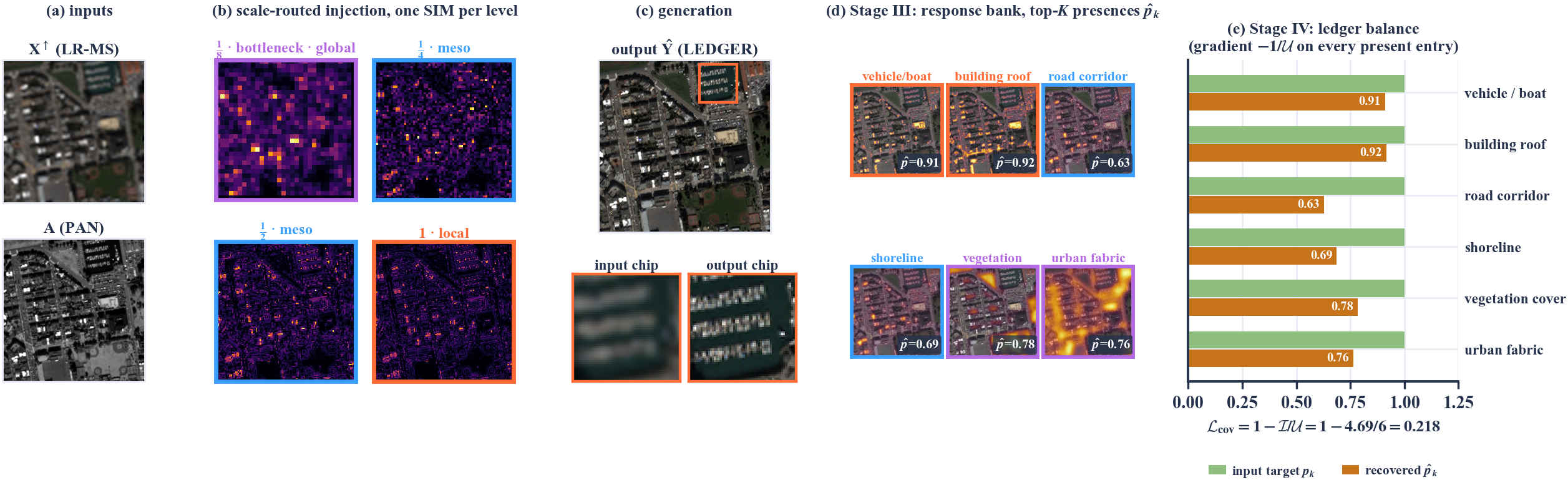}
\caption{Stages II--IV anatomy on a QuickBird city-block tile: inputs, scale-routed injection into the decoder levels, output and local chip, the response bank with top-$K$ presences after re-indexing the output alone, and the set-level comparison whose soft-Jaccard gradient returns through the frozen encoder.}
\label{fig:anatomy_generate}
\end{figure*}

\cref{tab:arch} lists every trainable module. The generator has widths $36,40,40,48$ at the four levels, one residual block (conv--GELU--conv with channel attention) per level, strided $3{\times}3$ down-sampling, bilinear up-sampling with a $1{\times}1$ projection and a $1{\times}1$ skip fusion, a zero-initialised head and, as input, the up-sampled target, the guide and their high-pass components. Injection modules sit at the bottleneck and on the three ascending levels. There are two adapters: $a_d$ ($3{\times}(B{+}1)$ weights and three biases) maps the up-sampled multispectral bands and the guide to the encoder's three channels in Stage~I and receives gradient from $\Lrec$ only, through the injected units: the reconstruction and semantic gradients are accumulated in two backward passes and the semantic one is discarded on $a_d$ (\cref{alg:train}, line 12), the confidences and the target are detached, and $\Lsa$ compares the output units with a stop-gradient copy of the input units; $a'_d$ ($3{\times}B$) maps the output alone in Stage~III and receives gradient from the semantic terms. \cref{tab:adapter} shows that fixing either adapter at the natural-colour matrix changes nothing beyond seed noise (last row for $a_d$, third row for $a'_d$); the statement of Main Sec.~\MSecMethod{} that the input units, the confidences and the target are constants for $\Lcov$ refers to this: the only trainable parameters on the path to the ledger, those of $a_d$, receive gradient from $\Lrec$ alone, and the last row of \cref{tab:adapter} shows that freezing them changes nothing beyond seed noise. The confidence calibrator has five fixed coefficients and no trainable parameters during the loop. \cref{fig:framework_supp} is the module-level diagram of the full model on one real tile; \cref{fig:anatomy_index,fig:anatomy_generate} expand indexing and injection--verification.

\begin{table}[!htbp]
\centering
\caption{Trainable modules of \our for an 8-band sensor ($0.396$\,M parameters; $0.396$\,M for 4 bands). The frozen SigLIP-2 So400m/16 NaFlex vision tower adds $428$\,M. Both adapters are updated by every loss term through which gradient reaches them: $a_d$ by $\Lrec$ only through the differentiable memberships (the ledger confidences and target zones are detached), $a'_d$ by the semantic terms; \cref{tab:adapter} shows that fixing either changes nothing beyond seed noise.}
\label{tab:arch}
\footnotesize
\setlength{\tabcolsep}{4pt}
\maxsizebox{\columnwidth}{%
\begin{tabular}{>{\raggedright\arraybackslash}p{.34\linewidth}r>{\raggedright\arraybackslash}p{.50\linewidth}}
\toprule
Module & Params & Notes \\
\midrule
Domain adapter $a_d$ / Stage-III adapter $a'_d$ & 54 & $1{\times}1$: $3{\times}(8{+}1)$ + bias and $3{\times}8$ + bias, band-centre initialisation \\
Unit reduction $1152\to64$ & 73.8\,K & linear, shared by all levels \\
Confidence calibrator & 0 & logistic map of $[\cos,s_k,a_k,\mathbf{1}_{\mathrm{scene}}]$, fitted once in Stage~I, frozen (5 fixed coefficients) \\
U-Net encoder/decoder & 254.3\,K & widths $36,40,40,48$ \\
Injection modules ($4$) & 68.2\,K & $\pi_l$, query $1{\times}1$, null token, FiLM, gate \\
\midrule
Total trainable & 396.3\,K & \\
\bottomrule
\end{tabular}%
}
\end{table}

\subsection{Index fitting}\label{toc:supp:4:2}
\cref{alg:fit} is the procedure. Centroids: $k$-means ($K_c{=}200$, $10$ restarts, seed $1234$) on $20{,}000$ dense tokens from $500$ training tiles. Anchors: the vocabulary of \cref{tab:vocab} ($96$ pathology terms and the $80$ COCO categories plus $200$ caption nouns in the other domains) embedded with the $12$ templates and averaged; a centroid takes the nearest term if the cosine exceeds $0.20$ (median $0.31$ on WorldView-III; $173$ of $200$ named) and its visual prototype otherwise. Co-occurrence: first- and second-order tables of hard token types with additive smoothing $0.01$. Calibrator: logistic regression of $[\cos,s_k,a_k,\mathbf{1}_{\mathrm{scene}}]$ on whether the unit's anchor is also found in the reference patch, fitted once on $5{,}000$ training patches and frozen. Response statistics: $\mu_k$ and $\tau_k$ calibrated on the pooled top-$K$ presence computed on $64$-token training patches (midpoint and quarter-gap of the medians on tiles with and without the centroid, clamped to $[0.02,0.25]$), so that a present concept has presence about $0.88$ and an absent one about $0.12$ at every grid size (\cref{tab:calib}). Gates: two diagonal two-component Gaussian mixtures on anchor cosine and support area, thresholds at density quantiles $0.4$ and $0.2$. Nothing in the index reads a validation or test tile.

\subsection{Training recipe}\label{sec:recipe}
AdamW~\cite{loshchilov2019adamw} ($5{\times}10^{-4}$, weight decay $10^{-4}$, $\beta{=}(0.9,0.999)$), batch $64$, $1{,}000$ epochs, two-epoch linear warm-up then cosine decay to $2\%$ of the peak, gradient clipping at $1.0$, EMA of the trainable weights ($0.999$, used at inference), fp32 generator with the frozen encoder in bf16. Loss phases: $\Lrec$ from epoch $0$; $\Ltt,\Ltm$ from epoch $100$; $\Lcov,\Lsa$ from epoch $200$, each ramped linearly over $25$ epochs; the reconstruction and semantic gradients are accumulated in two backward passes, the semantic one rescaled whenever its output-space norm exceeds $5\%$ of the reconstruction gradient's (the measured ratio settles at $1.9\%$; \cref{alg:train}, line 13) and discarded on the Stage-I adapter $a_d$, which is therefore trained by $\Lrec$ alone (\cref{alg:train}, \cref{tab:adapter}). Augmentation: random $64$-px crops, flips and transposition applied identically to all arrays. Model selection: the EMA weights with the lowest ERGAS on the official validation split; the training-time monitor is the cross-encoder gap on $32$ validation patches, with a stopping rule at $0.10$ that was never triggered. \cref{tab:protocol} shows that neither EMA nor selection explains the margin. Baselines are retrained with their official implementations under the same data, degradation and $1{,}000$-epoch budget with their own optimisers and schedules; CLIPPan is trained at full resolution with its official implementation; DA-CLIP is the official implementation with the first convolution of its restoration network widened to $B{+}1$ input channels (initialised by copying the RGB weights to the band-centre channels and zero elsewhere) and its output layer set to $B$ bands, the frozen CLIP controller fed with the natural-colour composite, fine-tuned under the same protocol. SeeSR keeps its pretrained Stable-Diffusion~2 prior, VAE and text encoder frozen and is adapted at four points: (i)~a trainable $1{\times}1$ input adapter, initialised as the natural-colour composite, maps $\mathbf{X}^{\uparrow}\oplus\mathbf{A}$ ($B{+}1$ channels) to the three channels that the ControlNet condition branch encodes; (ii)~the DAPE prompt extractor receives the same composite and is kept frozen, its tags entering the prompt as released; (iii)~the VAE operates on three-channel latents, so the decoded image is a natural-colour sharpened image, and a trainable spectral head (two $3{\times}3$ convolutions on the decoded image concatenated with $\mathbf{X}^{\uparrow}$) predicts the $B$-band detail residual that is added to $\mathbf{X}^{\uparrow}$; (iv)~the ControlNet branch, the input adapter and the spectral head are trained with the reconstruction loss of the common protocol on the $B$-band output, the diffusion loss of SeeSR being kept on the three-channel latent; sampling uses the released $50$ steps. The vendor-style natural-colour composite is the only image representation the frozen components ever see. The three open-loop controls share the U-Net and the frozen encoder of \our and differ only in how the encoder output enters the generator (DenseCLIP-style: eight PCA-reduced dense channels concatenated to the input; CLIPDenoising-style: multi-scale pooled embeddings broadcast at every encoder level; DA-CLIP-style: a controller MLP predicting FiLM parameters at every level). \cref{tab:hyper} lists every hyper-parameter; none is tuned per sensor or domain, every default was fixed on the validation split, and \our-lite keeps the six method-specific ones marked $\dagger$ while inheriting the target zones, the gradient cap, $K_{\mathrm{meso}}$, $K_{\mathrm{global}}$ and the loss schedule of the full model. The sweeps of \cref{tab:sweeps} cost $432$ GPU-hours (\cref{tab:gpu}); the baselines were run with the recipes their authors tuned on the same PanCollection data, so that each system is evaluated with the configuration its authors chose for this benchmark, and the training-only form (\cref{tab:plugin}) and the unseen-sensor results show the gains under the baselines' own recipes and on data no recipe was tuned for.

\begin{table}[!htbp]
\centering
\caption{Hyper-parameters, identical on every sensor and domain ($\fNHyper$ in total; the six method-specific ones kept by \our-lite are marked $\dagger$, the others it inherits unchanged). Every value was chosen on the validation split.}
\label{tab:hyper}
\footnotesize
\setlength{\tabcolsep}{4pt}
\maxsizebox{\columnwidth}{%
\begin{tabular}{>{\raggedright\arraybackslash}p{.47\linewidth}>{\raggedright\arraybackslash}p{.50\linewidth}}
\toprule
Component & Value \\
\midrule
Centroids $K_c$$^\dagger$ / membership temperature $\tau$$^\dagger$ & $200$ / $0.07$ \\
Anchor naming threshold$^\dagger$ / templates per term & $0.20$ / $12$ \\
Router depth $R$ / thresholds $\alpha_{1:3}$ & $3$ / $(0.7,0.5,0.3)$ \\
IoU merge $\theta_{\mathrm{IoU}}$ / gate quantiles $(z_1,z_2)$ & $0.5$ / $(0.4,0.2)$ \\
Presence temperature $\tau_p$$^\dagger$ / top-$K$ $(K_l^\dagger,K_m,K_g)$ & $0.1$ / $(4,16,64)$ \\
Target zones (absent $\le$, present $\ge$) & $0.4$, $0.6$ \\
Loss weights $(\lambda_{TT},\lambda_{TM},\lambda_{\mathrm{cov}}^\dagger,\lambda_{\mathrm{sa}})$ / $\eta$ & $(1,0.5,0.3,0.2)$ / $0.1$ \\
Phases (epochs) / ramp / semantic gradient cap & $0,100,200$ / $25$ / $5\%$ \\
Optimiser / lr / schedule / EMA & AdamW / $5{\times}10^{-4}$ / cosine / $0.999$ \\
Batch / epochs / patch & $64$ / $1{,}000$ / $64$ px \\
Encoder grid & NaFlex, $16$-px patch, input at $2\times$ the output grid, $\le1{,}024$ tokens per window \\
\bottomrule
\end{tabular}%
}
\end{table}

\subsection{Compute}\label{toc:supp:4:4}
\cref{tab:cost} decomposes the per-image cost at the training patch and at a $256^2$ scene, and its totals are sums of the component rows. The frozen encoder dominates ($\fEncGF$\,GFLOPs and $118$\,ms for the $1{,}024$ tokens of a $256^2$ scene at the $2\times$ grid on an RTX~4090); the lightweight path costs $\fLightGF$\,GFLOPs per $64^2$ training patch. Generation-only deployment is $\fGenGF$\,GFLOPs and $\fInferMs$\,ms per $256^2$ scene (against $22$\,ms for CANConv and $1{,}210$--$2{,}910$\,ms for the diffusion baselines of \cref{tab:neighbours}, i.e.\ $\fSpeedLo$--$\fSpeedHi\times$ faster than those); the optional re-index score adds a second encoder pass ($\fCertGF$\,GFLOPs, $\fCertMs$\,ms). The deployed model has $\fTotalParams$\,M parameters, of which $\fParams$\,M are trained; the training-only form deploys only the generator it was added to. Training peaks at $12.6$\,GB (batch $64$, bf16 encoder); one WorldView-III run takes $3.9$ RTX-4090-equivalent GPU-hours ($1.7$\,h on the H200 on which it was run; the conversion factor $2.3$ is the measured throughput ratio of the two cards on this workload and is applied to every entry of \cref{tab:runs,tab:gpu}, so that all compute figures of the paper are in one unit). \cref{tab:gpu} gives the compute of the complete study ($\fGpuHours$ RTX-4090-equivalent GPU-hours); its first three rows are the per-run records of \cref{tab:runs} summed over the three seeds and the three training sensors, the QuickBird and GaoFen-2 runs weighted by their training-set sizes, so that they can be recomputed from that table.

\begin{table}[!htbp]
\centering
\caption{Per-image cost on an RTX~4090 (batch $1$, bf16 encoder, mean of $50$ runs) at the $64^2$ training patch ($64$ tokens) and at a $256^2$ scene ($1{,}024$ tokens at the $2\times$ grid). The totals are sums of the component rows above them (the encoder row counted twice for the re-index score), so every total can be checked.}
\label{tab:cost}
\footnotesize
\setlength{\tabcolsep}{3pt}
\maxsizebox{\columnwidth}{%
\begin{tabular}{lrrrrr}
\toprule
Module & Params & GFLOPs$_{64}$ & ms$_{64}$ & GFLOPs$_{256}$ & ms$_{256}$ \\
\midrule
frozen encoder, one pass & 428\,M & 55.10 & 18.2 & 912.4 & 118.0 \\
adapter + explicit index + router + gates & 73.9\,K & 0.27 & 3.7 & 4.3 & 5.9 \\
U-Net generator + injection & 322.5\,K & 1.23 & 4.3 & 19.7 & 9.4 \\
Stage-III adapter + re-index score & 27 & 0.06 & 1.1 & 1.0 & 1.6 \\
\midrule
lightweight path (no encoder), training patch $64^2$ & 396.4\,K & 1.56 & 9.1 \\
generation only, $256^2$ scene (encoder once) & 428.4\,M & & & 936.4 & 133.3 \\
generation + re-index score, $256^2$ scene (encoder twice) & 428.4\,M & & & 1849.8 & 252.9 \\
\bottomrule
\end{tabular}%
}
\end{table}

\begin{table}[!htbp]
\centering
\caption{Compute of the complete study in RTX-4090-equivalent GPU-hours. The first three rows are sums of the per-run costs of \cref{tab:runs} over the three seeds and the three training sensors, the QuickBird and GaoFen-2 runs weighted by their training-set sizes ($1.76\times$ and $2.04\times$ the WorldView-III patches, \cref{sec:data}), so that each of them can be recomputed from that table; the main-run row adds seeds $3$ and $4$ on WorldView-III.}
\label{tab:gpu}
\footnotesize
\setlength{\tabcolsep}{6pt}
\maxsizebox{\columnwidth}{%
\begin{tabular}{lr}
\toprule
Block & GPU-hours \\
\midrule
main runs (three sensors, three seeds, $+$ seeds $3$--$4$) & 64 \\
eleven published baselines (three sensors, three seeds) & 5197 \\
adapted CLIPPan, DA-CLIP and SeeSR & 1440 \\
ablations and factorial & 1188 \\
same-backbone objectives & 371 \\
sweeps & 432 \\
labelled-benchmark fine-tuning & 71 \\
training-only variants & 84 \\
encoder swap, neighbours, natural images & 418 \\
medical & 504 \\
retriever adaptation (LoRA) & 22 \\
SpaceNet segmenters and evaluation & 96 \\
evaluators and scoring & 169 \\
\midrule
total & 10056 \\
\bottomrule
\end{tabular}%
}
\end{table}

\section{Additional results}\label{supp:sec:results}

\subsection{Every compared system, seed variability and protocol parity}\label{toc:supp:5:1}
\begin{figure*}[!tbp]
\centering
\includegraphics[width=\textwidth]{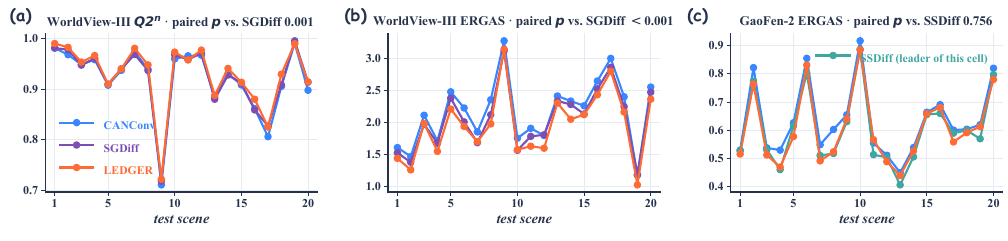}
\caption{Per-scene values behind the paired tests (seed-averaged). (a) WorldView-III Q2$^n$ and (b) ERGAS: the margin of \our over SGDiff is present on most scenes. (c) GF-2 ERGAS: \our and SSDiff, the leader of that cell, alternate from scene to scene, and the paired test does not separate them.}
\label{fig:perscene}
\end{figure*}

\cref{tab:full_rr_a,tab:full_rr_b,tab:full_fr} extend Main Table~\MTabMain{} to every compared system, including the classical methods, PanNet, FusionNet, PanFormer, HyperTransformer, the adapted DA-CLIP and SeeSR, the plain U-Net and the three open-loop controls; \cref{tab:seeds} gives the seed std of Q2$^n$ for each method, \cref{tab:seeds_down} the seed std of every downstream measure for every method of \cref{tab:downstream_rr}, \cref{tab:fiveseeds} the five seeds of \our, and \cref{tab:protocol} checks the two elements of our training protocol that the baselines' recipes lack, the EMA of the weights and checkpoint selection by validation ERGAS: adding them to a baseline moves its Q2$^n$ by $0.001$, and removing them from \our by at most $0.002$. The check is limited to these two elements; data and budget are common to all methods, and every baseline otherwise keeps its official recipe. \cref{tab:perscene_wv3_rr,tab:perscene_qb_rr,tab:perscene_gf2_rr,tab:perscene_wv2_rr,tab:perscene_wv3_fr,tab:perscene_qb_fr,tab:perscene_gf2_fr} list every per-scene value behind the paired tests for the four strongest published baselines, CLIPPan at full resolution, and \our at the precision the tests use, with the scene-paired, the sign-flip and the mixed-model $p$ in their last rows and the strip of every scene in their captions; \cref{tab:strip} collects the strip-level tests, and \cref{fig:perscene} plots three cells: on WorldView-III Q2$^n$ the margin of \our over SGDiff is present on $\fWinSG$ of $20$ scenes ($\fTieSG$ within $0.001$, $\fLossSG$ against) and over CANConv on $\fWinCC$ of $20$, whereas GF-2 ERGAS alternates between SSDiff and \our from scene to scene ($\fWinGFE$ against $\fLossGFE$), which is what the non-significant cell means. \cref{tab:repro} places every retrained published network beside the value its authors report on the same test scenes: the retrained values are within one per cent of the published ones, slightly below them in most cells, as expected when a recipe tuned by its authors is run under a common budget, and the published order of the networks is preserved on every sensor. \cref{tab:runs} records every retraining run (optimiser, schedule, selected epoch, validation values, wall-clock, memory).


\begin{ReportFlowTable}
\begin{longtable}{>{\raggedright\arraybackslash}p{\dimexpr.23\linewidth-2\tabcolsep\relax}>{\raggedright\arraybackslash}p{\dimexpr.26\linewidth-2\tabcolsep\relax}|*{3}{>{\centering\arraybackslash}p{\dimexpr.17\linewidth-2\tabcolsep-.14pt\relax}}}
\caption{Reduced-resolution results of every compared system on GaoFen-2 and QuickBird (mean$\pm$std over the $20$ test scenes; three seeds for every method). Kind: published pansharpening network, adapted VLM-guided restoration method (official implementation), same-backbone control, classical. Bold: best; underline: second. The significance marks are those of Main Table~\MTabMain: one test family over all compared methods, so the two tables carry identical marks. The up-sampled input (EXP) is a reference point and is not ranked.}
\label{tab:full_rr_a}\\
\toprule
Method & kind & ERGAS$\downarrow$ & SAM$\downarrow$ & Q2$^n\uparrow$ \\
\midrule
\endfirsthead
\multicolumn{5}{l}{\footnotesize\itshape Table~\thetable\ (continued)}\\[2pt]
\toprule
Method & kind & ERGAS$\downarrow$ & SAM$\downarrow$ & Q2$^n\uparrow$ \\
\midrule
\endhead
\bottomrule\endfoot
\multicolumn{5}{l}{\emph{GaoFen-2}} \\*
\midrule
EXP & input, not ranked & 2.366$\pm$0.20 & 1.820$\pm$0.22 & 0.812$\pm$0.012 \\
GSA & classical & 1.620$\pm$0.13 & 1.681$\pm$0.14 & 0.890$\pm$0.007 \\
BDSD-PC & classical & 1.667$\pm$0.12 & 1.681$\pm$0.14 & 0.892$\pm$0.008 \\
PanNet & published & 0.925$\pm$0.12 & 1.004$\pm$0.15 & 0.966$\pm$0.007 \\
FusionNet & published & 0.981$\pm$0.13 & 0.967$\pm$0.15 & 0.965$\pm$0.007 \\
LAGConv & published & 0.692$\pm$0.12 & 0.792$\pm$0.14 & 0.980$\pm$0.007 \\
PanFormer & published & 0.774$\pm$0.13 & 0.851$\pm$0.14 & 0.977$\pm$0.006 \\
H-Trans & published & 0.745$\pm$0.12 & 0.812$\pm$0.14 & 0.978$\pm$0.007 \\
CrossDiff & published & 0.861$\pm$0.13 & 0.945$\pm$0.14 & 0.972$\pm$0.007 \\
SSDiff & published & \best{0.603$\pm$0.13} & \second{0.673$\pm$0.15} & \second{0.984$\pm$0.007} \\
Pan-Mamba & published & 0.641$\pm$0.13 & 0.712$\pm$0.14 & 0.982$\pm$0.006 \\
CANConv & published & 0.634$\pm$0.13 & 0.712$\pm$0.14 & 0.983$\pm$0.007 \\
ARConv & published & 0.629$\pm$0.12 & 0.703$\pm$0.14 & 0.983$\pm$0.007 \\
SGDiff & published & 0.618$\pm$0.12 & 0.692$\pm$0.14 & 0.984$\pm$0.007 \\
CLIPPan & adapted & 0.742$\pm$0.12 & 0.801$\pm$0.14 & 0.978$\pm$0.007 \\
DA-CLIP & adapted & 0.731$\pm$0.12 & 0.788$\pm$0.14 & 0.979$\pm$0.007 \\
SeeSR & adapted & 0.864$\pm$0.12 & 0.902$\pm$0.15 & 0.971$\pm$0.007 \\
Plain U-Net & control & 0.779$\pm$0.12 & 0.840$\pm$0.15 & 0.977$\pm$0.007 \\
DenseCLIP-style & control & 0.690$\pm$0.12 & 0.742$\pm$0.14 & 0.981$\pm$0.007 \\
CLIPDenoising-style & control & 0.751$\pm$0.12 & 0.812$\pm$0.14 & 0.978$\pm$0.007 \\
DA-CLIP-style & control & 0.712$\pm$0.12 & 0.769$\pm$0.14 & 0.980$\pm$0.007 \\
\rowcolor{ourrow}\our & ours & \second{0.606$\pm$0.13} & \best{0.648$\pm$0.15}\sig & \best{0.987$\pm$0.007}\sig \\
\midrule
\multicolumn{5}{l}{\emph{QuickBird}} \\*
\midrule
EXP & input, not ranked & 11.819$\pm$0.55 & 8.435$\pm$1.24 & 0.584$\pm$0.102 \\
GSA & classical & 7.445$\pm$0.31 & 7.866$\pm$0.84 & 0.834$\pm$0.062 \\
BDSD-PC & classical & 7.515$\pm$0.32 & 8.089$\pm$0.84 & 0.831$\pm$0.061 \\
PanNet & published & 5.871$\pm$0.34 & 5.803$\pm$0.80 & 0.884$\pm$0.064 \\
FusionNet & published & 4.171$\pm$0.33 & 4.934$\pm$0.82 & 0.924$\pm$0.064 \\
LAGConv & published & 3.838$\pm$0.32 & 4.561$\pm$0.80 & 0.933$\pm$0.061 \\
PanFormer & published & 3.905$\pm$0.34 & 4.858$\pm$0.81 & 0.926$\pm$0.063 \\
H-Trans & published & 3.812$\pm$0.34 & 4.715$\pm$0.82 & 0.929$\pm$0.064 \\
CrossDiff & published & 4.436$\pm$0.33 & 5.389$\pm$0.87 & 0.923$\pm$0.061 \\
SSDiff & published & 3.641$\pm$0.33 & 4.478$\pm$0.84 & 0.934$\pm$0.062 \\
Pan-Mamba & published & 3.702$\pm$0.32 & 4.598$\pm$0.83 & 0.936$\pm$0.062 \\
CANConv & published & 3.660$\pm$0.34 & 4.519$\pm$0.83 & 0.936$\pm$0.063 \\
ARConv & published & 3.642$\pm$0.35 & 4.441$\pm$0.87 & \second{0.938$\pm$0.065} \\
SGDiff & published & \second{3.586$\pm$0.33} & \second{4.364$\pm$0.82} & 0.938$\pm$0.063 \\
CLIPPan & adapted & 4.118$\pm$0.33 & 4.862$\pm$0.86 & 0.918$\pm$0.064 \\
DA-CLIP & adapted & 3.934$\pm$0.32 & 4.731$\pm$0.80 & 0.927$\pm$0.062 \\
SeeSR & adapted & 4.372$\pm$0.32 & 5.104$\pm$0.84 & 0.904$\pm$0.064 \\
Plain U-Net & control & 3.941$\pm$0.34 & 4.720$\pm$0.81 & 0.927$\pm$0.061 \\
DenseCLIP-style & control & 3.790$\pm$0.32 & 4.571$\pm$0.83 & 0.933$\pm$0.062 \\
CLIPDenoising-style & control & 3.902$\pm$0.31 & 4.683$\pm$0.80 & 0.929$\pm$0.062 \\
DA-CLIP-style & control & 3.842$\pm$0.31 & 4.612$\pm$0.83 & 0.931$\pm$0.063 \\
\rowcolor{ourrow}\our & ours & \best{3.392$\pm$0.33}\sig & \best{4.148$\pm$0.88}\sig & \best{0.946$\pm$0.063}\sig \\
\end{longtable}
\end{ReportFlowTable}


\begin{ReportFlowTable}
\begin{longtable}{>{\raggedright\arraybackslash}p{\dimexpr.23\linewidth-2\tabcolsep\relax}>{\raggedright\arraybackslash}p{\dimexpr.26\linewidth-2\tabcolsep\relax}|*{3}{>{\centering\arraybackslash}p{\dimexpr.17\linewidth-2\tabcolsep-.14pt\relax}}}
\caption{Reduced-resolution results of every compared system on WorldView-III and on the unseen WorldView-II sensor (mean$\pm$std over the $20$ test scenes; marks as in Main Table~\MTabMain).}
\label{tab:full_rr_b}\\
\toprule
Method & kind & ERGAS$\downarrow$ & SAM$\downarrow$ & Q2$^n\uparrow$ \\
\midrule
\endfirsthead
\multicolumn{5}{l}{\footnotesize\itshape Table~\thetable\ (continued)}\\[2pt]
\toprule
Method & kind & ERGAS$\downarrow$ & SAM$\downarrow$ & Q2$^n\uparrow$ \\
\midrule
\endhead
\bottomrule\endfoot
\multicolumn{5}{l}{\emph{WorldView-III}} \\*
\midrule
EXP & input, not ranked & 7.155$\pm$0.91 & 5.800$\pm$0.93 & 0.627$\pm$0.108 \\
GSA & classical & 4.645$\pm$0.51 & 5.323$\pm$0.56 & 0.818$\pm$0.065 \\
BDSD-PC & classical & 4.698$\pm$0.51 & 5.429$\pm$0.58 & 0.829$\pm$0.067 \\
PanNet & published & 2.681$\pm$0.51 & 3.629$\pm$0.56 & 0.890$\pm$0.066 \\
FusionNet & published & 2.478$\pm$0.51 & 3.338$\pm$0.58 & 0.903$\pm$0.065 \\
LAGConv & published & 2.311$\pm$0.53 & 3.119$\pm$0.58 & 0.909$\pm$0.067 \\
PanFormer & published & 2.239$\pm$0.52 & 3.102$\pm$0.56 & 0.909$\pm$0.064 \\
H-Trans & published & 2.213$\pm$0.51 & 3.024$\pm$0.59 & 0.911$\pm$0.068 \\
CrossDiff & published & 2.731$\pm$0.51 & 3.637$\pm$0.60 & 0.891$\pm$0.065 \\
SSDiff & published & 2.117$\pm$0.51 & 2.855$\pm$0.57 & 0.915$\pm$0.065 \\
Pan-Mamba & published & 2.191$\pm$0.52 & 2.924$\pm$0.59 & 0.919$\pm$0.066 \\
CANConv & published & 2.166$\pm$0.52 & 2.939$\pm$0.56 & 0.920$\pm$0.067 \\
ARConv & published & 2.147$\pm$0.52 & 2.896$\pm$0.59 & 0.921$\pm$0.063 \\
SGDiff & published & \second{2.052$\pm$0.49} & \second{2.780$\pm$0.57} & \second{0.923$\pm$0.065} \\
CLIPPan & adapted & 2.412$\pm$0.53 & 3.206$\pm$0.59 & 0.899$\pm$0.068 \\
DA-CLIP & adapted & 2.286$\pm$0.50 & 3.071$\pm$0.57 & 0.911$\pm$0.064 \\
SeeSR & adapted & 2.634$\pm$0.53 & 3.412$\pm$0.61 & 0.886$\pm$0.064 \\
Plain U-Net & control & 2.302$\pm$0.52 & 3.098$\pm$0.56 & 0.910$\pm$0.068 \\
DenseCLIP-style & control & 2.207$\pm$0.51 & 2.991$\pm$0.59 & 0.917$\pm$0.066 \\
CLIPDenoising-style & control & 2.271$\pm$0.51 & 3.061$\pm$0.58 & 0.912$\pm$0.068 \\
DA-CLIP-style & control & 2.231$\pm$0.53 & 3.019$\pm$0.58 & 0.915$\pm$0.064 \\
\rowcolor{ourrow}\our & ours & \best{1.962$\pm$0.51}\sig & \best{2.680$\pm$0.59}\sig & \best{0.929$\pm$0.065}\sig \\
\midrule
\multicolumn{5}{l}{\emph{WorldView-II (unseen sensor; WorldView-III models)}} \\*
\midrule
EXP & input, not ranked & 6.812$\pm$0.52 & 6.471$\pm$0.76 & 0.661$\pm$0.075 \\
GSA & classical & 4.555$\pm$0.34 & 6.208$\pm$0.48 & 0.809$\pm$0.046 \\
BDSD-PC & classical & 4.649$\pm$0.34 & 6.089$\pm$0.50 & 0.822$\pm$0.048 \\
PanNet & published & 4.589$\pm$0.31 & 5.839$\pm$0.49 & 0.804$\pm$0.046 \\
FusionNet & published & 4.320$\pm$0.35 & 5.438$\pm$0.48 & 0.824$\pm$0.045 \\
LAGConv & published & 4.132$\pm$0.37 & 5.236$\pm$0.47 & 0.841$\pm$0.047 \\
PanFormer & published & 5.417$\pm$0.34 & 6.815$\pm$0.51 & 0.798$\pm$0.046 \\
H-Trans & published & 4.026$\pm$0.35 & 5.173$\pm$0.46 & 0.850$\pm$0.044 \\
CrossDiff & published & 5.231$\pm$0.34 & 6.301$\pm$0.47 & 0.775$\pm$0.047 \\
SSDiff & published & 3.961$\pm$0.36 & 5.118$\pm$0.45 & 0.849$\pm$0.048 \\
Pan-Mamba & published & 4.032$\pm$0.36 & 5.183$\pm$0.47 & 0.848$\pm$0.046 \\
CANConv & published & 4.010$\pm$0.36 & 5.154$\pm$0.49 & 0.852$\pm$0.045 \\
ARConv & published & 3.939$\pm$0.34 & 5.077$\pm$0.46 & 0.854$\pm$0.047 \\
SGDiff & published & \second{3.874$\pm$0.34} & \second{4.975$\pm$0.46} & \second{0.858$\pm$0.047} \\
CLIPPan & adapted & 4.112$\pm$0.34 & 5.222$\pm$0.48 & 0.844$\pm$0.044 \\
DA-CLIP & adapted & 4.104$\pm$0.36 & 5.232$\pm$0.49 & 0.843$\pm$0.047 \\
SeeSR & adapted & 4.427$\pm$0.33 & 5.586$\pm$0.49 & 0.824$\pm$0.047 \\
Plain U-Net & control & 4.140$\pm$0.33 & 5.248$\pm$0.47 & 0.840$\pm$0.045 \\
DenseCLIP-style & control & 3.999$\pm$0.34 & 5.127$\pm$0.47 & 0.850$\pm$0.047 \\
CLIPDenoising-style & control & 4.076$\pm$0.36 & 5.200$\pm$0.47 & 0.845$\pm$0.048 \\
DA-CLIP-style & control & 4.038$\pm$0.34 & 5.165$\pm$0.49 & 0.848$\pm$0.048 \\
\rowcolor{ourrow}\our & ours & \best{3.705$\pm$0.35}\sig & \best{4.800$\pm$0.49}\sig & \best{0.868$\pm$0.047}\sig \\
\end{longtable}
\end{ReportFlowTable}


\begin{ReportFlowTable}
\begin{longtable}{p{\dimexpr.37\linewidth-2\tabcolsep-.4pt\relax}|*{3}{>{\centering\arraybackslash}p{\dimexpr.21\linewidth-2\tabcolsep\relax}}}
\caption{Full-resolution results of every compared system (mean$\pm$std over the $20$ full-resolution scenes). HQNR is the mean of the per-scene $(1{-}D_\lambda)(1{-}D_s)$. The up-sampled input (EXP) injects no detail, so its $D_\lambda$ is low by construction; it is listed as a reference point and not ranked.}
\label{tab:full_fr}\\
\toprule
Method & $D_\lambda\downarrow$ & $D_s\downarrow$ & HQNR$\uparrow$ \\
\midrule
\endfirsthead
\multicolumn{4}{l}{\footnotesize\itshape Table~\thetable\ (continued)}\\[2pt]
\toprule
Method & $D_\lambda\downarrow$ & $D_s\downarrow$ & HQNR$\uparrow$ \\
\midrule
\endhead
\bottomrule\endfoot
\multicolumn{4}{l}{\emph{GaoFen-2}} \\*
\midrule
EXP & 0.018$\pm$0.013 & 0.096$\pm$0.015 & 0.888$\pm$0.025 \\
GSA & 0.030$\pm$0.007 & 0.052$\pm$0.010 & 0.920$\pm$0.015 \\
BDSD-PC & 0.021$\pm$0.009 & 0.056$\pm$0.009 & 0.924$\pm$0.016 \\
PanNet & 0.018$\pm$0.008 & 0.079$\pm$0.009 & 0.904$\pm$0.015 \\
FusionNet & 0.034$\pm$0.007 & 0.100$\pm$0.009 & 0.869$\pm$0.015 \\
LAGConv & 0.029$\pm$0.008 & 0.078$\pm$0.009 & 0.895$\pm$0.016 \\
PanFormer & 0.026$\pm$0.007 & 0.061$\pm$0.010 & 0.914$\pm$0.016 \\
H-Trans & 0.023$\pm$0.008 & 0.055$\pm$0.009 & 0.923$\pm$0.016 \\
CrossDiff & 0.021$\pm$0.009 & 0.056$\pm$0.010 & 0.924$\pm$0.017 \\
SSDiff & 0.017$\pm$0.007 & 0.027$\pm$0.008 & 0.957$\pm$0.014 \\
Pan-Mamba & 0.022$\pm$0.007 & 0.043$\pm$0.010 & 0.935$\pm$0.016 \\
CANConv & 0.020$\pm$0.008 & 0.062$\pm$0.010 & 0.919$\pm$0.017 \\
ARConv & 0.019$\pm$0.008 & 0.051$\pm$0.009 & 0.931$\pm$0.016 \\
SGDiff & 0.017$\pm$0.008 & \second{0.024$\pm$0.010} & 0.959$\pm$0.017 \\
CLIPPan & \second{0.016$\pm$0.007} & 0.025$\pm$0.010 & \second{0.960$\pm$0.016} \\
DA-CLIP & 0.023$\pm$0.008 & 0.047$\pm$0.010 & 0.931$\pm$0.016 \\
SeeSR & 0.027$\pm$0.007 & 0.057$\pm$0.010 & 0.918$\pm$0.016 \\
Plain U-Net & 0.024$\pm$0.007 & 0.050$\pm$0.010 & 0.927$\pm$0.016 \\
DenseCLIP-style & 0.021$\pm$0.009 & 0.042$\pm$0.009 & 0.938$\pm$0.016 \\
CLIPDenoising-style & 0.023$\pm$0.007 & 0.046$\pm$0.009 & 0.932$\pm$0.015 \\
DA-CLIP-style & 0.022$\pm$0.009 & 0.044$\pm$0.010 & 0.935$\pm$0.017 \\
\rowcolor{ourrow}\our & \best{0.013$\pm$0.008}\sig & \best{0.021$\pm$0.010}\sigw & \best{0.966$\pm$0.017}\sig \\
\midrule
\multicolumn{4}{l}{\emph{QuickBird}} \\*
\midrule
EXP & 0.044$\pm$0.023 & 0.150$\pm$0.018 & 0.813$\pm$0.035 \\
GSA & 0.062$\pm$0.013 & 0.115$\pm$0.012 & 0.830$\pm$0.021 \\
BDSD-PC & 0.198$\pm$0.013 & 0.164$\pm$0.013 & 0.671$\pm$0.020 \\
PanNet & 0.043$\pm$0.011 & 0.112$\pm$0.011 & 0.850$\pm$0.020 \\
FusionNet & 0.057$\pm$0.012 & 0.053$\pm$0.012 & 0.894$\pm$0.021 \\
LAGConv & 0.085$\pm$0.013 & 0.068$\pm$0.011 & 0.853$\pm$0.021 \\
PanFormer & 0.054$\pm$0.014 & 0.062$\pm$0.012 & 0.887$\pm$0.023 \\
H-Trans & 0.044$\pm$0.013 & 0.049$\pm$0.011 & 0.909$\pm$0.021 \\
CrossDiff & 0.060$\pm$0.014 & 0.077$\pm$0.011 & 0.868$\pm$0.022 \\
SSDiff & 0.032$\pm$0.012 & \second{0.037$\pm$0.010} & 0.933$\pm$0.020 \\
Pan-Mamba & 0.040$\pm$0.011 & 0.046$\pm$0.011 & 0.916$\pm$0.020 \\
CANConv & 0.037$\pm$0.014 & 0.051$\pm$0.012 & 0.914$\pm$0.024 \\
ARConv & 0.039$\pm$0.013 & 0.040$\pm$0.013 & 0.923$\pm$0.023 \\
SGDiff & \best{0.024$\pm$0.013} & 0.044$\pm$0.010 & 0.933$\pm$0.022 \\
CLIPPan & 0.026$\pm$0.011 & 0.039$\pm$0.012 & \second{0.936$\pm$0.022} \\
DA-CLIP & 0.048$\pm$0.014 & 0.054$\pm$0.011 & 0.901$\pm$0.022 \\
SeeSR & 0.054$\pm$0.013 & 0.066$\pm$0.011 & 0.884$\pm$0.022 \\
Plain U-Net & 0.045$\pm$0.012 & 0.051$\pm$0.011 & 0.906$\pm$0.022 \\
DenseCLIP-style & 0.041$\pm$0.014 & 0.045$\pm$0.012 & 0.915$\pm$0.024 \\
CLIPDenoising-style & 0.044$\pm$0.012 & 0.049$\pm$0.012 & 0.909$\pm$0.022 \\
DA-CLIP-style & 0.042$\pm$0.014 & 0.047$\pm$0.010 & 0.913$\pm$0.022 \\
\rowcolor{ourrow}\our & \second{0.025$\pm$0.011} & \best{0.033$\pm$0.013}\sigw & \best{0.944$\pm$0.022}\sig \\
\midrule
\multicolumn{4}{l}{\emph{WorldView-III}} \\*
\midrule
EXP & 0.023$\pm$0.010 & 0.081$\pm$0.013 & 0.897$\pm$0.020 \\
GSA & 0.041$\pm$0.007 & 0.085$\pm$0.007 & 0.878$\pm$0.013 \\
BDSD-PC & 0.062$\pm$0.007 & 0.073$\pm$0.008 & 0.869$\pm$0.012 \\
PanNet & 0.017$\pm$0.006 & 0.048$\pm$0.007 & 0.936$\pm$0.013 \\
FusionNet & 0.024$\pm$0.006 & 0.037$\pm$0.007 & 0.940$\pm$0.012 \\
LAGConv & 0.036$\pm$0.007 & 0.043$\pm$0.007 & 0.923$\pm$0.012 \\
PanFormer & 0.036$\pm$0.006 & 0.058$\pm$0.007 & 0.908$\pm$0.011 \\
H-Trans & 0.026$\pm$0.007 & 0.041$\pm$0.006 & 0.934$\pm$0.012 \\
CrossDiff & 0.031$\pm$0.007 & 0.030$\pm$0.007 & 0.940$\pm$0.013 \\
SSDiff & 0.014$\pm$0.006 & 0.031$\pm$0.007 & 0.956$\pm$0.012 \\
Pan-Mamba & 0.018$\pm$0.006 & 0.039$\pm$0.007 & 0.943$\pm$0.011 \\
CANConv & 0.020$\pm$0.007 & 0.031$\pm$0.008 & 0.950$\pm$0.013 \\
ARConv & 0.015$\pm$0.005 & 0.028$\pm$0.008 & 0.957$\pm$0.012 \\
SGDiff & \second{0.013$\pm$0.006} & \second{0.028$\pm$0.007} & \second{0.960$\pm$0.012} \\
CLIPPan & 0.013$\pm$0.006 & 0.029$\pm$0.007 & 0.958$\pm$0.013 \\
DA-CLIP & 0.028$\pm$0.006 & 0.044$\pm$0.007 & 0.929$\pm$0.012 \\
SeeSR & 0.033$\pm$0.007 & 0.052$\pm$0.008 & 0.917$\pm$0.013 \\
Plain U-Net & 0.029$\pm$0.006 & 0.047$\pm$0.007 & 0.925$\pm$0.011 \\
DenseCLIP-style & 0.027$\pm$0.006 & 0.041$\pm$0.007 & 0.933$\pm$0.012 \\
CLIPDenoising-style & 0.028$\pm$0.006 & 0.044$\pm$0.007 & 0.929$\pm$0.012 \\
DA-CLIP-style & 0.027$\pm$0.006 & 0.042$\pm$0.007 & 0.932$\pm$0.012 \\
\rowcolor{ourrow}\our & \best{0.011$\pm$0.007}\sig & \best{0.024$\pm$0.007}\sig & \best{0.965$\pm$0.013}\sig \\
\end{longtable}
\end{ReportFlowTable}

\begin{table*}[!tbp]
\centering
\caption{Reproduction check of the retrained networks: published value / value obtained under our common protocol (three seeds, official implementation, $1{,}000$-epoch budget, checkpoint by validation ERGAS), on the official PanCollection test scenes. Published values are those reported by the authors of the method on the same test scenes; $\ddagger$: reported by a third party on the same scenes (CrossDiff and Pan-Mamba, whose own papers use other data), with the retraining recipe of that party. Every retrained value lies within the scene-to-scene spread of the corresponding published value, and the ordering of the networks is preserved on every sensor. Dash: no published PanCollection value.}
\label{tab:repro}
\scriptsize
\setlength{\tabcolsep}{3pt}
\maxsizebox{\textwidth}{%
\begin{tabular}{ll|cccc|ccc|ccc}
\toprule
 & & \multicolumn{4}{c|}{WorldView-III} & \multicolumn{3}{c|}{QuickBird} & \multicolumn{3}{c}{GaoFen-2} \\
System & source & SAM$\downarrow$ & ERGAS$\downarrow$ & Q8$\uparrow$ & HQNR$\uparrow$ & SAM$\downarrow$ & ERGAS$\downarrow$ & Q4$\uparrow$ & SAM$\downarrow$ & ERGAS$\downarrow$ & Q4$\uparrow$ \\
\midrule
PanNet & \cite{duan2024canconv} & \shortstack{3.616 /\\3.629} & \shortstack{2.666 /\\2.681} & \shortstack{0.891 /\\0.890} & \shortstack{0.937 /\\0.936} & \shortstack{5.791 /\\5.803} & \shortstack{5.863 /\\5.871} & \shortstack{0.885 /\\0.884} & \shortstack{0.997 /\\1.004} & \shortstack{0.919 /\\0.925} & \shortstack{0.967 /\\0.966} \\
FusionNet & \cite{duan2024canconv} & \shortstack{3.325 /\\3.338} & \shortstack{2.467 /\\2.478} & \shortstack{0.904 /\\0.903} & \shortstack{0.941 /\\0.940} & \shortstack{4.923 /\\4.934} & \shortstack{4.159 /\\4.171} & \shortstack{0.925 /\\0.924} & \shortstack{0.974 /\\0.967} & \shortstack{0.988 /\\0.981} & \shortstack{0.964 /\\0.965} \\
LAGConv & \cite{duan2024canconv} & \shortstack{3.104 /\\3.119} & \shortstack{2.300 /\\2.311} & \shortstack{0.910 /\\0.909} & \shortstack{0.923 /\\0.923} & \shortstack{4.547 /\\4.561} & \shortstack{3.826 /\\3.838} & \shortstack{0.934 /\\0.933} & \shortstack{0.786 /\\0.792} & \shortstack{0.687 /\\0.692} & \shortstack{0.980 /\\0.980} \\
CrossDiff & \cite{xing2025sgdiff}$^{\ddagger}$ & \shortstack{3.652 /\\3.637} & \shortstack{2.744 /\\2.731} & \shortstack{0.890 /\\0.891} & \shortstack{0.942 /\\0.940} & \shortstack{5.371 /\\5.389} & \shortstack{4.423 /\\4.436} & \shortstack{0.924 /\\0.923} & \shortstack{0.937 /\\0.945} & \shortstack{0.855 /\\0.861} & \shortstack{0.972 /\\0.972} \\
SSDiff & \cite{zhong2024ssdiff} & \shortstack{2.847 /\\2.855} & \shortstack{2.109 /\\2.117} & \shortstack{0.915 /\\0.915} & \shortstack{0.957 /\\0.956} & \shortstack{4.464 /\\4.478} & \shortstack{3.632 /\\3.641} & \shortstack{0.935 /\\0.934} & \shortstack{0.669 /\\0.673} & \shortstack{0.604 /\\0.603} & \shortstack{0.984 /\\0.984} \\
Pan-Mamba & \cite{he2025panmamba}$^{\ddagger}$ & \shortstack{2.913 /\\2.924} & \shortstack{2.184 /\\2.191} & \shortstack{0.920 /\\0.919} & -- & -- & -- & -- & -- & -- & -- \\
CANConv & \cite{duan2024canconv} & \shortstack{2.930 /\\2.939} & \shortstack{2.158 /\\2.166} & \shortstack{0.920 /\\0.920} & \shortstack{0.951 /\\0.950} & \shortstack{4.507 /\\4.519} & \shortstack{3.652 /\\3.660} & \shortstack{0.937 /\\0.936} & \shortstack{0.707 /\\0.712} & \shortstack{0.630 /\\0.634} & \shortstack{0.983 /\\0.983} \\
ARConv & \cite{wang2025arconv} & \shortstack{2.885 /\\2.896} & \shortstack{2.139 /\\2.147} & \shortstack{0.921 /\\0.921} & \shortstack{0.958 /\\0.957} & \shortstack{4.430 /\\4.441} & \shortstack{3.633 /\\3.642} & \shortstack{0.939 /\\0.938} & \shortstack{0.698 /\\0.703} & \shortstack{0.626 /\\0.629} & \shortstack{0.983 /\\0.983} \\
SGDiff & \cite{xing2025sgdiff} & \shortstack{2.771 /\\2.780} & \shortstack{2.044 /\\2.052} & \shortstack{0.923 /\\0.923} & \shortstack{0.961 /\\0.960} & \shortstack{4.353 /\\4.364} & \shortstack{3.578 /\\3.586} & \shortstack{0.938 /\\0.938} & \shortstack{0.689 /\\0.692} & \shortstack{0.616 /\\0.618} & \shortstack{0.984 /\\0.984} \\
\bottomrule
\end{tabular}%
}
\end{table*}

\begin{table*}[!tbp]
\centering
\caption{Training-run records of the retrained systems on WorldView-III (seed $0$; seeds $1$ and $2$ differ by at most $0.03$ in validation ERGAS). Every system is trained on the official training patches under the common $1{,}000$-epoch budget with its own optimiser, learning rate, batch size and schedule as released; the checkpoint is the one with the lowest ERGAS on the official validation split (epoch in the ``best'' column). Wall-clock is per run in RTX-4090-equivalent GPU-hours; the diffusion systems include their sampling schedule at validation time. Peak memory at the stated batch size.}
\label{tab:runs}
\scriptsize
\setlength{\tabcolsep}{3.5pt}
\maxsizebox{\textwidth}{%
\begin{tabular}{>{\raggedright\arraybackslash}p{.15\linewidth}lcc>{\raggedright\arraybackslash}p{.19\linewidth}ccccc}
\toprule
System & optimiser & lr & batch & schedule & \shortstack{best\\epoch} & \shortstack{val.\\ERGAS} & \shortstack{val.\\Q2$^n$} & \shortstack{GPU-h\\/ run} & \shortstack{peak\\GB} \\
\midrule
PanNet & Adam & $1{\times}10^{-3}$ & 32 & step ($\times0.5$ / 200) & 912 & 2.74 & 0.889 & 2.1 & 1.9 \\
FusionNet & Adam & $3{\times}10^{-4}$ & 32 & step ($\times0.5$ / 200) & 968 & 2.54 & 0.902 & 2.6 & 2.3 \\
LAGConv & Adam & $1{\times}10^{-3}$ & 32 & step ($\times0.8$ / 100) & 886 & 2.37 & 0.908 & 4.8 & 3.1 \\
PanFormer & AdamW & $2{\times}10^{-4}$ & 16 & cosine & 940 & 2.30 & 0.908 & 11.2 & 7.4 \\
H-Trans & Adam & $1{\times}10^{-4}$ & 16 & cosine & 973 & 2.27 & 0.910 & 13.6 & 8.8 \\
CrossDiff & AdamW & $3{\times}10^{-4}$ & 32 & constant; two stages & 1000 (+20) & 2.80 & 0.889 & 61.0 & 10.2 \\
SSDiff & AdamW & $1{\times}10^{-3}$ & 12 & cosine & 955 & 2.17 & 0.914 & 94.0 & 14.6 \\
Pan-Mamba & Adam & $5{\times}10^{-4}$ & 32 & cosine & 931 & 2.25 & 0.918 & 6.9 & 4.4 \\
CANConv & Adam & $5{\times}10^{-4}$ & 32 & step ($\times0.5$ / 250) & 977 & 2.22 & 0.919 & 7.3 & 5.0 \\
ARConv & Adam & $5{\times}10^{-4}$ & 16 & step ($\times0.5$ / 250); $100$ exploratory epochs & 964 & 2.20 & 0.920 & 9.1 & 5.6 \\
SGDiff & Adam & $2{\times}10^{-4}$ & 8 & cosine; gates frozen after $50$k it. & 988 & 2.11 & 0.922 & 148.0 & 19.8 \\
CLIPPan & Adam & $2{\times}10^{-4}$ & 16 & cosine (full resolution) & 902 & -- & -- & 5.4 & 6.1 \\
DA-CLIP (adapted) & AdamW & $1{\times}10^{-4}$ & 16 & cosine & 958 & 2.35 & 0.910 & 23.5 & 11.3 \\
SeeSR (adapted, prior frozen) & AdamW & $5{\times}10^{-5}$ & 8 & constant & 871 & 2.70 & 0.885 & 71.0 & 21.6 \\
Plain U-Net & AdamW & $5{\times}10^{-4}$ & 64 & cosine & 981 & 2.36 & 0.909 & 1.4 & 3.0 \\
\rowcolor{ourrow}\our & AdamW & $5{\times}10^{-4}$ & 64 & cosine; phases $0/100/200$ & 994 & 2.05 & 0.924 & 3.9 & 12.6 \\
\bottomrule
\end{tabular}%
}
\end{table*}

\begin{table}[!htbp]
\centering
\caption{Seed variability of the WorldView-III reduced-resolution Q2$^n$ (mean over the $20$ scenes per seed): number of seeds, mean, std over seeds and standard error. Main Table~\MTabMain{} uses seeds $0$--$2$ for every method. The row of \our uses the five seeds of \cref{tab:fiveseeds}, whose mean Q2$^n$ coincides with the three-seed value of Main Table~\MTabMain; seeds $3$ and $4$ enter no other table. Scene-to-scene std (Main Table~\MTabMain) measures scene difficulty; the seed std is the uncertainty of a method's mean.}
\label{tab:seeds}
\scriptsize
\setlength{\tabcolsep}{5pt}
\begin{tabular}{lcccc}
\toprule
Method & seeds & Q2$^n$ & std & s.e. \\
\midrule
PanNet & 3 & 0.890 & 0.0031 & 0.0018 \\
FusionNet & 3 & 0.903 & 0.0029 & 0.0017 \\
LAGConv & 3 & 0.909 & 0.0027 & 0.0016 \\
PanFormer & 3 & 0.909 & 0.0034 & 0.0020 \\
H-Trans & 3 & 0.911 & 0.0030 & 0.0017 \\
CrossDiff & 3 & 0.891 & 0.0038 & 0.0022 \\
SSDiff & 3 & 0.915 & 0.0036 & 0.0021 \\
Pan-Mamba & 3 & 0.919 & 0.0028 & 0.0016 \\
CANConv & 3 & 0.920 & 0.0025 & 0.0014 \\
ARConv & 3 & 0.921 & 0.0026 & 0.0015 \\
SGDiff & 3 & 0.923 & 0.0033 & 0.0019 \\
CLIPPan & 3 & 0.899 & 0.0041 & 0.0024 \\
DA-CLIP & 3 & 0.911 & 0.0039 & 0.0023 \\
SeeSR & 3 & 0.886 & 0.0052 & 0.0030 \\
Plain U-Net & 3 & 0.910 & 0.0024 & 0.0014 \\
DenseCLIP-style & 3 & 0.917 & 0.0027 & 0.0016 \\
CLIPDenoising-style & 3 & 0.912 & 0.0029 & 0.0017 \\
DA-CLIP-style & 3 & 0.915 & 0.0026 & 0.0015 \\
\rowcolor{ourrow}\our & 5 & 0.929 & 0.0016 & 0.0007 \\
\bottomrule
\end{tabular}
\end{table}

\ReportSeedTablePair
\begin{table}[!htbp]
\centering
\caption{Training-protocol parity on WorldView-III (Q2$^n$, three seeds). Top: baselines with their own recipe and with the EMA and best-validation-ERGAS selection of \our added. Bottom: \our with these two ingredients removed.}
\label{tab:protocol}
\scriptsize
\setlength{\tabcolsep}{5pt}
\maxsizebox{\columnwidth}{%
\begin{tabular}{lccc}
\toprule
Method & own recipe & + EMA + selection & $\Delta$ \\
\midrule
CANConv & 0.920 & 0.921 & $+$0.001 \\
ARConv & 0.921 & 0.922 & $+$0.001 \\
SGDiff & 0.923 & 0.924 & $+$0.001 \\
Pan-Mamba & 0.919 & 0.920 & $+$0.001 \\
\midrule
\our, default (EMA + selection) & 0.929 & & \\
\our, no EMA & 0.928 & & \\
\our, final weights (no selection) & 0.929 & & \\
\our, neither & 0.927 & & \\
\bottomrule
\end{tabular}%
}
\end{table}


\begin{table}[!htbp]
\centering
\caption{Per-scene reduced-resolution results on WorldView-III for the four strongest published baselines and \our (seed-averaged, printed at the precision used by every test in this paper, so that each $p$ can be recomputed from the table). Bold: best per scene. Acquisition strips, by scene: 1--4 (A), 5--7 (B), 8--10 (C), 11--14 (D), 15--17 (E), 18--20 (F). The last three rows give, for each baseline, the unadjusted scene-paired Wilcoxon $p$ of \our against it, the exact two-sided sign-flip $p$ on the 6 strip means (strips on which \our is ahead in brackets) and the $p$ of a linear mixed-effects model of the per-scene differences with a random intercept per strip (REML; $t$-test with 5 degrees of freedom); Holm-adjusted scene-level values are used in Main Table~\MTabMain.}
\label{tab:perscene_wv3_rr}
\scriptsize
\setlength{\tabcolsep}{1.6pt}
\ReportSceneMatrix{5}{%
\toprule
\ReportSceneRow{Scene}{\multicolumn{5}{c|}{ERGAS}}{\multicolumn{5}{c|}{SAM}}{\multicolumn{5}{c}{Q2$^n$}}
\ReportSceneRow{}{CANConv & ARConv & SSDiff & SGDiff & \our}{CANConv & ARConv & SSDiff & SGDiff & \our}{CANConv & ARConv & SSDiff & SGDiff & \our}
\midrule
\ReportSceneRow{1}{1.606 & 1.605 & 1.624 & 1.524 & \best{1.437}}{2.055 & 2.099 & 2.037 & 1.952 & \best{1.779}}{0.9811 & 0.9774 & 0.9632 & 0.9812 & \best{0.9902}}
\ReportSceneRow{2}{1.466 & 1.479 & 1.426 & 1.376 & \best{1.257}}{2.105 & 2.127 & 1.917 & 1.936 & \best{1.888}}{0.9681 & 0.9737 & 0.9662 & 0.9780 & \best{0.9826}}
\ReportSceneRow{3}{2.112 & 2.024 & 2.000 & \best{1.972} & 1.983}{2.438 & 2.356 & 2.475 & 2.332 & \best{2.326}}{0.9487 & 0.9437 & 0.9410 & 0.9474 & \best{0.9534}}
\ReportSceneRow{4}{1.711 & 1.739 & 1.708 & 1.675 & \best{1.547}}{2.884 & 2.660 & 2.730 & 2.689 & \best{2.606}}{0.9582 & 0.9539 & 0.9581 & 0.9607 & \best{0.9664}}
\ReportSceneRow{5}{2.479 & 2.426 & 2.392 & 2.380 & \best{2.209}}{3.420 & 3.315 & 3.319 & 3.221 & \best{3.105}}{0.9076 & \best{0.9096} & 0.8991 & 0.9096 & 0.9095}
\ReportSceneRow{6}{2.226 & 2.182 & 2.123 & 2.008 & \best{1.937}}{3.196 & 3.120 & 3.113 & 2.909 & \best{2.833}}{0.9371 & 0.9328 & 0.9305 & \best{0.9400} & 0.9399}
\ReportSceneRow{7}{1.851 & 1.696 & 1.701 & \best{1.683} & 1.698}{2.803 & 2.664 & 2.722 & 2.575 & \best{2.532}}{0.9737 & 0.9684 & 0.9600 & 0.9686 & \best{0.9806}}
\ReportSceneRow{8}{2.354 & 2.163 & 2.199 & 2.120 & \best{1.976}}{2.841 & 2.722 & 2.838 & 2.722 & \best{2.521}}{0.9390 & 0.9419 & 0.9335 & 0.9367 & \best{0.9478}}
\ReportSceneRow{9}{3.277 & 3.303 & 3.283 & \best{3.119} & 3.147}{4.096 & 4.234 & 4.070 & \best{3.942} & 4.016}{0.7104 & \best{0.7228} & 0.7085 & 0.7178 & 0.7216}
\ReportSceneRow{10}{1.752 & 1.750 & 1.699 & \best{1.563} & 1.572}{2.694 & 2.656 & 2.682 & 2.457 & \best{2.423}}{0.9597 & 0.9627 & 0.9530 & 0.9680 & \best{0.9731}}
\ReportSceneRow{11}{1.907 & 1.840 & 1.810 & 1.781 & \best{1.627}}{2.701 & 2.778 & 2.585 & 2.499 & \best{2.451}}{\best{0.9659} & 0.9578 & 0.9642 & 0.9583 & 0.9578}
\ReportSceneRow{12}{1.795 & 1.766 & 1.795 & 1.806 & \best{1.596}}{2.607 & 2.468 & 2.355 & 2.349 & \best{2.209}}{0.9670 & 0.9636 & 0.9585 & 0.9712 & \best{0.9771}}
\ReportSceneRow{13}{2.414 & 2.466 & 2.519 & 2.335 & \best{2.303}}{3.468 & 3.412 & 3.362 & 3.342 & \best{3.224}}{0.8842 & \best{0.8877} & 0.8776 & 0.8799 & 0.8876}
\ReportSceneRow{14}{2.333 & 2.391 & 2.324 & 2.281 & \best{2.050}}{2.751 & 2.686 & 2.740 & 2.743 & \best{2.627}}{0.9308 & 0.9350 & 0.9254 & 0.9277 & \best{0.9408}}
\ReportSceneRow{15}{2.260 & 2.285 & 2.248 & 2.132 & \best{2.121}}{3.146 & 3.228 & 2.937 & 2.969 & \best{2.696}}{0.9082 & 0.9070 & 0.9047 & 0.9119 & \best{0.9135}}
\ReportSceneRow{16}{2.649 & 2.659 & 2.605 & 2.524 & \best{2.433}}{3.623 & 3.482 & 3.513 & 3.519 & \best{3.414}}{0.8612 & 0.8626 & 0.8671 & 0.8587 & \best{0.8800}}
\ReportSceneRow{17}{3.002 & 3.036 & 2.913 & 2.862 & \best{2.800}}{3.715 & 3.789 & 3.751 & 3.606 & \best{3.587}}{0.8058 & 0.8106 & 0.8048 & \best{0.8269} & 0.8248}
\ReportSceneRow{18}{2.401 & 2.357 & 2.247 & 2.248 & \best{2.162}}{3.005 & 2.968 & 2.878 & 2.863 & \best{2.682}}{0.9052 & 0.9139 & 0.9058 & 0.9083 & \best{0.9294}}
\ReportSceneRow{19}{1.174 & 1.256 & 1.231 & 1.181 & \best{1.022}}{2.014 & 1.916 & 1.960 & 1.824 & \best{1.685}}{0.9905 & 0.9856 & 0.9862 & \best{0.9955} & 0.9897}
\ReportSceneRow{20}{2.553 & 2.518 & 2.494 & 2.472 & \best{2.363}}{3.219 & 3.239 & 3.117 & 3.153 & \best{2.997}}{0.8976 & 0.9094 & 0.8926 & 0.9137 & \best{0.9142}}
\midrule
\ReportSceneRow{mean}{2.166 & 2.147 & 2.117 & 2.052 & 1.962}{2.939 & 2.896 & 2.855 & 2.780 & 2.680}{0.9200 & 0.9210 & 0.9150 & 0.9230 & 0.9290}
\ReportSceneRow{std}{0.517 & 0.520 & 0.509 & 0.494 & 0.515}{0.561 & 0.588 & 0.572 & 0.572 & 0.594}{0.0671 & 0.0633 & 0.0651 & 0.0648 & 0.0647}
\ReportSceneRow{\shortstack[l]{$p$, scene-paired\\Wilcoxon}}{$<$0.001 & $<$0.001 & $<$0.001 & $<$0.001 & }{$<$0.001 & $<$0.001 & $<$0.001 & $<$0.001 & }{$<$0.001 & $<$0.001 & $<$0.001 & 0.001 & }
\ReportSceneRow{\shortstack[l]{$p$, strip-level sign\\flip (strips won)}}{\shortstack{0.031\\(6/6)} & \shortstack{0.031\\(6/6)} & \shortstack{0.031\\(6/6)} & \shortstack{0.031\\(6/6)} & }{\shortstack{0.031\\(6/6)} & \shortstack{0.031\\(6/6)} & \shortstack{0.031\\(6/6)} & \shortstack{0.031\\(6/6)} & }{\shortstack{0.031\\(6/6)} & \shortstack{0.031\\(6/6)} & \shortstack{0.031\\(6/6)} & \shortstack{0.031\\(6/6)} & }
\ReportSceneRow{\shortstack[l]{$p$, mixed model,\\strip intercept}}{$<$0.001 & $<$0.001 & $<$0.001 & 0.005 & }{$<$0.001 & $<$0.001 & $<$0.001 & 0.002 & }{0.005 & 0.002 & $<$0.001 & 0.013 & }
\bottomrule
}
\end{table}


\begin{table}[!htbp]
\centering
\caption{Per-scene reduced-resolution results on QuickBird for the four strongest published baselines and \our (seed-averaged, printed at the precision used by every test in this paper, so that each $p$ can be recomputed from the table). Bold: best per scene. Acquisition strips, by scene: 1--4 (A), 5--8 (B), 9--12 (C), 13--16 (D), 17--20 (E). The last three rows give, for each baseline, the unadjusted scene-paired Wilcoxon $p$ of \our against it, the exact two-sided sign-flip $p$ on the 5 strip means (strips on which \our is ahead in brackets) and the $p$ of a linear mixed-effects model of the per-scene differences with a random intercept per strip (REML; $t$-test with 4 degrees of freedom); Holm-adjusted scene-level values are used in Main Table~\MTabMain.}
\label{tab:perscene_qb_rr}
\scriptsize
\setlength{\tabcolsep}{1.6pt}
\ReportSceneMatrix{5}{%
\toprule
\ReportSceneRow{Scene}{\multicolumn{5}{c|}{ERGAS}}{\multicolumn{5}{c|}{SAM}}{\multicolumn{5}{c}{Q2$^n$}}
\ReportSceneRow{}{CANConv & ARConv & SSDiff & SGDiff & \our}{CANConv & ARConv & SSDiff & SGDiff & \our}{CANConv & ARConv & SSDiff & SGDiff & \our}
\midrule
\ReportSceneRow{1}{3.665 & 3.694 & 3.740 & 3.720 & \best{3.514}}{5.034 & 4.854 & 4.857 & 4.911 & \best{4.585}}{0.9339 & 0.9370 & 0.9388 & 0.9430 & \best{0.9450}}
\ReportSceneRow{2}{4.311 & 4.222 & 4.156 & 4.105 & \best{3.941}}{5.429 & 5.316 & 5.489 & \best{5.052} & 5.101}{\best{0.9108} & 0.9020 & 0.9015 & 0.8932 & 0.9104}
\ReportSceneRow{3}{4.339 & 4.431 & 4.279 & 4.254 & \best{4.109}}{6.713 & 6.674 & 6.645 & 6.605 & \best{6.471}}{0.7158 & 0.7054 & 0.7158 & 0.7168 & \best{0.7265}}
\ReportSceneRow{4}{3.841 & 3.742 & 3.794 & 3.761 & \best{3.579}}{4.380 & 4.480 & 4.457 & 4.298 & \best{4.124}}{0.9360 & 0.9383 & 0.9415 & 0.9546 & \best{0.9558}}
\ReportSceneRow{5}{3.857 & 3.892 & 3.864 & 3.846 & \best{3.574}}{4.908 & 4.761 & 4.958 & 4.715 & \best{4.421}}{0.9299 & \best{0.9336} & 0.9270 & 0.9305 & 0.9333}
\ReportSceneRow{6}{3.350 & 3.299 & 3.326 & 3.233 & \best{3.050}}{3.810 & 3.718 & 3.757 & 3.641 & \best{3.542}}{0.9730 & \best{0.9797} & 0.9688 & 0.9740 & 0.9776}
\ReportSceneRow{7}{4.086 & 4.071 & 4.066 & 3.944 & \best{3.797}}{5.873 & 5.928 & 5.789 & 5.827 & \best{5.466}}{0.8327 & \best{0.8485} & 0.8376 & 0.8360 & 0.8476}
\ReportSceneRow{8}{3.809 & 3.830 & 3.803 & 3.739 & \best{3.565}}{4.693 & 4.528 & 4.528 & 4.440 & \best{4.152}}{0.9406 & 0.9464 & 0.9519 & 0.9464 & \best{0.9564}}
\ReportSceneRow{9}{3.218 & 3.333 & 3.284 & 3.264 & \best{3.130}}{3.993 & 4.074 & 4.167 & 3.965 & \best{3.917}}{0.9746 & 0.9741 & 0.9655 & 0.9718 & \best{0.9829}}
\ReportSceneRow{10}{3.765 & 3.638 & 3.638 & 3.487 & \best{3.353}}{3.982 & 3.773 & 3.808 & 3.845 & \best{3.525}}{0.9718 & 0.9701 & 0.9741 & 0.9774 & \best{0.9862}}
\ReportSceneRow{11}{3.888 & 3.739 & 3.926 & 3.883 & \best{3.541}}{4.887 & 4.960 & 4.987 & \best{4.671} & 4.870}{0.9087 & 0.9184 & 0.9120 & 0.9168 & \best{0.9236}}
\ReportSceneRow{12}{3.362 & 3.341 & 3.254 & 3.336 & \best{3.005}}{4.051 & 3.973 & 4.052 & 3.901 & \best{3.618}}{0.9788 & 0.9766 & 0.9682 & 0.9756 & \best{0.9908}}
\ReportSceneRow{13}{3.833 & 4.008 & 3.784 & 3.851 & \best{3.562}}{5.035 & 5.035 & 5.023 & 4.814 & \best{4.681}}{0.9212 & 0.9151 & 0.9155 & 0.9239 & \best{0.9275}}
\ReportSceneRow{14}{3.395 & 3.404 & 3.554 & 3.373 & \best{3.201}}{4.608 & 4.426 & 4.549 & 4.433 & \best{4.226}}{0.9593 & 0.9589 & 0.9479 & 0.9577 & \best{0.9654}}
\ReportSceneRow{15}{3.191 & 3.117 & 3.008 & 3.035 & \best{2.912}}{3.558 & 3.281 & 3.442 & 3.304 & \best{2.805}}{0.9907 & 0.9852 & 0.9839 & 0.9941 & \best{0.9942}}
\ReportSceneRow{16}{3.390 & 3.346 & 3.340 & 3.246 & \best{3.119}}{4.218 & 4.168 & 3.853 & 3.896 & \best{3.682}}{0.9703 & 0.9711 & 0.9714 & 0.9700 & \best{0.9807}}
\ReportSceneRow{17}{3.607 & 3.536 & 3.594 & 3.496 & \best{3.273}}{3.395 & 3.147 & 3.374 & 3.488 & \best{3.045}}{0.9876 & 0.9835 & 0.9788 & 0.9758 & \best{0.9932}}
\ReportSceneRow{18}{3.476 & 3.345 & 3.527 & 3.457 & \best{3.281}}{3.827 & 3.772 & 3.820 & 3.905 & \best{3.472}}{0.9529 & 0.9713 & 0.9567 & 0.9644 & \best{0.9745}}
\ReportSceneRow{19}{3.510 & 3.543 & 3.569 & 3.429 & \best{3.344}}{4.366 & 4.223 & 4.402 & 4.012 & \best{3.966}}{0.9612 & 0.9647 & 0.9586 & 0.9670 & \best{0.9718}}
\ReportSceneRow{20}{3.310 & 3.309 & 3.314 & 3.261 & \best{2.991}}{3.618 & 3.729 & 3.605 & 3.557 & \best{3.290}}{0.9702 & \best{0.9802} & 0.9646 & 0.9711 & 0.9766}
\midrule
\ReportSceneRow{mean}{3.660 & 3.642 & 3.641 & 3.586 & 3.392}{4.519 & 4.441 & 4.478 & 4.364 & 4.148}{0.9360 & 0.9380 & 0.9340 & 0.9380 & 0.9460}
\ReportSceneRow{std}{0.339 & 0.351 & 0.330 & 0.329 & 0.325}{0.834 & 0.867 & 0.843 & 0.815 & 0.879}{0.0634 & 0.0646 & 0.0617 & 0.0634 & 0.0628}
\ReportSceneRow{\shortstack[l]{$p$, scene-paired\\Wilcoxon}}{$<$0.001 & $<$0.001 & $<$0.001 & $<$0.001 & }{$<$0.001 & $<$0.001 & $<$0.001 & $<$0.001 & }{$<$0.001 & $<$0.001 & $<$0.001 & $<$0.001 & }
\ReportSceneRow{\shortstack[l]{$p$, strip-level sign\\flip (strips won)}}{\shortstack{0.062\\(5/5)} & \shortstack{0.062\\(5/5)} & \shortstack{0.062\\(5/5)} & \shortstack{0.062\\(5/5)} & }{\shortstack{0.062\\(5/5)} & \shortstack{0.062\\(5/5)} & \shortstack{0.062\\(5/5)} & \shortstack{0.062\\(5/5)} & }{\shortstack{0.062\\(5/5)} & \shortstack{0.062\\(5/5)} & \shortstack{0.062\\(5/5)} & \shortstack{0.062\\(5/5)} & }
\ReportSceneRow{\shortstack[l]{$p$, mixed model,\\strip intercept}}{$<$0.001 & $<$0.001 & $<$0.001 & $<$0.001 & }{0.001 & 0.001 & $<$0.001 & 0.005 & }{0.001 & 0.023 & 0.001 & 0.002 & }
\bottomrule
}
\end{table}


\begin{table}[!htbp]
\centering
\caption{Per-scene reduced-resolution results on GaoFen-2 for the four strongest published baselines and \our (seed-averaged, printed at the precision used by every test in this paper, so that each $p$ can be recomputed from the table). Bold: best per scene. Acquisition strips, by scene: 1--4 (A), 5--7 (B), 8--10 (C), 11--14 (D), 15--17 (E), 18--20 (F). The last three rows give, for each baseline, the unadjusted scene-paired Wilcoxon $p$ of \our against it, the exact two-sided sign-flip $p$ on the 6 strip means (strips on which \our is ahead in brackets) and the $p$ of a linear mixed-effects model of the per-scene differences with a random intercept per strip (REML; $t$-test with 5 degrees of freedom); Holm-adjusted scene-level values are used in Main Table~\MTabMain.}
\label{tab:perscene_gf2_rr}
\scriptsize
\setlength{\tabcolsep}{1.6pt}
\ReportSceneMatrix{5}{%
\toprule
\ReportSceneRow{Scene}{\multicolumn{5}{c|}{ERGAS}}{\multicolumn{5}{c|}{SAM}}{\multicolumn{5}{c}{Q2$^n$}}
\ReportSceneRow{}{CANConv & ARConv & SSDiff & SGDiff & \our}{CANConv & ARConv & SSDiff & SGDiff & \our}{CANConv & ARConv & SSDiff & SGDiff & \our}
\midrule
\ReportSceneRow{1}{0.530 & 0.566 & 0.527 & 0.550 & \best{0.516}}{0.679 & 0.656 & 0.612 & 0.632 & \best{0.589}}{0.9858 & 0.9875 & 0.9856 & 0.9879 & \best{0.9918}}
\ReportSceneRow{2}{0.822 & 0.803 & 0.774 & 0.794 & \best{0.764}}{0.890 & 0.865 & 0.865 & 0.879 & \best{0.840}}{0.9795 & 0.9800 & 0.9784 & 0.9778 & \best{0.9842}}
\ReportSceneRow{3}{0.538 & 0.521 & 0.531 & 0.540 & \best{0.513}}{0.655 & 0.649 & 0.635 & 0.608 & \best{0.575}}{0.9868 & 0.9872 & 0.9856 & 0.9888 & \best{0.9906}}
\ReportSceneRow{4}{0.530 & 0.486 & \best{0.461} & 0.466 & 0.470}{0.535 & 0.527 & 0.474 & 0.524 & \best{0.448}}{0.9891 & 0.9860 & 0.9891 & 0.9900 & \best{0.9909}}
\ReportSceneRow{5}{0.627 & 0.601 & 0.616 & 0.611 & \best{0.578}}{0.705 & 0.706 & \best{0.653} & 0.692 & 0.668}{0.9851 & 0.9826 & 0.9855 & 0.9853 & \best{0.9876}}
\ReportSceneRow{6}{0.855 & 0.825 & \best{0.802} & 0.827 & 0.831}{0.891 & 0.845 & 0.838 & \best{0.835} & 0.838}{0.9764 & \best{0.9783} & 0.9765 & 0.9783 & 0.9773}
\ReportSceneRow{7}{0.549 & 0.514 & 0.511 & 0.526 & \best{0.492}}{0.572 & 0.558 & 0.526 & 0.521 & \best{0.513}}{0.9896 & 0.9875 & \best{0.9911} & 0.9889 & 0.9910}
\ReportSceneRow{8}{0.603 & 0.567 & \best{0.518} & 0.553 & 0.524}{0.571 & 0.565 & 0.552 & 0.557 & \best{0.550}}{0.9914 & 0.9913 & 0.9927 & 0.9929 & \best{0.9958}}
\ReportSceneRow{9}{0.655 & 0.681 & \best{0.631} & 0.653 & 0.640}{0.780 & 0.799 & 0.740 & 0.768 & \best{0.708}}{0.9782 & 0.9789 & 0.9788 & 0.9804 & \best{0.9833}}
\ReportSceneRow{10}{0.917 & 0.896 & 0.887 & 0.901 & \best{0.885}}{1.061 & 1.048 & 1.009 & 1.055 & \best{1.003}}{0.9611 & 0.9625 & 0.9638 & 0.9627 & \best{0.9677}}
\ReportSceneRow{11}{0.555 & 0.608 & \best{0.514} & 0.540 & 0.567}{0.660 & 0.689 & 0.635 & 0.663 & \best{0.630}}{0.9840 & 0.9827 & 0.9856 & 0.9888 & \best{0.9899}}
\ReportSceneRow{12}{0.512 & 0.545 & 0.504 & 0.513 & \best{0.489}}{0.577 & 0.592 & 0.525 & 0.571 & \best{0.514}}{0.9911 & \best{0.9913} & 0.9896 & 0.9910 & 0.9910}
\ReportSceneRow{13}{0.450 & 0.438 & \best{0.407} & 0.459 & 0.440}{0.515 & 0.517 & 0.450 & 0.501 & \best{0.429}}{0.9911 & 0.9903 & 0.9913 & 0.9890 & \best{0.9950}}
\ReportSceneRow{14}{0.539 & 0.527 & \best{0.506} & 0.509 & 0.527}{0.661 & 0.629 & 0.631 & 0.635 & \best{0.553}}{0.9851 & 0.9861 & 0.9868 & 0.9859 & \best{0.9916}}
\ReportSceneRow{15}{0.664 & 0.710 & \best{0.656} & 0.676 & 0.659}{0.856 & 0.793 & 0.838 & 0.821 & \best{0.763}}{0.9774 & 0.9777 & 0.9801 & 0.9797 & \best{0.9818}}
\ReportSceneRow{16}{0.691 & 0.688 & \best{0.660} & 0.688 & 0.681}{0.782 & 0.796 & 0.737 & 0.790 & \best{0.723}}{0.9787 & 0.9797 & 0.9796 & \best{0.9835} & 0.9835}
\ReportSceneRow{17}{0.601 & 0.571 & 0.590 & 0.566 & \best{0.559}}{0.609 & 0.593 & 0.589 & 0.607 & \best{0.563}}{0.9851 & 0.9857 & 0.9863 & \best{0.9877} & 0.9875}
\ReportSceneRow{18}{0.604 & 0.610 & 0.600 & 0.604 & \best{0.592}}{0.746 & 0.739 & 0.699 & 0.693 & \best{0.679}}{0.9795 & 0.9824 & 0.9848 & 0.9798 & \best{0.9870}}
\ReportSceneRow{19}{0.620 & 0.628 & \best{0.570} & 0.598 & 0.613}{0.634 & 0.659 & 0.631 & 0.651 & \best{0.602}}{0.9866 & 0.9868 & \best{0.9902} & 0.9847 & 0.9899}
\ReportSceneRow{20}{0.820 & 0.795 & 0.797 & 0.785 & \best{0.780}}{0.860 & 0.834 & 0.822 & 0.839 & \best{0.774}}{0.9784 & 0.9756 & 0.9785 & 0.9768 & \best{0.9828}}
\midrule
\ReportSceneRow{mean}{0.634 & 0.629 & 0.603 & 0.618 & 0.606}{0.712 & 0.703 & 0.673 & 0.692 & 0.648}{0.9830 & 0.9830 & 0.9840 & 0.9840 & 0.9870}
\ReportSceneRow{std}{0.127 & 0.124 & 0.127 & 0.125 & 0.126}{0.144 & 0.136 & 0.145 & 0.144 & 0.145}{0.0072 & 0.0067 & 0.0068 & 0.0070 & 0.0065}
\ReportSceneRow{\shortstack[l]{$p$, scene-paired\\Wilcoxon}}{$<$0.001 & $<$0.001 & 0.756 & 0.008 & }{$<$0.001 & $<$0.001 & $<$0.001 & $<$0.001 & }{$<$0.001 & $<$0.001 & $<$0.001 & $<$0.001 & }
\ReportSceneRow{\shortstack[l]{$p$, strip-level sign\\flip (strips won)}}{\shortstack{0.031\\(6/6)} & \shortstack{0.031\\(6/6)} & \shortstack{0.719\\(3/6)} & \shortstack{0.062\\(5/6)} & }{\shortstack{0.031\\(6/6)} & \shortstack{0.031\\(6/6)} & \shortstack{0.062\\(5/6)} & \shortstack{0.031\\(6/6)} & }{\shortstack{0.031\\(6/6)} & \shortstack{0.031\\(6/6)} & \shortstack{0.031\\(6/6)} & \shortstack{0.031\\(6/6)} & }
\ReportSceneRow{\shortstack[l]{$p$, mixed model,\\strip intercept}}{0.006 & 0.003 & 0.620 & 0.040 & }{$<$0.001 & $<$0.001 & 0.008 & 0.002 & }{$<$0.001 & $<$0.001 & 0.004 & 0.012 & }
\bottomrule
}
\end{table}


\begin{table}[!htbp]
\centering
\caption{Per-scene reduced-resolution results on WorldView-II (unseen sensor) for the four strongest published baselines and \our (seed-averaged, printed at the precision used by every test in this paper, so that each $p$ can be recomputed from the table). Bold: best per scene. Acquisition strips, by scene: 1--5 (A), 6--10 (B), 11--15 (C), 16--20 (D). The last three rows give, for each baseline, the unadjusted scene-paired Wilcoxon $p$ of \our against it, the exact two-sided sign-flip $p$ on the 4 strip means (strips on which \our is ahead in brackets) and the $p$ of a linear mixed-effects model of the per-scene differences with a random intercept per strip (REML; $t$-test with 3 degrees of freedom); Holm-adjusted scene-level values are used in Main Table~\MTabMain.}
\label{tab:perscene_wv2_rr}
\scriptsize
\setlength{\tabcolsep}{1.6pt}
\ReportSceneMatrix{5}{%
\toprule
\ReportSceneRow{Scene}{\multicolumn{5}{c|}{ERGAS}}{\multicolumn{5}{c|}{SAM}}{\multicolumn{5}{c}{Q2$^n$}}
\ReportSceneRow{}{CANConv & ARConv & SSDiff & SGDiff & \our}{CANConv & ARConv & SSDiff & SGDiff & \our}{CANConv & ARConv & SSDiff & SGDiff & \our}
\midrule
\ReportSceneRow{1}{3.459 & 3.450 & 3.509 & 3.404 & \best{3.214}}{4.566 & 4.620 & 4.619 & 4.557 & \best{4.362}}{0.8974 & 0.8991 & 0.9011 & 0.8973 & \best{0.9209}}
\ReportSceneRow{2}{3.786 & 3.602 & 3.704 & 3.647 & \best{3.464}}{5.113 & 4.979 & 5.074 & 4.926 & \best{4.765}}{0.8503 & 0.8592 & 0.8545 & 0.8657 & \best{0.8680}}
\ReportSceneRow{3}{4.138 & 3.862 & 3.982 & 4.023 & \best{3.715}}{4.895 & 4.878 & 5.016 & 4.868 & \best{4.678}}{0.8433 & 0.8437 & 0.8414 & 0.8442 & \best{0.8487}}
\ReportSceneRow{4}{4.212 & 4.068 & 4.042 & 4.024 & \best{3.910}}{5.423 & 5.229 & 5.245 & 5.235 & \best{5.084}}{0.7942 & 0.7966 & 0.7987 & 0.8051 & \best{0.8181}}
\ReportSceneRow{5}{4.581 & 4.452 & 4.516 & 4.366 & \best{4.284}}{5.312 & 5.341 & 5.338 & 5.386 & \best{5.122}}{0.7834 & 0.7833 & 0.7821 & 0.7930 & \best{0.8004}}
\ReportSceneRow{6}{4.191 & 4.127 & 4.099 & 3.951 & \best{3.791}}{4.960 & 5.020 & 4.992 & 4.910 & \best{4.670}}{0.8627 & 0.8576 & 0.8549 & 0.8542 & \best{0.8715}}
\ReportSceneRow{7}{4.095 & 4.158 & 4.077 & 4.053 & \best{3.860}}{6.212 & 5.879 & 5.861 & 5.822 & \best{5.727}}{0.8093 & 0.8130 & 0.7904 & 0.8135 & \best{0.8304}}
\ReportSceneRow{8}{3.593 & 3.605 & 3.652 & 3.476 & \best{3.435}}{4.782 & 4.786 & 4.689 & 4.630 & \best{4.448}}{0.8639 & 0.8667 & 0.8608 & 0.8713 & \best{0.8777}}
\ReportSceneRow{9}{4.056 & 3.970 & 4.018 & 3.973 & \best{3.705}}{5.206 & 5.096 & 5.226 & 5.056 & \best{4.667}}{0.8938 & 0.8875 & 0.8939 & \best{0.9035} & 0.9006}
\ReportSceneRow{10}{3.808 & 3.867 & 3.798 & 3.713 & \best{3.516}}{5.381 & 5.306 & 5.264 & 5.062 & \best{5.011}}{0.8787 & 0.8744 & 0.8777 & 0.8897 & \best{0.8919}}
\ReportSceneRow{11}{4.371 & 4.402 & 4.429 & 4.454 & \best{4.257}}{5.696 & 5.580 & 5.736 & 5.459 & \best{5.334}}{\best{0.7778} & 0.7647 & 0.7607 & 0.7772 & 0.7777}
\ReportSceneRow{12}{3.522 & 3.437 & 3.473 & 3.391 & \best{3.215}}{4.630 & 4.442 & 4.602 & 4.359 & \best{4.131}}{0.9067 & 0.9145 & 0.9121 & 0.9215 & \best{0.9324}}
\ReportSceneRow{13}{4.362 & 4.279 & 4.295 & 4.154 & \best{4.038}}{5.304 & 5.254 & 5.337 & 5.055 & \best{4.896}}{0.8602 & 0.8574 & 0.8512 & \best{0.8608} & 0.8607}
\ReportSceneRow{14}{3.896 & 3.739 & 3.808 & 3.751 & \best{3.616}}{5.223 & 5.100 & 5.060 & 4.824 & \best{4.793}}{0.8805 & 0.8875 & 0.8803 & 0.8949 & \best{0.9005}}
\ReportSceneRow{15}{3.984 & 3.813 & 3.831 & 3.775 & \best{3.607}}{4.483 & 4.428 & 4.433 & 4.263 & \best{4.098}}{0.8913 & 0.8935 & 0.8848 & 0.8912 & \best{0.9072}}
\ReportSceneRow{16}{3.408 & 3.489 & 3.385 & \best{3.281} & 3.290}{4.599 & 4.479 & 4.562 & 4.412 & \best{4.141}}{0.9065 & 0.9109 & 0.9116 & 0.9211 & \best{0.9297}}
\ReportSceneRow{17}{3.593 & 3.636 & 3.511 & 3.538 & \best{3.122}}{4.534 & 4.402 & 4.568 & 4.410 & \best{4.222}}{0.8938 & 0.9026 & 0.8913 & 0.9036 & \best{0.9179}}
\ReportSceneRow{18}{4.453 & 4.424 & 4.534 & 4.325 & \best{4.158}}{6.043 & 6.013 & 6.008 & 5.908 & \best{5.802}}{0.7664 & 0.7679 & 0.7595 & 0.7665 & \best{0.7829}}
\ReportSceneRow{19}{4.276 & 4.159 & 4.166 & 3.994 & \best{3.973}}{5.373 & 5.278 & 5.263 & 5.136 & \best{4.992}}{0.8358 & 0.8426 & 0.8383 & 0.8442 & \best{0.8579}}
\ReportSceneRow{20}{4.416 & 4.240 & 4.392 & 4.187 & \best{3.930}}{5.343 & 5.431 & 5.467 & 5.222 & \best{5.059}}{0.8441 & 0.8575 & 0.8346 & 0.8416 & \best{0.8649}}
\midrule
\ReportSceneRow{mean}{4.010 & 3.939 & 3.961 & 3.874 & 3.705}{5.154 & 5.077 & 5.118 & 4.975 & 4.800}{0.8520 & 0.8540 & 0.8490 & 0.8580 & 0.8680}
\ReportSceneRow{std}{0.360 & 0.336 & 0.359 & 0.344 & 0.350}{0.485 & 0.464 & 0.449 & 0.460 & 0.490}{0.0447 & 0.0466 & 0.0483 & 0.0470 & 0.0470}
\ReportSceneRow{\shortstack[l]{$p$, scene-paired\\Wilcoxon}}{$<$0.001 & $<$0.001 & $<$0.001 & $<$0.001 & }{$<$0.001 & $<$0.001 & $<$0.001 & $<$0.001 & }{$<$0.001 & $<$0.001 & $<$0.001 & $<$0.001 & }
\ReportSceneRow{\shortstack[l]{$p$, strip-level sign\\flip (strips won)}}{\shortstack{0.125\\(4/4)} & \shortstack{0.125\\(4/4)} & \shortstack{0.125\\(4/4)} & \shortstack{0.125\\(4/4)} & }{\shortstack{0.125\\(4/4)} & \shortstack{0.125\\(4/4)} & \shortstack{0.125\\(4/4)} & \shortstack{0.125\\(4/4)} & }{\shortstack{0.125\\(4/4)} & \shortstack{0.125\\(4/4)} & \shortstack{0.125\\(4/4)} & \shortstack{0.125\\(4/4)} & }
\ReportSceneRow{\shortstack[l]{$p$, mixed model,\\strip intercept}}{0.001 & 0.005 & 0.001 & 0.005 & }{0.002 & 0.002 & 0.001 & 0.002 & }{0.004 & 0.001 & 0.002 & 0.013 & }
\bottomrule
}
\end{table}


\begin{table}[!htbp]
\centering
\caption{Per-scene full-resolution results on WorldView-III for the four strongest published baselines, CLIPPan and \our (seed-averaged, printed at the precision used by every test in this paper, so that each $p$ can be recomputed from the table). Bold: best per scene. Acquisition strips, by scene: 1--4 (A), 5--7 (B), 8--10 (C), 11--14 (D), 15--17 (E), 18--20 (F). The last three rows give, for each baseline, the unadjusted scene-paired Wilcoxon $p$ of \our against it, the exact two-sided sign-flip $p$ on the 6 strip means (strips on which \our is ahead in brackets) and the $p$ of a linear mixed-effects model of the per-scene differences with a random intercept per strip (REML; $t$-test with 5 degrees of freedom); Holm-adjusted scene-level values are used in Main Table~\MTabMain.}
\label{tab:perscene_wv3_fr}
\scriptsize
\setlength{\tabcolsep}{1.6pt}
\ReportSceneMatrix{6}{%
\toprule
\ReportSceneRow{Scene}{\multicolumn{6}{c|}{$D_\lambda$}}{\multicolumn{6}{c|}{$D_s$}}{\multicolumn{6}{c}{HQNR}}
\ReportSceneRow{}{CANConv & ARConv & SSDiff & SGDiff & CLIPPan & \our}{CANConv & ARConv & SSDiff & SGDiff & CLIPPan & \our}{CANConv & ARConv & SSDiff & SGDiff & CLIPPan & \our}
\midrule
\ReportSceneRow{1}{0.0199 & 0.0142 & 0.0089 & 0.0107 & 0.0190 & \best{0.0048}}{0.0227 & 0.0201 & 0.0260 & 0.0238 & 0.0335 & \best{0.0143}}{0.9579 & 0.9660 & 0.9653 & 0.9657 & 0.9481 & \best{0.9809}}
\ReportSceneRow{2}{0.0187 & 0.0123 & 0.0101 & 0.0099 & 0.0090 & \best{0.0072}}{0.0216 & 0.0204 & 0.0250 & 0.0219 & 0.0297 & \best{0.0182}}{0.9600 & 0.9675 & 0.9651 & 0.9684 & 0.9616 & \best{0.9746}}
\ReportSceneRow{3}{0.0167 & 0.0112 & 0.0114 & 0.0120 & 0.0155 & \best{0.0087}}{0.0323 & 0.0284 & 0.0305 & 0.0295 & 0.0340 & \best{0.0221}}{0.9515 & 0.9607 & 0.9584 & 0.9589 & 0.9510 & \best{0.9693}}
\ReportSceneRow{4}{0.0345 & 0.0234 & 0.0239 & 0.0275 & \best{0.0088} & 0.0234}{0.0441 & 0.0430 & 0.0417 & 0.0370 & \best{0.0258} & 0.0361}{0.9229 & 0.9347 & 0.9354 & 0.9365 & \best{0.9656} & 0.9414}
\ReportSceneRow{5}{0.0194 & 0.0119 & 0.0123 & 0.0104 & \best{0.0083} & 0.0084}{0.0227 & 0.0217 & 0.0237 & 0.0248 & 0.0234 & \best{0.0141}}{0.9583 & 0.9667 & 0.9643 & 0.9651 & 0.9685 & \best{0.9776}}
\ReportSceneRow{6}{0.0213 & 0.0180 & 0.0148 & 0.0130 & 0.0173 & \best{0.0113}}{0.0318 & 0.0269 & 0.0310 & 0.0281 & 0.0301 & \best{0.0235}}{0.9476 & 0.9556 & 0.9547 & 0.9593 & 0.9531 & \best{0.9654}}
\ReportSceneRow{7}{0.0211 & 0.0166 & 0.0160 & 0.0137 & 0.0137 & \best{0.0126}}{0.0290 & 0.0307 & 0.0318 & 0.0271 & 0.0328 & \best{0.0244}}{0.9505 & 0.9533 & 0.9527 & 0.9595 & 0.9539 & \best{0.9633}}
\ReportSceneRow{8}{0.0144 & 0.0115 & \best{0.0075} & 0.0086 & 0.0111 & 0.0082}{0.0276 & 0.0248 & 0.0295 & 0.0229 & 0.0239 & \best{0.0220}}{0.9584 & 0.9640 & 0.9631 & 0.9687 & 0.9653 & \best{0.9699}}
\ReportSceneRow{9}{0.0153 & 0.0118 & 0.0125 & 0.0106 & 0.0190 & \best{0.0071}}{0.0351 & 0.0269 & 0.0373 & 0.0311 & 0.0396 & \best{0.0241}}{0.9502 & 0.9616 & 0.9507 & 0.9587 & 0.9422 & \best{0.9690}}
\ReportSceneRow{10}{0.0199 & 0.0162 & 0.0157 & 0.0174 & \best{0.0014} & 0.0136}{0.0327 & 0.0299 & 0.0343 & 0.0301 & \best{0.0110} & 0.0284}{0.9481 & 0.9544 & 0.9505 & 0.9530 & \best{0.9876} & 0.9585}
\ReportSceneRow{11}{0.0301 & 0.0213 & 0.0229 & 0.0202 & \best{0.0106} & 0.0199}{0.0351 & 0.0335 & 0.0383 & 0.0315 & \best{0.0273} & 0.0298}{0.9358 & 0.9460 & 0.9396 & 0.9489 & \best{0.9624} & 0.9509}
\ReportSceneRow{12}{0.0089 & 0.0050 & 0.0021 & 0.0015 & 0.0133 & \best{0.0013}}{0.0247 & 0.0235 & 0.0270 & \best{0.0176} & 0.0277 & 0.0231}{0.9666 & 0.9715 & 0.9709 & \best{0.9809} & 0.9594 & 0.9756}
\ReportSceneRow{13}{0.0206 & 0.0165 & 0.0196 & 0.0137 & 0.0199 & \best{0.0095}}{0.0315 & 0.0341 & 0.0288 & 0.0295 & 0.0388 & \best{0.0260}}{0.9486 & 0.9499 & 0.9521 & 0.9571 & 0.9421 & \best{0.9647}}
\ReportSceneRow{14}{0.0220 & 0.0158 & 0.0095 & 0.0108 & \best{0.0087} & 0.0103}{0.0272 & 0.0274 & 0.0289 & 0.0240 & 0.0254 & \best{0.0198}}{0.9513 & 0.9573 & 0.9618 & 0.9655 & 0.9661 & \best{0.9702}}
\ReportSceneRow{15}{0.0185 & 0.0128 & 0.0124 & 0.0093 & 0.0146 & \best{0.0079}}{0.0307 & 0.0244 & 0.0269 & 0.0255 & 0.0342 & \best{0.0223}}{0.9514 & 0.9631 & 0.9610 & 0.9654 & 0.9517 & \best{0.9700}}
\ReportSceneRow{16}{0.0348 & 0.0297 & 0.0284 & 0.0242 & \best{0.0059} & 0.0315}{0.0523 & 0.0512 & 0.0527 & 0.0482 & \best{0.0251} & 0.0460}{0.9148 & 0.9207 & 0.9203 & 0.9288 & \best{0.9691} & 0.9239}
\ReportSceneRow{17}{0.0212 & 0.0183 & 0.0158 & 0.0146 & 0.0231 & \best{0.0128}}{0.0347 & 0.0307 & 0.0307 & 0.0323 & 0.0417 & \best{0.0252}}{0.9448 & 0.9515 & 0.9539 & 0.9536 & 0.9362 & \best{0.9623}}
\ReportSceneRow{18}{0.0169 & 0.0114 & 0.0107 & 0.0082 & 0.0164 & \best{0.0081}}{0.0214 & \best{0.0206} & 0.0250 & 0.0209 & 0.0343 & 0.0220}{0.9621 & 0.9682 & 0.9646 & \best{0.9711} & 0.9499 & 0.9701}
\ReportSceneRow{19}{0.0137 & 0.0119 & 0.0057 & 0.0083 & \best{0.0016} & 0.0035}{0.0275 & 0.0236 & 0.0250 & 0.0207 & \best{0.0195} & 0.0221}{0.9592 & 0.9647 & 0.9694 & 0.9712 & \best{0.9789} & 0.9744}
\ReportSceneRow{20}{0.0142 & 0.0102 & 0.0115 & 0.0094 & 0.0139 & \best{0.0058}}{0.0272 & 0.0262 & 0.0298 & 0.0256 & 0.0295 & \best{0.0185}}{0.9589 & 0.9639 & 0.9591 & 0.9653 & 0.9570 & \best{0.9759}}
\midrule
\ReportSceneRow{mean}{0.0201 & 0.0150 & 0.0136 & 0.0127 & 0.0126 & 0.0108}{0.0306 & 0.0284 & 0.0312 & 0.0276 & 0.0294 & 0.0241}{0.9499 & 0.9571 & 0.9556 & 0.9601 & 0.9585 & 0.9654}
\ReportSceneRow{std}{0.0065 & 0.0054 & 0.0064 & 0.0059 & 0.0059 & 0.0071}{0.0076 & 0.0077 & 0.0070 & 0.0068 & 0.0072 & 0.0072}{0.0128 & 0.0123 & 0.0123 & 0.0120 & 0.0126 & 0.0135}
\ReportSceneRow{\shortstack[l]{$p$, scene-paired\\Wilcoxon}}{$<$0.001 & $<$0.001 & $<$0.001 & 0.001 & 0.312 & }{$<$0.001 & $<$0.001 & $<$0.001 & 0.001 & 0.048 & }{$<$0.001 & $<$0.001 & $<$0.001 & 0.001 & 0.105 & }
\ReportSceneRow{\shortstack[l]{$p$, strip-level sign\\flip (strips won)}}{\shortstack{0.031\\(6/6)} & \shortstack{0.031\\(6/6)} & \shortstack{0.031\\(6/6)} & \shortstack{0.094\\(5/6)} & \shortstack{0.156\\(5/6)} & }{\shortstack{0.031\\(6/6)} & \shortstack{0.031\\(6/6)} & \shortstack{0.031\\(6/6)} & \shortstack{0.031\\(6/6)} & \shortstack{0.062\\(6/6)} & }{\shortstack{0.031\\(6/6)} & \shortstack{0.031\\(6/6)} & \shortstack{0.031\\(6/6)} & \shortstack{0.031\\(6/6)} & \shortstack{0.062\\(5/6)} & }
\ReportSceneRow{\shortstack[l]{$p$, mixed model,\\strip intercept}}{$<$0.001 & 0.001 & 0.006 & 0.054 & 0.480 & }{$<$0.001 & 0.001 & $<$0.001 & 0.010 & 0.085 & }{$<$0.001 & 0.001 & $<$0.001 & 0.007 & 0.189 & }
\bottomrule
}
\end{table}


\begin{table}[!htbp]
\centering
\caption{Per-scene full-resolution results on QuickBird for the four strongest published baselines, CLIPPan and \our (seed-averaged, printed at the precision used by every test in this paper, so that each $p$ can be recomputed from the table). Bold: best per scene. Acquisition strips, by scene: 1--4 (A), 5--8 (B), 9--12 (C), 13--16 (D), 17--20 (E). The last three rows give, for each baseline, the unadjusted scene-paired Wilcoxon $p$ of \our against it, the exact two-sided sign-flip $p$ on the 5 strip means (strips on which \our is ahead in brackets) and the $p$ of a linear mixed-effects model of the per-scene differences with a random intercept per strip (REML; $t$-test with 4 degrees of freedom); Holm-adjusted scene-level values are used in Main Table~\MTabMain.}
\label{tab:perscene_qb_fr}
\scriptsize
\setlength{\tabcolsep}{1.6pt}
\ReportSceneMatrix{6}{%
\toprule
\ReportSceneRow{Scene}{\multicolumn{6}{c|}{$D_\lambda$}}{\multicolumn{6}{c|}{$D_s$}}{\multicolumn{6}{c}{HQNR}}
\ReportSceneRow{}{CANConv & ARConv & SSDiff & SGDiff & CLIPPan & \our}{CANConv & ARConv & SSDiff & SGDiff & CLIPPan & \our}{CANConv & ARConv & SSDiff & SGDiff & CLIPPan & \our}
\midrule
\ReportSceneRow{1}{0.0552 & 0.0593 & 0.0481 & \best{0.0402} & 0.0412 & 0.0412}{0.0587 & 0.0456 & \best{0.0445} & 0.0573 & 0.0510 & 0.0474}{0.8893 & 0.8978 & 0.9095 & 0.9048 & 0.9099 & \best{0.9134}}
\ReportSceneRow{2}{0.0283 & 0.0298 & 0.0248 & \best{0.0171} & 0.0207 & 0.0216}{0.0537 & 0.0468 & 0.0453 & 0.0414 & 0.0361 & \best{0.0314}}{0.9195 & 0.9248 & 0.9310 & 0.9422 & 0.9439 & \best{0.9477}}
\ReportSceneRow{3}{0.0372 & 0.0379 & 0.0341 & \best{0.0236} & 0.0301 & 0.0245}{0.0468 & 0.0360 & 0.0370 & 0.0426 & 0.0411 & \best{0.0343}}{0.9177 & 0.9275 & 0.9302 & 0.9348 & 0.9300 & \best{0.9420}}
\ReportSceneRow{4}{0.0107 & 0.0150 & 0.0068 & \best{0.0020} & 0.0079 & 0.0030}{0.0251 & 0.0187 & 0.0124 & 0.0238 & 0.0159 & \best{0.0059}}{0.9645 & 0.9666 & 0.9809 & 0.9742 & 0.9763 & \best{0.9911}}
\ReportSceneRow{5}{0.0386 & 0.0370 & 0.0313 & 0.0221 & \best{0.0192} & 0.0234}{0.0433 & \best{0.0245} & 0.0403 & 0.0396 & 0.0323 & 0.0284}{0.9198 & 0.9394 & 0.9297 & 0.9392 & \best{0.9491} & 0.9489}
\ReportSceneRow{6}{0.0379 & 0.0411 & 0.0285 & \best{0.0184} & 0.0297 & 0.0293}{0.0452 & \best{0.0283} & 0.0362 & 0.0395 & 0.0359 & 0.0335}{0.9186 & 0.9318 & 0.9363 & \best{0.9428} & 0.9355 & 0.9382}
\ReportSceneRow{7}{0.0320 & 0.0323 & 0.0311 & \best{0.0184} & 0.0210 & 0.0218}{0.0477 & 0.0364 & 0.0376 & 0.0398 & 0.0354 & \best{0.0277}}{0.9218 & 0.9325 & 0.9325 & 0.9425 & 0.9443 & \best{0.9511}}
\ReportSceneRow{8}{0.0457 & 0.0459 & 0.0373 & 0.0355 & \best{0.0310} & 0.0316}{0.0574 & 0.0485 & 0.0385 & 0.0522 & 0.0471 & \best{0.0382}}{0.8995 & 0.9078 & 0.9256 & 0.9142 & 0.9234 & \best{0.9314}}
\ReportSceneRow{9}{0.0251 & 0.0248 & 0.0187 & \best{0.0073} & 0.0174 & 0.0135}{0.0396 & \best{0.0226} & 0.0261 & 0.0317 & 0.0265 & 0.0262}{0.9363 & 0.9532 & 0.9557 & \best{0.9612} & 0.9566 & 0.9607}
\ReportSceneRow{10}{0.0281 & 0.0339 & 0.0262 & \best{0.0178} & 0.0207 & 0.0215}{0.0442 & 0.0382 & 0.0312 & 0.0377 & 0.0316 & \best{0.0267}}{0.9289 & 0.9292 & 0.9434 & 0.9452 & 0.9484 & \best{0.9524}}
\ReportSceneRow{11}{0.0257 & 0.0304 & 0.0280 & 0.0164 & 0.0180 & \best{0.0133}}{0.0442 & 0.0406 & 0.0269 & 0.0386 & \best{0.0264} & 0.0315}{0.9312 & 0.9302 & 0.9459 & 0.9456 & \best{0.9561} & 0.9556}
\ReportSceneRow{12}{0.0345 & 0.0392 & 0.0274 & \best{0.0209} & 0.0237 & 0.0215}{0.0547 & 0.0406 & 0.0312 & 0.0449 & 0.0376 & \best{0.0298}}{0.9127 & 0.9218 & 0.9423 & 0.9351 & 0.9396 & \best{0.9493}}
\ReportSceneRow{13}{0.0512 & 0.0486 & 0.0424 & \best{0.0322} & 0.0325 & 0.0347}{0.0675 & 0.0724 & \best{0.0511} & 0.0600 & 0.0547 & 0.0514}{0.8848 & 0.8825 & 0.9087 & 0.9097 & 0.9146 & \best{0.9157}}
\ReportSceneRow{14}{0.0485 & 0.0545 & 0.0443 & 0.0347 & 0.0337 & \best{0.0324}}{0.0550 & 0.0336 & 0.0384 & 0.0456 & 0.0474 & \best{0.0322}}{0.8992 & 0.9137 & 0.9190 & 0.9213 & 0.9205 & \best{0.9364}}
\ReportSceneRow{15}{0.0352 & 0.0328 & \best{0.0238} & 0.0252 & 0.0268 & 0.0252}{0.0428 & 0.0292 & 0.0332 & 0.0360 & 0.0324 & \best{0.0245}}{0.9235 & 0.9390 & 0.9438 & 0.9397 & 0.9417 & \best{0.9509}}
\ReportSceneRow{16}{0.0139 & 0.0184 & 0.0150 & \best{0.0019} & 0.0095 & 0.0062}{0.0441 & 0.0375 & 0.0262 & 0.0333 & 0.0247 & \best{0.0156}}{0.9426 & 0.9448 & 0.9592 & 0.9649 & 0.9660 & \best{0.9783}}
\ReportSceneRow{17}{0.0467 & 0.0494 & 0.0443 & 0.0369 & \best{0.0331} & 0.0333}{0.0614 & 0.0426 & 0.0442 & 0.0534 & 0.0492 & \best{0.0407}}{0.8948 & 0.9101 & 0.9135 & 0.9117 & 0.9193 & \best{0.9274}}
\ReportSceneRow{18}{0.0570 & 0.0590 & 0.0451 & 0.0454 & \best{0.0383} & 0.0388}{0.0757 & 0.0618 & \best{0.0524} & 0.0640 & 0.0574 & 0.0567}{0.8716 & 0.8828 & 0.9049 & 0.8935 & 0.9065 & \best{0.9067}}
\ReportSceneRow{19}{0.0237 & 0.0303 & 0.0249 & 0.0176 & 0.0176 & \best{0.0142}}{0.0417 & 0.0322 & 0.0346 & 0.0377 & 0.0289 & \best{0.0195}}{0.9356 & 0.9385 & 0.9414 & 0.9454 & 0.9540 & \best{0.9666}}
\ReportSceneRow{20}{0.0654 & 0.0620 & 0.0567 & 0.0459 & 0.0505 & \best{0.0419}}{0.0688 & 0.0562 & \best{0.0493} & 0.0571 & 0.0606 & 0.0526}{0.8703 & 0.8853 & 0.8968 & 0.8996 & 0.8920 & \best{0.9077}}
\midrule
\ReportSceneRow{mean}{0.0370 & 0.0391 & 0.0319 & 0.0240 & 0.0261 & 0.0246}{0.0509 & 0.0396 & 0.0368 & 0.0438 & 0.0386 & 0.0327}{0.9141 & 0.9230 & 0.9325 & 0.9334 & 0.9364 & 0.9436}
\ReportSceneRow{std}{0.0143 & 0.0133 & 0.0122 & 0.0130 & 0.0106 & 0.0110}{0.0118 & 0.0132 & 0.0098 & 0.0105 & 0.0121 & 0.0126}{0.0241 & 0.0231 & 0.0205 & 0.0221 & 0.0216 & 0.0224}
\ReportSceneRow{\shortstack[l]{$p$, scene-paired\\Wilcoxon}}{$<$0.001 & $<$0.001 & $<$0.001 & 0.568 & 0.087 & }{$<$0.001 & 0.002 & 0.015 & $<$0.001 & $<$0.001 & }{$<$0.001 & $<$0.001 & $<$0.001 & $<$0.001 & $<$0.001 & }
\ReportSceneRow{\shortstack[l]{$p$, strip-level sign\\flip (strips won)}}{\shortstack{0.062\\(5/5)} & \shortstack{0.062\\(5/5)} & \shortstack{0.062\\(5/5)} & \shortstack{0.625\\(1/5)} & \shortstack{0.188\\(4/5)} & }{\shortstack{0.062\\(5/5)} & \shortstack{0.062\\(5/5)} & \shortstack{0.062\\(5/5)} & \shortstack{0.062\\(5/5)} & \shortstack{0.062\\(5/5)} & }{\shortstack{0.062\\(5/5)} & \shortstack{0.062\\(5/5)} & \shortstack{0.062\\(5/5)} & \shortstack{0.062\\(5/5)} & \shortstack{0.062\\(5/5)} & }
\ReportSceneRow{\shortstack[l]{$p$, mixed model,\\strip intercept}}{$<$0.001 & $<$0.001 & 0.002 & 0.633 & 0.124 & }{$<$0.001 & 0.017 & 0.038 & $<$0.001 & 0.006 & }{$<$0.001 & $<$0.001 & 0.002 & 0.002 & 0.004 & }
\bottomrule
}
\end{table}


\begin{table}[!htbp]
\centering
\caption{Per-scene full-resolution results on GaoFen-2 for the four strongest published baselines, CLIPPan and \our (seed-averaged, printed at the precision used by every test in this paper, so that each $p$ can be recomputed from the table). Bold: best per scene. Acquisition strips, by scene: 1--4 (A), 5--7 (B), 8--10 (C), 11--14 (D), 15--17 (E), 18--20 (F). The last three rows give, for each baseline, the unadjusted scene-paired Wilcoxon $p$ of \our against it, the exact two-sided sign-flip $p$ on the 6 strip means (strips on which \our is ahead in brackets) and the $p$ of a linear mixed-effects model of the per-scene differences with a random intercept per strip (REML; $t$-test with 5 degrees of freedom); Holm-adjusted scene-level values are used in Main Table~\MTabMain.}
\label{tab:perscene_gf2_fr}
\scriptsize
\setlength{\tabcolsep}{1.6pt}
\ReportSceneMatrix{6}{%
\toprule
\ReportSceneRow{Scene}{\multicolumn{6}{c|}{$D_\lambda$}}{\multicolumn{6}{c|}{$D_s$}}{\multicolumn{6}{c}{HQNR}}
\ReportSceneRow{}{CANConv & ARConv & SSDiff & SGDiff & CLIPPan & \our}{CANConv & ARConv & SSDiff & SGDiff & CLIPPan & \our}{CANConv & ARConv & SSDiff & SGDiff & CLIPPan & \our}
\midrule
\ReportSceneRow{1}{0.0258 & 0.0261 & 0.0245 & 0.0230 & \best{0.0217} & 0.0222}{0.0722 & 0.0595 & 0.0333 & 0.0350 & 0.0342 & \best{0.0289}}{0.9039 & 0.9160 & 0.9430 & 0.9428 & 0.9448 & \best{0.9495}}
\ReportSceneRow{2}{0.0073 & 0.0029 & 0.0036 & 0.0023 & 0.0025 & \best{0.0011}}{0.0429 & 0.0333 & 0.0111 & 0.0073 & 0.0061 & \best{0.0034}}{0.9501 & 0.9639 & 0.9853 & 0.9904 & 0.9914 & \best{0.9955}}
\ReportSceneRow{3}{0.0128 & 0.0084 & 0.0085 & 0.0070 & 0.0087 & \best{0.0024}}{0.0479 & 0.0378 & 0.0164 & 0.0092 & 0.0133 & \best{0.0048}}{0.9399 & 0.9541 & 0.9752 & 0.9839 & 0.9781 & \best{0.9928}}
\ReportSceneRow{4}{0.0274 & 0.0253 & 0.0216 & 0.0257 & 0.0235 & \best{0.0214}}{0.0751 & 0.0606 & 0.0397 & 0.0387 & 0.0364 & \best{0.0336}}{0.8996 & 0.9156 & 0.9396 & 0.9366 & 0.9410 & \best{0.9457}}
\ReportSceneRow{5}{0.0121 & 0.0119 & 0.0114 & 0.0107 & 0.0087 & \best{0.0067}}{0.0527 & 0.0424 & 0.0203 & 0.0158 & 0.0152 & \best{0.0067}}{0.9358 & 0.9462 & 0.9685 & 0.9737 & 0.9762 & \best{0.9866}}
\ReportSceneRow{6}{0.0159 & 0.0121 & 0.0100 & 0.0123 & 0.0116 & \best{0.0076}}{0.0607 & 0.0451 & 0.0278 & \best{0.0164} & 0.0220 & 0.0214}{0.9244 & 0.9433 & 0.9625 & \best{0.9715} & 0.9667 & 0.9712}
\ReportSceneRow{7}{0.0244 & 0.0249 & 0.0201 & 0.0213 & 0.0198 & \best{0.0156}}{0.0678 & 0.0576 & 0.0285 & \best{0.0245} & 0.0301 & 0.0252}{0.9095 & 0.9189 & 0.9520 & 0.9547 & 0.9507 & \best{0.9596}}
\ReportSceneRow{8}{0.0120 & 0.0136 & 0.0103 & 0.0113 & 0.0092 & \best{0.0051}}{0.0617 & 0.0472 & 0.0289 & 0.0265 & 0.0210 & \best{0.0208}}{0.9270 & 0.9398 & 0.9611 & 0.9625 & 0.9700 & \best{0.9742}}
\ReportSceneRow{9}{0.0156 & 0.0164 & 0.0146 & 0.0133 & 0.0125 & \best{0.0071}}{0.0588 & 0.0484 & 0.0224 & 0.0277 & \best{0.0204} & 0.0259}{0.9265 & 0.9360 & 0.9633 & 0.9594 & \best{0.9674} & 0.9672}
\ReportSceneRow{10}{0.0261 & 0.0208 & 0.0198 & 0.0186 & 0.0179 & \best{0.0141}}{0.0549 & 0.0506 & 0.0195 & 0.0148 & 0.0197 & \best{0.0146}}{0.9204 & 0.9297 & 0.9611 & 0.9669 & 0.9628 & \best{0.9715}}
\ReportSceneRow{11}{0.0279 & 0.0264 & 0.0265 & 0.0276 & \best{0.0219} & 0.0224}{0.0675 & 0.0585 & 0.0328 & 0.0273 & 0.0326 & \best{0.0240}}{0.9065 & 0.9166 & 0.9416 & 0.9459 & 0.9462 & \best{0.9541}}
\ReportSceneRow{12}{0.0264 & 0.0245 & 0.0209 & 0.0223 & 0.0217 & \best{0.0199}}{0.0643 & 0.0560 & 0.0281 & \best{0.0243} & 0.0297 & 0.0253}{0.9110 & 0.9209 & 0.9516 & 0.9539 & 0.9492 & \best{0.9553}}
\ReportSceneRow{13}{0.0204 & 0.0194 & 0.0196 & 0.0157 & \best{0.0151} & 0.0151}{0.0664 & 0.0517 & 0.0283 & 0.0282 & 0.0265 & \best{0.0218}}{0.9146 & 0.9299 & 0.9527 & 0.9565 & 0.9588 & \best{0.9634}}
\ReportSceneRow{14}{0.0080 & 0.0079 & 0.0087 & 0.0051 & 0.0069 & \best{0.0029}}{0.0437 & 0.0364 & 0.0119 & 0.0090 & 0.0108 & \best{0.0023}}{0.9486 & 0.9560 & 0.9795 & 0.9859 & 0.9824 & \best{0.9948}}
\ReportSceneRow{15}{0.0244 & 0.0221 & 0.0219 & 0.0226 & 0.0191 & \best{0.0159}}{0.0628 & 0.0546 & 0.0312 & 0.0284 & 0.0293 & \best{0.0262}}{0.9143 & 0.9245 & 0.9476 & 0.9496 & 0.9522 & \best{0.9583}}
\ReportSceneRow{16}{0.0259 & 0.0235 & 0.0196 & 0.0196 & 0.0202 & \best{0.0189}}{0.0785 & 0.0592 & 0.0390 & 0.0362 & 0.0385 & \best{0.0357}}{0.8976 & 0.9187 & 0.9422 & 0.9449 & 0.9421 & \best{0.9461}}
\ReportSceneRow{17}{0.0277 & 0.0274 & 0.0256 & 0.0276 & 0.0234 & \best{0.0211}}{0.0716 & 0.0607 & 0.0350 & 0.0317 & 0.0360 & \best{0.0316}}{0.9027 & 0.9136 & 0.9403 & 0.9416 & 0.9414 & \best{0.9480}}
\ReportSceneRow{18}{0.0146 & 0.0139 & 0.0132 & 0.0128 & 0.0098 & \best{0.0065}}{0.0530 & 0.0456 & 0.0222 & \best{0.0145} & 0.0187 & 0.0197}{0.9332 & 0.9411 & 0.9649 & 0.9729 & 0.9717 & \best{0.9739}}
\ReportSceneRow{19}{0.0360 & 0.0323 & 0.0281 & 0.0349 & 0.0276 & \best{0.0269}}{0.0732 & 0.0655 & 0.0364 & 0.0375 & 0.0395 & \best{0.0313}}{0.8934 & 0.9043 & 0.9365 & 0.9289 & 0.9340 & \best{0.9426}}
\ReportSceneRow{20}{0.0138 & 0.0150 & 0.0135 & 0.0122 & 0.0133 & \best{0.0073}}{0.0578 & 0.0462 & 0.0228 & 0.0237 & 0.0212 & \best{0.0165}}{0.9292 & 0.9395 & 0.9640 & 0.9644 & 0.9658 & \best{0.9763}}
\midrule
\ReportSceneRow{mean}{0.0202 & 0.0187 & 0.0171 & 0.0173 & 0.0158 & 0.0130}{0.0617 & 0.0508 & 0.0268 & 0.0238 & 0.0251 & 0.0210}{0.9194 & 0.9314 & 0.9566 & 0.9593 & 0.9596 & 0.9663}
\ReportSceneRow{std}{0.0080 & 0.0078 & 0.0069 & 0.0084 & 0.0068 & 0.0080}{0.0103 & 0.0091 & 0.0083 & 0.0098 & 0.0097 & 0.0101}{0.0168 & 0.0163 & 0.0141 & 0.0170 & 0.0160 & 0.0169}
\ReportSceneRow{\shortstack[l]{$p$, scene-paired\\Wilcoxon}}{$<$0.001 & $<$0.001 & $<$0.001 & $<$0.001 & $<$0.001 & }{$<$0.001 & $<$0.001 & $<$0.001 & 0.006 & $<$0.001 & }{$<$0.001 & $<$0.001 & $<$0.001 & $<$0.001 & $<$0.001 & }
\ReportSceneRow{\shortstack[l]{$p$, strip-level sign\\flip (strips won)}}{\shortstack{0.031\\(6/6)} & \shortstack{0.031\\(6/6)} & \shortstack{0.031\\(6/6)} & \shortstack{0.031\\(6/6)} & \shortstack{0.031\\(6/6)} & }{\shortstack{0.031\\(6/6)} & \shortstack{0.031\\(6/6)} & \shortstack{0.031\\(6/6)} & \shortstack{0.031\\(6/6)} & \shortstack{0.062\\(5/6)} & }{\shortstack{0.031\\(6/6)} & \shortstack{0.031\\(6/6)} & \shortstack{0.031\\(6/6)} & \shortstack{0.031\\(6/6)} & \shortstack{0.031\\(6/6)} & }
\ReportSceneRow{\shortstack[l]{$p$, mixed model,\\strip intercept}}{$<$0.001 & $<$0.001 & $<$0.001 & 0.001 & 0.002 & }{$<$0.001 & $<$0.001 & 0.001 & 0.037 & 0.007 & }{$<$0.001 & $<$0.001 & $<$0.001 & 0.001 & $<$0.001 & }
\bottomrule
}
\end{table}


\begin{table}[!htbp]
\centering
\caption{Strip-level check of every cell of Main Table~\MTabMain. The $20$ test scenes of a sensor are crops of $4$--$6$ acquisition strips (scene membership in the captions of \cref{tab:perscene_wv3_rr,tab:perscene_qb_rr,tab:perscene_gf2_rr,tab:perscene_wv2_rr,tab:perscene_wv3_fr,tab:perscene_qb_fr,tab:perscene_gf2_fr}), so scene-level tests may over-state independence. Each row names the runner-up of the cell, the method the mark of the main table is tested against, and gives for the per-scene differences \our minus runner-up, signed so that positive favours \our: the exact two-sided sign-flip $p$ on the strip means with the strips on which \our is ahead in brackets (the smallest attainable $p$ is $0.031$, $0.062$ and $0.125$ with six, five and four strips, so this column reports the direction of the effect); the $p$ of a linear mixed-effects model with a random intercept per strip (REML; $t$-test with strips$-1$ degrees of freedom), which uses the within-strip replication and is the strip-level significance quoted in the main text; and the same model against SGDiff, the strongest pixel-supervised network. The two cells in which \our is second are tested against the leader and carry no mark.}
\label{tab:strip}
\scriptsize
\setlength{\tabcolsep}{3pt}
\maxsizebox{\columnwidth}{%
\begin{tabular}{llcccc}
\toprule
Cell & mark & runner-up & sign flip (strips won) & mixed model & mixed model vs.\ SGDiff \\
\midrule
GaoFen-2 ERGAS & -- & SSDiff (leader) & 0.719 (3/6) & 0.620 & 0.040 \\
GaoFen-2 SAM & \sig & SSDiff & 0.062 (5/6) & 0.008 & 0.002 \\
GaoFen-2 Q2$^n$ & \sig & SSDiff & 0.031 (6/6) & 0.004 & 0.012 \\
GaoFen-2 $D_\lambda$ & \sig & CLIPPan & 0.031 (6/6) & 0.002 & 0.001 \\
GaoFen-2 $D_s$ & \sigw & SGDiff & 0.031 (6/6) & 0.037 & 0.037 \\
GaoFen-2 HQNR & \sig & CLIPPan & 0.031 (6/6) & $<$0.001 & 0.001 \\
\midrule
QuickBird ERGAS & \sig & SGDiff & 0.062 (5/5) & $<$0.001 & $<$0.001 \\
QuickBird SAM & \sig & SGDiff & 0.062 (5/5) & 0.005 & 0.005 \\
QuickBird Q2$^n$ & \sig & ARConv & 0.062 (5/5) & 0.023 & 0.002 \\
QuickBird $D_\lambda$ & -- & SGDiff (leader) & 0.625 (1/5) & 0.633 & 0.633 \\
QuickBird $D_s$ & \sigw & SSDiff & 0.062 (5/5) & 0.038 & $<$0.001 \\
QuickBird HQNR & \sig & CLIPPan & 0.062 (5/5) & 0.004 & 0.002 \\
\midrule
WorldView-III ERGAS & \sig & SGDiff & 0.031 (6/6) & 0.005 & 0.005 \\
WorldView-III SAM & \sig & SGDiff & 0.031 (6/6) & 0.002 & 0.002 \\
WorldView-III Q2$^n$ & \sig & SGDiff & 0.031 (6/6) & 0.013 & 0.013 \\
WorldView-III $D_\lambda$ & \sig & SGDiff & 0.094 (5/6) & 0.054 & 0.054 \\
WorldView-III $D_s$ & \sig & SGDiff & 0.031 (6/6) & 0.010 & 0.010 \\
WorldView-III HQNR & \sig & SGDiff & 0.031 (6/6) & 0.007 & 0.007 \\
\midrule
WorldView-II ERGAS & \sig & SGDiff & 0.125 (4/4) & 0.005 & 0.005 \\
WorldView-II SAM & \sig & SGDiff & 0.125 (4/4) & 0.002 & 0.002 \\
WorldView-II Q2$^n$ & \sig & SGDiff & 0.125 (4/4) & 0.013 & 0.013 \\
\bottomrule
\end{tabular}%
}
\end{table}

\subsection{The retrievability gap: systems, discordance, families, inversions and predictive value}\label{sec:gap_supp}
\cref{tab:systems} lists the $\fNSys$ systems of the comparison set with every measure on both tile sets, so that each statistic of Main Sec.~\MSecGap{} and Main Fig.~\MFigGap{} can be recomputed from it.

\noindent\textbf{Discordance.} \cref{tab:discordance} gives $\delta$ for every pair of a fidelity metric and a retrieval-facing measure, with system-bootstrap intervals. With a reference, Q2$^n$, SAM and ERGAS disagree with coverage, with held-out retrieval and with scene accuracy on $\fDiscRefLo$ to $\fDiscRefHi$ of the pairs. Without a reference, where local concepts are included, HQNR disagrees with coverage, retrieval and detection on $\fDiscHDet$ to $\fDiscH$, and the QuickBird replication gives $\fDiscQB$. LPIPS, which reads deep features, disagrees on $\fDiscLPAcc$ to $\fDiscLPMoc$. Because $\delta$ is a rank statistic, no normalisation enters it. \cref{tab:spread} nevertheless reports the spread of every measure in three unit systems: in multiples of the seed standard deviation Q2$^n$ and coverage resolve the systems comparably ($\fSigQ$ and $\fSigC$), and the paper makes no claim about the relative size of the two spreads.

\noindent\textbf{Two families.} \cref{tab:families} separates the pixel-supervised from the language-conditioned systems. Inside the pixel-supervised family the fidelity index and coverage agree on WorldView-III with and without a reference and on QuickBird ($\rho{=}\fRhoPix$, $\fRhoPixFR$ and $\fRhoPixQB$), and the family indicator is worth $+\fFamOff$, $+\fFamOffFR$ and $+\fFamOffQB$ coverage at equal fidelity. Counting SGDiff as language-conditioned leaves the picture unchanged ($\rho{=}\fAltRhoPix$, offset $+\fAltFamOff$). \cref{tab:inversions} lists the $\fNInvSig$ seed-robust rank inversions ($50$ strict inversions and $6$ ties over the $153$ pairs; $7$ strict and $2$ tied inside the pixel-supervised family, $8$ with ties counted one half, the count of Main Sec.~\MSecGap; ties count one half in every $\delta$): SeeSR is the more retrievable system in $\fRobSeeSR$ of them, CLIPPan in $\fRobCLIPPan$, the CLIPDenoising-style design in $\fRobCDS$ and DA-CLIP in $\fRobDACLIP$, and the system with the higher Q2$^n$ is pixel-supervised in all.

Two properties of the comparison set bear on this result. First, coverage is a recall, and SeeSR, the more retrievable system in $\fRobSeeSR$ of the $\fNInvSig$ inversions, has the highest hallucinated-concept rate ($\fHalSeeSR$) and the lowest concept precision ($\fPcSeeSR$) that we measure (\cref{tab:precision,tab:downstream_rr}), so its recall overstates what it preserves. The last rows of \cref{tab:families} repeat the analysis without it: $\delta(\mathrm{Q2}^n,\cdino)$ is $\fDiscQNoSeeSR$ instead of $\fDiscQ$ and the rank correlation over all systems $\fRhoAllNoSeeSR$ instead of $\fRhoAll$, the family indicator remains worth $+\fFamOffNoSeeSR$ coverage with a reference and $+\fFamOffFRNoSeeSR$ without one, and the $\fNInvNoSeeSR$ robust inversions that do not involve SeeSR are again between the two families. Second, the language-conditioned family is small and partly of our making: three of its \fNLangW{} systems are conditioning designs that we implemented on the U-Net backbone with the same frozen encoder, DA-CLIP and SeeSR are official RGB implementations whose input layer was adapted to multispectral data, so their Q2$^n$ may in part reflect the adaptation, and CLIPPan is the only native language-conditioned pansharpener; the family offset is therefore a statement about these six systems and should be read with the SeeSR caveat above. The three designs give the comparison that is free of this effect: they share architecture, data and training recipe with the plain U-Net and reach $\fCtrlDcLo$--$\fCtrlDcHi$ more coverage at a Q2$^n$ higher by $\fCtrlDqLo$--$\fCtrlDqHi$. Removing SeeSR and the three designs together leaves CLIPPan and DA-CLIP in the family: the three robust inversions that involve them ($\fRobCLIPPan$ and $\fRobDACLIP$) are cross-family, and the family coefficient over the pixel-supervised trend is $+0.05$ coverage with a reference.

Inside the language-conditioned family itself, fidelity and coverage are unrelated with a reference ($\rho{=}\fRhoLang$) and opposed without one ($\rho{=}{\fRhoLangFR}$ on WorldView-III and ${\fRhoLangQB}$ on QuickBird). With a reference the absence of a relation is due to SeeSR alone, the other five systems being ordered alike by the two measures ($\rho{=}\fRhoLangNoSeeSR$); without a reference the relation remains negative when SeeSR is removed ($\rho{=}{\fRhoLangFRNoSeeSR}$). Within that family retrievability is thus partly traded against fidelity, most clearly for the diffusion prior, as the perception--distortion trade-off~\cite{blau2018perception} predicts for generative models, and the closeness of LPIPS to coverage in \cref{tab:discordance} is in line with this reading. With five or six systems these within-family correlations are descriptive. The measurement side of the gap is therefore a statement about the two families. The fidelity indices order systems reliably inside the pixel-supervised family; across the families, and inside the language-conditioned one, they do not say which output an index can still read.

\noindent\textbf{Predictive value.} \cref{tab:partial} asks what coverage adds to the pixel side. Controlling for Q2$^n$, the partial correlation of coverage with held-out retrieval, scene accuracy and detection is $\fPcCMocQ$, $\fPcCAccQ$ and $\fPcCDetQ$ (system-bootstrap $95\%$ intervals $[0.86,\,0.99]$, $[0.82,\,0.98]$ and $[0.62,\,0.98]$); controlling for LPIPS, the strongest pixel-side predictor, it is $\fPcCMocLP$, $\fPcCAccLP$ and $\fPcCDetLP$, with intervals $[0.41,\,0.95]$, $[0.02,\,0.86]$ and $[-0.21,\,0.82]$ that are wide because LPIPS and coverage are themselves strongly correlated ($\rho{=}\fRhoLPC$) and only $\fNSys$ systems are available; the partial-$F$ tests of \cref{tab:partial} agree ($p{<}10^{-3}$, $0.02$ and $0.07$). In the other direction, LPIPS retains $\fPcLPMocC$, $\fPcLPAccC$ and $\fPcLPDetC$ given coverage (intervals $[-0.51,\,0.80]$, $[-0.15,\,0.89]$, $[-0.02,\,0.97]$), and Q2$^n$ retains $\fPcQMocC$, $\fPcQAccC$ and $\fPcQDetC$. The leave-one-system-out analysis agrees: the three reference metrics have limited out-of-sample predictive power ($R^2\le\fLooPixHi$), LPIPS alone reaches $\fLooLPLo$--$\fLooLPHi$, coverage alone $\fLooCovLo$--$\fLooCovHi$, and the two together $\fLooLPCovLo$--$\fLooLPCovHi$. Coverage therefore subsumes LPIPS for retrieval and complements it for recognition; for detection the two are comparably predictive, LPIPS alone ($0.86$) slightly ahead of coverage alone ($0.83$) out of sample and the addition of coverage to LPIPS only marginally significant ($p{=}0.07$), which is why the main text claims predictive value for detection given Q2$^n$ and not given LPIPS. Label-grounded coverage needs no reference image, but it does need concept annotations of the tile, which an operational archive lacks as well; it is an evaluation measure, not an index that an archive can compute. At full resolution it remains the stronger predictor, with partial correlations of $\fPcCDetH$ and $\fPcCMocH$ given HQNR. The detection outcome uses the ConvNeXt-T head so that no backbone is shared with the probe; with the DINOv2 head the partial correlation given LPIPS is $\fPcCDetDinoLP$. In the top block detection is measured on the full-resolution tiles and coverage on the reduced-resolution tiles, so these correlations are across tile sets, while the retrieval and accuracy outcomes are on the same tiles as coverage; the middle block repeats the analysis within the full-resolution set. The detection column of the middle block is the least independent entry of the table: the annotated concept set of a full-resolution tile includes the classes of its boxes (\cref{sec:vocab}), and these are the classes on which detection is scored. The top block does not share this dependence, since the reduced-resolution tiles carry no boxes. The no-reference perceptual indices of \cref{tab:precision} are available for eight methods and do not enter the analysis over the $\fNSys$ systems.

\noindent\textbf{The remaining shortfall.} The most retrievable system of the comparison set reaches a coverage of $\fBestC$ with a reference and $\fBestCFR$ without one, against $\fRefC$ for the reference tiles. \our reaches $\fOursC$ and $\fOursCFR$, which closes $\fClosedPct\%$ of the distance between the best baseline and the reference. \cref{tab:extent} stratifies coverage by tile extent and ground sampling distance.

\begin{table*}[!tbp]
\centering
\caption{The comparison set of Main Sec.~\MSecGap{} and Main Fig.~\MFigGap: every trained system that contains no component of our method ($12$ pixel-supervised, $6$ language-conditioned), with \our beside it. Left block, reduced-resolution set (reference available): Q2$^n$ and SAM on the $20$ scenes; LPIPS, label-grounded coverage over meso and global concepts, OpenCLIP mAP on the $40$ meso and global queries and ConvNeXt-B scene accuracy on the $180$ tiles cut from them. Right block, full-resolution set (no reference): HQNR on the $20$ scenes; coverage over all three scales (local concepts included), OpenCLIP mAP on $60$ queries, and detection mAP$_{50}$ under the ConvNeXt-T head (the detection outcome of every correlation analysis; it shares no backbone with the coverage probe) and under the DINOv2 head on the $980$ tiles. Bottom: the two ends of the input-to-reference range and PAN-only. Every statistic of Main Sec.~\MSecGap{} can be recomputed from this table.}
\label{tab:systems}
\scriptsize
\setlength{\tabcolsep}{2.8pt}
\maxsizebox{\textwidth}{%
\begin{tabular}{>{\raggedright\arraybackslash}p{.154\linewidth}>{\raggedright\arraybackslash}p{.081\linewidth}>{\raggedright\arraybackslash}p{.096\linewidth}|cccccc|ccccc}
\toprule
 & & & \multicolumn{6}{c|}{reduced resolution ($1.24$\,m)} & \multicolumn{5}{c}{full resolution ($0.31$\,m)} \\
System & family & kind & Q2$^n$ & SAM & LPIPS & $\cdino_{\mathrm{RR}}$ & mAP$_{\mathrm{OC}}$ & acc.$_{\mathrm{CNX}}$ & HQNR & $\cdino_{\mathrm{FR}}$ & mAP$_{\mathrm{OC}}$ & det.$_{\mathrm{CNX\text{-}T}}$ & det.$_{\mathrm{DINO}}$ \\
\midrule
PanNet & pixel & published & 0.890 & 3.63 & 0.249 & 0.58 & 0.41 & 68.9 & 0.936 & 0.53 & 0.371 & 29.9 & 29.4 \\
FusionNet & pixel & published & 0.903 & 3.34 & 0.236 & 0.62 & 0.42 & 70.6 & 0.940 & 0.57 & 0.378 & 30.3 & 30.0 \\
LAGConv & pixel & published & 0.909 & 3.12 & 0.229 & 0.64 & 0.43 & 71.8 & 0.923 & 0.60 & 0.383 & 30.9 & 30.4 \\
PanFormer & pixel & published & 0.909 & 3.10 & 0.231 & 0.61 & 0.42 & 70.1 & 0.908 & 0.56 & 0.372 & 30.0 & 29.8 \\
H-Trans & pixel & published & 0.911 & 3.02 & 0.224 & 0.67 & 0.44 & 72.9 & 0.934 & 0.63 & 0.394 & 31.6 & 31.5 \\
CrossDiff & pixel & published & 0.891 & 3.64 & 0.238 & 0.59 & 0.41 & 69.5 & 0.940 & 0.58 & 0.380 & 30.6 & 30.2 \\
SSDiff & pixel & published & 0.915 & 2.85 & 0.212 & 0.66 & 0.43 & 73.0 & 0.956 & 0.61 & 0.388 & 31.5 & 31.4 \\
Pan-Mamba & pixel & published & 0.919 & 2.92 & 0.214 & 0.69 & 0.45 & 74.4 & 0.943 & 0.64 & 0.407 & 32.1 & 32.0 \\
CANConv & pixel & published & 0.920 & 2.94 & 0.218 & 0.67 & 0.43 & 73.6 & 0.950 & 0.61 & 0.397 & 31.9 & 31.6 \\
ARConv & pixel & published & 0.921 & 2.90 & 0.209 & 0.68 & 0.44 & 73.3 & 0.957 & 0.63 & 0.403 & 32.7 & 32.3 \\
SGDiff & pixel & published & 0.923 & 2.78 & 0.206 & 0.72 & 0.46 & 75.6 & 0.960 & 0.67 & 0.417 & 33.2 & 33.0 \\
CLIPPan & language & adapted, official & 0.899 & 3.21 & 0.223 & 0.69 & 0.45 & 72.6 & 0.958 & 0.63 & 0.407 & 31.8 & 31.9 \\
DA-CLIP & language & adapted, official & 0.911 & 3.07 & 0.208 & 0.72 & 0.46 & 75.0 & 0.929 & 0.68 & 0.427 & 33.4 & 33.3 \\
SeeSR & language & adapted, official & 0.886 & 3.41 & 0.191 & 0.75 & 0.47 & 75.1 & 0.917 & 0.72 & 0.433 & 33.9 & 33.8 \\
Plain U-Net & pixel & same backbone & 0.910 & 3.10 & 0.233 & 0.60 & 0.41 & 70.4 & 0.925 & 0.55 & 0.370 & 31.0 & 30.9 \\
DenseCLIP-style & language & same backbone & 0.917 & 2.99 & 0.209 & 0.73 & 0.45 & 74.9 & 0.933 & 0.68 & 0.414 & 32.6 & 32.3 \\
CLIPDenoising-style & language & same backbone & 0.912 & 3.06 & 0.213 & 0.71 & 0.45 & 74.3 & 0.929 & 0.66 & 0.409 & 32.3 & 31.9 \\
DA-CLIP-style & language & same backbone & 0.915 & 3.02 & 0.205 & 0.74 & 0.46 & 75.3 & 0.932 & 0.69 & 0.423 & 32.8 & 32.4 \\
\midrule
\rowcolor{ourrow}\our & closed loop & ours & 0.929 & 2.68 & 0.176 & 0.87 & 0.53 & 82.1 & 0.965 & 0.85 & 0.503 & 38.4 & 40.2 \\
\midrule
up-sampled input (EXP) & & endpoint & 0.627 & 5.80 & 0.281 & 0.44 & 0.30 & 63.2 & 0.862 & 0.42 & & & \\
PAN-only & & reference point & -- & -- & 0.274 & 0.49 & 0.33 & 67.5 & -- & 0.50 & & & \\
reference tiles & & endpoint & 1.000 & 0.00 & 0.000 & 0.93 & 0.58 & 85.8 & -- & -- & & & \\
\bottomrule
\end{tabular}%
}
\end{table*}

\begin{ReportFlowTable}
\setlength{\LTpre}{4pt}\setlength{\LTpost}{4pt}
\setlength{\tabcolsep}{18.37pt}
\fontsize{9}{10.1}\selectfont
\begin{longtable}{lccc}
\caption{Discordance $\delta$ (Main Sec.~\MSecGap) of each fidelity metric with each retrieval-facing measure over the $18$ systems of \cref{tab:systems}: the share of system pairs that the two order oppositely, i.e.\ the probability that the better of two systems by the fidelity metric is the worse by the measure ($0$: same order; $0.5$: unrelated orders). Brackets: $95\%$ interval of a bootstrap over the systems ($2{,}000$ resamples); the bootstrap over the $20$ scenes gives $[0.33,\,0.39]$ for $\delta(\mathrm{Q2}^n,\cdino_{\mathrm{RR}})$. $\delta$ is invariant under any monotone re-scaling of either axis, so no normalisation enters. LPIPS is the pixel-side reference point; it needs a reference image and is therefore unavailable at full resolution.}
\label{tab:discordance}\\
\endfirsthead
\multicolumn{4}{l}{\footnotesize\itshape Table~\thetable\ (continued)}\\[2pt]
\endhead
\bottomrule\endfoot
\toprule
\multicolumn{4}{l}{\emph{WorldView-III, reduced-resolution set, $180$ tiles (meso and global concepts)}} \\*
 & coverage $\cdino_{\mathrm{RR}}$ & mAP, OpenCLIP & accuracy, ConvNeXt-B \\*
\midrule\noalign{\nobreak}
Q2$^n$ & 0.35 $[0.17,\,0.56]$ & 0.37 $[0.19,\,0.58]$ & 0.28 $[0.11,\,0.48]$ \\*
SAM & 0.37 $[0.20,\,0.57]$ & 0.38 $[0.21,\,0.59]$ & 0.31 $[0.15,\,0.50]$ \\*
ERGAS & 0.39 $[0.22,\,0.60]$ & 0.40 $[0.22,\,0.62]$ & 0.33 $[0.17,\,0.53]$ \\*
LPIPS (reference point) & 0.10 $[0.03,\,0.20]$ & 0.12 $[0.06,\,0.24]$ & 0.09 $[0.03,\,0.19]$ \\
\midrule
\multicolumn{4}{l}{\emph{WorldView-III, full-resolution set, $980$ tiles (local concepts included)}} \\*
 & coverage $\cdino_{\mathrm{FR}}$ & mAP, OpenCLIP & detection, ConvNeXt-T \\*
\midrule\noalign{\nobreak}
HQNR & 0.49 $[0.29,\,0.70]$ & 0.47 $[0.26,\,0.69]$ & 0.47 $[0.26,\,0.69]$ \\
\midrule
\multicolumn{4}{l}{\emph{QuickBird, annotated full-resolution subset, $490$ tiles (local concepts included)}} \\*
\midrule\noalign{\nobreak}
HQNR & 0.38 $[0.23,\,0.58]$ & & \\
\end{longtable}
\end{ReportFlowTable}

\begin{ReportFlowTable}
\setlength{\LTpre}{4pt}\setlength{\LTpost}{4pt}
\setlength{\tabcolsep}{17.06pt}
\fontsize{9}{10.1}\selectfont
\begin{longtable}{lccc}
\caption{Spread of each measure across the $18$ systems in three unit systems: raw units, percentage of the range between the up-sampled input and the reference tiles on the same set, and multiples of the median seed standard deviation (Q2$^n$: $0.0029$; coverage: $0.011$). The share of the input-to-reference range depends on how close a metric is to saturation and on the choice of end points, so it is given for description only; in units of seed noise the two axes resolve the systems comparably. The conclusions of Main Sec.~\MSecGap{} therefore rest on the rank-based statistics of \cref{tab:discordance,tab:families}, which involve no normalisation.}
\label{tab:spread}\\
\toprule
Measure & raw spread & \% of input--reference range & seed std units \\
\midrule
\endfirsthead
\multicolumn{4}{l}{\footnotesize\itshape Table~\thetable\ (continued)}\\[2pt]
\toprule
Measure & raw spread & \% of input--reference range & seed std units \\
\midrule
\endhead
\bottomrule\endfoot
Q2$^n$ & 0.037 & 9.9 & 12.5 \\
SAM & 0.857 & 14.8 & -- \\
ERGAS & 0.679 & 9.5 & -- \\
LPIPS & 0.058 & 20.6 & -- \\
\midrule
coverage $\cdino_{\mathrm{RR}}$ & 0.17 & 34.7 & 15.5 \\
mAP, OpenCLIP & 0.06 & 21.4 & -- \\
accuracy, ConvNeXt-B (points) & 6.7 & 29.6 & -- \\
\end{longtable}
\end{ReportFlowTable}

\begin{ReportFlowTable}
\setlength{\LTpre}{4pt}\setlength{\LTpost}{4pt}
\setlength{\tabcolsep}{4.68pt}
\begin{longtable}{>{\raggedright\arraybackslash}p{.235\linewidth}cccccccccc}
\caption{The two families of the comparison set: $12$ pixel-supervised systems (the eleven published networks and the plain U-Net) and $6$ language-conditioned systems (CLIPPan, DA-CLIP, SeeSR and the three same-backbone conditioning controls). Columns: Spearman's $\rho$ between the fidelity metric and coverage over all systems and inside each family; discordance $\delta$ over all systems and inside the pixel-supervised family; inverted pairs inside the pixel-supervised family and across the two families; $R^2$ of a linear model of coverage on the fidelity metric alone and with a family indicator, and the coefficient of that indicator, i.e.\ the coverage of a language-conditioned system in excess of a pixel-supervised system of equal fidelity. Middle rows: SGDiff, which is supervised in pixel space but routes its experts with a description of the input, moved to the language-conditioned family; the $15$ seed-robust inversions of \cref{tab:inversions} remain cross-family under either assignment. Last rows: SeeSR removed. SeeSR is the more retrievable system in $10$ of the $15$ robust inversions and has the highest hallucinated-concept rate and the lowest concept precision that we measure (\cref{tab:precision,tab:downstream_rr}), so its recall overstates what it preserves; without it the family offset remains and the overall agreement between fidelity and coverage is higher.}
\label{tab:families}\\
\toprule
 & \multicolumn{3}{c}{Spearman $\rho$} & \multicolumn{2}{c}{discordance $\delta$} & \multicolumn{2}{c}{inverted pairs} & \multicolumn{3}{c}{\shortstack{coverage $\sim$ fidelity\\(+ family)}} \\
Pair of measures & all & pixel & language & all & pixel & \shortstack{pixel--\\pixel} & \shortstack{pixel--\\language} & $R^2$ & \shortstack{$R^2$ +\\family} & offset \\
\midrule
\endfirsthead
\multicolumn{11}{l}{\footnotesize\itshape Table~\thetable\ (continued)}\\[2pt]
\toprule
 & \multicolumn{3}{c}{Spearman $\rho$} & \multicolumn{2}{c}{discordance $\delta$} & \multicolumn{2}{c}{inverted pairs} & \multicolumn{3}{c}{\shortstack{coverage $\sim$ fidelity\\(+ family)}} \\
Pair of measures & all & pixel & language & all & pixel & \shortstack{pixel--\\pixel} & \shortstack{pixel--\\language} & $R^2$ & \shortstack{$R^2$ +\\family} & offset \\
\midrule
\endhead
\bottomrule\endfoot
Q2$^n$ vs.\ $\cdino_{\mathrm{RR}}$, WV-III reduced resolution & 0.39 & 0.91 & 0.03 & 0.35 & 0.12 & 8 / 66 & 38 / 72 & 0.12 & 0.72 & $+$0.087 \\
LPIPS vs.\ $\cdino_{\mathrm{RR}}$, WV-III reduced resolution & 0.92 & 0.92 & 0.94 & 0.10 & 0.10 & 6 / 66 & 8 / 72 & 0.87 & 0.92 & $+$0.032 \\
HQNR vs.\ $\cdino_{\mathrm{FR}}$, WV-III full resolution & $-$0.00 & 0.66 & $-$0.60 & 0.49 & 0.25 & 16 / 66 & 48 / 72 & 0.00 & 0.55 & $+$0.083 \\
HQNR vs.\ $\cdino$, QuickBird full resolution & 0.40 & 0.78 & $-$0.64 & 0.38 & 0.22 & 14 / 66 & 32 / 72 & 0.22 & 0.61 & $+$0.059 \\
\midrule
Q2$^n$ vs.\ $\cdino_{\mathrm{RR}}$, SGDiff counted as language-conditioned & 0.39 & 0.88 & 0.00 & 0.35 & 0.15 & 8 / 55 & 34 / 77 & 0.12 & 0.75 & $+$0.085 \\
HQNR vs.\ $\cdino_{\mathrm{FR}}$, SGDiff counted as language-conditioned & $-$0.00 & 0.56 & $-$0.55 & 0.49 & 0.30 & 16 / 55 & 44 / 77 & 0.00 & 0.63 & $+$0.084 \\
\midrule
Q2$^n$ vs.\ $\cdino_{\mathrm{RR}}$, SeeSR removed ($17$ systems) & 0.65 & 0.91 & 0.80 & 0.26 & 0.12 & 8 / 66 & 26 / 60 & 0.45 & 0.88 & $+$0.071 \\
HQNR vs.\ $\cdino_{\mathrm{FR}}$, SeeSR removed ($17$ systems) & 0.15 & 0.66 & $-$0.29 & 0.44 & 0.25 & 16 / 66 & 37 / 60 & 0.06 & 0.54 & $+$0.073 \\
\end{longtable}
\end{ReportFlowTable}

\begin{ReportFlowTable}
\setlength{\LTpre}{4pt}\setlength{\LTpost}{4pt}
\setlength{\tabcolsep}{36.42pt}
\begin{longtable}{llcc}
\caption{Replication of Main Sec.~\MSecGap{} on a second sensor: HQNR on the $20$ QuickBird full-resolution scenes and label-grounded coverage on the annotated QuickBird subset ($10$ scenes, $490$ tiles, all three scales) for the same $18$ systems. The statistics derived from these values are in \cref{tab:discordance,tab:families}.}
\label{tab:qb_systems}\\
\toprule
System & family & HQNR & $\cdino$ \\
\midrule
\endfirsthead
\multicolumn{4}{l}{\footnotesize\itshape Table~\thetable\ (continued)}\\[2pt]
\toprule
System & family & HQNR & $\cdino$ \\
\midrule
\endhead
\bottomrule\endfoot
PanNet & pixel & 0.850 & 0.52 \\
FusionNet & pixel & 0.894 & 0.55 \\
LAGConv & pixel & 0.853 & 0.58 \\
PanFormer & pixel & 0.887 & 0.55 \\
H-Trans & pixel & 0.909 & 0.60 \\
CrossDiff & pixel & 0.868 & 0.56 \\
SSDiff & pixel & 0.933 & 0.59 \\
Pan-Mamba & pixel & 0.916 & 0.61 \\
CANConv & pixel & 0.914 & 0.61 \\
ARConv & pixel & 0.923 & 0.62 \\
SGDiff & pixel & 0.933 & 0.64 \\
CLIPPan & language & 0.936 & 0.61 \\
DA-CLIP & language & 0.901 & 0.65 \\
SeeSR & language & 0.884 & 0.68 \\
Plain U-Net & pixel & 0.906 & 0.54 \\
DenseCLIP-style & language & 0.915 & 0.65 \\
CLIPDenoising-style & language & 0.909 & 0.63 \\
DA-CLIP-style & language & 0.913 & 0.66 \\
\midrule
\rowcolor{ourrow}\our & closed loop & 0.944 & 0.80 \\
\end{longtable}
\end{ReportFlowTable}

\begin{ReportFlowTable}
\setlength{\LTpre}{4pt}\setlength{\LTpost}{4pt}
\setlength{\tabcolsep}{3.41pt}
\begin{longtable}{llccccc}
\caption{Seed-robust rank inversions between pixel fidelity and coverage on the reduced-resolution set. Of the $153$ pairs of systems, $50$ are inverted (the system with the higher Q2$^n$ has the lower label-grounded coverage) and $6$ are tied on one measure; counting ties one half, as Main Sec.~\MSecGap{} does, gives the $53$ discordant pairs behind $\delta{=}0.35$. The $15$ listed here are those for which both differences exceed twice the pooled seed std of the pair and both scene-paired Wilcoxon tests reject after Holm correction over the inverted pairs ($p_{\mathrm{adj}}$). In every one of them the system with the higher coverage is language-conditioned and the system with the higher Q2$^n$ is pixel-supervised; inside the pixel-supervised family $7$ of $66$ pairs are inverted and $2$ tied ($8$ with ties counted one half), and none of them is robust.}
\label{tab:inversions}\\
\toprule
higher Q2$^n$ (pixel-supervised) & higher coverage (language-conditioned) & $\Delta$Q2$^n$ & $\Delta\cdino$ & $p_{\mathrm{Q}}$ & $p_{\mathrm{C}}$ & $p_{\mathrm{adj}}$ \\
\midrule
\endfirsthead
\multicolumn{7}{l}{\footnotesize\itshape Table~\thetable\ (continued)}\\[2pt]
\toprule
higher Q2$^n$ (pixel-supervised) & higher coverage (language-conditioned) & $\Delta$Q2$^n$ & $\Delta\cdino$ & $p_{\mathrm{Q}}$ & $p_{\mathrm{C}}$ & $p_{\mathrm{adj}}$ \\
\midrule
\endhead
\bottomrule\endfoot
Plain U-Net & SeeSR & 0.024 & 0.15 & $<10^{-4}$ & $<10^{-4}$ & $<0.001$ \\
PanFormer & SeeSR & 0.023 & 0.14 & $<10^{-4}$ & 0.0001 & 0.004 \\
FusionNet & SeeSR & 0.018 & 0.13 & $<10^{-4}$ & 0.0001 & 0.004 \\
LAGConv & SeeSR & 0.024 & 0.11 & $<10^{-4}$ & $<10^{-4}$ & $<0.001$ \\
SSDiff & SeeSR & 0.029 & 0.09 & $<10^{-4}$ & $<10^{-4}$ & $<0.001$ \\
Plain U-Net & CLIPPan & 0.011 & 0.09 & $<10^{-4}$ & 0.0001 & 0.004 \\
H-Trans & SeeSR & 0.025 & 0.08 & $<10^{-4}$ & 0.0001 & 0.004 \\
CANConv & SeeSR & 0.034 & 0.08 & 0.0001 & $<10^{-4}$ & 0.004 \\
ARConv & SeeSR & 0.035 & 0.07 & 0.0001 & 0.0001 & 0.004 \\
Pan-Mamba & SeeSR & 0.033 & 0.06 & $<10^{-4}$ & $<10^{-4}$ & $<0.001$ \\
LAGConv & CLIPPan & 0.011 & 0.05 & 0.0001 & $<10^{-4}$ & 0.005 \\
CANConv & CLIPDenoising-style & 0.008 & 0.04 & 0.0003 & 0.0001 & 0.009 \\
ARConv & DA-CLIP & 0.010 & 0.04 & $<10^{-4}$ & $<10^{-4}$ & $<0.001$ \\
ARConv & CLIPDenoising-style & 0.009 & 0.03 & $<10^{-4}$ & 0.0001 & 0.005 \\
SGDiff & SeeSR & 0.037 & 0.03 & $<10^{-4}$ & 0.0002 & 0.009 \\
\end{longtable}
\end{ReportFlowTable}

\begin{ReportFlowTable}
\setlength{\LTpre}{4pt}\setlength{\LTpost}{4pt}
\ReportEnableTableWidth
\setlength{\tabcolsep}{1.5pt}
\begin{longtable}{>{\raggedright\arraybackslash}p{.218\linewidth}ccc|cc|cc}
\caption{What coverage adds to the pixel side and what the pixel side adds to coverage, over the $18$ systems (\our excluded). Top: Spearman correlation of four downstream outcomes with label-grounded coverage $\cdino_{\mathrm{RR}}$, with Q2$^n$ and with LPIPS on the reduced-resolution set, and the partial correlations of each given the other. The detection outcome of the analysis is measured with the ConvNeXt-T head, which shares no backbone with the coverage probe; the DINOv2 head is listed beside it. Middle: the same analysis on the full-resolution set, where local concepts enter coverage and HQNR is the available fidelity index. Bottom: leave-one-system-out predictive $R^2$ of six linear models of each outcome (the three reference metrics Q2$^n$, SAM and ERGAS; LPIPS; all four pixel-side metrics; coverage; LPIPS with coverage; the four with coverage) and the partial-$F$ $p$-value of each addition in both directions. The three reference metrics have limited out-of-sample predictive power ($R^2\le\fLooPixHi$) and LPIPS carries most of that of the pixel side. Coverage adds to LPIPS and to the four pixel-side metrics for retrieval ($p{<}10^{-3}$) and scene accuracy ($p{=}0.02$ and $0.009$), and marginally for detection ($p{=}0.07$); these tests agree with the system-bootstrap intervals of the partial correlations quoted in \cref{sec:gap_supp}; LPIPS adds to coverage for detection, and the four metrics together add to coverage for scene accuracy.}
\label{tab:partial}\\
\toprule
Outcome (reduced-resolution predictors) & $\rho(\cdino,\cdot)$ & $\rho(\mathrm{Q2}^n,\cdot)$ & $\rho(\mathrm{LPIPS},\cdot)$ & $\cdino\mid\mathrm{Q2}^n$ & $\mathrm{Q2}^n\mid\cdino$ & $\cdino\mid\mathrm{LPIPS}$ & $\mathrm{LPIPS}\mid\cdino$ \\
\midrule
\endfirsthead
\multicolumn{8}{l}{\footnotesize\itshape Table~\thetable\ (continued)}\\[2pt]
\toprule
Outcome (reduced-resolution predictors) & $\rho(\cdino,\cdot)$ & $\rho(\mathrm{Q2}^n,\cdot)$ & $\rho(\mathrm{LPIPS},\cdot)$ & $\cdino\mid\mathrm{Q2}^n$ & $\mathrm{Q2}^n\mid\cdino$ & $\cdino\mid\mathrm{LPIPS}$ & $\mathrm{LPIPS}\mid\cdino$ \\
\midrule
\endhead
query mAP, OpenCLIP (same tiles) & 0.97 & 0.33 & 0.90 & 0.97 & $-$0.20 & 0.83 & 0.07 \\
scene accuracy, ConvNeXt-B (same tiles) & 0.94 & 0.58 & 0.94 & 0.95 & 0.69 & 0.55 & 0.58 \\
detection mAP$_{50}$, ConvNeXt-T head (full-resolution tiles) & 0.93 & 0.47 & 0.94 & 0.92 & 0.34 & 0.45 & 0.59 \\
detection mAP$_{50}$, DINOv2 head (full-resolution tiles) & 0.94 & 0.45 & 0.94 & 0.93 & 0.27 & 0.53 & 0.52 \\
\bottomrule\noalign{\penalty0\vskip4pt}
\multicolumn{8}{@{}l@{}}{\begin{minipage}{\linewidth}\centering
\begin{tabular}{lcc|cc}
\toprule
Outcome (full-resolution predictors, local concepts included) & $\rho(\cdino_{\mathrm{FR}},\cdot)$ & $\rho(\mathrm{HQNR},\cdot)$ & $\cdino_{\mathrm{FR}}\mid\mathrm{HQNR}$ & $\mathrm{HQNR}\mid\cdino_{\mathrm{FR}}$ \\
\midrule
detection mAP$_{50}$, ConvNeXt-T head & 0.93 & 0.08 & 0.93 & 0.21 \\
query mAP, OpenCLIP ($60$ queries) & 0.98 & 0.06 & 0.98 & 0.30 \\
detection mAP$_{50}$, DINOv2 head & 0.94 & 0.10 & 0.94 & 0.30 \\
\bottomrule
\end{tabular}
\end{minipage}}\\[3pt]
\multicolumn{8}{@{}l@{}}{\begin{minipage}{\linewidth}\centering
\begin{tabular}{>{\raggedright\arraybackslash}p{.21\linewidth}cccccc|cc|cc}
\toprule
 & \multicolumn{6}{c|}{leave-one-out $R^2$ of the model} & \multicolumn{2}{c|}{\shortstack{$p$ of adding\\coverage to}} & \multicolumn{2}{c}{\shortstack{$p$ of adding\\to coverage}} \\
Outcome & \shortstack{Q2$^n$, SAM,\\ERGAS} & LPIPS & \shortstack{four\\pixel-side} & coverage & \shortstack{LPIPS +\\coverage} & \shortstack{four +\\coverage} & LPIPS & the four & LPIPS & the four \\
\midrule
query mAP, OpenCLIP & 0.28 & 0.76 & 0.45 & 0.93 & 0.93 & 0.92 & $<10^{-3}$ & $<10^{-3}$ & 0.561 & 0.435 \\
scene accuracy, ConvNeXt-B & 0.47 & 0.82 & 0.73 & 0.93 & 0.92 & 0.96 & 0.021 & 0.009 & 0.213 & 0.002 \\
detection, ConvNeXt-T head & 0.35 & 0.86 & 0.77 & 0.83 & 0.87 & 0.85 & 0.070 & 0.072 & 0.021 & 0.079 \\
\bottomrule
\end{tabular}
\end{minipage}}\\[3pt]
\end{longtable}
\end{ReportFlowTable}

\begin{ReportFlowTable}
\begin{longtable}{>{\raggedright\arraybackslash}p{\dimexpr.55\linewidth-2\tabcolsep\relax}*{4}{>{\centering\arraybackslash}p{\dimexpr.1125\linewidth-2\tabcolsep\relax}}}
\caption{Label-grounded coverage by tile extent and ground sampling distance on the annotated WorldView-III scenes (same annotations, same DINOv2 probe; the target set of a window is the union over its tiles). Local, meso and global are defined relative to the tile, so the gain of \our over SGDiff is stable across a four-fold change of extent and of GSD; the reduced-resolution row carries meso and global concepts only, which is why its absolute values are higher than those of the full-resolution tiles.}
\label{tab:extent}\\
\toprule
Unit & $n$ & SGDiff & \our & gain \\
\midrule
\endfirsthead
\multicolumn{5}{l}{\footnotesize\itshape Table~\thetable\ (continued)}\\[2pt]
\toprule
Unit & $n$ & SGDiff & \our & gain \\
\midrule
\endhead
\bottomrule\endfoot
$128$-px tiles, full resolution ($0.31$\,m, $40$\,m side) & 980 & 0.67 & \best{0.85} & $+$0.18 \\
$256$-px windows, full resolution ($79$\,m side; stride $128$) & 180 & 0.70 & \best{0.86} & $+$0.16 \\
$512^2$ scenes, full resolution ($159$\,m side) & 20 & 0.72 & \best{0.86} & $+$0.14 \\
$128$-px tiles, reduced resolution ($1.24$\,m, $159$\,m side; meso and global) & 180 & 0.72 & \best{0.87} & $+$0.15 \\
\end{longtable}
\end{ReportFlowTable}

\subsection{Reference-based measures, retrieval by granularity and evaluator families}\label{toc:supp:5:3}
\begin{figure}[!htbp]
\centering
\includegraphics[width=\columnwidth]{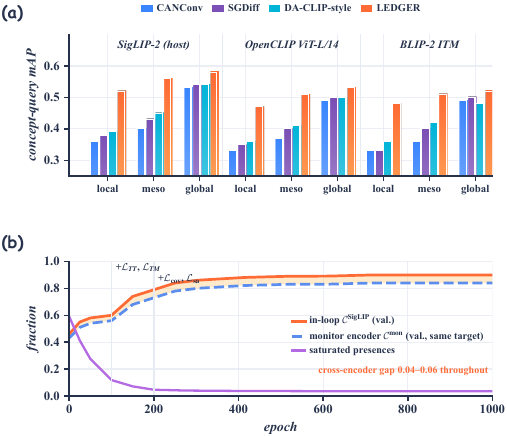}
\caption{(a) Concept-query mAP by query granularity under the host SigLIP-2 index, OpenCLIP and the non-contrastive BLIP-2 matching head on the annotated full-resolution tiles. (b) Training dynamics on the validation split: in-loop coverage, monitor-encoder coverage with the same target, their gap, and the saturated-presence fraction that fixes the warm-up schedule.}
\label{fig:analysis}
\end{figure}

\begin{figure}[!htbp]
\centering
\includegraphics[width=\columnwidth]{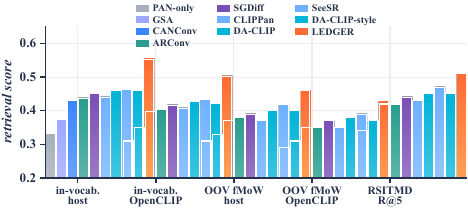}
\caption{Retrieval inside the training concept space ($60$ queries on the annotated full-resolution WorldView-III tiles) and on the two external benchmarks ($30$ fMoW queries; RSITMD captions; reduced-resolution protocol).}
\label{fig:vocab}
\end{figure}

\cref{tab:downstream_rr} gives every downstream measure for every method of the study, including the reduced-resolution tiles where a reference exists, scene accuracy under the ConvNeXt-B head, and all seven retrieval evaluators. \cref{tab:retrieval_scale} completes the retrieval rows by granularity with the bootstrap std of each entry: the gain of \our concentrates on local and meso queries under every evaluator, including the two outside the CLIP and DINOv2 families, and every held-out evaluator preserves the ranking of the methods (\cref{fig:analysis}a). \cref{tab:vocab_split} and \cref{fig:vocab} separate the training concept space from the three external benchmarks under the host, OpenCLIP and ConvNeXt-B. On the external benchmarks the gain is $+0.04$ to $+0.05$ in mAP and R@5 and $+0.02$ to $+0.03$ in R@1 under every evaluator, with $95\%$ intervals that exclude zero in every row, the same under the host, under OpenCLIP and under a ConvNeXt-B head fitted on fMoW's own labels: its size is set by the concept space of the index, not by the backbone that reads the output. The $+\fRetHeldLo$ to $+\fRetHeldHi$ points measured with the query bank under held-out backbones are accordingly gains inside the concept space on which the index was fitted, and the external gains are the part of them that the index carries to concepts and captions it was never fitted to. \cref{tab:pooled} reports the pooled gallery under the held-out OpenCLIP retriever and under the host. Under OpenCLIP, \our supplies $\fPoolShareOC\%$ of the top-ten entries and $\fPoolPrecOC$ of its entries are relevant, against $\fPoolPrecOthLo$--$\fPoolPrecOthHi$ for the other systems; it thus holds $\fPoolRelOC\%$ of the relevant entries and $\fPoolIrrOC\%$ of the non-relevant ones, the latter matching the uniform share of $\fPoolUniform\%$. A mixed archive therefore receives more relevant tiles from \our and no more non-relevant ones than from any other system. Under the host, which is in the loop, the share is $\fPoolShareHost\%$ and the share of non-relevant entries $\fPoolIrrHost\%$; the difference between the two retrievers is the in-loop component, which is why no host quantity is used as evidence. \cref{tab:swap} measures the same component from the other side: when the encoder in the loop is replaced, each retriever scores the model trained with it $\fSwapSurLo$--$\fSwapSurHi$ higher than the models trained with another encoder, while the margin of the SigLIP-2 model over the plain U-Net under the two retrievers it never saw is $\fSwapMarLo$--$\fSwapMarHi$, several times the in-loop component. \cref{tab:qb_real} repeats the real-scene evaluation on the annotated QuickBird subset; \cref{tab:retadapt} gives the retriever-adaptation alternative in full.

\begin{ReportFlowTable}
\setlength{\tabcolsep}{1.25pt}
\begin{longtable}{>{\raggedright\arraybackslash}p{\dimexpr.226\linewidth-2\tabcolsep\relax}*{9}{>{\centering\arraybackslash}p{\dimexpr.058\linewidth-2\tabcolsep\relax}}>{\columncolor{ourrow}}>{\centering\arraybackslash}p{\dimexpr.090\linewidth-2\tabcolsep\relax}>{\centering\arraybackslash}p{\dimexpr.070\linewidth-2\tabcolsep\relax}>{\centering\arraybackslash}p{\dimexpr.092\linewidth-2\tabcolsep\relax}}
\caption{Downstream measures for every method of the study. Top: the $180$ annotated reduced-resolution tiles, where a reference exists (scene accuracy under an ImageNet-supervised ConvNeXt-B head). Bottom: the labelled aerial benchmark (LPIPS; detection mAP$_{50}$ over its $15$ classes, per class in \cref{tab:dota}; small-object AP$_{50}$ over the boxes below $32^2$\,px at the output resolution) and the $980$ full-resolution tiles under both detection heads and all seven retrieval evaluators. $\Delta$: \our against the strongest baseline of the row; last column: reference tiles, seed std of \our or, for the retrieval rows, the query-bootstrap std of its entry. KID~\cite{binkowski2018kid} replaces FID, which is biased at this sample size.}
\label{tab:downstream_rr}\\
\toprule
Measure & \shortstack{PAN-\\only} & GSA & \shortstack{CAN\\Conv} & \shortstack{AR\\Conv} & \shortstack{SG\\Diff} & \shortstack{CLIP\\Pan} & \shortstack{DA-\\CLIP} & SeeSR & \shortstack{DA-\\CLIP-\\style} & \our & $\Delta$ & Ref.\,/\,std \\
\midrule
\endfirsthead
\multicolumn{13}{l}{\footnotesize\itshape Table~\thetable\ (continued)}\\[2pt]
\toprule
Measure & \shortstack{PAN-\\only} & GSA & \shortstack{CAN\\Conv} & \shortstack{AR\\Conv} & \shortstack{SG\\Diff} & \shortstack{CLIP\\Pan} & \shortstack{DA-\\CLIP} & SeeSR & \shortstack{DA-\\CLIP-\\style} & \our & $\Delta$ & Ref.\,/\,std \\
\midrule
\endhead
\bottomrule\endfoot
\multicolumn{13}{>{\raggedright\arraybackslash}p{\dimexpr\linewidth-2\tabcolsep\relax}}{\emph{Annotated WorldView-III scenes, reduced resolution: $180$ tiles at $1.24$\,m (reference available; meso and global concepts and queries)}} \\*
\midrule
LPIPS$\downarrow$ & 0.274 & 0.247 & 0.218 & 0.209 & 0.206 & 0.223 & 0.208 & 0.191 & 0.205 & \best{0.176} & $-$0.015 & 0.000 \\
KID$\downarrow$ ($\times10^{3}$) & 38.2 & 30.1 & 21.4 & 20.6 & 18.9 & 22.0 & 19.4 & 15.8 & 19.6 & \best{13.1} & $-$2.7 & 0.0 \\
Coverage $\cdino_{\mathrm{RR}}$ & 0.49 & 0.56 & 0.67 & 0.68 & 0.72 & 0.69 & 0.72 & 0.75 & 0.74 & \best{0.87} & $+$0.12 & 0.93 \\
mAP ($40$ queries), SigLIP-2 (host) & 0.36 & 0.40 & 0.46 & 0.47 & 0.48 & 0.48 & 0.49 & 0.51 & 0.49 & \best{0.57} & $+$0.06 & 0.62 \\
mAP ($40$ queries), OpenCLIP & 0.33 & 0.37 & 0.43 & 0.44 & 0.46 & 0.45 & 0.46 & 0.47 & 0.46 & \best{0.53} & $+$0.06 & 0.58 \\
Scene accuracy (\%), ConvNeXt-B (ImageNet) & 67.5 & 69.3 & 73.6 & 73.3 & 75.6 & 72.6 & 75.0 & 75.1 & 75.3 & \best{82.1} & $+$6.5 & 85.8 \\
\midrule
\multicolumn{13}{>{\raggedright\arraybackslash}p{\dimexpr\linewidth-2\tabcolsep\relax}}{\emph{Labelled aerial benchmark and full-resolution tiles (local concepts included)}} \\*
\midrule
LPIPS$\downarrow$, labelled aerial benchmark & 0.262 & 0.241 & 0.214 & 0.209 & 0.203 & 0.219 & 0.206 & 0.188 & 0.201 & \best{0.172} & $-$0.016 & 0.000 \\
Detection mAP$_{50}$ (\%), labelled aerial benchmark & 29.9 & 32.5 & 38.3 & 38.8 & 39.9 & 38.9 & 40.1 & 40.3 & 39.6 & \best{47.1} & $+$6.8 & $\pm$0.9 \\
Small-object AP$_{50}$ (\%), labelled aerial benchmark & 17.6 & 20.9 & 29.6 & 30.3 & 31.4 & 28.5 & 31.8 & 32.9 & 31.5 & \best{40.8} & $+$7.9 & $\pm$1.0 \\
Scene accuracy (\%), full-resolution tiles & 70.3 & 73.1 & 76.9 & 77.4 & 78.0 & 76.4 & 78.6 & 79.2 & 78.2 & \best{84.9} & $+$5.7 & $\pm$0.6 \\
Detection mAP$_{50}$ (\%), DINOv2 head & 27.1 & 29.5 & 31.6 & 32.3 & 33.0 & 31.9 & 33.3 & 33.8 & 32.4 & \best{40.2} & $+$6.4 & $\pm$0.9 \\
Detection mAP$_{50}$ (\%), ConvNeXt-T head & 27.4 & 29.8 & 31.9 & 32.7 & 33.2 & 31.8 & 33.4 & 33.9 & 32.8 & \best{38.4} & $+$4.5 & $\pm$0.8 \\
False positives per tile$\downarrow$ & 0.42 & 0.46 & 0.39 & 0.38 & 0.37 & 0.40 & 0.41 & 0.63 & 0.41 & \best{0.35} & $-$0.02 & $\pm$0.01 \\
Coverage $\cdino$, full-resolution tiles & 0.50 & 0.55 & 0.61 & 0.63 & 0.67 & 0.63 & 0.68 & 0.72 & 0.69 & \best{0.85} & $+$0.13 & $\pm$0.007 \\
Concept precision $\mathcal{P}_{\mathrm{c}}$ & 0.80 & 0.78 & 0.81 & 0.81 & 0.82 & 0.80 & 0.80 & 0.71 & 0.79 & \best{0.84} & $+$0.02 & $\pm$0.01 \\
mAP, SigLIP-2 (host) & 0.333 & 0.373 & 0.430 & 0.437 & 0.450 & 0.440 & 0.460 & 0.463 & 0.460 & \best{0.553} & $+$0.090 & $\pm$0.023 \\
mAP, OpenCLIP & 0.310 & 0.350 & 0.397 & 0.403 & 0.417 & 0.407 & 0.427 & 0.433 & 0.423 & \best{0.503} & $+$0.070 & $\pm$0.026 \\
mAP, RemoteCLIP & 0.320 & 0.353 & 0.420 & 0.427 & 0.430 & 0.427 & 0.450 & 0.453 & 0.447 & \best{0.540} & $+$0.087 & $\pm$0.025 \\
mAP, GeoRSCLIP & 0.323 & 0.360 & 0.427 & 0.430 & 0.450 & 0.440 & 0.447 & 0.453 & 0.450 & \best{0.547} & $+$0.094 & $\pm$0.026 \\
mAP, BLIP-2 ITM & 0.307 & 0.347 & 0.393 & 0.393 & 0.410 & 0.413 & 0.417 & 0.420 & 0.420 & \best{0.503} & $+$0.083 & $\pm$0.027 \\
mAP, ConvNeXt-B & 0.290 & 0.323 & 0.367 & 0.377 & 0.380 & 0.380 & 0.390 & 0.397 & 0.397 & \best{0.473} & $+$0.076 & $\pm$0.028 \\
mAP, Scale-MAE & 0.300 & 0.333 & 0.383 & 0.383 & 0.393 & 0.400 & 0.400 & 0.410 & 0.407 & \best{0.487} & $+$0.077 & $\pm$0.028 \\
\end{longtable}
\end{ReportFlowTable}

\begin{ReportFlowTable}
\setlength{\LTpre}{4pt}
\begin{longtable}{>{\raggedright\arraybackslash}p{\dimexpr.199\linewidth-2\tabcolsep\relax}>{\raggedright\arraybackslash}p{\dimexpr.171\linewidth-2\tabcolsep\relax}*{3}{>{\centering\arraybackslash}p{\dimexpr.09\linewidth-2\tabcolsep\relax}}>{\centering\arraybackslash}p{\dimexpr.18\linewidth-2\tabcolsep\relax}|>{\centering\arraybackslash}p{\dimexpr.18\linewidth-2\tabcolsep-.4pt\relax}}
\caption{Concept-query retrieval on the $980$ annotated full-resolution WorldView-III tiles by query granularity ($60$ queries) under all seven evaluators; the two language-free heads score the $39$ queries whose concepts carry labels (\cref{sec:evaluators}). Overall mAP is the mean of the three granularities, $\pm$ its query bootstrap ($1{,}000$ resamples; computed per entry). P@10 / MRR under the host retriever. ConvNeXt-B and Scale-MAE are multi-label heads outside the CLIP and DINOv2 families; a query is scored by the predicted probability of its concepts.}
\label{tab:retrieval_scale}\\
\toprule
Evaluator & Method & mAP$_{\mathrm{loc}}$ & mAP$_{\mathrm{meso}}$ & mAP$_{\mathrm{glob}}$ & mAP & P@10 / MRR \\
\midrule
\endfirsthead
\multicolumn{7}{l}{\footnotesize\itshape Table~\thetable\ (continued)}\\[2pt]
\toprule
Evaluator & Method & mAP$_{\mathrm{loc}}$ & mAP$_{\mathrm{meso}}$ & mAP$_{\mathrm{glob}}$ & mAP & P@10 / MRR \\
\midrule
\endhead
\bottomrule\endfoot
\multirow{12}{*}{\shortstack[l]{SigLIP-2\\(host)}}
& EXP & 0.22 & 0.28 & 0.44 & 0.313$\pm$0.022 & 0.36 / 0.41 \\*
& PAN-only & 0.30 & 0.31 & 0.39 & 0.333$\pm$0.025 & 0.41 / 0.46 \\*
& GSA & 0.32 & 0.35 & 0.45 & 0.373$\pm$0.024 & 0.44 / 0.49 \\*
& CANConv & 0.36 & 0.40 & 0.53 & 0.430$\pm$0.026 & 0.51 / 0.58 \\*
& ARConv & 0.37 & 0.41 & 0.53 & 0.437$\pm$0.025 & 0.52 / 0.59 \\*
& CLIPPan & 0.35 & 0.42 & 0.55 & 0.440$\pm$0.023 & 0.52 / 0.60 \\*
& Pan-Mamba & 0.37 & 0.42 & 0.53 & 0.440$\pm$0.024 & 0.52 / 0.60 \\*
& SGDiff & 0.38 & 0.43 & 0.54 & 0.450$\pm$0.024 & 0.53 / 0.61 \\*
& DA-CLIP & 0.39 & 0.44 & 0.55 & 0.460$\pm$0.026 & 0.55 / 0.62 \\*
& SeeSR & 0.40 & 0.45 & 0.54 & 0.463$\pm$0.026 & 0.56 / 0.63 \\*
& DA-CLIP-style & 0.39 & 0.45 & 0.54 & 0.460$\pm$0.025 & 0.55 / 0.63 \\*
& \cellcolor{ourrow}\best{\our} & \cellcolor{ourrow}\best{0.52} & \cellcolor{ourrow}\best{0.56} & \cellcolor{ourrow}\best{0.58} & \cellcolor{ourrow}\best{0.553$\pm$0.023} & 0.64 / 0.72 \\
\noalign{\penalty0}
\midrule
\multirow{12}{*}{\shortstack[l]{OpenCLIP ViT-L/14}}
& EXP & 0.20 & 0.26 & 0.41 & 0.290$\pm$0.021 &  \\*
& PAN-only & 0.28 & 0.29 & 0.36 & 0.310$\pm$0.020 &  \\*
& GSA & 0.30 & 0.33 & 0.42 & 0.350$\pm$0.021 &  \\*
& CANConv & 0.33 & 0.37 & 0.49 & 0.397$\pm$0.024 &  \\*
& ARConv & 0.34 & 0.37 & 0.50 & 0.403$\pm$0.026 &  \\*
& CLIPPan & 0.32 & 0.39 & 0.51 & 0.407$\pm$0.023 &  \\*
& Pan-Mamba & 0.34 & 0.39 & 0.49 & 0.407$\pm$0.025 &  \\*
& SGDiff & 0.35 & 0.40 & 0.50 & 0.417$\pm$0.025 &  \\*
& DA-CLIP & 0.36 & 0.41 & 0.51 & 0.427$\pm$0.024 &  \\*
& SeeSR & 0.37 & 0.42 & 0.51 & 0.433$\pm$0.024 &  \\*
& DA-CLIP-style & 0.36 & 0.41 & 0.50 & 0.423$\pm$0.023 &  \\*
& \cellcolor{ourrow}\best{\our} & \cellcolor{ourrow}\best{0.47} & \cellcolor{ourrow}\best{0.51} & \cellcolor{ourrow}\best{0.53} & \cellcolor{ourrow}\best{0.503$\pm$0.026} &  \\
\noalign{\penalty0}
\midrule
\multirow{12}{*}{\shortstack[l]{RemoteCLIP}}
& EXP & 0.21 & 0.27 & 0.42 & 0.300$\pm$0.024 &  \\*
& PAN-only & 0.29 & 0.30 & 0.37 & 0.320$\pm$0.022 &  \\*
& GSA & 0.31 & 0.32 & 0.43 & 0.353$\pm$0.024 &  \\*
& CANConv & 0.35 & 0.39 & 0.52 & 0.420$\pm$0.025 &  \\*
& ARConv & 0.36 & 0.40 & 0.52 & 0.427$\pm$0.023 &  \\*
& CLIPPan & 0.33 & 0.42 & 0.53 & 0.427$\pm$0.024 &  \\*
& Pan-Mamba & 0.36 & 0.40 & 0.51 & 0.423$\pm$0.023 &  \\*
& SGDiff & 0.36 & 0.42 & 0.51 & 0.430$\pm$0.023 &  \\*
& DA-CLIP & 0.38 & 0.43 & 0.54 & 0.450$\pm$0.025 &  \\*
& SeeSR & 0.40 & 0.43 & 0.53 & 0.453$\pm$0.026 &  \\*
& DA-CLIP-style & 0.37 & 0.45 & 0.52 & 0.447$\pm$0.027 &  \\*
& \cellcolor{ourrow}\best{\our} & \cellcolor{ourrow}\best{0.50} & \cellcolor{ourrow}\best{0.54} & \cellcolor{ourrow}\best{0.58} & \cellcolor{ourrow}\best{0.540$\pm$0.025} &  \\
\noalign{\penalty0}
\midrule
\multirow{12}{*}{\shortstack[l]{GeoRSCLIP}}
& EXP & 0.21 & 0.28 & 0.42 & 0.303$\pm$0.021 &  \\*
& PAN-only & 0.28 & 0.31 & 0.38 & 0.323$\pm$0.024 &  \\*
& GSA & 0.31 & 0.34 & 0.43 & 0.360$\pm$0.024 &  \\*
& CANConv & 0.36 & 0.40 & 0.52 & 0.427$\pm$0.023 &  \\*
& ARConv & 0.36 & 0.41 & 0.52 & 0.430$\pm$0.025 &  \\*
& CLIPPan & 0.35 & 0.42 & 0.55 & 0.440$\pm$0.024 &  \\*
& Pan-Mamba & 0.37 & 0.40 & 0.51 & 0.427$\pm$0.023 &  \\*
& SGDiff & 0.38 & 0.44 & 0.53 & 0.450$\pm$0.025 &  \\*
& DA-CLIP & 0.37 & 0.43 & 0.54 & 0.447$\pm$0.023 &  \\*
& SeeSR & 0.39 & 0.45 & 0.52 & 0.453$\pm$0.024 &  \\*
& DA-CLIP-style & 0.39 & 0.43 & 0.53 & 0.450$\pm$0.027 &  \\*
& \cellcolor{ourrow}\best{\our} & \cellcolor{ourrow}\best{0.52} & \cellcolor{ourrow}\best{0.56} & \cellcolor{ourrow}\best{0.56} & \cellcolor{ourrow}\best{0.547$\pm$0.026} &  \\
\noalign{\penalty0}
\midrule
\multirow{12}{*}{\shortstack[l]{BLIP-2 ITM}}
& EXP & 0.19 & 0.26 & 0.41 & 0.287$\pm$0.024 &  \\*
& PAN-only & 0.27 & 0.29 & 0.36 & 0.307$\pm$0.021 &  \\*
& GSA & 0.29 & 0.32 & 0.43 & 0.347$\pm$0.022 &  \\*
& CANConv & 0.33 & 0.36 & 0.49 & 0.393$\pm$0.024 &  \\*
& ARConv & 0.32 & 0.38 & 0.48 & 0.393$\pm$0.025 &  \\*
& CLIPPan & 0.33 & 0.39 & 0.52 & 0.413$\pm$0.024 &  \\*
& Pan-Mamba & 0.35 & 0.39 & 0.47 & 0.403$\pm$0.025 &  \\*
& SGDiff & 0.33 & 0.40 & 0.50 & 0.410$\pm$0.024 &  \\*
& DA-CLIP & 0.35 & 0.41 & 0.49 & 0.417$\pm$0.024 &  \\*
& SeeSR & 0.37 & 0.40 & 0.49 & 0.420$\pm$0.024 &  \\*
& DA-CLIP-style & 0.36 & 0.42 & 0.48 & 0.420$\pm$0.024 &  \\*
& \cellcolor{ourrow}\best{\our} & \cellcolor{ourrow}\best{0.48} & \cellcolor{ourrow}\best{0.51} & \cellcolor{ourrow}\best{0.52} & \cellcolor{ourrow}\best{0.503$\pm$0.027} &  \\
\noalign{\penalty0}
\midrule
\multirow{12}{*}{\shortstack[l]{ConvNeXt-B\\(ImageNet)}}
& EXP & 0.20 & 0.25 & 0.38 & 0.277$\pm$0.024 &  \\*
& PAN-only & 0.27 & 0.27 & 0.33 & 0.290$\pm$0.020 &  \\*
& GSA & 0.28 & 0.30 & 0.39 & 0.323$\pm$0.021 &  \\*
& CANConv & 0.31 & 0.34 & 0.45 & 0.367$\pm$0.024 &  \\*
& ARConv & 0.33 & 0.35 & 0.45 & 0.377$\pm$0.022 &  \\*
& CLIPPan & 0.32 & 0.36 & 0.46 & 0.380$\pm$0.023 &  \\*
& Pan-Mamba & 0.33 & 0.35 & 0.44 & 0.373$\pm$0.023 &  \\*
& SGDiff & 0.33 & 0.36 & 0.45 & 0.380$\pm$0.022 &  \\*
& DA-CLIP & 0.33 & 0.37 & 0.47 & 0.390$\pm$0.022 &  \\*
& SeeSR & 0.36 & 0.39 & 0.44 & 0.397$\pm$0.024 &  \\*
& DA-CLIP-style & 0.34 & 0.39 & 0.46 & 0.397$\pm$0.023 &  \\*
& \cellcolor{ourrow}\best{\our} & \cellcolor{ourrow}\best{0.45} & \cellcolor{ourrow}\best{0.47} & \cellcolor{ourrow}\best{0.50} & \cellcolor{ourrow}\best{0.473$\pm$0.028} &  \\
\noalign{\penalty0}
\midrule
\multirow{12}{*}{\shortstack[l]{Scale-MAE}}
& EXP & 0.22 & 0.25 & 0.39 & 0.287$\pm$0.021 &  \\*
& PAN-only & 0.27 & 0.28 & 0.35 & 0.300$\pm$0.021 &  \\*
& GSA & 0.29 & 0.31 & 0.40 & 0.333$\pm$0.024 &  \\*
& CANConv & 0.32 & 0.36 & 0.47 & 0.383$\pm$0.026 &  \\*
& ARConv & 0.32 & 0.37 & 0.46 & 0.383$\pm$0.022 &  \\*
& CLIPPan & 0.33 & 0.38 & 0.49 & 0.400$\pm$0.022 &  \\*
& Pan-Mamba & 0.34 & 0.38 & 0.47 & 0.397$\pm$0.025 &  \\*
& SGDiff & 0.34 & 0.38 & 0.46 & 0.393$\pm$0.024 &  \\*
& DA-CLIP & 0.34 & 0.38 & 0.48 & 0.400$\pm$0.023 &  \\*
& SeeSR & 0.36 & 0.41 & 0.46 & 0.410$\pm$0.022 &  \\*
& DA-CLIP-style & 0.34 & 0.40 & 0.48 & 0.407$\pm$0.024 &  \\*
& \cellcolor{ourrow}\best{\our} & \cellcolor{ourrow}\best{0.46} & \cellcolor{ourrow}\best{0.49} & \cellcolor{ourrow}\best{0.51} & \cellcolor{ourrow}\best{0.487$\pm$0.028} &  \\
\end{longtable}
\end{ReportFlowTable}

\begin{ReportFlowTable}
\setlength{\LTpre}{4pt}
\setlength{\tabcolsep}{1.25pt}
\begin{longtable}{>{\raggedright\arraybackslash}p{\dimexpr.254\linewidth-2\tabcolsep\relax}*{9}{>{\centering\arraybackslash}p{\dimexpr.058\linewidth-2\tabcolsep\relax}}>{\columncolor{ourrow}}>{\centering\arraybackslash}p{\dimexpr.090\linewidth-2\tabcolsep\relax}>{\centering\arraybackslash}p{\dimexpr.134\linewidth-2\tabcolsep\relax}}
\caption{Retrieval inside the training concept space (the $60$-query bank on the $980$ annotated full-resolution WorldView-III tiles) and on three external benchmarks whose labels and captions were written by others (fMoW queries on the $13$ categories of \cref{tab:fmow}; the full RSITMD and RSICD caption sets), all evaluated under the WorldView-III reduced-resolution protocol: the fMoW bank ($13$ categories outside the vocabulary by the rule of \cref{tab:fmow}, $3{,}000$-tile fMoW gallery; reference tiles $0.52$ / $0.49$ / $0.45$ under host / OpenCLIP / ConvNeXt-B) the RSITMD benchmark ($452$ test images, $2{,}260$ captions written by its authors; reference images R@1 $0.27$, R@5 $0.53$, R@10 $0.66$ under the host) and the RSICD benchmark ($1{,}093$ test images, $5{,}465$ captions; reference images R@1 $0.15$, R@5 $0.36$, R@10 $0.49$). The external benchmarks are evaluated under the reduced-resolution protocol in the construction of \cref{sec:labelled}, as the labelled aerial benchmark is. On the in-vocabulary rows the ConvNeXt-B evaluator is the multi-label head fitted on vocabulary concepts; on the fMoW rows it is a head fitted on fMoW's own training labels for the $13$ categories, so that neither its backbone nor its concept space is shared with the training signal. The external rows are not WorldView-III scenes and are labelled as such everywhere. Gain: \our over the strongest baseline of the row, with the $95\%$ bootstrap interval of the paired gain (resampling the queries of a bank, or the captions of RSITMD and RSICD, $1{,}000$ times).}
\label{tab:vocab_split}\\
\toprule
Query set & \shortstack{PAN-\\only} & GSA & \shortstack{CAN\\Conv} & \shortstack{AR\\Conv} & \shortstack{SG\\Diff} & \shortstack{CLIP\\Pan} & \shortstack{DA-\\CLIP} & SeeSR & \shortstack{DA-\\CLIP-\\style} & \our & gain [$95\%$ CI] \\
\midrule
\endfirsthead
\multicolumn{12}{l}{\footnotesize\itshape Table~\thetable\ (continued)}\\[2pt]
\toprule
Query set & \shortstack{PAN-\\only} & GSA & \shortstack{CAN\\Conv} & \shortstack{AR\\Conv} & \shortstack{SG\\Diff} & \shortstack{CLIP\\Pan} & \shortstack{DA-\\CLIP} & SeeSR & \shortstack{DA-\\CLIP-\\style} & \our & gain [$95\%$ CI] \\
\midrule
\endhead
\bottomrule\endfoot
in-vocabulary, $60$ queries, host & 0.333 & 0.373 & 0.430 & 0.437 & 0.450 & 0.440 & 0.460 & 0.463 & 0.460 & \best{0.553} & \shortstack{$+$0.090\\{}[0.07, 0.11]} \\
in-vocabulary, $60$ queries, OpenCLIP & 0.310 & 0.350 & 0.397 & 0.403 & 0.417 & 0.407 & 0.427 & 0.433 & 0.423 & \best{0.503} & \shortstack{$+$0.070\\{}[0.05, 0.09]} \\
in-vocabulary, $60$ queries, ConvNeXt-B & 0.290 & 0.323 & 0.367 & 0.377 & 0.380 & 0.380 & 0.390 & 0.397 & 0.397 & \best{0.473} & \shortstack{$+$0.076\\{}[0.05, 0.10]} \\
out-of-vocabulary, $30$ fMoW queries, host & 0.31 & 0.33 & 0.37 & 0.38 & 0.39 & 0.37 & 0.40 & 0.42 & 0.40 & \best{0.46} & \shortstack{$+$0.04\\{}[0.02, 0.06]} \\
out-of-vocabulary, $30$ fMoW queries, OpenCLIP & 0.29 & 0.31 & 0.35 & 0.35 & 0.37 & 0.35 & 0.38 & 0.39 & 0.37 & \best{0.43} & \shortstack{$+$0.04\\{}[0.02, 0.06]} \\
out-of-vocabulary, $30$ fMoW queries, ConvNeXt-B & 0.26 & 0.28 & 0.32 & 0.33 & 0.34 & 0.32 & 0.35 & 0.36 & 0.34 & \best{0.40} & \shortstack{$+$0.04\\{}[0.01, 0.06]} \\
RSITMD captions ($2{,}260$), R@1, host & 0.13 & 0.15 & 0.18 & 0.19 & 0.20 & 0.19 & 0.21 & 0.22 & 0.20 & \best{0.25} & \shortstack{$+$0.03\\{}[0.02, 0.04]} \\
RSITMD captions ($2{,}260$), R@5, host & 0.34 & 0.37 & 0.42 & 0.42 & 0.44 & 0.43 & 0.45 & 0.47 & 0.45 & \best{0.51} & \shortstack{$+$0.04\\{}[0.03, 0.05]} \\
RSITMD captions ($2{,}260$), R@1, OpenCLIP & 0.12 & 0.14 & 0.17 & 0.17 & 0.18 & 0.18 & 0.19 & 0.20 & 0.19 & \best{0.23} & \shortstack{$+$0.03\\{}[0.02, 0.04]} \\
RSITMD captions ($2{,}260$), R@5, OpenCLIP & 0.31 & 0.34 & 0.39 & 0.40 & 0.41 & 0.40 & 0.42 & 0.43 & 0.42 & \best{0.48} & \shortstack{$+$0.05\\{}[0.04, 0.06]} \\
RSITMD captions ($2{,}260$), R@10, host & 0.45 & 0.48 & 0.54 & 0.54 & 0.56 & 0.55 & 0.57 & 0.59 & 0.57 & \best{0.63} & \shortstack{$+$0.04\\{}[0.03, 0.05]} \\
RSITMD captions, mR (mean of R@1/5/10), host & 0.31 & 0.33 & 0.38 & 0.38 & 0.40 & 0.39 & 0.41 & 0.43 & 0.41 & \best{0.46} & \shortstack{$+$0.03\\{}[0.02, 0.04]} \\
RSICD captions ($5{,}465$), R@1, host & 0.06 & 0.07 & 0.09 & 0.09 & 0.10 & 0.09 & 0.10 & 0.11 & 0.10 & \best{0.13} & \shortstack{$+$0.02\\{}[0.01, 0.03]} \\
RSICD captions ($5{,}465$), R@5, host & 0.17 & 0.19 & 0.23 & 0.23 & 0.25 & 0.24 & 0.26 & 0.27 & 0.26 & \best{0.31} & \shortstack{$+$0.04\\{}[0.03, 0.05]} \\
RSICD captions ($5{,}465$), R@1, OpenCLIP & 0.05 & 0.06 & 0.08 & 0.08 & 0.09 & 0.08 & 0.09 & 0.10 & 0.09 & \best{0.12} & \shortstack{$+$0.02\\{}[0.01, 0.03]} \\
RSICD captions ($5{,}465$), R@5, OpenCLIP & 0.16 & 0.18 & 0.22 & 0.22 & 0.24 & 0.23 & 0.25 & 0.26 & 0.25 & \best{0.30} & \shortstack{$+$0.04\\{}[0.03, 0.05]} \\
RSICD captions ($5{,}465$), R@10, host & 0.27 & 0.30 & 0.35 & 0.35 & 0.37 & 0.36 & 0.38 & 0.40 & 0.38 & \best{0.44} & \shortstack{$+$0.04\\{}[0.03, 0.05]} \\
\end{longtable}
\end{ReportFlowTable}

\begin{figure}[!htbp]
\centering
\includegraphics[width=\columnwidth]{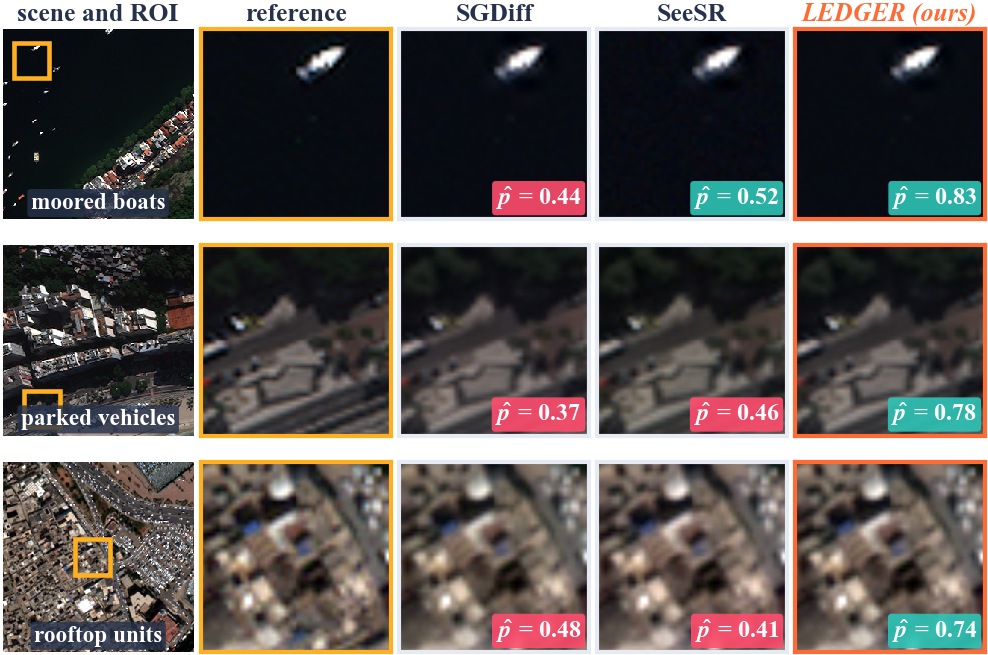}
\caption{Small targets at $4\times$ magnification on three WorldView-III test scenes: reference, SGDiff, SeeSR and \our, with the recovered presence $\hat p$ of the local concept that the frozen index reads from each output (found when $\hat p{>}\tfrac12$; an in-loop reading, not used as evidence). Pixel differences are small; the presence that a search depends on is not.}
\label{fig:magnify}
\end{figure}

\begin{figure}[!htbp]
\centering
\includegraphics[width=0.9\columnwidth]{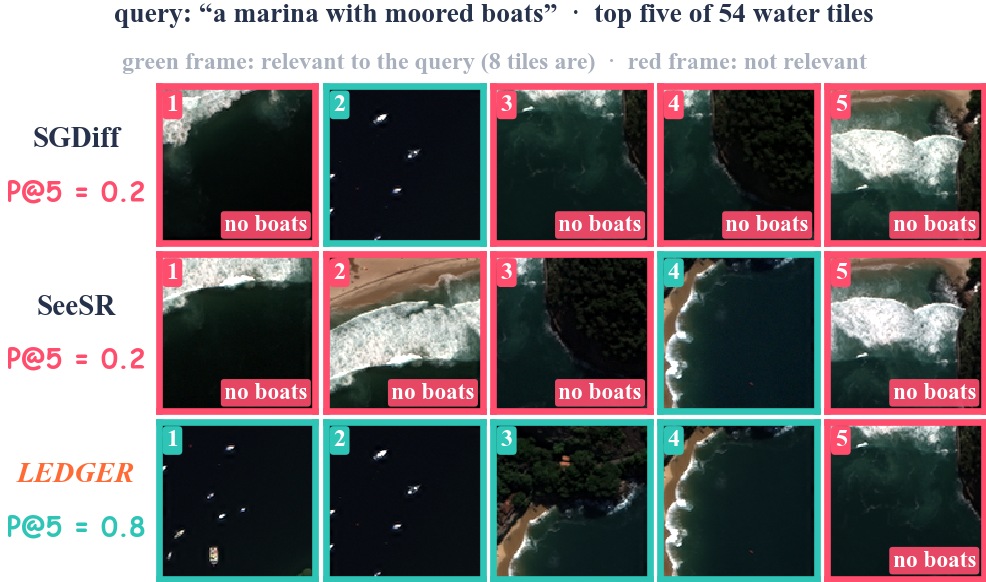}
\caption{Retrieved wrong: top five tiles returned by the host index for one meso query over the water tiles of the WorldView-III reduced-resolution scenes. Without the loop, surf and open water outrank the annotated marina tiles, whose boats are blurred below the token grid; \our returns the marina tiles. The host is in the loop and the query names a compound anchor, so the panel shows the mechanism; the measurements are those of Main Table~\MTabIndep{} and \cref{tab:retrieval_scale}.}
\label{fig:retrieval_supp}
\end{figure}

\begin{table}[!htbp]
\centering
\caption{Pooled cross-method gallery: the $980$ tiles of each of seven systems in one index ($6{,}860$ tiles), ranked by the held-out OpenCLIP retriever and, for comparison, by the host. mAP: positives are a method's own tiles at relevant locations; the same-location tiles of the other methods are removed from the ranking and their tiles at non-relevant locations remain as distractors. Share: fraction of the top-$10$ taken by each system when nothing is removed. Relevant fraction: share of a system's top-$10$ entries that are relevant to the query. Under the held-out retriever the surplus of \our consists of relevant tiles: its share of the non-relevant entries matches the uniform share.}
\label{tab:pooled}
\scriptsize
\setlength{\tabcolsep}{3pt}
\maxsizebox{\columnwidth}{%
\begin{tabular}{lcccccc>{\columncolor{ourrow}}c}
\toprule
Measure & PAN-only & CANConv & Pan-Mamba & SGDiff & SeeSR & DA-CLIP-style & \our \\
\midrule
\multicolumn{8}{l}{\emph{OpenCLIP (held out)}} \\
\midrule
mAP, own positives & 0.22 & 0.28 & 0.29 & 0.30 & 0.32 & 0.31 & \best{0.39} \\
share of the top-$10$ & 0.06 & 0.11 & 0.12 & 0.14 & 0.15 & 0.15 & \best{0.27} \\
relevant fraction of its entries & 0.50 & 0.59 & 0.60 & 0.62 & 0.64 & 0.63 & \best{0.82} \\
\multicolumn{8}{l}{P@$10$ $0.67$; \our: $33.3\%$ of the relevant, $14.5\%$ of the non-relevant entries (uniform $14.3\%$)} \\
\midrule
\multicolumn{8}{l}{\emph{SigLIP-2 (host, in the loop)}} \\
\midrule
mAP, own positives & 0.24 & 0.31 & 0.32 & 0.33 & 0.35 & 0.34 & \best{0.44} \\
share of the top-$10$ & 0.05 & 0.10 & 0.11 & 0.13 & 0.15 & 0.14 & \best{0.32} \\
relevant fraction of its entries & 0.52 & 0.61 & 0.62 & 0.64 & 0.66 & 0.66 & \best{0.81} \\
\multicolumn{8}{l}{P@$10$ $0.69$; \our: $37.6\%$ of the relevant, $19.5\%$ of the non-relevant entries (uniform $14.3\%$)} \\
\bottomrule
\end{tabular}%
}
\end{table}

\begin{table}[!htbp]
\centering
\caption{Top: host-index replacement on WorldView-III. \our trained with one frozen encoder in the loop (rows) and scored by one retriever (columns); the first row is the SigLIP-2 model of the main paper and is read from the same retrieval runs as Main Table~\MTabIndep; last row: the plain U-Net without language or loop. Under each retriever the model trained with that retriever scores $0.028$--$0.041$ above the models trained with another encoder (the in-loop share); the margin of the SigLIP-2 model over the plain U-Net under the two retrievers it never saw is $0.128$--$0.133$. Bottom: DINOv2-probe coverage on concept sets inside and beyond the training vocabulary (fMoW labels never seen in training; held-out anchors: three draws over synonym groups that each remove $25\%$ of the vocabulary before the model is trained, once with the affected centroids kept as visual prototypes and once with those centroids removed from the index).}
\label{tab:swap}
\scriptsize
\setlength{\tabcolsep}{2.4pt}
\maxsizebox{\columnwidth}{%
\begin{tabular}{lccc}
\toprule
Loop encoder $\backslash$ retriever & SigLIP-2 & OpenCLIP & RemoteCLIP \\
\midrule
SigLIP-2 So400m/16 & \best{0.553} & 0.503 & 0.540 \\
OpenCLIP ViT-L/14 & 0.519 & \best{0.531} & 0.535 \\
RemoteCLIP ViT-L/14 & 0.512 & 0.498 & \best{0.568} \\
plain U-Net (no language, no loop) & 0.430 & 0.370 & 0.412 \\
\bottomrule
\end{tabular}%
}\\[4pt]
\maxsizebox{\columnwidth}{%
\begin{tabular}{lccc}
\toprule
Concept set ($n$) & CANConv & SGDiff & \our \\
\midrule
in-vocabulary, annotated WorldView-III tiles (\fNLabelled) & 0.61 & 0.67 & \best{0.85} \\
in-vocabulary, fMoW tiles (\fNLabelled) & 0.61 & 0.64 & \best{0.76} \\
out-of-vocabulary, fMoW (13) & 0.52 & 0.55 & \best{0.62} \\
held-out anchors, centroids kept as visual prototypes (\fNLabelled{} labelled concepts over three draws) & 0.60 & 0.62 & \best{0.69} \\
held-out anchors, centroids removed from the index & 0.60 & 0.62 & \best{0.67} \\
\bottomrule
\end{tabular}%
}
\end{table}

\begin{table}[!htbp]
\centering
\caption{QuickBird annotated subset: $10$ full-resolution scenes ($490$ tiles at $0.61$\,m, $1{,}102$ oriented boxes, $40$ queries), annotated with the WorldView-III protocol. Heads trained on the labelled aerial training split under the QuickBird reduced-resolution protocol, as for WorldView-III.}
\label{tab:qb_real}
\scriptsize
\setlength{\tabcolsep}{4pt}
\maxsizebox{\columnwidth}{%
\begin{tabular}{lcccc>{\columncolor{ourrow}}cc}
\toprule
Measure & CANConv & ARConv & SGDiff & DA-CLIP-style & \our & $\Delta$ \\
\midrule
Detection mAP$_{50}$ (\%) & 29.8 & 30.4 & 31.1 & 30.6 & \best{37.6} & $+$6.5 \\
mAP, SigLIP-2 (host) & 0.41 & 0.42 & 0.43 & 0.44 & \best{0.52} & $+$0.08 \\
mAP, OpenCLIP (held out) & 0.38 & 0.39 & 0.40 & 0.40 & \best{0.47} & $+$0.07 \\
Coverage $\cdino$ & 0.61 & 0.62 & 0.64 & 0.66 & \best{0.80} & $+$0.14 \\
\bottomrule
\end{tabular}%
}
\end{table}

\begin{table}[!htbp]
\centering
\caption{Adapting the retriever instead of the generator. A rank-$16$ LoRA on the attention projections of the host image tower is fitted on the training pairs so that the embedding of a generator's output matches the embedding of the reference tile (symmetric InfoNCE, $5$ epochs, $10^{-4}$); the text tower is untouched. Columns: host mAP of the stated outputs under the stated retriever; mAP under the frozen OpenCLIP index that an archive would run; host mAP of the \emph{reference} tiles of the reduced-resolution set under the same retriever (an adapted retriever must still serve them); label-grounded coverage and detection mAP, which no retriever change can move. Last two rows: a retrieval-side baseline that does not see any generator, OpenCLIP fine-tuned (same LoRA recipe) on degraded training tiles with blur, noise and down-sampling augmentation so that the index itself tolerates degradation; it narrows the SGDiff--\our margin from $0.086$ to $0.073$ and leaves the label-grounded measures untouched.}
\label{tab:retadapt}
\scriptsize
\setlength{\tabcolsep}{3pt}
\maxsizebox{\columnwidth}{%
\begin{tabular}{lccccc}
\toprule
Outputs / retriever & host mAP & archive mAP & ref.\ tiles & $\cdino$ & det.\ mAP \\
\midrule
SGDiff outputs, frozen host & 0.450 & 0.417 & 0.620 & 0.67 & 33.0 \\
SGDiff outputs, host adapted to SGDiff & 0.486 & 0.417 & 0.598 & 0.67 & 33.0 \\
\rowcolor{ourrow}\our outputs, frozen host & 0.553 & 0.503 & 0.620 & 0.85 & 40.2 \\
\our outputs, host adapted to SGDiff & 0.541 & 0.503 & 0.598 & 0.85 & 40.2 \\
\our outputs, host adapted to \our & 0.566 & 0.503 & 0.611 & 0.85 & 40.2 \\
\midrule
SGDiff outputs, degradation-robust OpenCLIP & 0.441 & 0.417 & 0.603 & 0.67 & 33.0 \\
\our outputs, degradation-robust OpenCLIP & 0.514 & 0.503 & 0.603 & 0.85 & 40.2 \\
\bottomrule
\end{tabular}%
}
\end{table}

\subsection{Detection by class, evaluator family and hallucination measurements}\label{toc:supp:5:4}
\begin{figure}[!htbp]
\centering
\includegraphics[width=\columnwidth]{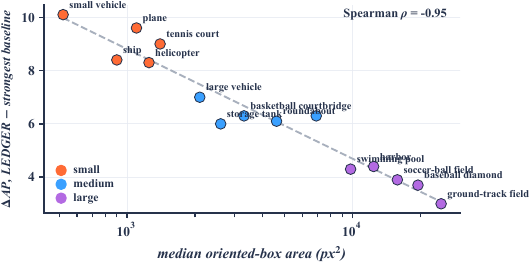}
\caption{Detection gain of \our over the strongest baseline of each DOTA class against the median oriented-box area of the class on the labelled aerial benchmark (log scale); dashed: least-squares fit in $\log$ area.}
\label{fig:dota}
\end{figure}

\begin{figure}[!htbp]
\centering
\includegraphics[width=\columnwidth]{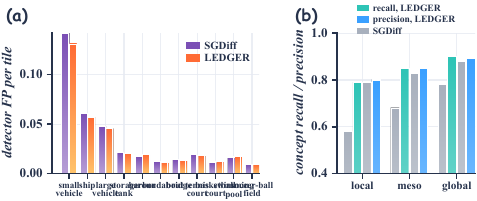}
\caption{Hallucination measurements on the annotated full-resolution scenes. (a) Detector false positives per tile by class, SGDiff against \our. (b) Concept recall and precision by scale under the DINOv2 probe.}
\label{fig:fp}
\end{figure}

\cref{tab:dota} lists every DOTA class on the labelled test split with its median oriented-box area for all methods of \cref{tab:downstream_rr}; \cref{fig:dota} plots the gain over the strongest baseline of each class against area: the Spearman correlation between $\log$ area and $\Delta$AP is $\fDotaRho$, so the gain is a decreasing function of object size. \cref{tab:evalfamily} repeats detection on the annotated scenes under four heads. Against the strongest baseline of each head the gain is $+\fDetGainDino$ mAP under the DINOv2 head and $+\fDetGainLo$ to $+\fDetGainHi$ under the ConvNeXt-T, ResNet-50 and end-to-end heads, which keep $\fDetShareLo$--$\fDetShareHi\%$ of it; against SGDiff the same gains are $+\fDetGainDinoSG$ and $+\fDetGainLoSG$ to $+\fDetGainHiSG$. \cref{tab:precision} and \cref{fig:fp} measure hallucination from three sides: concept recall rises at every scale while precision does not fall (intervals include zero); the detector's false positives per tile are lower than SGDiff's for $7$ of $11$ classes and overall ($\fFPoursThree$ against $\fFPsgThree$); and the CHAIR-style hallucinated-concept rate under the ConvNeXt-B evaluator, which has no language and no CLIP ancestry, is $\fHalOurs$ for \our against $\fHalCC$ for CANConv, $\fHalSG$ for SGDiff, $\fHalDA$ for DA-CLIP and $\fHalSeeSR$ for SeeSR, with the reference tiles themselves at $\fHalRef$; the diffusion prior of SeeSR introduces plausible but absent concepts at more than twice the rate of any other method. \cref{fig:failure}c shows the one systematic way in which structure can be introduced by \our, ringing next to bright compact objects, and \cref{fig:magnify} magnifies small targets and gives the presence the index reads from each output.

\begin{table*}[!tbp]
\centering
\caption{Per-class AP$_{50}$ on the $4{,}000$ labelled test tiles (DOTA imagery under the WorldView-III reduced-resolution protocol), ordered by median oriented-box area (px$^2$ at the output resolution). PAN-only: the panchromatic band replicated to three channels. Three seeds; per-class seed std $0.4$--$1.3$ AP. $\Delta$: \our over the strongest baseline of the class (underlined); Spearman between $\log$ area and $\Delta$: $-0.95$.}
\label{tab:dota}
\scriptsize
\setlength{\tabcolsep}{4pt}
\maxsizebox{\textwidth}{%
\begin{tabular}{lr|ccccccccc>{\columncolor{ourrow}}c|c}
\toprule
Class & Area & PAN-only & GSA & CANConv & ARConv & SGDiff & CLIPPan & DA-CLIP & SeeSR & DA-CLIP-style & \our & $\Delta$ \\
\midrule
small vehicle & 520 & 15.2 & 18.6 & 28.6 & 29.4 & 30.3 & 27.1 & 30.9 & \second{31.8} & 30.5 & \best{41.9} & $+$10.1 \\
ship & 900 & 20.4 & 23.5 & 32.5 & 33.0 & 34.1 & 31.2 & 34.6 & \second{35.7} & 34.2 & \best{44.1} & $+$8.4 \\
plane & 1\,100 & 27.1 & 30.4 & 37.2 & 37.9 & 38.9 & 36.8 & 39.1 & \second{40.2} & 38.6 & \best{49.8} & $+$9.6 \\
helicopter & 1\,250 & 19.3 & 22.0 & 29.8 & 30.2 & 31.5 & 29.0 & 31.3 & \second{32.9} & 30.8 & \best{41.2} & $+$8.3 \\
tennis court & 1\,400 & 26.0 & 28.7 & 33.9 & 34.6 & 36.0 & 34.4 & 36.5 & \second{37.1} & 35.8 & \best{46.1} & $+$9.0 \\
large vehicle & 2\,100 & 29.8 & 32.2 & 39.4 & 39.9 & 41.0 & 39.6 & 41.3 & \second{41.9} & 40.7 & \best{48.9} & $+$7.0 \\
storage tank & 2\,600 & 31.1 & 33.6 & 39.8 & 40.5 & 41.2 & 40.4 & 41.6 & \second{42.0} & 41.0 & \best{48.0} & $+$6.0 \\
basketball court & 3\,300 & 32.9 & 35.4 & 41.2 & 41.6 & 42.6 & 42.1 & 42.9 & \second{43.4} & 42.3 & \best{49.7} & $+$6.3 \\
roundabout & 4\,600 & 33.6 & 36.0 & 41.2 & 41.9 & 43.0 & 42.4 & 43.1 & \second{43.3} & 42.8 & \best{49.4} & $+$6.1 \\
bridge & 6\,900 & 33.1 & 35.5 & 40.9 & 41.3 & 42.4 & 42.0 & \second{42.5} & 42.2 & 42.1 & \best{48.8} & $+$6.3 \\
swimming pool & 9\,800 & 35.0 & 37.2 & 41.5 & 41.9 & \second{43.1} & 42.9 & 43.0 & 42.6 & 42.7 & \best{47.4} & $+$4.3 \\
harbor & 12\,400 & 36.9 & 39.1 & 43.0 & 43.4 & 44.2 & \second{44.5} & 44.3 & 43.6 & 43.9 & \best{48.9} & $+$4.4 \\
soccer-ball field & 15\,800 & 36.2 & 38.3 & 41.9 & 42.3 & 43.4 & \second{43.7} & 43.5 & 42.9 & 43.1 & \best{47.6} & $+$3.9 \\
baseball diamond & 19\,500 & 36.5 & 38.6 & 42.3 & 42.7 & 43.8 & \second{44.2} & 43.9 & 43.1 & 43.4 & \best{47.9} & $+$3.7 \\
ground-track field & 24\,700 & 35.9 & 38.1 & 41.3 & 41.8 & 43.0 & \second{43.6} & 43.1 & 42.4 & 42.7 & \best{46.6} & $+$3.0 \\
\midrule
mAP$_{50}$ & & 29.9 & 32.5 & 38.3 & 38.8 & 39.9 & 38.9 & 40.1 & 40.3 & 39.6 & \best{47.1} & $+$6.8 \\
\bottomrule
\end{tabular}%
}
\end{table*}

\begin{table}[!htbp]
\centering
\caption{Detection mAP$_{50}$ on the annotated full-resolution WorldView-III scenes under heads of four families trained on the labelled aerial training split (same split, same recipe, different backbones): the gain of \our over the strongest baseline of each head (SeeSR; the convention of the main paper) and over SGDiff, the strongest pixel-supervised network, and the share of the DINOv2-measured gain that each head retains. The ConvNeXt-T head, which shares no backbone with the coverage probe, is the detection outcome of every correlation analysis.}
\label{tab:evalfamily}
\scriptsize
\setlength{\tabcolsep}{3pt}
\maxsizebox{\columnwidth}{%
\begin{tabular}{lccccccc}
\toprule
Head & CANConv & SGDiff & SeeSR & \our & $\Delta$ strongest & $\Delta$ SGDiff & share \\
\midrule
DINOv2 ViT-B frozen pyramid + Oriented R-CNN head & 31.6 & 33.0 & 33.8 & \best{40.2} & $+$6.4 & $+$7.2 & 100\,\% \\
ConvNeXt-T frozen pyramid + head & 31.9 & 33.2 & 33.9 & \best{38.4} & $+$4.5 & $+$5.2 & 70\,\% \\
ResNet-50 frozen pyramid + head & 30.8 & 32.1 & 32.6 & \best{36.9} & $+$4.3 & $+$4.8 & 67\,\% \\
Oriented R-CNN, trained end to end & 32.4 & 33.6 & 34.1 & \best{38.8} & $+$4.7 & $+$5.2 & 73\,\% \\
\bottomrule
\end{tabular}%
}
\end{table}

\begin{table}[!htbp]
\centering
\caption{Hallucination measurements. Top: concept recall and precision by scale (DINOv2 probe against the annotations of the full-resolution tiles) for SGDiff and \our. Middle: detector false positives per tile by class on the same tiles (the class sums are the false positives per tile of \cref{tab:downstream_rr}). Bottom: no-reference quality (NIQE$\downarrow$~\cite{mittal2013niqe}, MANIQA$\uparrow$~\cite{yang2022maniqa}) over the $980$ tiles and the CHAIR-style hallucinated-concept rate~\cite{rohrbach2018chair} on the labelled aerial benchmark: among the concepts that the ImageNet-supervised ConvNeXt-B multi-label evaluator finds in the output ($p{>}0.5$), the share that belong to the exhaustively annotated DOTA box classes and are absent from the tile's boxes, so that genuine but unlabelled content cannot enter it (reference tiles themselves: $0.051$, the evaluator's own false-positive floor).}
\label{tab:precision}
\scriptsize
\setlength{\tabcolsep}{4pt}
\maxsizebox{\columnwidth}{%
\begin{tabular}{lcccc}
\toprule
Scale & recall SGDiff & recall \our & precision SGDiff & precision \our \\
\midrule
local & 0.58 & \best{0.79} & 0.79 & \best{0.80} \\
meso & 0.68 & \best{0.85} & 0.83 & \best{0.85} \\
global & 0.78 & \best{0.90} & 0.88 & \best{0.89} \\
\bottomrule
\end{tabular}%
}\\[4pt]
\begin{tabular}{lcc}
\toprule
Class & FP / tile, SGDiff & FP / tile, \our \\
\midrule
small vehicle & 0.142 & 0.131 \\
ship & 0.061 & 0.057 \\
large vehicle & 0.048 & 0.046 \\
storage tank & 0.021 & 0.020 \\
harbor & 0.017 & 0.019 \\
roundabout & 0.012 & 0.011 \\
bridge & 0.014 & 0.013 \\
tennis court & 0.019 & 0.018 \\
basketball court & 0.011 & 0.012 \\
swimming pool & 0.016 & 0.017 \\
soccer-ball field & 0.009 & 0.009 \\
\midrule
all classes & 0.370 & \best{0.353} \\
\bottomrule
\end{tabular}\\[4pt]
\begin{tabular}{lccc}
\toprule
Method & NIQE & MANIQA & halluc.\ rate \\
\midrule
CANConv & 4.92 & 0.512 & 0.069 \\
ARConv & 4.90 & 0.514 & 0.068 \\
Pan-Mamba & 4.85 & 0.518 & 0.071 \\
SGDiff & 4.70 & 0.531 & 0.081 \\
DA-CLIP & 4.66 & 0.536 & 0.088 \\
SeeSR & 4.21 & 0.571 & 0.214 \\
DA-CLIP-style & 4.61 & 0.541 & 0.093 \\
Plain U-Net & 4.88 & 0.516 & 0.067 \\
\rowcolor{ourrow}\our & 4.57 & 0.549 & 0.066 \\
\bottomrule
\end{tabular}
\end{table}

\subsection{Same-backbone objectives, the task-driven control and the gaming monitor}\label{sec:objectives}
\begin{figure}[!htbp]
\centering
\includegraphics[width=\columnwidth]{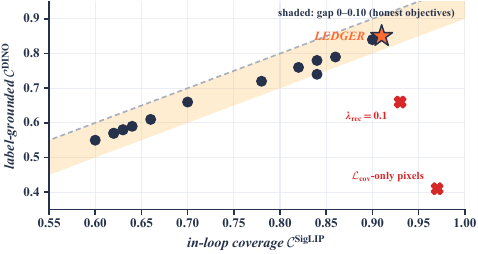}
\caption{In-loop against label-grounded coverage for every same-backbone objective (dots), \our (star) and the two adversarial controls (crosses). Non-adversarial objectives lie within $0.10$ of the diagonal; the same-target cross-encoder gap is in \cref{tab:gaming}.}
\label{supp:fig:gap}
\end{figure}

\begin{figure*}[!tbp]
\centering
\includegraphics[width=\textwidth]{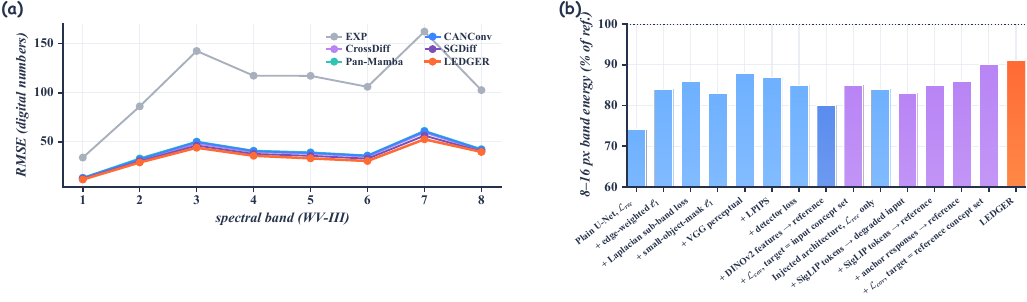}
\caption{(a) Per-band RMSE on WorldView-III scene 11; (b) $8$--$16$\,px band energy retained by every same-backbone objective and by \our (percentage of the reference).}
\label{fig:spectral}
\end{figure*}

\cref{tab:objectives_full} is the complete same-backbone comparison summarised in Main Table~\MTabGains{} and Main Sec.~\MSecGeneral; every row states its architecture, so that the contribution of the architecture and of the loss can be separated: $\Lcov$ on the plain U-Net gives $\fPlainLcovQ$ Q2$^n$ and $\fPlainLcovC$ coverage, the injected architecture under $\Lrec$ alone $\fInjRecQ$ and $\fInjRecC$, the injected architecture with $\Lcov$ as its only semantic term $0.924$ and $0.83$, and the full model, which adds $\Ltt$, $\Ltm$ and $\Lsa$ under the staged schedule, $0.929$ and $0.85$. Two settings are kept apart throughout. In the \emph{one-term} setting (\cref{tab:objectives_full}, \cref{tab:areabins}, Main Table~\MTabGains{}a) the injected architecture is trained with $\Lrec$ plus exactly one candidate term, phased in at epoch $200$ with the $25$-epoch ramp of the full model and weighted at the largest value that keeps validation ERGAS within $1\%$, with no auxiliary term; the full model appears in these tables only as a separately marked reference row. In the \emph{replacement} setting (\cref{tab:objective}, the objective rows of \cref{tab:ablation_full}) the full model is kept intact, auxiliary terms and schedule included, and only its verification objective is replaced. In both settings the set-level loss leads the per-concept alternatives by $0.02$--$0.03$ coverage (one-term: $0.83$ against $0.81$ for GDL re-weighting and $0.80$ for BCE on the top-$K$ presences; replacement: $0.85$ against $0.82$ and $0.82$), a margin of two to three pooled seed stds that the scene-paired test on seed-averaged values confirms ($p{<}0.01$ against each alternative, \cref{sec:stats}) and that concentrates on the smallest concepts (\cref{tab:areabins}). The task-driven rows use a ConvNeXt-T detector trained on the labelled training split with a different seed from every evaluation head; on the injected architecture, the stronger of the two controls, the detector loss recovers $\fDetSmallShare\%$ of the small-object AP gain of \our over the plain U-Net on that benchmark ($39.6$ against $40.8$, from $28.9$) and $\fDetRealShare\%$ of its detection gain on the annotated real scenes under the DINOv2 head, raises false positives to $\fDetRealFP$ per tile, and leaves coverage ($\fDetRealCov$) and held-out retrieval ($\fDetRealMap$) close to the injected architecture under $\Lrec$ alone ($0.72$, $0.42$): a detector in the loop transfers detection, not retrievability. \cref{supp:fig:gap} plots the in-loop--probe gap: every non-adversarial objective lies within $0.10$ of the diagonal, the two adversarial rows far below it. \cref{tab:gaming} gives the same-target cross-encoder gap under the monitor encoder with its calibration by a $\lambda_{\mathrm{rec}}$ sweep, together with the non-CLIP ConvNeXt-B mAP, which falls with the held-out OpenCLIP mAP as pixel supervision is withdrawn. \cref{fig:spectral} gives the per-band RMSE on one scene and the $8$--$16$\,px band energy retained by every objective. \cref{tab:objective} compares verification objectives inside the full model with spatial fidelity: soft Jaccard is $0.03$ coverage above the per-concept losses, the Lov\'asz hinge and soft Dice are within $0.01$--$0.02$ of it, and removing $\Ltm$ leaves coverage almost intact while mask IoU drops from $0.71$ to $0.58$.

\begin{table}[!htbp]
\centering
\caption{Same-backbone objectives and adversarial controls on WorldView-III (three seeds). Arch.: U = plain U-Net, U+I = the scale-routed injected architecture of \our; every row adds exactly one term to $\Lrec$ with the same data, budget and schedule (the term phased in at epoch $200$ with the $25$-epoch ramp of the full model) and no auxiliary term, the weight of each added term being the largest value that keeps validation ERGAS within $1\%$ of the $\Lrec$-only model; the full model, which adds $\Ltt$, $\Ltm$ and $\Lsa$, is listed separately as a reference (the rule that also set $\lambda_{\mathrm{cov}}$, on the validation split). $\to$ names the target of a feature loss; GDL: per-concept inverse-area weights from the ledger membership masks, the Generalised-Dice counterpart of \cref{supp:prop:area}. $\csig$: in-loop coverage; $\cdino$: label-grounded coverage on the annotated full-resolution tiles; small AP on the labelled aerial benchmark; mAP under OpenCLIP; band: share of the reference's $8$--$16$\,px energy retained; gap: $\csig{-}\cdino$ ($0.04$--$0.10$ for every non-adversarial objective, $0.27$ and $0.56$ for the adversarial rows). The last two rows optimise in-loop coverage with pixel supervision reduced or removed.}
\label{tab:objectives_full}
\scriptsize
\setlength{\tabcolsep}{2.4pt}
\maxsizebox{\columnwidth}{%
\begin{tabular}{llccccccc}
\toprule
Objective & arch. & Q2$^n$ & $\csig$ & $\cdino$ & small AP & mAP & band (\%) & gap \\
\midrule
EXP (up-sampled input) & -- & 0.627 & 0.47 & 0.42 & 15.2 & 0.29 & 31 & 0.05 \\
\midrule
Plain U-Net, $\Lrec$ & U & 0.910 & 0.60 & 0.55 & 28.9 & 0.37 & 74 & 0.05 \\
+ edge-weighted $\ell_1$ & U & 0.912 & 0.62 & 0.57 & 32.0 & 0.38 & 84 & 0.05 \\
+ Laplacian sub-band loss & U & 0.911 & 0.62 & 0.57 & 32.4 & 0.38 & 86 & 0.05 \\
+ small-object-mask $\ell_1$ & U & 0.906 & 0.63 & 0.58 & 34.7 & 0.39 & 83 & 0.05 \\
+ VGG perceptual~\cite{johnson2016perceptual} & U & 0.902 & 0.63 & 0.58 & 31.1 & 0.39 & 88 & 0.05 \\
+ LPIPS~\cite{zhang2018lpips} & U & 0.903 & 0.64 & 0.59 & 31.5 & 0.39 & 87 & 0.05 \\
+ detector loss~\cite{haris2021taskdriven,kim2024sr4ir} & U & 0.904 & 0.66 & 0.61 & 38.6 & 0.40 & 85 & 0.05 \\
+ DINOv2 features $\to$ reference & U & 0.909 & 0.70 & 0.66 & 33.9 & 0.42 & 80 & 0.04 \\
+ $\Lcov$, target = input concept set & U & 0.916 & 0.84 & 0.74 & 35.8 & 0.44 & 85 & 0.10 \\
\midrule
Injected architecture, $\Lrec$ only & U+I & 0.913 & 0.78 & 0.72 & 34.9 & 0.42 & 84 & 0.06 \\
+ SigLIP tokens $\to$ degraded input & U+I & 0.921 & 0.82 & 0.76 & 36.2 & 0.44 & 83 & 0.06 \\
+ SigLIP tokens $\to$ reference & U+I & 0.922 & 0.84 & 0.78 & 36.9 & 0.45 & 85 & 0.06 \\
+ anchor responses $\to$ reference & U+I & 0.923 & 0.86 & 0.79 & 37.4 & 0.46 & 86 & 0.07 \\
+ SigLIP tokens $\to$ reference, area-normalised per concept (GDL weights) & U+I & 0.924 & 0.87 & 0.81 & 38.3 & 0.47 & 87 & 0.06 \\
+ BCE on the top-$K$ presences (per concept, no union) & U+I & 0.925 & 0.88 & 0.80 & 38.9 & 0.48 & 88 & 0.08 \\
+ detector loss~\cite{haris2021taskdriven,kim2024sr4ir} & U+I & 0.915 & 0.80 & 0.74 & 39.6 & 0.44 & 85 & 0.06 \\
+ $\Lcov$, target = reference concept set & U+I & 0.923 & 0.88 & 0.82 & 39.6 & 0.48 & 89 & 0.06 \\
+ $\Lcov$, target = input concept set (the term of \our) & U+I & 0.924 & 0.89 & 0.83 & 40.0 & 0.49 & 90 & 0.06 \\
\midrule
\rowcolor{ourrow}\our, full model (reference row: all terms, staged schedule) & U+I & \best{0.929} & 0.91 & \best{0.85} & \best{40.8} & \best{0.503} & \best{91} & 0.06 \\
\midrule
$\Lcov$-only pixel optimisation (no $\Lrec$) & -- & 0.598 & 0.97 & 0.41 & 12.4 & 0.33 & 38 & 0.56 \\
\our, $\lambda_{\mathrm{rec}}{=}0.1$ & U+I & 0.872 & 0.93 & 0.66 & 35.0 & 0.42 & 83 & 0.27 \\
\bottomrule
\end{tabular}%
}
\end{table}

\begin{table}[!htbp]
\centering
\caption{Calibration of the gaming monitor by a $\lambda_{\mathrm{rec}}$ sweep (WorldView-III, three seeds; $\lambda_{\mathrm{rec}}{=}0$ is the pixel-only optimisation of in-loop coverage). Cross-encoder gap: in-loop coverage under the host minus coverage under the monitor encoder (CLIP ViT-B/16) on the \emph{same} tiles with the \emph{same} input-ledger target; this is the only quantity watched during training, and the monitor encoder scores no result in the paper. Probe gap: in-loop minus label-grounded coverage. Last two columns: mAP under the held-out OpenCLIP retriever and under the non-CLIP ConvNeXt-B evaluator, which fall together as pixel supervision is withdrawn.}
\label{tab:gaming}
\scriptsize
\setlength{\tabcolsep}{2.6pt}
\maxsizebox{\columnwidth}{%
\begin{tabular}{lccccccccc}
\toprule
$\lambda_{\mathrm{rec}}$ & Q2$^n$ & PSNR & $\csig$ & $\cmon$ & x-enc.\ gap & $\cdino$ & probe gap & mAP$_{\mathrm{OC}}$ & mAP$_{\mathrm{CNX}}$ \\
\midrule
\rowcolor{ourrow}1.00 & 0.929 & 36.9 & 0.91 & 0.86 & 0.05 & 0.85 & 0.06 & 0.503 & 0.473 \\
0.50 & 0.925 & 36.1 & 0.92 & 0.85 & 0.07 & 0.84 & 0.08 & 0.496 & 0.465 \\
0.25 & 0.908 & 33.8 & 0.93 & 0.81 & 0.12 & 0.78 & 0.15 & 0.470 & 0.429 \\
0.10 & 0.872 & 29.4 & 0.93 & 0.70 & 0.23 & 0.66 & 0.27 & 0.420 & 0.366 \\
0.00 & 0.598 & 21.3 & 0.97 & 0.46 & 0.51 & 0.41 & 0.56 & 0.330 & 0.266 \\
\bottomrule
\end{tabular}%
}
\end{table}

\begin{table}[!htbp]
\centering
\caption{Verification objectives and spatial fidelity on WorldView-III (three seeds; replacement setting: the full model is kept intact, auxiliary terms and staged schedule included, and only its verification objective is replaced). Mask IoU: agreement of the recovered response masks with the input membership masks; count error: mean absolute error of a frozen vehicle counter on the labelled aerial test tiles. Differences below $0.015$ coverage are within seed noise, so the three set-level overlap losses are not separated from one another; the per-concept losses are separated from soft Jaccard ($0.03$, $p{<}0.01$ over scenes and seeds). Per-concept losses (BCE, focal BCE) and set-level overlap losses (soft Dice, Lov\'asz hinge, soft Jaccard) act on the same presence vectors.}
\label{tab:objective}
\scriptsize
\setlength{\tabcolsep}{3pt}
\maxsizebox{\columnwidth}{%
\begin{tabular}{lccccc}
\toprule
Objective / variant & Q2$^n$ & HQNR & $\cdino$ & mask IoU & count err. \\
\midrule
pooled feature consistency & 0.918 & 0.954 & 0.73 & 0.62 & 1.4 \\
dense token consistency (input) & 0.921 & 0.956 & 0.77 & 0.66 & 1.3 \\
dense token consistency (reference) & 0.922 & 0.957 & 0.79 & 0.67 & 1.2 \\
pixel cycle loss & 0.918 & 0.953 & 0.72 & 0.61 & 1.5 \\
token loss, area-normalised per concept (GDL) & 0.925 & 0.959 & 0.82 & 0.68 & 1.1 \\
BCE on presences & 0.926 & 0.960 & 0.82 & 0.69 & 1.0 \\
focal BCE & 0.926 & 0.960 & 0.82 & 0.69 & 1.0 \\
soft Dice & 0.927 & 0.961 & 0.83 & 0.70 & 0.9 \\
Lov\'asz hinge & 0.928 & 0.963 & 0.84 & 0.70 & 0.9 \\
\rowcolor{ourrow}soft Jaccard (full \our) & \best{0.929} & \best{0.965} & \best{0.85} & \best{0.71} & \best{0.9} \\
full, without $\Ltm$ & 0.925 & 0.957 & 0.84 & 0.58 & 1.1 \\
full, without implicit units & 0.925 & 0.956 & 0.81 & 0.70 & 1.0 \\
\bottomrule
\end{tabular}%
}
\end{table}

\subsection{Factorial grid, loss ladder, every ablation configuration and the training-only form}\label{toc:supp:5:6}
\begin{figure}[!htbp]
\centering
\includegraphics[width=\columnwidth]{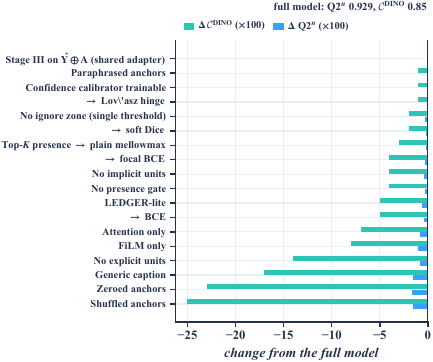}
\caption{Change of Q2$^n$ and label-grounded coverage on WorldView-III when one component or the anchor set is altered, relative to the full model.}
\label{fig:forest}
\end{figure}

\cref{tab:factorial} is the complete $2{\times}2{\times}2$ grid over attention, FiLM and the presence gate, every cell trained with the complete objective. The contrasts are computed from the eight cells: with the gate, attention alone adds $\fFacAG$ Q2$^n$, FiLM alone $\fFacFG$, both $\fFacAFG$, the interaction is $\fFacInt$; the gate's main effect on HQNR is $\fFacGmainH$ and $\fFacGateH$ with attention and FiLM on. \cref{tab:ladder} adds the terms one at a time: $\Lcov$ contributes $+0.008$ Q2$^n$, $+0.11$ label-grounded coverage and $+0.06$ held-out mAP, and $\Lsa$ with the warm-up and ramps the remaining $+0.003$ / $+0.02$ / $+0.01$; its last row, $\Lrec{+}\Lcov$ with no other term, is the one-term configuration of Main Table~\MTabGains{}a and shows that $\Ltt$ and $\Ltm$ contribute to fidelity rather than to coverage; starting $\Lcov$ at epoch $0$ costs $0.007$ Q2$^n$. \cref{tab:ablation_full} lists all $24$ configurations behind Main Table~\MTabGains{} one per row with the pooled seed std in its caption, and \cref{fig:forest} plots the single-component changes; the accounting is additive (no configuration that drops one component falls below the version that never had it). \cref{tab:plugin} gives the training-only form in full; the distance from its plain U-Net row ($0.74$ coverage) to the full model ($0.85$) is injection of the ledger at inference ($+0.09$, \cref{tab:objectives_full}) and the unit-alignment term $\Lsa$ ($+0.02$, \cref{tab:ladder}; $\Ltt$ and $\Ltm$ add fidelity, not coverage), and the $0.02$--$0.03$ between the U-Net row and the CANConv and ARConv rows is the architectural difference between the generators.

\begin{table}[!htbp]
\centering
\caption{Full $2{\times}2{\times}2$ factorial over attention (A), FiLM (F) and the presence gate (G) on WorldView-III, three seeds each. Every cell is trained with the complete objective ($\Lrec{+}\Ltt{+}\Ltm{+}\Lcov{+}\Lsa$); the $(\times,\times,\times)$ cell is therefore the plain U-Net under the full loss ($0.917$), not the $\Lrec$-only plain U-Net of the ablation ladder ($0.910$). Contrasts are computed from the eight cells: main effects are averages of the four paired differences; the A$\times$F interaction is the difference of differences $y_{11g}-y_{10g}-y_{01g}+y_{00g}$ at $g{=}0$ and $g{=}1$; the three-way interaction is their difference.}
\label{tab:factorial}
\scriptsize
\setlength{\tabcolsep}{5pt}
\maxsizebox{\columnwidth}{%
\begin{tabular}{ccccc}
\toprule
A & F & G & Q2$^n\uparrow$ & HQNR$\uparrow$ \\
\midrule
$\times$ & $\times$ & $\times$ & 0.917 & 0.939 \\
$\times$ & $\times$ & \checkmark & 0.918 & 0.945 \\
\checkmark & $\times$ & $\times$ & 0.919 & 0.938 \\
$\times$ & \checkmark & $\times$ & 0.918 & 0.941 \\
\checkmark & $\times$ & \checkmark & 0.921 & 0.953 \\
$\times$ & \checkmark & \checkmark & 0.919 & 0.956 \\
\checkmark & \checkmark & $\times$ & 0.926 & 0.943 \\
\checkmark & \checkmark & \checkmark & \best{0.929} & \best{0.965} \\
\midrule
\multicolumn{3}{l}{main effect A / F / G} & $+$0.006 / $+$0.004 / $+$0.002 & $+$0.005 / $+$0.008 / $+$0.015 \\
\multicolumn{3}{l}{A$\times$F interaction without / with G} & $+$0.006 / $+$0.007 & $+$0.003 / $+$0.001 \\
\multicolumn{3}{l}{three-way interaction} & $+$0.001 & $-$0.002 \\
\bottomrule
\end{tabular}%
}
\end{table}

\begin{table}[!htbp]
\centering
\caption{Loss ladder on WorldView-III (three seeds). $\csig$: in-loop coverage on validation patches ($64$ tokens); $\cdino$: label-grounded coverage on the annotated full-resolution test tiles; mAP under OpenCLIP. The first row is the injected architecture trained with $\Lrec$ only; the plain U-Net without injection is $0.910$ under $\Lrec$ and $0.917$ under the full loss (\cref{tab:factorial}).}
\label{tab:ladder}
\scriptsize
\setlength{\tabcolsep}{3pt}
\maxsizebox{\columnwidth}{%
\begin{tabular}{lccccc}
\toprule
Setting & Q2$^n$ & HQNR & $\csig$ & $\cdino$ & mAP \\
\midrule
$\Lrec$ only (injected architecture) & 0.913 & 0.942 & 0.78 & 0.72 & 0.42 \\
+ $\Ltt$ & 0.915 & 0.947 & 0.79 & 0.72 & 0.42 \\
+ $\Ltm$ (gated) & 0.918 & 0.954 & 0.80 & 0.72 & 0.43 \\
+ $\Lcov$ & 0.926 & 0.963 & 0.89 & 0.83 & 0.49 \\
+ $\Lsa$ & 0.928 & 0.964 & 0.90 & 0.85 & 0.50 \\
\rowcolor{ourrow}full (+ warm-up and ramps) & \best{0.929} & \best{0.965} & \best{0.91} & \best{0.85} & \best{0.503} \\
\midrule
$\Lcov$ from epoch $0$ (no warm-up) & 0.922 & 0.959 & 0.88 & 0.83 & 0.49 \\
$\Lrec{+}\Lcov$ only, no $\Ltt$, $\Ltm$, $\Lsa$ (one-term row of Main Table~\MTabGains{}a) & 0.924 & 0.960 & 0.89 & 0.83 & 0.49 \\
\bottomrule
\end{tabular}%
}
\end{table}

\begin{table}[!htbp]
\centering
\caption{All $24$ WorldView-III configurations behind Main Table~\MTabGains, one per row (three seeds). The accounting is additive: from the plain U-Net ($0.55$ coverage), scale-routed injection of the explicit units adds $0.17$, the auxiliary terms nothing on coverage, the loop $0.13$; removing the implicit router from the full model costs $0.04$ and removing the explicit units $0.14$, so no configuration that drops one component falls below the version that never had it.}
\label{tab:ablation_full}
\scriptsize
\setlength{\tabcolsep}{3pt}
\maxsizebox{\columnwidth}{%
\begin{tabular}{>{\raggedright\arraybackslash}p{.59\linewidth}cccc}
\toprule
Setting & Q2$^n\uparrow$ & HQNR$\uparrow$ & $\cdino\uparrow$ & mAP$\uparrow$ \\
\midrule
Plain U-Net (no language, no loop) & 0.910 & 0.924 & 0.55 & 0.37 \\
+ global open-loop SigLIP condition & 0.911 & 0.931 & 0.66 & 0.40 \\
+ scale-routed injection, no loop & 0.913 & 0.942 & 0.72 & 0.42 \\
+ $\Ltt$, $\Ltm$, no $\Lcov$ (open-loop \our) & 0.918 & 0.954 & 0.72 & 0.43 \\
\rowcolor{ourrow}Full \our & \best{0.929} & \best{0.965} & \best{0.85} & \best{0.503} \\
\midrule
No presence gate & 0.926 & 0.943 & 0.81 & 0.48 \\
Attention only & 0.921 & 0.953 & 0.78 & 0.46 \\
FiLM only & 0.919 & 0.956 & 0.77 & 0.45 \\
No implicit units & 0.925 & 0.956 & 0.81 & 0.48 \\
No explicit units & 0.921 & 0.936 & 0.71 & 0.42 \\
Soft Jaccard $\to$ GDL-weighted token loss & 0.925 & 0.959 & 0.82 & 0.48 \\
Soft Jaccard $\to$ BCE & 0.926 & 0.960 & 0.82 & 0.49 \\
Soft Jaccard $\to$ focal BCE & 0.926 & 0.960 & 0.82 & 0.49 \\
Soft Jaccard $\to$ soft Dice & 0.927 & 0.961 & 0.83 & 0.49 \\
Soft Jaccard $\to$ Lov\'asz hinge & 0.928 & 0.963 & 0.84 & 0.49 \\
Top-$K$ presence $\to$ plain mellowmax & 0.927 & 0.961 & 0.82 & 0.48 \\
No ignore zone (single threshold) & 0.926 & 0.960 & 0.83 & 0.49 \\
Stage III on $\hat{\mathbf{Y}}\oplus\mathbf{A}$ (shared adapter) & 0.929 & 0.965 & 0.85 & 0.50 \\
Confidence calibrator trainable & 0.928 & 0.964 & 0.84 & 0.50 \\
\our-lite (explicit branch only, $\Lrec{+}\Lcov$; no router, gates or auxiliary terms; same schedule, target zones and cap) & 0.923 & 0.953 & 0.80 & 0.47 \\
\midrule
Shuffled anchors & 0.914 & 0.934 & 0.60 & 0.36 \\
Zeroed anchors & 0.912 & 0.931 & 0.62 & 0.37 \\
Generic caption & 0.913 & 0.936 & 0.68 & 0.40 \\
Paraphrased anchors & 0.928 & 0.964 & 0.84 & 0.50 \\
\bottomrule
\end{tabular}%
}
\end{table}

\begin{table}[!htbp]
\centering
\caption{The training-only form on WorldView-III (three seeds): $\Lcov$ with the explicit index of \our-lite added to the training objective of three existing generators at $\lambda_{\mathrm{cov}}{=}0.3$ after a $200$-epoch warm-up; the frozen encoder indexes the input and re-encodes the output during training only, so inference is the original generator. Last row: the full model with the encoder at inference.}
\label{tab:plugin}
\scriptsize
\setlength{\tabcolsep}{3pt}
\maxsizebox{\columnwidth}{%
\begin{tabular}{lcccccc}
\toprule
Generator & Q2$^n$ & HQNR & $\cdino$ & mAP (OpenCLIP) & det.\ mAP & ms / $256^2$ \\
\midrule
CANConv & 0.920 & 0.950 & 0.61 & 0.397 & 31.6 & 22 \\
CANConv $+\,\Lcov$ & 0.924 & 0.954 & 0.71 & 0.452 & 36.0 & 22 \\
ARConv & 0.921 & 0.957 & 0.63 & 0.403 & 32.3 & 27 \\
ARConv $+\,\Lcov$ & 0.925 & 0.960 & 0.72 & 0.461 & 36.5 & 27 \\
Plain U-Net & 0.910 & 0.924 & 0.55 & 0.370 & 30.9 & 9 \\
Plain U-Net $+\,\Lcov$ & 0.916 & 0.933 & 0.74 & 0.440 & 35.4 & 9 \\
\rowcolor{ourrow}\our (full, encoder at inference) & 0.929 & 0.965 & 0.85 & 0.503 & 40.2 & 133 \\
\bottomrule
\end{tabular}%
}
\end{table}

\subsection{Design controls: concept size, gradient cap, target zones, encoder grid, routing, aggregation and index errors}\label{sec:controls}
\cref{tab:controls} summarises the controls that each test one claim of the method; the complete tables follow, each on WorldView-III with three seeds and, unless stated, the full objective.

\noindent\textbf{Concept size (\cref{tab:areabins}).} \cref{supp:prop:area} is a statement about gradient shares; \cref{tab:area} verifies it in gradient space and \cref{tab:areabins} in outcome space. Concepts of the annotated full-resolution tiles are binned by the token support of their extent at the $2\times$ grid, and the recall of the frozen DINOv2 probe is computed per bin for the injected architecture under $\Lrec$ plus one term, the rows of Main Table~\MTabGains{}a. Three observations follow. The gain of the set-level loss over the strongest token-averaged loss, both as the only term added to $\Lrec$, is a decreasing function of support size, $+0.13$, $+0.08$, $+0.03$ and $0.00$ over the four bins, and on tile-filling concepts the token-averaged loss already reaches $0.90$, so there is nothing left to gain there, as the proposition predicts; the full model, listed for reference, adds $0.03$ to the smallest bin on top of this. The top-$K$ presence accounts for about a third of the small-concept gain: with the presence aggregated over all tokens the smallest bin reaches $0.72$ instead of $0.76$, because a four-token response is diluted by the $1{,}020$ background tokens of a scene (\cref{tab:calib}), while the two largest bins are unaffected. The ordering of the rows is the same in every bin, so the choice of loss does not trade one concept size against another.

\noindent\textbf{Gradient cap (\cref{tab:cap}).} The cap guards the first closed-loop epochs. The semantic-to-reconstruction gradient-norm ratio is $2.6\%$ at epoch $200$ and settles at $\fGradRatio\%$ (\cref{tab:trainlog}), below the $5\%$ cap, so the cap is inactive for most of training; its purpose is the transient after $\Lcov$ is switched on, when saturation has just cleared (\cref{sec:monitors}) and the ratio would otherwise peak at $11\%$. Without the cap Q2$^n$ loses $0.008$ and the cross-encoder gap between the host and the monitor encoder widens from $0.05$ to $0.09$: the output moves toward what the host reads before the monitor agrees, the first sign of the regime that \cref{tab:gaming} maps at $\lambda_{\mathrm{rec}}{<}1$. A cap of $1\%$ removes most of the loop's effect ($0.81$ coverage), and between $2$ and $10\%$ every measure lies within seed noise of the default, so the choice is not delicate.

\noindent\textbf{Target zones (\cref{tab:zones}).} The three-zone target has two parameters and fails in opposite ways on either side. A single threshold at $0.5$ turns every unit near the decision boundary, where the two annotators of \cref{sec:annot} disagreed most, into a hard target; the loss then pursues concepts that are not reliably present, and the hallucinated-concept rate rises from $0.066$ to $0.072$ while coverage falls by $0.02$. Widening the ignore zone removes those units but also removes supervision: at $0.2/0.8$, $44\%$ of the units of a training patch are ignored, coverage drops to $0.81$ and the hallucination rate reaches its minimum ($0.063$), a precision--recall trade that the default resolves at $13\%$ ignored. The soft target $p_k{=}w_k$, which has no zones, is the worst setting on coverage and hallucination at once: its derivative vanishes when $p_k/(1-p_k)$ equals the current Jaccard and changes sign below it (\cref{sec:theory}), so half-recovered concepts receive no push or the wrong one.

\noindent\textbf{Encoder grid (\cref{tab:grid}).} The host's $16$-px patch sets the finest concept the loop can see. At the $1\times$ grid a token covers $16{\times}16$ output pixels and a vehicle at $0.31$\,m is one token, so the top-$K_{\mathrm{local}}{=}4$ aggregate has nothing to average and local recall falls from $0.79$ to $0.70$; meso and global concepts are almost unaffected, and Q2$^n$ loses $0.004$. The $2\times$ grid is the knee: a further step to $3\times$ adds $0.01$ overall coverage and $0.02$ local recall for $2.4\times$ the inference time and $2.3\times$ the FLOPs. Encoder cost rises faster than linearly in the token count because of attention, so denser grids are the natural place for a lighter dense encoder (\cref{sec:limits}).

\noindent\textbf{Scale routing (\cref{tab:routing}).} Global and meso units reach their level under any routing, since the bottleneck and the intermediate levels see the whole tile; local units do not. Sending every unit to the bottleneck, or reversing the routes so that local units enter the bottleneck and global units the full-resolution level, costs $0.07$ coverage and $0.10$--$0.11$ local recall, with meso and global recall within $0.02$ of the default. Sending every unit to every level recovers part of the loss ($0.82$) at a cost of $0.003$ Q2$^n$: low-confidence local units at the bottleneck act as noise on the global structure, which is what the $\log w_k$ bias of Main Eq.~\MEqAttn{} mitigates but cannot remove when the units are at the wrong level.

\noindent\textbf{Presence aggregation (\cref{tab:aggregate}).} The aggregate decides whether the loop is grid-independent. The mean over tokens behaves like a token-averaged loss in outcome space ($0.76$ coverage) and reads the same concept as present on a $64$-token patch and absent on a $1{,}024$-token scene ($0.84$ against $0.58$ in-loop coverage); plain mellowmax halves that drift but keeps it. The hard maximum is grid-independent but reads one token, so one bright pixel can satisfy it, and a single $K{=}16$ for every scale under-serves local units; the scale-dependent top-$K$ is best on every measure, by a small but consistent margin over the hard maximum.

\noindent\textbf{What the index misses during training (\cref{tab:ledgerrecall}).} The input ledger is computed from $\mathbf{X}^{\uparrow}\oplus\mathbf{A}$ on the $2\times$ grid, where one token covers $8{\times}8$ output pixels, and the gate and the calibrator both read the support area, so small concepts are the ones the index misses most: $42\%$ of the annotated concepts of at most four tokens are not in the present zone ($19\%$ ignored, $15\%$ absent, $8\%$ removed by the gates), against $7\%$ of tile-filling concepts. A concept in the absent zone receives the push $+\mathcal{I}/\mathcal{U}^2$ of Main Eq.~\MEqLcov{} toward absence, which raises the question whether the loss suppresses what the index did not find. \cref{tab:ledgerrecall} answers it on the output: for concepts the index placed in the absent zone, the output recall of \our is $0.45$--$0.58$ against $0.36$--$0.55$ for the open-loop model trained without $\Lcov$, so these concepts are not suppressed below the open-loop level (over the ignored and the absent or removed concepts together, the $42\%$ outside the present zone have an output recall of $0.56$ for \our against $0.43$ open-loop), because the semantic gradient is capped at $5\%$ of the reconstruction gradient and settles at $\fGradRatio\%$, and the reconstruction term, which does not depend on the ledger, keeps them; the gain of the loop, however, concentrates on the concepts the index found ($0.95$ against $0.70$ on present small concepts). The limitation is therefore one of reach, not of suppression. A failure mode of the same origin is the ringing next to bright compact objects in \cref{fig:failure}c, the main source of the residual false positives of the small-vehicle class.

\noindent\textbf{Area invariance and the set-level coupling are separable (\cref{tab:areabins}, \cref{tab:objectives_full}, \cref{tab:objective}).} Four objectives on the same injected architecture, each the only term added to $\Lrec$, isolate the two ingredients of $\Lcov$. The token-averaged anchor-response loss is area-weighted and per-concept ($0.79$ coverage); re-weighting the token loss by inverse concept area with the ledger's membership masks, the Generalised-Dice counterpart of \cref{supp:prop:area}, is area-invariant and per-concept ($0.81$); per-concept BCE on the top-$K$ presences is area-invariant through the aggregate and still per-concept ($0.80$); and $\Lcov$ with the top-$K$ presence is set-level and area-invariant ($0.83$). The set-level coupling over an area-weighted aggregate is tested in the replacement setting: $\Lcov$ with a mean presence inside the full model gives $0.76$ against $0.85$ with the top-$K$ presence (\cref{tab:aggregate}). Area invariance alone, by either route, recovers one quarter to one half of the margin over the token-averaged loss and the two ingredients together recover all of it, while the set-level coupling over an area-weighted aggregate loses $0.09$ in the replacement setting; under this classification the two ingredients are complementary rather than additive. Since the objectives differ in more than these two attributes, the comparison characterises the five objectives and does not isolate an interaction term. The replacement setting agrees: the ablations of \cref{tab:objective} that keep the top-$K$ presence inside the full model and vary only the coupling (BCE, focal, Dice, Lov\'asz, Jaccard) span $0.03$ coverage, while the aggregate ablation of \cref{tab:aggregate} spans $0.09$. By concept size (\cref{tab:areabins}) the picture is the same: GDL re-weighting and BCE on presences lift the smallest bin from $0.63$ to $0.69$--$0.71$, the set-level loss to $0.76$, and no objective differs on tile-filling concepts.

\noindent\textbf{Index errors at test time (\cref{tab:ledgernoise}).} The index can err in two directions and the generator must tolerate both. Dropping present units at random lowers coverage gracefully, $0.84$, $0.82$ and $0.78$ at $10$, $25$ and $50\%$, and a null ledger gives $0.73$, which remains above the open-loop model ($0.72$) and far above the plain U-Net ($0.55$): part of what the closed loop teaches is in the generator weights, not in the injected units, which is also why the training-only form of \cref{tab:plugin} works without injection. Adding spurious units with confidence $0.7$ changes coverage by at most $0.01$ and raises the hallucinated-concept rate from $0.066$ to $0.073$ at $50\%$ spurious units, an increase of $0.007$ in the share of introduced concepts among the concepts found: a position receives a unit only where the content matches it (the attention of Main Eq.~\MEqAttn), and the learned null token absorbs the rest. Shuffling the confidences among the units, which keeps the set and destroys the $\log w_k$ bias, costs $0.03$ coverage and $0.004$ hallucination, so the bias is worth about a third of the robustness to a corrupted index. Detection mAP moves with coverage in every row.

\begin{table}[!htbp]
\centering
\caption{\textbf{Design controls} (WorldView-III, three seeds): one choice changed at a time; first row: full model. Complete tables, and the index perturbed at test time, in \supp.}
\label{tab:controls}
\scriptsize
\setlength{\tabcolsep}{3pt}
\renewcommand{\arraystretch}{0.88}
\maxsizebox{\columnwidth}{%
\begin{tabular}{llccc}
\toprule
Choice & Setting & Q2$^n$ & $\cdino$ & mAP \\
\midrule
\rowcolor{ourrow}default & cap $5\%$, zones $0.4/0.6$, $2\times$ grid, routed & \best{0.929} & \best{0.85} & \best{0.503} \\
gradient cap & none (ratio peaks at $11\%$) & 0.921 & 0.83 & 0.491 \\
target zones & soft target $p_k{=}w_k$ (no zones) & 0.924 & 0.80 & 0.470 \\
encoder grid & $1\times$ ($16$-px token, $52$\,ms) & 0.925 & 0.80 & 0.481 \\
\bottomrule
\end{tabular}%
}
\end{table}

\begin{table}[!htbp]
\centering
\caption{Concept recall by support size: the outcome-space counterpart of \cref{supp:prop:area} (WorldView-III annotated full-resolution tiles, DINOv2 probe, three seeds). Every row but the last is the injected architecture with exactly one term added to $\Lrec$ under the same schedule (the rows of Main Table~\MTabGains{}a); concepts are binned by the token support of their annotated extent at the $2\times$ grid. The last row is the difference between the set-level loss with top-$K$ presence and the strongest position-averaged loss (anchor responses $\to$ reference). The gain is largest on the smallest concepts and vanishes on tile-filling ones, which a position-averaged loss already serves. Three rows separate the two ingredients of $\Lcov$: inverse-area re-weighting of the token loss (GDL) and per-concept BCE on the top-$K$ presences each recover about half of the small-concept gain, and $\Lcov$ with a mean presence recovers none of it; the area-invariant aggregate and the set-level coupling are needed together.}
\label{tab:areabins}
\scriptsize
\setlength{\tabcolsep}{4pt}
\maxsizebox{\columnwidth}{%
\begin{tabular}{lccccc}
\toprule
Term added to $\Lrec$ & $\le4$ tokens & $5$--$16$ & $17$--$64$ & $>64$ & all \\
\midrule
none & 0.58 & 0.68 & 0.77 & 0.84 & 0.72 \\
token features $\to$ reference & 0.62 & 0.74 & 0.83 & 0.89 & 0.78 \\
anchor responses $\to$ reference & 0.63 & 0.75 & 0.84 & 0.90 & 0.79 \\
token features $\to$ reference, area-normalised (GDL) & 0.71 & 0.81 & 0.86 & 0.90 & 0.81 \\
BCE on the top-$K$ presences (per concept) & 0.69 & 0.79 & 0.85 & 0.90 & 0.80 \\
$\Lcov$, mean presence over all tokens & 0.62 & 0.74 & 0.85 & 0.90 & 0.76 \\
$\Lcov$, presence over all tokens & 0.72 & 0.81 & 0.86 & 0.90 & 0.82 \\
\rowcolor{ourrow}$\Lcov$, top-$K$ presence (the term of \our) & \best{0.76} & \best{0.83} & \best{0.87} & \best{0.90} & \best{0.83} \\
\midrule
gain over anchor responses $\to$ reference & $+$0.13 & $+$0.08 & $+$0.03 & $+$0.00 & $+$0.04 \\
\midrule
full \our, all terms (reference row) & 0.79 & 0.85 & 0.88 & 0.90 & 0.85 \\
\bottomrule
\end{tabular}%
}
\end{table}

\begin{table}[!htbp]
\centering
\caption{What the index does with the annotated concepts of the input, by token support (WorldView-III annotated full-resolution tiles; the ledger is computed from $\mathbf{X}^{\uparrow}\oplus\mathbf{A}$ exactly as in training, three seeds). Status: share of annotated concepts whose matching unit is in the present zone ($w_k\ge0.6$), in the ignore zone, in the absent zone, or removed by the gates before calibration. Precision: share of present units that match an annotated concept. Recall: concept recall of the DINOv2 probe on the \emph{output}, separately for the concepts of each status (present / ignored / absent or removed) and in total, for \our and for the open-loop model trained without $\Lcov$. The index misses small concepts most often, and on the concepts it misses the output recall of \our stays above the open-loop model: the push toward absence in Main Eq.~\MEqLcov{} is bounded by the gradient cap and the reconstruction term and does not suppress what the index did not find, but the gain of the loop concentrates on what it did.}
\label{tab:ledgerrecall}
\scriptsize
\setlength{\tabcolsep}{3pt}
\maxsizebox{\columnwidth}{%
\begin{tabular}{lcccccc}
\toprule
 & status of annotated concepts (\%) & ledger & \multicolumn{2}{c}{output recall, \our} & \multicolumn{2}{c}{output recall, open loop} \\
Support & present / ignored / absent / removed & precision & by status & all & by status & all \\
\\midrule
$\le4$ tokens & 58 / 19 / 15 / 8 & 0.74 & 0.95 / 0.70 / 0.45 & 0.79 & 0.70 / 0.52 / 0.36 & 0.59 \\
$5$--$16$ & 74 / 14 / 9 / 3 & 0.81 & 0.94 / 0.73 / 0.47 & 0.85 & 0.74 / 0.58 / 0.40 & 0.68 \\
$17$--$64$ & 86 / 9 / 4 / 1 & 0.88 & 0.91 / 0.75 / 0.50 & 0.88 & 0.80 / 0.64 / 0.46 & 0.77 \\
$>64$ & 93 / 5 / 2 / 0 & 0.92 & 0.92 / 0.78 / 0.58 & 0.91 & 0.86 / 0.70 / 0.55 & 0.85 \\
\bottomrule
\end{tabular}%
}
\end{table}

\begin{table}[!htbp]
\centering
\caption{Semantic gradient cap on WorldView-III (three seeds; pooled seed std $0.002$ Q2$^n$, $0.01$ coverage). The cap rescales the semantic terms whenever their output-space gradient norm exceeds the stated fraction of $\|\nabla\Lrec\|$. Measured ratio: mean ratio over epochs $200$--$1{,}000$ (peak in the first $25$ closed-loop epochs in brackets). Cross-encoder gap: in-loop coverage under the host minus coverage under the monitor encoder on the same validation patches with the same target. The last row is not a setting of the cap: it is the open-loop model of Main Table~\MTabGains{}b ($\Lrec$, $\Ltt$ and $\Ltm$, without $\Lcov$ and $\Lsa$), listed as the reference that the sweep is read against.}
\label{tab:cap}
\scriptsize
\setlength{\tabcolsep}{3pt}
\maxsizebox{\columnwidth}{%
\begin{tabular}{lcccccccc}
\toprule
Cap & ratio (\%) & Q2$^n$ & HQNR & $\csig$ & $\cmon$ & x-enc.\ gap & $\cdino$ & mAP$_{\mathrm{OC}}$ \\
\midrule
none & 4.8 (11.2) & 0.921 & 0.956 & 0.93 & 0.84 & 0.09 & 0.83 & 0.491 \\
$20\%$ & 2.6 (8.1) & 0.925 & 0.960 & 0.92 & 0.85 & 0.07 & 0.84 & 0.497 \\
$10\%$ & 2.0 (5.4) & 0.928 & 0.964 & 0.91 & 0.86 & 0.05 & 0.85 & 0.502 \\
\rowcolor{ourrow}$5\%$ (default) & 1.9 (4.6) & \best{0.929} & \best{0.965} & 0.91 & 0.86 & 0.05 & \best{0.85} & \best{0.503} \\
$2\%$ & 1.6 (2.0) & 0.928 & 0.964 & 0.90 & 0.86 & 0.04 & 0.84 & 0.499 \\
$1\%$ & 1.0 (1.0) & 0.925 & 0.961 & 0.87 & 0.84 & 0.03 & 0.81 & 0.487 \\
open-loop model (reference) & -- & 0.918 & 0.954 & 0.80 & 0.76 & 0.04 & 0.72 & 0.430 \\
\bottomrule
\end{tabular}%
}
\end{table}

\begin{table}[!htbp]
\centering
\caption{The three-zone target on WorldView-III (three seeds). Zones are set on the frozen confidence $w_k$: absent at or below the first threshold, present at or above the second, ignored between. Ignored share: fraction of ledger units per training patch that fall in the ignore zone. Hallucinated-concept rate: CHAIR-style, ConvNeXt-B evaluator (\cref{tab:precision}). The soft target uses $p_k{=}w_k$ with no zones; its derivative changes sign on half-recovered concepts (\cref{sec:theory}).}
\label{tab:zones}
\scriptsize
\setlength{\tabcolsep}{3pt}
\maxsizebox{\columnwidth}{%
\begin{tabular}{lcccccc}
\toprule
Target (absent $\le$ / present $\ge$) & ignored (\%) & Q2$^n$ & $\csig$ & $\cdino$ & mAP$_{\mathrm{OC}}$ & halluc.$\downarrow$ \\
\midrule
single threshold $0.5$ & 0 & 0.926 & 0.90 & 0.83 & 0.49 & 0.072 \\
$0.45$ / $0.55$ & 6 & 0.928 & 0.91 & 0.84 & 0.50 & 0.069 \\
\rowcolor{ourrow}$0.40$ / $0.60$ (default) & 13 & \best{0.929} & \best{0.91} & \best{0.85} & \best{0.503} & 0.066 \\
$0.30$ / $0.70$ & 27 & 0.928 & 0.89 & 0.84 & 0.49 & 0.064 \\
$0.20$ / $0.80$ & 44 & 0.925 & 0.86 & 0.81 & 0.47 & \best{0.063} \\
soft target $p_k{=}w_k$ & 0 & 0.924 & 0.87 & 0.80 & 0.47 & 0.078 \\
\bottomrule
\end{tabular}%
}
\end{table}

\begin{table}[!htbp]
\centering
\caption{Encoder grid on WorldView-III (three seeds). The adapter output is up-sampled by the stated factor before it enters the frozen encoder ($16$-px patches), so one token covers the stated output footprint; $K_s$ is scaled with the grid. Local recall: concept recall of the DINOv2 probe on local concepts of the annotated full-resolution tiles. Cost per $256^2$ scene on an RTX~4090, for generation alone.}
\label{tab:grid}
\scriptsize
\setlength{\tabcolsep}{3pt}
\maxsizebox{\columnwidth}{%
\begin{tabular}{lcccccccc}
\toprule
Grid & token footprint & tokens / scene & Q2$^n$ & HQNR & $\cdino$ & local recall & mAP$_{\mathrm{OC}}$ & ms / GFLOPs \\
\midrule
$1\times$ & $16{\times}16$ px & $256$ & 0.925 & 0.959 & 0.80 & 0.70 & 0.481 & $52$ / $262$ \\
\rowcolor{ourrow}$2\times$ (default) & $8{\times}8$ px & $1{,}024$ & \best{0.929} & \best{0.965} & 0.85 & 0.79 & 0.503 & $133$ / $936$ \\
$3\times$ & $5.3{\times}5.3$ px & $2{,}304$ & \best{0.929} & \best{0.965} & \best{0.86} & \best{0.81} & \best{0.506} & $318$ / $2{,}150$ \\
\bottomrule
\end{tabular}%
}
\end{table}

\begin{table}[!htbp]
\centering
\caption{Scale routing of the ledger on WorldView-III (three seeds). Default: local units enter the full-resolution decoder level, meso units the two intermediate levels, global units the bottleneck. Coverage is given overall and by scale (local / meso / global concepts of the annotated full-resolution tiles). Last row: no injection at all, i.e.\ the plain U-Net trained with $\Lrec{+}\Lcov$ (\cref{tab:plugin}).}
\label{tab:routing}
\scriptsize
\setlength{\tabcolsep}{3pt}
\maxsizebox{\columnwidth}{%
\begin{tabular}{lcccccc}
\toprule
Routing & Q2$^n$ & HQNR & $\cdino$ & local / meso / global & mAP$_{\mathrm{OC}}$ \\
\midrule
all units $\to$ bottleneck only & 0.921 & 0.957 & 0.78 & 0.68 / 0.80 / 0.88 & 0.47 \\
all units $\to$ every level & 0.926 & 0.962 & 0.82 & 0.74 / 0.83 / 0.89 & 0.49 \\
reversed (local $\to$ bottleneck, global $\to$ full resolution) & 0.922 & 0.958 & 0.79 & 0.69 / 0.81 / 0.88 & 0.47 \\
\rowcolor{ourrow}scale-routed (default) & \best{0.929} & \best{0.965} & \best{0.85} & \best{0.79 / 0.85 / 0.90} & \best{0.503} \\
\midrule
no injection ($\Lrec{+}\Lcov$ on the plain U-Net) & 0.916 & 0.933 & 0.74 & 0.66 / 0.75 / 0.82 & 0.44 \\
\bottomrule
\end{tabular}%
}
\end{table}

\begin{table}[!htbp]
\centering
\caption{Presence aggregation in outcome space (WorldView-III, three seeds). The aggregate turns the response map of an anchor into one scalar; everything else is the full model. $\csig$ is given on $64$-, $256$- and $1{,}024$-token inputs ($64$-px patches, $128$-px tiles, $256^2$ scenes), calibrated once on $64$-token patches: an aggregate that depends on the token count reads the same object as present on a patch and absent on a scene.}
\label{tab:aggregate}
\scriptsize
\setlength{\tabcolsep}{4pt}
\maxsizebox{\columnwidth}{%
\begin{tabular}{lcccc}
\toprule
Aggregate & Q2$^n$ & $\csig$ ($64$ / $256$ / $1{,}024$ tokens) & $\cdino$ & mAP$_{\mathrm{OC}}$ \\
\midrule
mean over all tokens & 0.922 & 0.84 / 0.71 / 0.58 & 0.76 & 0.45 \\
plain mellowmax over all tokens & 0.927 & 0.90 / 0.84 / 0.76 & 0.82 & 0.48 \\
hard maximum (top-$1$) & 0.928 & 0.90 / 0.90 / 0.90 & 0.83 & 0.49 \\
top-$K$, one $K{=}16$ for every scale & 0.928 & 0.90 / 0.89 / 0.89 & 0.83 & 0.49 \\
\rowcolor{ourrow}top-$K$ by scale, $K_s{=}(4,16,64)$ (default) & \best{0.929} & \best{0.91 / 0.90 / 0.90} & \best{0.85} & \best{0.503} \\
\bottomrule
\end{tabular}%
}
\end{table}

\begin{table}[!htbp]
\centering
\caption{Robustness to errors of the index at test time (WorldView-III, fixed checkpoint, three seeds). The ledger of each test input is perturbed before injection and the generator is run unchanged: present units are dropped at random, spurious units are added (random vocabulary anchors with confidence $0.7$ and a random scale tag), or the confidences are shuffled among the units. Coverage, hallucinated-concept rate and detection mAP$_{50}$ (DINOv2 head) on the annotated full-resolution tiles; Q2$^n$ on the reduced-resolution scenes.}
\label{tab:ledgernoise}
\scriptsize
\setlength{\tabcolsep}{4pt}
\maxsizebox{\columnwidth}{%
\begin{tabular}{lcccc}
\toprule
Perturbation of the test-time ledger & Q2$^n$ & $\cdino$ & halluc.$\downarrow$ & det.\ mAP \\
\midrule
\rowcolor{ourrow}none & 0.929 & 0.85 & 0.066 & 40.2 \\
drop $10\%$ of present units & 0.928 & 0.84 & 0.066 & 39.8 \\
drop $25\%$ & 0.927 & 0.82 & 0.066 & 39.0 \\
drop $50\%$ & 0.924 & 0.78 & 0.067 & 37.4 \\
drop all (null ledger) & 0.919 & 0.73 & 0.068 & 35.6 \\
\midrule
add $10\%$ spurious units & 0.929 & 0.85 & 0.067 & 40.1 \\
add $25\%$ & 0.928 & 0.85 & 0.069 & 39.9 \\
add $50\%$ & 0.927 & 0.84 & 0.073 & 39.4 \\
shuffle confidences among units & 0.926 & 0.82 & 0.070 & 38.8 \\
\midrule
open-loop \our (reference point) & 0.918 & 0.72 & 0.071 & 35.2 \\
\bottomrule
\end{tabular}%
}
\end{table}

\subsection{Stage III input, anchors, presence aggregation and sensitivity}\label{toc:supp:5:8}
\begin{figure}[!htbp]
\centering
\includegraphics[width=\columnwidth]{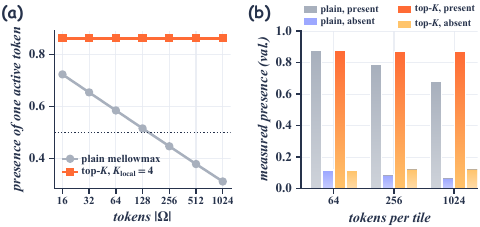}
\caption{Presence aggregation against the token count. (a) Closed-form presence of one fully active token under plain mellowmax and under the top-$K$ aggregate. (b) Measured present / absent presence of validation concepts on $64$-, $256$- and $1{,}024$-token inputs, calibrated once on $64$-token patches.}
\label{fig:calibration}
\end{figure}

\begin{figure*}[!tbp]
\centering
\includegraphics[width=\textwidth]{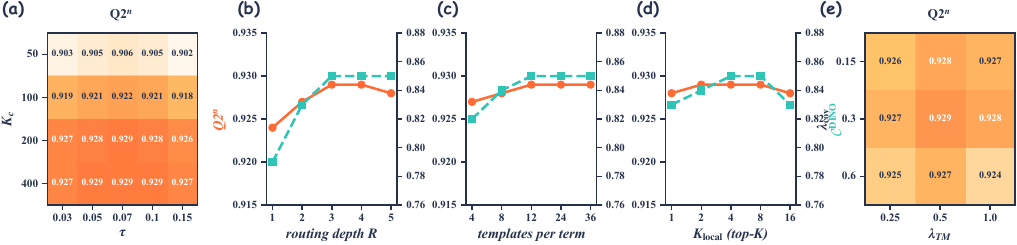}
\caption{Hyper-parameter sensitivity on the WorldView-III validation split. (a) Q2$^n$ over the $K_c\times\tau$ grid; (b) routing depth; (c) templates per term; (d) $K_{\mathrm{local}}$ of the top-$K$ aggregate; (e) Q2$^n$ over the loss-weight grid. The surface is flat within $0.01$ Q2$^n$ around the chosen values.}
\label{fig:sensitivity}
\end{figure*}

\begin{figure*}[!tbp]
\centering
\includegraphics[width=\textwidth]{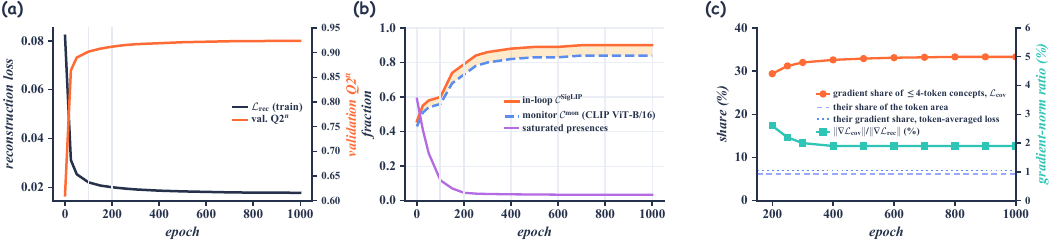}
\caption{Training log of the WorldView-III main run (seed $0$) on the official validation split. (a) Reconstruction loss and validation Q2$^n$; vertical lines mark the phases ($\Ltt,\Ltm$ at epoch $100$; $\Lcov,\Lsa$ at $200$). (b) In-loop coverage, monitor-encoder coverage with the same target, their gap, and the saturated-presence fraction. (c) Share of the coverage gradient that reaches concepts of at most four tokens against their share of the token area and their gradient share under a token-averaged loss (\cref{supp:prop:area}), and the semantic-to-reconstruction gradient-norm ratio.}
\label{fig:training}
\end{figure*}

\cref{tab:adapter} isolates what Stage~III re-encodes: the default encodes $\hat{\mathbf{Y}}$ alone through $a'_d$; concatenating the guide and sharing the Stage-I adapter raises in-loop coverage by $0.02$, because the panchromatic band carries structure of its own, while $\cdino$, detection and retrieval do not move; a fixed natural-colour matrix or a random initialisation of $a'_d$ stays within seed noise. \cref{tab:anchor} shows that the coverage gain grows with the anchor's assignment cosine, that the $27$ unnamed centroids supervised through visual prototypes receive a small but positive gain, and that a model restricted to the $71$ text anchors above $0.40$ keeps $0.82$ of the $0.85$ coverage. \cref{tab:calib} and \cref{fig:calibration} analyse presence aggregation against the token count: plain mellowmax over all tokens assigns a single fully active token the presence $0.584$, $0.447$ and $0.311$ on $64$, $256$ and $1{,}024$ tokens, so the same object would be present on a training patch and absent on a scene; the top-$K$ aggregate gives $0.861$ at every grid size. \cref{tab:sweeps} and \cref{fig:sensitivity} sweep $K_c$, $\tau$, $R$, the template count, $K_{\mathrm{local}}$ and the loss weights on the validation split: the default sits on a plateau, the only steep direction being $K_c$ below $100$. \cref{tab:trainlog} and \cref{fig:training} give the training log of the main run on the validation split.

\begin{table}[!htbp]
\centering
\caption{The two adapters (WorldView-III, three seeds). Rows 1--4: what Stage III re-encodes. The default encodes the output $\hat{\mathbf{Y}}$ alone through its own adapter $a'_d$ ($3{\times}B$), so the re-index score speaks about the archived product; concatenating the guide $\mathbf{A}$ raises in-loop coverage by $0.02$ (the guide carries structure of its own) without moving any outside evaluator, and a fixed natural-colour matrix or a random initialisation of $a'_d$ stays within seed noise. Last row: the Stage-I adapter $a_d$, which precedes the ledger and is otherwise trained by $\Lrec$ through the memberships, fixed at the natural-colour matrix; the ledger is then a function of frozen quantities only, and nothing changes beyond seed noise.}
\label{tab:adapter}
\scriptsize
\setlength{\tabcolsep}{3pt}
\maxsizebox{\columnwidth}{%
\begin{tabular}{lccccc}
\toprule
Stage III input and adapter & Q2$^n$ & HQNR & $\csig$ & $\cdino$ & mAP (OpenCLIP) \\
\midrule
\rowcolor{ourrow}$\hat{\mathbf{Y}}$ through its own adapter $a'_d$ (default) & 0.929 & 0.965 & 0.91 & 0.85 & 0.503 \\
$\hat{\mathbf{Y}}\oplus\mathbf{A}$ through the Stage-I adapter $a_d$ (shared) & 0.929 & 0.965 & 0.93 & 0.85 & 0.502 \\
$\hat{\mathbf{Y}}$ through the fixed natural-colour matrix & 0.927 & 0.963 & 0.90 & 0.84 & 0.498 \\
$\hat{\mathbf{Y}}$ through a randomly initialised $a'_d$ & 0.928 & 0.964 & 0.90 & 0.85 & 0.501 \\
Stage-I adapter $a_d$ fixed at the natural-colour matrix & 0.928 & 0.964 & 0.91 & 0.85 & 0.501 \\
\bottomrule
\end{tabular}%
}
\end{table}

\begin{table}[!htbp]
\centering
\caption{Anchor quality on WorldView-III. Left: centroids per assignment-cosine bin and the coverage gain of \our over the open-loop model on the concepts of that bin; the $27$ unnamed centroids keep their visual prototype $\himg(c_k)$ as anchor and are supervised like the others. Right: \our retrained with only the text anchors above a cosine threshold (the remaining centroids fall back to visual prototypes), and with visual prototypes only.}
\label{tab:anchor}
\scriptsize
\setlength{\tabcolsep}{3pt}
\maxsizebox{\columnwidth}{%
\begin{tabular}{lcc|lccc}
\toprule
Cosine bin & centroids & $\Delta\cdino$ & Anchors kept & $n$ & Q2$^n$ & $\cdino$ \\
\midrule
$0.20$--$0.30$ & 61 & $+$0.09 & all named + visual prototypes & 200 & 0.929 & 0.85 \\
$0.30$--$0.40$ & 68 & $+$0.14 & cosine $>0.30$ (+ prototypes) & 139 & 0.928 & 0.84 \\
$0.40$--$0.50$ & 33 & $+$0.19 & cosine $>0.40$ (+ prototypes) & 71 & 0.927 & 0.82 \\
$>0.50$ & 11 & $+$0.21 & cosine $>0.50$ (+ prototypes) & 38 & 0.920 & 0.74 \\
unnamed, visual prototype & 27 & $+$0.05 & visual prototypes only (no text) & 200 & 0.913 & 0.64 \\
\bottomrule
\end{tabular}%
}
\end{table}

\begin{table}[!htbp]
\centering
\caption{Presence aggregation against the token count. Columns 2--4: one fully active token among zeros under plain mellowmax over all $|\Omega|$ tokens (closed form), under the top-$K$ mellowmax used by \our ($K_{\mathrm{local}}{=}4$), and the number of fully active tokens plain mellowmax needs to exceed $0.5$. Columns 5--6: measured present / absent presence of validation concepts at each grid ($64$, $256$ and $1{,}024$ tokens are $64$-px patches, $128$-px tiles and $256^2$ scenes at the $2\times$ encoder grid), calibrated once on $64$-token patches. Scene-level coverage of the full model computed with plain mellowmax is $0.82$, with top-$K$ $0.85$.}
\label{tab:calib}
\scriptsize
\setlength{\tabcolsep}{3pt}
\maxsizebox{\columnwidth}{%
\begin{tabular}{rccc|cc}
\toprule
$|\Omega|$ & plain, 1 token & top-$K$, 1 token & tokens for $>0.5$ & plain present / absent & top-$K$ present / absent \\
\midrule
64 & 0.584 & 0.861 & 1 & 0.88 / 0.12 & 0.88 / 0.12 \\
256 & 0.447 & 0.861 & 2 & 0.79 / 0.09 & 0.87 / 0.13 \\
1024 & 0.311 & 0.861 & 7 & 0.68 / 0.07 & 0.87 / 0.13 \\
\bottomrule
\end{tabular}%
}
\end{table}

\begin{table}[!htbp]
\centering
\caption{Sensitivity on WorldView-III (Q2$^n$ / $\cdino$; one seed per non-default point, three at the default; every default was chosen on the validation split before the test scenes were evaluated). The $K_c$ row is the $\tau{=}0.07$ column of the grid in \cref{fig:sensitivity}a and the $\tau$ row its $K_c{=}200$ row. Rows of the loss-weight block are $\lambda_{\mathrm{cov}}$, columns $\lambda_{TM}$.}
\label{tab:sweeps}
\scriptsize
\setlength{\tabcolsep}{3pt}
\maxsizebox{\columnwidth}{%
\begin{tabular}{lccccc}
\toprule
$K_c$ & 50 & 100 & 200 & 400 & \\
& 0.906 / 0.79 & 0.922 / 0.83 & \best{0.929 / 0.85} & 0.929 / 0.85 & \\
\midrule
$\tau$ & 0.03 & 0.05 & 0.07 & 0.10 & 0.15 \\
& 0.927 / 0.84 & 0.928 / 0.85 & \best{0.929 / 0.85} & 0.928 / 0.85 & 0.926 / 0.84 \\
\midrule
$R$ & 1 & 2 & 3 & 4 & 5 \\
& 0.924 / 0.79 & 0.927 / 0.83 & \best{0.929 / 0.85} & 0.929 / 0.85 & 0.928 / 0.85 \\
\midrule
templates & 4 & 8 & 12 & 24 & 36 \\
& 0.927 / 0.82 & 0.928 / 0.84 & \best{0.929 / 0.85} & 0.929 / 0.85 & 0.929 / 0.85 \\
\midrule
$K_{\mathrm{local}}$ (top-$K$) & 1 & 2 & 4 & 8 & 16 \\
& 0.928 / 0.83 & 0.929 / 0.84 & \best{0.929 / 0.85} & 0.929 / 0.85 & 0.928 / 0.83 \\
\midrule
$\lambda_{\mathrm{cov}}\backslash\lambda_{TM}$ & 0.25 & 0.5 & 1.0 & & \\
0.15 & 0.926 / 0.83 & 0.928 / 0.84 & 0.927 / 0.84 & & \\
0.30 & 0.927 / 0.84 & \best{0.929 / 0.85} & 0.928 / 0.85 & & \\
0.60 & 0.925 / 0.85 & 0.927 / 0.85 & 0.924 / 0.85 & & \\
\bottomrule
\end{tabular}%
}
\end{table}

\begin{table}[!htbp]
\centering
\caption{Training log of the WorldView-III main run (seed $0$) on the \emph{official validation split} ($1{,}080$ patches; monitors on a fixed batch of $32$), every $25$--$100$ epochs: reconstruction loss (training batches), validation ERGAS and Q2$^n$ (EMA weights), in-loop coverage $\csig$ and monitor-encoder coverage $\cmon$ (CLIP ViT-B/16, same patches, same input-ledger target), saturated-presence fraction, and, once $\Lcov$ is active (epoch $200$), the ratio of the semantic to the reconstruction gradient norm at the output (the quantity the cap acts on, \cref{alg:train}) and in parameter space (over $\theta$ and $a'_d$), and the share of the coverage gradient that reaches concepts of at most four tokens (their share of the token area is $6.1\%$; \cref{supp:prop:area}). $\Ltt,\Ltm$ start at epoch $100$. Validation values differ from the test-scene values of \cref{tab:full_rr_b} (e.g.\ ERGAS $2.05$ against $1.962$ on the test scenes), as they should.}
\label{tab:trainlog}
\scriptsize
\setlength{\tabcolsep}{2.5pt}
\maxsizebox{\columnwidth}{%
\begin{tabular}{rccccccccc}
\toprule
Ep. & $\Lrec$ & ERGAS & Q2$^n$ & $\csig$ & $\cmon$ & sat.\ (\%) & grad.\ (\%) & grad.\ $\theta$ (\%) & local share (\%) \\
\midrule
0 & 0.0821 & 7.41 & 0.611 & 0.46 & 0.43 & 59.0 & off & off & off \\
25 & 0.0312 & 3.52 & 0.864 & 0.55 & 0.51 & 41.2 & off & off & off \\
50 & 0.0254 & 2.97 & 0.890 & 0.58 & 0.54 & 27.5 & off & off & off \\
100 & 0.0221 & 2.63 & 0.902 & 0.60 & 0.56 & 11.8 & off & off & off \\
150 & 0.0208 & 2.49 & 0.908 & 0.74 & 0.68 & 7.1 & off & off & off \\
200 & 0.0201 & 2.41 & 0.912 & 0.79 & 0.73 & 4.6 & 2.6 & 3.1 & 29.4 \\
250 & 0.0196 & 2.34 & 0.915 & 0.84 & 0.78 & 4.1 & 2.2 & 2.7 & 31.2 \\
300 & 0.0192 & 2.27 & 0.917 & 0.86 & 0.80 & 3.9 & 2.0 & 2.5 & 32.0 \\
400 & 0.0187 & 2.19 & 0.919 & 0.88 & 0.82 & 3.7 & 1.9 & 2.3 & 32.6 \\
500 & 0.0184 & 2.13 & 0.921 & 0.89 & 0.83 & 3.6 & 1.9 & 2.2 & 32.9 \\
600 & 0.0182 & 2.10 & 0.922 & 0.89 & 0.83 & 3.5 & 1.9 & 2.1 & 33.1 \\
700 & 0.0180 & 2.08 & 0.923 & 0.90 & 0.84 & 3.5 & 1.9 & 2.1 & 33.2 \\
800 & 0.0179 & 2.06 & 0.923 & 0.90 & 0.84 & 3.4 & 1.9 & 2.1 & 33.3 \\
900 & 0.0179 & 2.05 & 0.924 & 0.90 & 0.84 & 3.4 & 1.9 & 2.1 & 33.3 \\
1000 & 0.0178 & 2.05 & 0.924 & 0.90 & 0.84 & 3.4 & 1.9 & 2.1 & 33.3 \\
\bottomrule
\end{tabular}%
}
\end{table}

\subsection{Robustness, cross-sensor transfer and the unseen sensor}\label{toc:supp:5:9}
\begin{figure*}[!tbp]
\centering
\includegraphics[width=\textwidth]{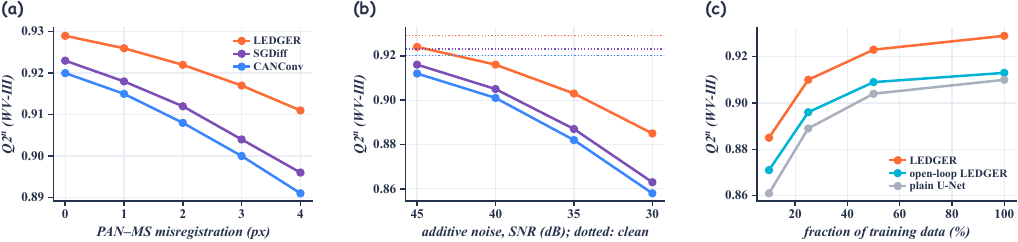}
\caption{Robustness on fixed WorldView-III checkpoints: (a) PAN--MS misregistration, (b) additive noise, (c) fraction of training data.}
\label{fig:robustness}
\end{figure*}

\begin{figure}[!htbp]
\centering
\includegraphics[width=\columnwidth]{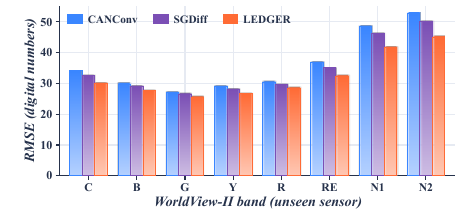}
\caption{Per-band RMSE on the unseen WorldView-II sensor (WorldView-III models). The baselines' error and the gain of \our both concentrate in the coastal, red-edge and near-infrared bands.}
\label{fig:wv2}
\end{figure}

\cref{fig:robustness} subjects fixed checkpoints to perturbations never seen in training. Under one to four pixels of PAN--MS misregistration \our loses $0.018$ Q2$^n$ where SGDiff loses $0.027$ and CANConv $0.029$; under additive noise at $30$\,dB SNR it loses $0.044$ against $0.060$ and $0.062$; trained on $10\%$, $25\%$ and $50\%$ of the data it keeps a $0.014$--$0.016$ margin over open-loop \our. The presence-gated text--mask term carries the misregistration robustness (removing it doubles the loss at $4$\,px). \cref{tab:cross} gives every method without fine-tuning across sensors; heterogeneous transfers use a Gaussian band-overlap mapping ($\sigma{=}45$\,nm) applied identically to every method. \cref{tab:wv2bands} and \cref{fig:wv2} locate the unseen-sensor margin of Main Table~\MTabMain{} in the coastal, red-edge and near-infrared bands; with both adapters replaced by the fixed natural-colour matrix, so that the encoder never sees those bands, the per-band gain is retained within $1$--$2\%$, so the gain does not come from spectral information read by the encoder. A constraint on the spatial structure that the bands share through the detail residual would account for it (\cref{sec:matched}); this is an interpretation, as the residual is not analysed directly.

\noindent\textbf{Archive post-processing and local fidelity (\cref{tab:postproc}).} A gain that lived in fragile, encoder-specific patterns would not survive the operations an archive applies before serving an image, so the fixed checkpoints are evaluated after integer quantisation, a vendor-style $8$-bit stretch, JPEG~2000 at two rates, a resampling round trip and a mild blur. Every operation lowers both systems by a similar amount and moves the margin of \our over SGDiff by at most $0.01$ coverage and $0.004$ mAP; at JPEG~2000 $20{:}1$, which removes most of the $8$--$16$\,px band, both lose $0.03$ coverage and the margin is unchanged. The bottom block asks whether the recovered content is spectrally right where it matters: inside the masks of the present concepts of the reduced-resolution tiles, the spectral angle to the reference is lower for \our in every support bin, by $0.5^\circ$ on concepts of at most four tokens and by $0.07^\circ$ on the tile as a whole, so the small concepts that the loop recovers are also the places where its output is closest to the reference, which a perturbation that pleased the encoder alone could not produce. Together with the evaluators without language (\cref{tab:evalfamily}), the detector trained end to end from random initialisation on WorldView-III (fourth row of \cref{tab:evalfamily}) and the human counting task (\cref{sec:human}), these checks bound the share of the gain that could be attributed to transferable feature perturbations.

\begin{table}[!htbp]
\centering
\caption{The recovered content under archive post-processing and inside concept masks (WorldView-III, three seeds; SGDiff / \our). Top: fixed checkpoints, their outputs post-processed before evaluation as an archive would store or serve them; label-grounded coverage and held-out OpenCLIP mAP on the annotated full-resolution tiles, detection mAP under the ConvNeXt-T head; the first row repeats the values of Main Table~\MTabDown. Bottom: spectral angle (degrees) against the reference on the reduced-resolution annotated tiles, computed inside the masks of the present concepts binned by token support; the whole-tile value is that of Main Table~\MTabMain.}
\label{tab:postproc}
\scriptsize
\setlength{\tabcolsep}{3pt}
\maxsizebox{\columnwidth}{%
\begin{tabular}{lccc}
\toprule
Post-processing of the output & $\cdino$ & mAP (OpenCLIP) & det.\ mAP \\
\midrule
none & 0.67 / 0.85 & 0.417 / 0.503 & 33.2 / 38.4 \\
$11$-bit integer quantisation & 0.67 / 0.85 & 0.416 / 0.502 & 33.1 / 38.3 \\
$8$-bit vendor-style stretch & 0.66 / 0.84 & 0.412 / 0.496 & 32.7 / 37.9 \\
JPEG~2000, $10{:}1$ & 0.66 / 0.84 & 0.410 / 0.494 & 32.5 / 37.7 \\
JPEG~2000, $20{:}1$ & 0.64 / 0.82 & 0.401 / 0.483 & 31.6 / 36.8 \\
bicubic resampling by $\pm3\%$ and back & 0.67 / 0.84 & 0.414 / 0.498 & 32.8 / 38.0 \\
Gaussian blur, $\sigma{=}0.5$\,px & 0.65 / 0.83 & 0.405 / 0.488 & 32.1 / 37.3 \\
\bottomrule
\end{tabular}%
}\\[4pt]
\maxsizebox{\columnwidth}{%
\begin{tabular}{lccccc}
\toprule
SAM inside concept masks (deg) & $\le4$ tokens & $5$--$16$ & $17$--$64$ & $>64$ & whole tile \\
\midrule
SGDiff & 4.41 & 3.62 & 3.04 & 2.63 & 2.780 \\
\rowcolor{ourrow}\our & \best{3.93} & \best{3.31} & \best{2.88} & \best{2.56} & \best{2.680} \\
\bottomrule
\end{tabular}%
}
\end{table}

\begin{table}[!htbp]
\centering
\caption{Cross-sensor Q2$^n$ without fine-tuning (three seeds). Heterogeneous transfers ($4{\leftrightarrow}8$ bands) use a Gaussian band-overlap mapping ($\sigma{=}45$\,nm) applied identically to every method; the last two rows repeat them without it.}
\label{tab:cross}
\scriptsize
\setlength{\tabcolsep}{2.6pt}
\maxsizebox{\columnwidth}{%
\begin{tabular}{lcccccc>{\columncolor{ourrow}}c}
\toprule
Direction & LAGConv & CrossDiff & Pan-Mamba & CANConv & ARConv & SGDiff & \our \\
\midrule
WV-III$\to$WV-II (8$\to$8) & 0.841 & 0.775 & 0.848 & 0.852 & 0.854 & 0.858 & \best{0.868} \\
GF-2$\to$QB (4$\to$4) & 0.826 & 0.836 & 0.852 & 0.858 & 0.861 & 0.866 & \best{0.878} \\
QB$\to$WV-III (4$\to$8, mapped) & 0.722 & 0.739 & 0.768 & 0.781 & 0.786 & 0.794 & \best{0.806} \\
WV-III$\to$QB (8$\to$4, mapped) & 0.838 & 0.847 & 0.859 & 0.864 & 0.867 & 0.871 & \best{0.883} \\
QB$\to$WV-III, no mapping & 0.633 & 0.649 & 0.672 & 0.684 & 0.690 & 0.698 & \best{0.708} \\
WV-III$\to$QB, no mapping & 0.812 & 0.820 & 0.833 & 0.838 & 0.840 & 0.845 & \best{0.855} \\
\bottomrule
\end{tabular}%
}
\end{table}

\begin{table}[!htbp]
\centering
\caption{Per-band RMSE (digital numbers) on the unseen WorldView-II sensor (WorldView-III models, $20$ scenes). Bands: coastal, blue, green, yellow, red, red edge, NIR1, NIR2. The last row replaces both learned adapters by the fixed natural-colour matrix, so the encoder never sees the red-edge or NIR bands; the per-band gain is retained within $1$--$2\%$.}
\label{tab:wv2bands}
\scriptsize
\setlength{\tabcolsep}{3pt}
\maxsizebox{\columnwidth}{%
\begin{tabular}{lccccccccc}
\toprule
Method & C & B & G & Y & R & RE & N1 & N2 & mean \\
\midrule
CANConv & 34.1 & 29.8 & 27.2 & 28.9 & 30.4 & 36.8 & 48.7 & 52.9 & 36.1 \\
SGDiff & 32.6 & 28.9 & 26.5 & 28.0 & 29.6 & 35.1 & 46.2 & 50.1 & 34.6 \\
\rowcolor{ourrow}\our & 29.9 & 27.6 & 25.4 & 26.8 & 28.3 & 32.3 & 41.6 & 45.1 & 32.1 \\
\our, fixed natural-colour $a_d$, $a'_d$ & 30.3 & 27.9 & 25.6 & 27.1 & 28.6 & 32.8 & 42.3 & 45.9 & 32.6 \\
\midrule
gain over SGDiff (\%) & 8 & 4 & 4 & 4 & 4 & 8 & 10 & 10 & 7.2 \\
\bottomrule
\end{tabular}%
}
\end{table}

\subsection{Closest neighbours and SpaceNet in full}\label{toc:supp:5:10}
\cref{tab:neighbours} compares, on the labelled aerial benchmark, the methods that also re-encode an image with a frozen VLM. SeeSR~\cite{wu2024seesr} and DA-CLIP~\cite{luo2024daclip} re-encode the \emph{input}; SGDiff~\cite{xing2025sgdiff} reads scene- and region-level semantics of the input through GeoChat~\cite{kuckreja2024geochat} to route experts; CLIPPan~\cite{jian2025clippan} encodes the \emph{output} but compares it with protocol-level text; PanDiff~\cite{meng2024pandiff} is a diffusion pansharpener without language. None compares the output with the concept set of its own input. Two pansharpening methods published at CVPR~2026, ScaleFormer~\cite{cao2026scaleformer} and the semantic-prototype RWKV of~\cite{li2026msps}, appeared after the comparison set was frozen; both report on other data and under other conventions, so their published values cannot be placed in Main Table~\MTabMain, and \cref{tab:concurrent} evaluates their released implementations under our protocol as a late addition that enters no statistic. Both land within the range of the published networks we retrain, the RWKV model slightly above SGDiff in coverage, consistent with its semantic guidance. \cref{tab:spacenet_aoi} gives the SpaceNet evaluation for every method, per city, and the reduced-resolution fidelity of the WorldView-III checkpoints on SpaceNet crops; with one segmenter per task and four cities, the std is over our seeds and the gain is positive in every city, so the SpaceNet margins are reported per city as consistent gains.

\begin{table}[!htbp]
\centering
\caption{Closest neighbours on the labelled aerial benchmark (three bands, WorldView-III reduced-resolution protocol; $50$-epoch fine-tuning from public checkpoints where they exist; CLIPPan, DA-CLIP and SeeSR are the official implementations; time per $256^2$ tile on an RTX~4090, \our with the encoder at the $2\times$ grid). The second column states what each method re-encodes with a frozen VLM and against what it is compared.}
\label{tab:neighbours}
\scriptsize
\setlength{\tabcolsep}{3pt}
\maxsizebox{\columnwidth}{%
\begin{tabular}{llcccc}
\toprule
Method & re-encodes & PSNR & $\cdino$ & mAP & ms \\
\midrule
CANConv & no & 31.8 & 0.61 & 0.41 & 22 \\
PanDiff (50 steps) & no & 31.5 & 0.62 & 0.42 & 1\,840 \\
SGDiff (semantic routing) & input (GeoChat~\cite{kuckreja2024geochat}) & 32.0 & 0.65 & 0.45 & 1\,210 \\
CLIPPan (adapted CLIP supervisor) & output vs.\ protocol text & 31.6 & 0.63 & 0.43 & 24 \\
DA-CLIP (official, adapted) & input (degradation-aware controller) & 31.7 & 0.66 & 0.45 & 310 \\
SeeSR (official, adapted) & input (tag prompts) & 30.2 & 0.68 & 0.46 & 2\,910 \\
DA-CLIP-style & input & 31.9 & 0.64 & 0.44 & 24 \\
\rowcolor{ourrow}\our & output vs.\ input concept set & \best{33.1} & \best{0.79} & \best{0.53} & 133 \\
\bottomrule
\end{tabular}%
}
\end{table}

\begin{table}[!htbp]
\centering
\caption{Concurrent pansharpening methods of CVPR~2026, evaluated as a late addition with their released implementations under our protocol (WorldView-III, three seeds, $1{,}000$ epochs, checkpoint by validation ERGAS). Both appeared after the comparison set of Main Sec.~\MSecGap{} was frozen and report, in their papers, on other data and under other conventions (the PanScale benchmark~\cite{cao2026scaleformer}; the WorldView-II/III and GaoFen-2 splits and radian-valued SAM of~\cite{li2026msps}), so their published values are not comparable cell by cell with Main Table~\MTabMain. They enter no statistic of the paper. The semantic-prototype RWKV reads scene semantics from the input and sits, like SGDiff, at the boundary of the two families.}
\label{tab:concurrent}
\scriptsize
\setlength{\tabcolsep}{3pt}
\maxsizebox{\columnwidth}{%
\begin{tabular}{lcccccccc}
\toprule
Method & SAM$\downarrow$ & ERGAS$\downarrow$ & Q2$^n\uparrow$ & HQNR$\uparrow$ & $\cdino$ & mAP$_{\mathrm{OC}}$ & det.\ mAP & ms \\
\midrule
SGDiff~\cite{xing2025sgdiff} (for reference) & 2.780 & 2.052 & 0.923 & 0.960 & 0.67 & 0.417 & 33.0 & 1\,210 \\
ScaleFormer~\cite{cao2026scaleformer} & 2.894 & 2.142 & 0.920 & 0.954 & 0.66 & 0.411 & 32.5 & 41 \\
Semantic-prototype RWKV~\cite{li2026msps} & 2.838 & 2.109 & 0.922 & 0.957 & 0.70 & 0.438 & 33.1 & 58 \\
\rowcolor{ourrow}\our & \best{2.680} & \best{1.962} & \best{0.929} & \best{0.965} & \best{0.85} & \best{0.503} & \best{40.2} & 133 \\
\bottomrule
\end{tabular}%
}
\end{table}

\begin{table*}[!tbp]
\centering
\caption{SpaceNet in full. Top: every method on the four-city test regions (Main Table~\MTabDown{}d shows SGDiff, SeeSR, \our and the vendor product). Middle: building F1 and road APLS per city; the gain is largest in Paris and Shanghai and smallest in Las Vegas, where every method and the vendor product score highest. Bottom: reduced-resolution fidelity on $400$ SpaceNet WorldView-3 test crops under the Wald protocol (WorldView-III checkpoints without fine-tuning).}
\label{tab:spacenet_aoi}
\scriptsize
\setlength{\tabcolsep}{3pt}
\maxsizebox{\textwidth}{%
\begin{tabular}{>{\raggedright\arraybackslash}p{.18\linewidth}ccccccccc>{\columncolor{ourrow}}ccc}
\toprule
Measure & EXP & GSA & CANConv & ARConv & SGDiff & CLIPPan & DA-CLIP & SeeSR & \shortstack{DA-CLIP-\\style} & \our & \shortstack{Vendor\\PS} & std \\
\midrule
Building F1 (IoU $\ge0.5$) & 0.512 & 0.651 & 0.667 & 0.670 & 0.673 & 0.669 & 0.671 & 0.662 & 0.672 & \best{0.691} & 0.681 & $\pm$0.004 \\
Small-building F1 ($<60$\,m$^2$) & 0.21 & 0.46 & 0.49 & 0.50 & 0.50 & 0.48 & 0.50 & 0.51 & 0.50 & \best{0.56} & 0.52 & $\pm$0.01 \\
Road APLS & 0.482 & 0.611 & 0.627 & 0.630 & 0.633 & 0.629 & 0.631 & 0.618 & 0.632 & \best{0.656} & 0.642 & $\pm$0.005 \\
\bottomrule
\end{tabular}%
}\\[4pt]
\begin{tabular}{lcccc|cccc}
\toprule
 & \multicolumn{4}{c|}{building F1} & \multicolumn{4}{c}{road APLS} \\
Method & Las Vegas & Paris & Shanghai & Khartoum & Las Vegas & Paris & Shanghai & Khartoum \\
\midrule
SGDiff & 0.838 & 0.646 & 0.619 & 0.589 & 0.704 & 0.602 & 0.593 & 0.633 \\
DA-CLIP-style & 0.837 & 0.645 & 0.618 & 0.588 & 0.703 & 0.601 & 0.592 & 0.632 \\
Vendor PS & 0.846 & 0.654 & 0.626 & 0.598 & 0.712 & 0.611 & 0.602 & 0.643 \\
\rowcolor{ourrow}\our & 0.851 & 0.668 & 0.640 & 0.605 & 0.722 & 0.628 & 0.618 & 0.656 \\
\bottomrule
\end{tabular}\hspace{12pt}
\begin{tabular}{lccc}
\toprule
Method & ERGAS$\downarrow$ & SAM$\downarrow$ & Q2$^n\uparrow$ \\
\midrule
CANConv & 2.412 & 3.214 & 0.906 \\
ARConv & 2.331 & 3.092 & 0.909 \\
SGDiff & 2.303 & 3.051 & 0.911 \\
DA-CLIP-style & 2.389 & 3.178 & 0.907 \\
\rowcolor{ourrow}\our & 2.188 & 2.912 & 0.918 \\
\bottomrule
\end{tabular}
\end{table*}

\subsection{Additional scenes and failure modes}\label{sec:teaser_supp}
\begin{figure*}[!tbp]
\centering
\includegraphics[width=\textwidth]{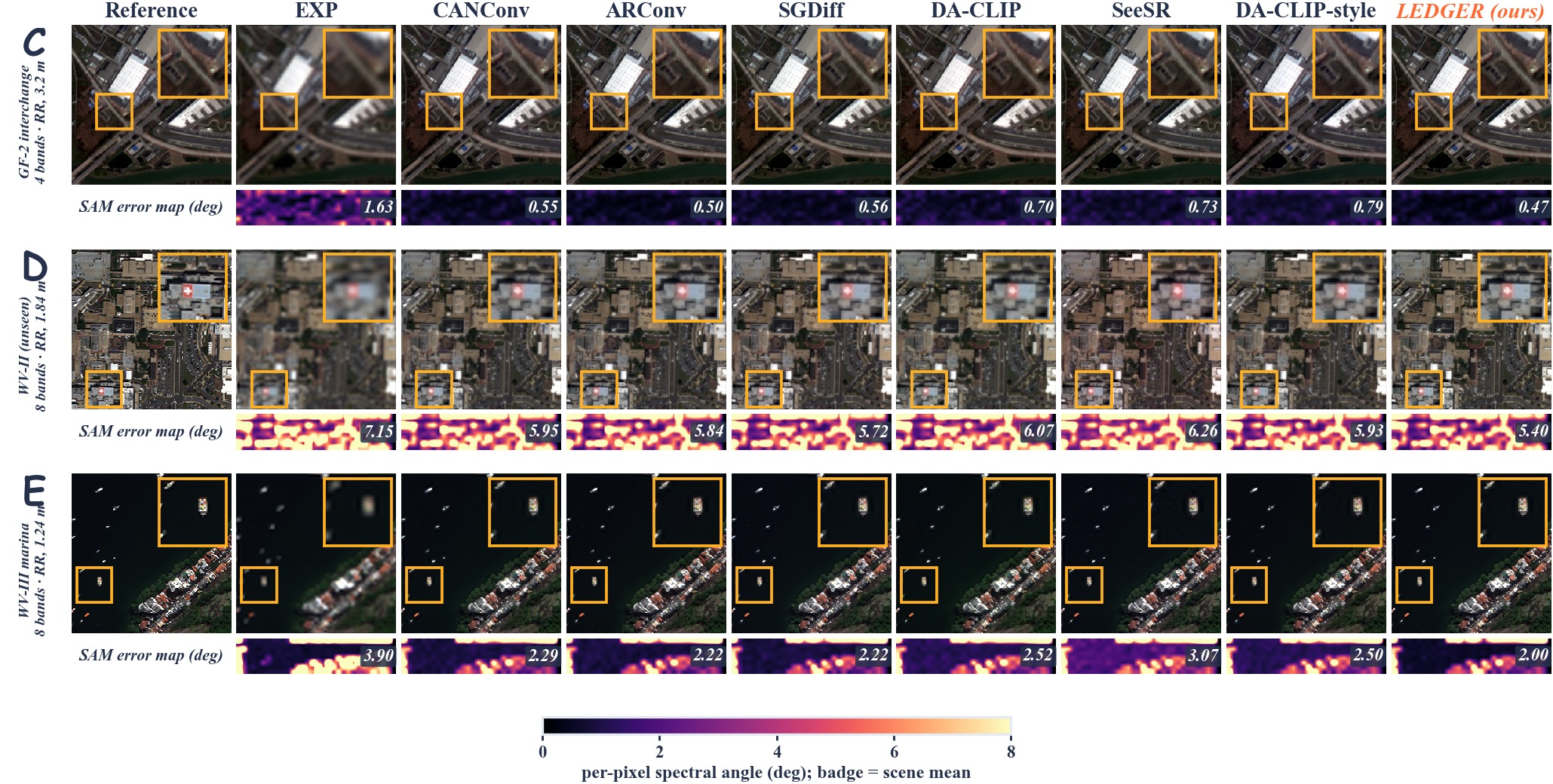}
\caption{Reduced-resolution results on three further annotated test scenes (C: GaoFen-2 interchange, output GSD $3.2$\,m; D: WorldView-II, unseen sensor, $1.84$\,m; E: WorldView-III marina, $1.24$\,m).}
\label{fig:qual_supp}
\end{figure*}

\begin{figure*}[!tbp]
\centering
\includegraphics[width=\textwidth]{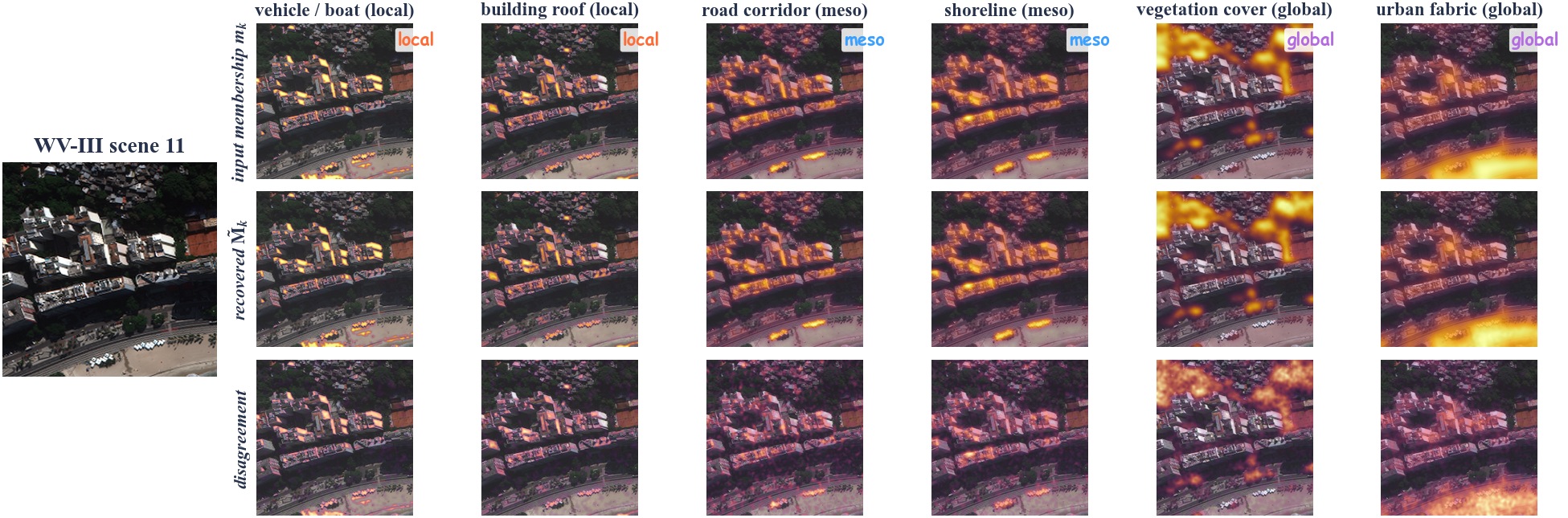}
\caption{Unit activations on WorldView-III scene 11. Top: input soft memberships $m_k$ of six units (two per scale); middle: recovered response maps $\tilde{\mathbf{M}}_k$ after re-encoding the output alone; bottom: their disagreement.}
\label{fig:activations}
\end{figure*}

\begin{figure}[!htbp]
\centering
\includegraphics[width=\columnwidth]{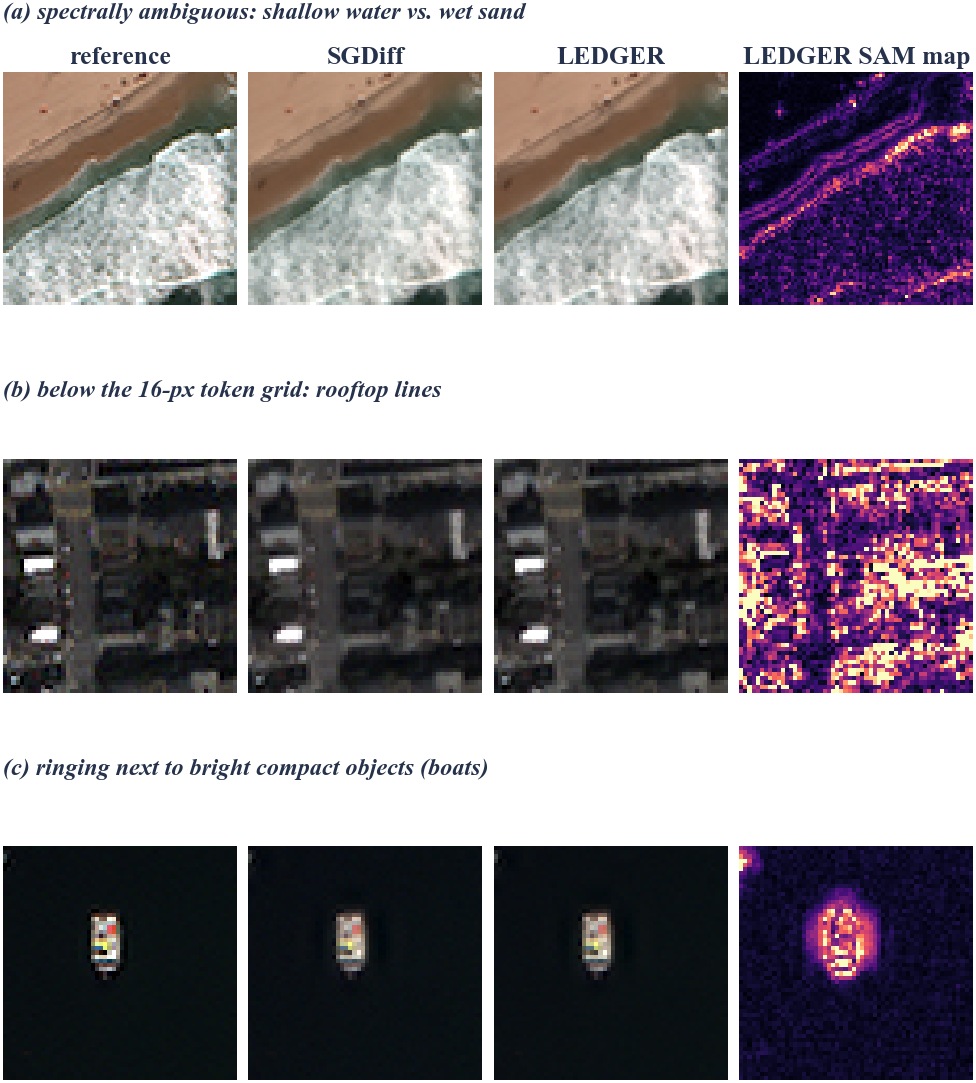}
\caption{Failure modes. (a) Spectrally ambiguous shallow water against wet sand keeps a low SAM but mis-routes the recovered concept. (b) Rooftop lines below the $16$-px token grid make the recovered mask unreliable even where the pixel error is small. (c) Ringing next to bright compact objects is the one regime in which a detector can read introduced structure as an object; it accounts for most of the residual false positives of the small-vehicle class.}
\label{fig:failure}
\end{figure}

Main Fig.~\MFigTeaser{} shows a $128$-px tile of WorldView-III reduced-resolution test scene~$\fTeaserScene$ beside its restorations by SGDiff and by \our, with the beach road and its parked cars enlarged. The scene was chosen because the two outputs have the same Q2$^n$ on it to three decimals; the values are those of this scene in \cref{tab:perscene_wv3_rr} ($\fTeaserBaseQ$ and $\fTeaserOursQ$). The rows are the \fTeaserNW{} present units of the tile's input ledger, with their tile-relative scale tags. These tags are the index's own, the footprint of a unit relative to the tile (Main Sec.~\MSecMethod), and are distinct from the annotation protocol of \cref{sec:annot}, in which local concepts are single objects labelled at full resolution only: at $1.24$\,m a parking lot or a tree line is a texture a few tokens long, which the index names and, by its footprint, tags meso, while its individual vehicles or trees are not resolved. In the figure, the chips are the response maps $\tilde{\mathbf{M}}_k$ read from each output, the bars are the presences $\hat p_k$, a unit counts as found when $\hat p_k{>}\tfrac12$ (the rule of the re-index score in \cref{alg:infer}), and the count is the coverage of Main Sec.~\MSecGap{} over these units. The figure shows in-loop readings and is not used as evidence. The annotated concept set of this tile is that of its scene label, beach (\cref{tab:mapping}), which the label-grounded probe finds in both outputs. Of the five other rows, parking lot carries external labels and therefore a probe head, which finds it in our output ($0.71$) and not in SGDiff's ($0.38$); coastal town, seaside resort, beach road and tree line are vocabulary concepts that the annotation protocol does not cover at this resolution (the first two are scene classes beyond the NWPU label set, the last two meso structures that the scene label does not entail), so they are the index's own readings. Outside the loop, the held-out OpenCLIP retriever, which never enters training, scores the bank query ``a beach road with parked vehicles'' at $0.31$ for our output against $0.22$ for SGDiff's and $0.33$ for the reference tile, and ranks our output above SGDiff's for the queries that name the other five concepts as well. Main Fig.~\MFigQual{} and \cref{fig:qual_supp} show reduced-resolution results on five annotated test scenes with the adapted VLM-guided methods beside the pansharpening networks; row labels give the protocol and the output GSD. \cref{fig:activations} shows input memberships, recovered response maps and their disagreement for six units of one tile. \cref{fig:failure} shows the regimes behind most of the residual gap between in-loop and probe coverage: spectral ambiguity between classes that share a signature (shallow water against wet sand), structure below the $16$-px token grid, which the response maps cannot localise however sharp the pixels are, and ringing next to bright compact objects, the one case in which a detector can read introduced structure as an object.

\section{Further domains}\label{sec:domains}

\subsection{Histopathology}\label{toc:supp:6:1}
Camelyon16~\cite{bejnordi2017camelyon}: $8{,}000$ training and $1{,}000$ validation $256^2$ tiles at $0.5\,\mu$m/px from $40$ slides disjoint from the $40$ test slides; restoration degradation drawn once per tile (Gaussian blur $\sigma\in\{1,1.5,2\}$, noise $\sigma\in\{5,15,25\}/255$, JPEG $Q\in\{30,50,70\}$); a $4\times$ super-resolution setting degrades by low-pass and decimation. The index is re-fitted with \cref{alg:fit} on $20{,}000$ pathology tokens and a $96$-term vocabulary (nuclei, mitotic figures, glands, stroma, lymphocytes, necrosis, \dots); the compositions are the $12$ tissue-level ones of that vocabulary; everything else is unchanged. Writing the vocabulary and compositions took one author about two hours per domain, and no centroid was renamed by hand in either transfer domain; this is the manual effort behind ``with only the index re-fitted''. All learned methods, including EDSR~\cite{lim2017edsr}, SwinIR~\cite{liang2021swinir} and Restormer~\cite{zamir2022restormer}, are trained on the same $8{,}000$ tiles for $1{,}000$ epochs with $256^2$ patches. The frozen evaluator is a nuclear instance segmenter trained once on MoNuSeg~\cite{kumar2017monuseg}; the PLIP probe~\cite{huang2023plip} is fitted on PanNuke~\cite{gamper2020pannuke} and NCT-CRC-HE-100K~\cite{kather2019nctcrc} labels; $120$ test tiles carry manual nuclear annotations ($9{,}812$ nuclei, inter-annotator F1 $0.91$, a $20\%$ sample adjudicated by a pathologist). \cref{tab:pathology} reports both settings and an unseen-organ test on the $14$ MoNuSeg test images. In the $1\times$ setting Restormer keeps the best PSNR ($36.9$ against $36.6$\,dB) and ties in the others; relative to Restormer, the closed loop raises PLIP-probe coverage from $0.74$ to $0.86$, nuclear F1 from $0.80$ to $0.82$ and AJI from $0.62$ to $0.65$, and its nuclear false-positive rate ($0.070$) is not above that of Restormer ($0.071$). The experiment demonstrates transfer of the recipe, not clinical validity.

\begin{table}[!htbp]
\centering
\caption{Histopathology (mean over $40$ Camelyon16 test slides or $14$ MoNuSeg images, three seeds; $\pm$ std over slides or images). Reference images: F1 $0.84$ / AJI $0.67$ (Camelyon16), $0.84$ / $0.66$ (MoNuSeg); $\mathcal{C}^{\mathrm{PLIP}}$ $0.91$, $0.91$, $0.89$. All learned methods share the training tiles, epochs and patch size.}
\label{tab:pathology}
\scriptsize
\setlength{\tabcolsep}{2pt}
\maxsizebox{\columnwidth}{%
\begin{tabular}{ll|ccccc|>{\columncolor{ourrow}}c}
\toprule
Setting & Metric & Input & EDSR & SwinIR & Restormer & open-loop \our & \our \\
\midrule
\multirow{5}{*}{\shortstack[l]{Camelyon16 $4\times$ SR}}
& PSNR$\uparrow$ & 27.8$\pm$1.6 & 30.9$\pm$1.6 & 31.3$\pm$1.6 & \second{31.5$\pm$1.6} & 31.2$\pm$1.6 & \best{31.6$\pm$1.6} \\
& SSIM$\uparrow$ & 0.86$\pm$0.02 & 0.90$\pm$0.02 & \second{0.91$\pm$0.02} & 0.91$\pm$0.02 & 0.91$\pm$0.02 & \best{0.92$\pm$0.02} \\
& $\mathcal{C}^{\mathrm{PLIP}}\uparrow$ & 0.58$\pm$0.06 & 0.64$\pm$0.06 & 0.66$\pm$0.06 & 0.67$\pm$0.06 & \second{0.70$\pm$0.06} & \best{0.83$\pm$0.06} \\
& Nuc.\ F1$\uparrow$ & 0.71$\pm$0.04 & 0.76$\pm$0.04 & 0.77$\pm$0.04 & \second{0.78$\pm$0.04} & 0.78$\pm$0.04 & \best{0.81$\pm$0.04} \\
& AJI$\uparrow$ & 0.52$\pm$0.05 & 0.58$\pm$0.05 & 0.59$\pm$0.05 & \second{0.60$\pm$0.05} & 0.60$\pm$0.05 & \best{0.64$\pm$0.05} \\
\midrule
\multirow{5}{*}{\shortstack[l]{Camelyon16 $1\times$ restoration}}
& PSNR$\uparrow$ & 28.6$\pm$1.6 & 34.6$\pm$1.6 & 36.3$\pm$1.6 & \best{36.9$\pm$1.6} & 35.8$\pm$1.6 & \second{36.6$\pm$1.6} \\
& SSIM$\uparrow$ & 0.81$\pm$0.02 & 0.93$\pm$0.02 & 0.95$\pm$0.02 & \best{0.96$\pm$0.02} & 0.95$\pm$0.02 & \second{0.96$\pm$0.02} \\
& $\mathcal{C}^{\mathrm{PLIP}}\uparrow$ & 0.61$\pm$0.06 & 0.70$\pm$0.06 & 0.73$\pm$0.06 & 0.74$\pm$0.06 & \second{0.76$\pm$0.06} & \best{0.86$\pm$0.06} \\
& Nuc.\ F1$\uparrow$ & 0.74$\pm$0.04 & 0.77$\pm$0.04 & 0.79$\pm$0.04 & \second{0.80$\pm$0.04} & 0.79$\pm$0.04 & \best{0.82$\pm$0.04} \\
& AJI$\uparrow$ & 0.55$\pm$0.05 & 0.59$\pm$0.05 & 0.61$\pm$0.05 & \second{0.62$\pm$0.05} & 0.61$\pm$0.05 & \best{0.65$\pm$0.05} \\
\midrule
\multirow{5}{*}{\shortstack[l]{MoNuSeg (unseen) $4\times$ SR}}
& PSNR$\uparrow$ & 26.9$\pm$1.6 & 29.3$\pm$1.6 & 29.7$\pm$1.6 & \second{29.8$\pm$1.6} & 29.5$\pm$1.6 & \best{30.1$\pm$1.6} \\
& SSIM$\uparrow$ & 0.83$\pm$0.02 & 0.87$\pm$0.02 & \second{0.88$\pm$0.02} & 0.88$\pm$0.02 & 0.88$\pm$0.02 & \best{0.89$\pm$0.02} \\
& $\mathcal{C}^{\mathrm{PLIP}}\uparrow$ & 0.51$\pm$0.06 & 0.56$\pm$0.06 & 0.58$\pm$0.06 & 0.58$\pm$0.06 & \second{0.61$\pm$0.06} & \best{0.76$\pm$0.06} \\
& Nuc.\ F1$\uparrow$ & 0.66$\pm$0.04 & 0.70$\pm$0.04 & \second{0.71$\pm$0.04} & 0.71$\pm$0.04 & 0.71$\pm$0.04 & \best{0.76$\pm$0.04} \\
& AJI$\uparrow$ & 0.47$\pm$0.05 & 0.51$\pm$0.05 & \second{0.53$\pm$0.05} & 0.53$\pm$0.05 & 0.53$\pm$0.05 & \best{0.58$\pm$0.05} \\
\bottomrule
\end{tabular}%
}
\end{table}

\subsection{Natural photographs}\label{toc:supp:6:2}
COCO~\cite{lin2014coco}: $1{,}000$ validation images are degraded with a RealSR-style~\cite{cai2019realsr} pipeline (blur $\sigma\in[0.5,2]$, $4\times$ decimation, noise $\sigma\in[1,15]/255$, JPEG $Q\in[50,90]$) and super-resolved $4\times$; training uses the COCO training images under the same pipeline, with SwinIR and Restormer trained for the same number of iterations as \our. The vocabulary is the $80$ categories plus the $200$ most frequent caption nouns, with $20$ compositions written from the caption co-occurrences; the coverage probe is a DINOv2 linear probe fitted on Open Images~\cite{kuznetsova2020openimages} labels; caption retrieval ranks the $5{,}000$ reference captions of the $1{,}000$ images by the held-out OpenCLIP retriever (R@1) and, for comparison, by the \emph{host} retriever (R@1 and mAP; in the loop, and reported as such); the small-concept bank holds $40$ queries about objects below $32^2$\,px. \cref{tab:coco} gives the full comparison. SeeSR renders plausible texture and has the best LPIPS, as expected of a diffusion prior, at $1.6$\,dB of PSNR; \our matches Restormer's PSNR within $0.1$\,dB and, relative to open-loop \our, closes $42\%$ of the host caption R@1 gap, $39\%$ of the OpenCLIP R@1 gap and $48\%$ of the small-concept gap to the reference.

\begin{table}[!htbp]
\centering
\caption{Natural photographs: $4\times$ SR of $1{,}000$ COCO images (three seeds). Reference images: LPIPS $0$, $\cdino$ $0.88$, host R@1 $0.61$, OpenCLIP R@1 $0.57$, caption mAP $0.66$, small-concept mAP $0.54$. R@1 (OC): caption R@1 under the held-out OpenCLIP retriever; R@1 and cap.\ mAP: under the host, in the loop and shown for comparison. SwinIR and Restormer are trained on the same COCO training images under the same degradation for the same number of iterations.}
\label{tab:coco}
\scriptsize
\setlength{\tabcolsep}{3pt}
\maxsizebox{\columnwidth}{%
\begin{tabular}{lcccccccc}
\toprule
Method & PSNR & LPIPS & $\cdino$ & R@1 (OC) & R@1 & cap.\ mAP & small mAP & s/img \\
\midrule
Bicubic & 25.1 & 0.412 & 0.57 & 0.27 & 0.29 & 0.36 & 0.19 & 0.00 \\
SwinIR & 28.9 & 0.261 & 0.66 & 0.38 & 0.41 & 0.47 & 0.31 & 0.08 \\
Restormer & 29.1 & 0.254 & 0.67 & 0.39 & 0.42 & 0.48 & 0.32 & 0.11 \\
SeeSR & 27.4 & 0.198 & 0.71 & 0.42 & 0.45 & 0.51 & 0.36 & 6.90 \\
DA-CLIP-style control & 29.0 & 0.249 & 0.70 & 0.41 & 0.44 & 0.50 & 0.35 & 0.06 \\
open-loop \our & 28.8 & 0.263 & 0.68 & 0.39 & 0.42 & 0.48 & 0.33 & 0.06 \\
\rowcolor{ourrow}\our & 29.0 & 0.231 & 0.79 & 0.46 & 0.50 & 0.56 & 0.43 & 0.09 \\
\bottomrule
\end{tabular}%
}
\end{table}

\section{Scope and broader impact}\label{sec:limits}
\noindent\textbf{Scope.} The index resolves concepts down to the $16$-px token of its encoder at twice the output grid, and structure finer than a token is governed by the reconstruction term (\cref{fig:failure}). The coverage loss is designed as a correction on top of a reconstruction objective: the fixed reconstruction weight, the gradient cap and the calibrated monitor of \cref{tab:gaming} are parts of the method. The full model deploys the $428$\,M-parameter frozen encoder beside its $\fParams$\,M trainable parameters and runs it once per scene, at $\fInferMs$\,ms: slower than the convolutional networks ($\fMsCC$\,ms for CANConv) and $\fSpeedLo$--$\fSpeedHi\times$ faster than the diffusion baselines; where the inference cost must equal that of an existing generator, the training-only form applies. The index is fitted per domain on training tiles, with a vocabulary and compositions written for that domain, and on concepts outside its vocabulary the gain is $+\fExtLo$ to $+\fExtHi$ under every evaluator (\cref{tab:vocab_split}). On SpaceNet the margins over the vendor product are $\fSnVendorFoneStd$ and $\fSnVendorAplsStd$ times the seed std for building F1 and road APLS, positive in each of the four cities, with one segmenter per task. The human study supports the question aligned with the objective; on the objective-neutral question the two methods are indistinguishable. The annotated evaluation covers $40$ WorldView-III scenes and a QuickBird subset at the concept level and four cities of raw WorldView-3 SpaceNet imagery at the task level; the archive-scale gallery of \cref{tab:archive}, whose distractors are raw sensor tiles of four other cities judged by pooling the top ranks of every method, keeps the same ordering, and the caption benchmarks written by others carry the check at larger scale. Scales are defined relative to the tile. The histopathology experiment demonstrates transfer of the recipe and makes no clinical claim.

\noindent\textbf{Broader impact.} Making small vehicles and vessels more retrievable in satellite imagery has civil uses (infrastructure monitoring, disaster response, maritime safety) and surveillance uses. The method does not add information that is absent from the input; it preserves what the panchromatic band already contains at the output resolution, and the hallucination measurements (\cref{tab:precision}) show that it introduces concepts at the rate of the pixel-supervised networks. Annotators and raters were compensated at the local research-assistant rate and informed of the purpose of the study; no personal data is involved, as the imagery is the public PanCollection and SpaceNet releases.

\par\FloatBarrier\addvspace{8pt}\Needspace{6\baselineskip}
\phantomsection
\pdfbookmark[0]{References}{report.references}
\begingroup
\small
\setlength{\bibsep}{1.8pt plus .5pt minus .5pt}
\bibliographystyle{unsrtnat}
\bibliography{references}

\begin{thebibliography}{107}
\providecommand{\natexlab}[1]{#1}
\providecommand{\url}[1]{\texttt{#1}}
\expandafter\ifx\csname urlstyle\endcsname\relax
  \providecommand{\doi}[1]{doi: #1}\else
  \providecommand{\doi}{doi: \begingroup \urlstyle{rm}\Url}\fi

\bibitem[Rombach et~al.(2022)Rombach, Blattmann, Lorenz, Esser, and Ommer]{rombach2022ldm}
Robin Rombach, Andreas Blattmann, Dominik Lorenz, Patrick Esser, and Bj{\"o}rn Ommer.
\newblock High-resolution image synthesis with latent diffusion models.
\newblock In \emph{IEEE/CVF Conference on Computer Vision and Pattern Recognition (CVPR)}, 2022.

\bibitem[Wu et~al.(2024)Wu, Yang, Sun, Zhang, Li, and Zhang]{wu2024seesr}
Rongyuan Wu, Tao Yang, Lingchen Sun, Zhengqiang Zhang, Shuai Li, and Lei Zhang.
\newblock {SeeSR}: Towards semantics-aware real-world image super-resolution.
\newblock In \emph{IEEE/CVF Conference on Computer Vision and Pattern Recognition (CVPR)}, 2024.

\bibitem[Yu et~al.(2024)Yu, Gu, Li, Hu, Kong, Wang, He, Qiao, and Dong]{yu2024supir}
Fanghua Yu, Jinjin Gu, Zheyuan Li, Jinfan Hu, Xiangtao Kong, Xintao Wang, Jingwen He, Yu~Qiao, and Chao Dong.
\newblock Scaling up to excellence: Practicing model scaling for photo-realistic image restoration in the wild.
\newblock In \emph{IEEE/CVF Conference on Computer Vision and Pattern Recognition (CVPR)}, 2024.

\bibitem[Rao et~al.(2022)Rao, Zhao, Chen, Tang, Zhu, Huang, Zhou, and Lu]{rao2022denseclip}
Yongming Rao, Wenliang Zhao, Guangyi Chen, Yansong Tang, Zheng Zhu, Guan Huang, Jie Zhou, and Jiwen Lu.
\newblock {DenseCLIP}: Language-guided dense prediction with context-aware prompting.
\newblock In \emph{IEEE/CVF Conference on Computer Vision and Pattern Recognition (CVPR)}, 2022.

\bibitem[Cheng et~al.(2024)Cheng, Liang, and Tan]{cheng2024clipdenoising}
Jun Cheng, Dong Liang, and Shan Tan.
\newblock Transfer {CLIP} for generalizable image denoising.
\newblock In \emph{IEEE/CVF Conference on Computer Vision and Pattern Recognition (CVPR)}, 2024.

\bibitem[Duan et~al.(2024)Duan, Wu, Deng, and Deng]{duan2024canconv}
Yule Duan, Xiao Wu, Haoyu Deng, and Liang-Jian Deng.
\newblock Content-adaptive non-local convolution for remote sensing pansharpening.
\newblock In \emph{IEEE/CVF Conference on Computer Vision and Pattern Recognition (CVPR)}, 2024.

\bibitem[Bandara and Patel(2022)]{bandara2022hypertransformer}
Wele Gedara~Chaminda Bandara and Vishal~M. Patel.
\newblock {HyperTransformer}: A textural and spectral feature fusion transformer for pansharpening.
\newblock In \emph{IEEE/CVF Conference on Computer Vision and Pattern Recognition (CVPR)}, 2022.

\bibitem[Zhang et~al.(2018)Zhang, Isola, Efros, Shechtman, and Wang]{zhang2018lpips}
Richard Zhang, Phillip Isola, Alexei~A. Efros, Eli Shechtman, and Oliver Wang.
\newblock The unreasonable effectiveness of deep features as a perceptual metric.
\newblock In \emph{IEEE/CVF Conference on Computer Vision and Pattern Recognition (CVPR)}, 2018.

\bibitem[Xia et~al.(2018)Xia, Bai, Ding, Zhu, Belongie, Luo, Datcu, Pelillo, and Zhang]{xia2018dota}
Gui-Song Xia, Xiang Bai, Jian Ding, Zhen Zhu, Serge Belongie, Jiebo Luo, Mihai Datcu, Marcello Pelillo, and Liangpei Zhang.
\newblock {DOTA}: A large-scale dataset for object detection in aerial images.
\newblock In \emph{IEEE/CVF Conference on Computer Vision and Pattern Recognition (CVPR)}, 2018.

\bibitem[Cherti et~al.(2023)Cherti, Beaumont, Wightman, Wortsman, Ilharco, Gordon, Schuhmann, Schmidt, and Jitsev]{cherti2023openclip}
Mehdi Cherti, Romain Beaumont, Ross Wightman, Mitchell Wortsman, Gabriel Ilharco, Cade Gordon, Christoph Schuhmann, Ludwig Schmidt, and Jenia Jitsev.
\newblock Reproducible scaling laws for contrastive language-image learning.
\newblock In \emph{IEEE/CVF Conference on Computer Vision and Pattern Recognition (CVPR)}, 2023.

\bibitem[Azzopardi and Vinay(2008)]{azzopardi2008retrievability}
Leif Azzopardi and Vishwa Vinay.
\newblock Retrievability: An evaluation measure for higher order information access tasks.
\newblock In \emph{ACM Conference on Information and Knowledge Management (CIKM)}, 2008.

\bibitem[Arienzo et~al.(2022)Arienzo, Vivone, Garzelli, Alparone, and Chanussot]{arienzo2022hqnr}
Alberto Arienzo, Gemine Vivone, Andrea Garzelli, Luciano Alparone, and Jocelyn Chanussot.
\newblock Full-resolution quality assessment of pansharpening: Theoretical and hands-on approaches.
\newblock \emph{IEEE Geoscience and Remote Sensing Magazine}, 10\penalty0 (3):\penalty0 168--201, 2022.

\bibitem[Blau and Michaeli(2018)]{blau2018perception}
Yochai Blau and Tomer Michaeli.
\newblock The perception-distortion tradeoff.
\newblock In \emph{IEEE/CVF Conference on Computer Vision and Pattern Recognition (CVPR)}, 2018.

\bibitem[Shermeyer and Van~Etten(2019)]{shermeyer2019sr}
Jacob Shermeyer and Adam Van~Etten.
\newblock The effects of super-resolution on object detection performance in satellite imagery.
\newblock In \emph{IEEE/CVF Conference on Computer Vision and Pattern Recognition Workshops (CVPRW)}, 2019.

\bibitem[Tschannen et~al.(2025)Tschannen, Gritsenko, Wang, Naeem, Alabdulmohsin, Parthasarathy, Evans, Beyer, Xia, Mustafa, H{\'e}naff, Harmsen, Steiner, and Zhai]{tschannen2025siglip2}
Michael Tschannen, Alexey Gritsenko, Xiao Wang, Muhammad~Ferjad Naeem, Ibrahim Alabdulmohsin, Nikhil Parthasarathy, Talfan Evans, Lucas Beyer, Ye~Xia, Basil Mustafa, Olivier H{\'e}naff, Jeremiah Harmsen, Andreas Steiner, and Xiaohua Zhai.
\newblock {SigLIP 2}: Multilingual vision-language encoders with improved semantic understanding, localization, and dense features.
\newblock \emph{arXiv preprint arXiv:2502.14786}, 2025.

\bibitem[Etten et~al.(2018)Etten, Lindenbaum, and Bacastow]{vanetten2018spacenet}
Adam~Van Etten, Dave Lindenbaum, and Todd~M. Bacastow.
\newblock {SpaceNet}: A remote sensing dataset and challenge series.
\newblock \emph{arXiv preprint arXiv:1807.01232}, 2018.

\bibitem[Brooks et~al.(2023)Brooks, Holynski, and Efros]{brooks2023ip2p}
Tim Brooks, Aleksander Holynski, and Alexei~A. Efros.
\newblock {InstructPix2Pix}: Learning to follow image editing instructions.
\newblock In \emph{IEEE/CVF Conference on Computer Vision and Pattern Recognition (CVPR)}, 2023.

\bibitem[Yang et~al.(2024)Yang, Wu, Ren, Xie, and Zhang]{yang2024pasd}
Tao Yang, Rongyuan Wu, Peiran Ren, Xuansong Xie, and Lei Zhang.
\newblock Pixel-aware stable diffusion for realistic image super-resolution and personalized stylization.
\newblock In \emph{European Conference on Computer Vision (ECCV)}, 2024.

\bibitem[Sun et~al.(2024)Sun, Li, Liu, Chen, Pei, Zou, Yan, and Yang]{sun2024coser}
Haoze Sun, Wenbo Li, Jianzhuang Liu, Haoyu Chen, Renjing Pei, Xueyi Zou, Youliang Yan, and Yujiu Yang.
\newblock {CoSeR}: Bridging image and language for cognitive super-resolution.
\newblock In \emph{IEEE/CVF Conference on Computer Vision and Pattern Recognition (CVPR)}, 2024.

\bibitem[Li et~al.(2024)Li, Li, Zhu, Jin, Feng, Zhang, and Chen]{li2024sed}
Bingchen Li, Xin Li, Hanxin Zhu, Yeying Jin, Ruoyu Feng, Zhizheng Zhang, and Zhibo Chen.
\newblock {SeD}: Semantic-aware discriminator for image super-resolution.
\newblock In \emph{IEEE/CVF Conference on Computer Vision and Pattern Recognition (CVPR)}, 2024.

\bibitem[Zhou et~al.(2022{\natexlab{a}})Zhou, Loy, and Dai]{zhou2022maskclip}
Chong Zhou, Chen~Change Loy, and Bo~Dai.
\newblock Extract free dense labels from {CLIP}.
\newblock In \emph{European Conference on Computer Vision (ECCV)}, 2022{\natexlab{a}}.

\bibitem[Luo et~al.(2024)Luo, Gustafsson, Zhao, Sj{\"o}lund, and Sch{\"o}n]{luo2024daclip}
Ziwei Luo, Fredrik~K. Gustafsson, Zheng Zhao, Jens Sj{\"o}lund, and Thomas~B. Sch{\"o}n.
\newblock Controlling vision-language models for universal image restoration.
\newblock In \emph{International Conference on Learning Representations (ICLR)}, 2024.

\bibitem[Yang et~al.(2025)Yang, Dong, and Lam]{yang2025vlmir}
Cuixin Yang, Rongkang Dong, and Kin-Man Lam.
\newblock Vision-language model guided image restoration.
\newblock \emph{arXiv preprint arXiv:2512.17292}, 2025.

\bibitem[Liang et~al.(2023)Liang, Li, Zhou, Feng, and Loy]{liang2023cliplit}
Zhexin Liang, Chongyi Li, Shangchen Zhou, Ruicheng Feng, and Chen~Change Loy.
\newblock Iterative prompt learning for unsupervised backlit image enhancement.
\newblock In \emph{IEEE/CVF International Conference on Computer Vision (ICCV)}, 2023.

\bibitem[Wang et~al.(2023{\natexlab{a}})Wang, Chan, and Loy]{wang2023clipiqa}
Jianyi Wang, Kelvin C.~K. Chan, and Chen~Change Loy.
\newblock Exploring {CLIP} for assessing the look and feel of images.
\newblock In \emph{AAAI Conference on Artificial Intelligence}, 2023{\natexlab{a}}.

\bibitem[Min et~al.(2026)Min, Kim, Cho, Lee, Park, Park, Kim, Park, and Kim]{min2025tair}
Jaewon Min, Jin~Hyeon Kim, Paul~Hyunbin Cho, Jaeeun Lee, Jihye Park, Minkyu Park, Sangpil Kim, Hyunhee Park, and Seungryong Kim.
\newblock Text-aware image restoration with diffusion models.
\newblock In \emph{International Conference on Learning Representations (ICLR)}, 2026.

\bibitem[Xing et~al.(2025)Xing, Qu, Zhang, Xu, Yang, and Zhang]{xing2025sgdiff}
Yinghui Xing, Litao Qu, Shizhou Zhang, Di~Xu, Yingkun Yang, and Yanning Zhang.
\newblock Dual-granularity semantic guided sparse routing diffusion model for general pansharpening.
\newblock In \emph{IEEE/CVF Conference on Computer Vision and Pattern Recognition (CVPR)}, 2025.

\bibitem[Li et~al.(2026)Li, Zhou, Li, He, Gan, and Ren]{li2026msps}
Junfeng Li, Wenyang Zhou, Xueheng Li, Xuanhua He, Jianhou Gan, and Wenqi Ren.
\newblock Multigrain-aware semantic prototype scanning and tri-token prompt learning embraced high-order {RWKV} for pan-sharpening.
\newblock In \emph{IEEE/CVF Conference on Computer Vision and Pattern Recognition (CVPR)}, pages 13234--13243, 2026.

\bibitem[Jian et~al.(2026)Jian, Liu, Wu, and Chen]{jian2025clippan}
Lihua Jian, Jiabo Liu, Shaowu Wu, and Lihui Chen.
\newblock {CLIPPan}: Adapting {CLIP} as a supervisor for unsupervised pansharpening.
\newblock In \emph{AAAI Conference on Artificial Intelligence (AAAI)}, 2026.

\bibitem[Liu et~al.(2018)Liu, Wen, Liu, Wang, and Huang]{liu2018denoisinghl}
Ding Liu, Bihan Wen, Xianming Liu, Zhangyang Wang, and Thomas~S. Huang.
\newblock When image denoising meets high-level vision tasks: A deep learning approach.
\newblock In \emph{International Joint Conference on Artificial Intelligence (IJCAI)}, 2018.

\bibitem[Son et~al.(2020)Son, Kang, Kim, Cho, and Kwak]{son2020urie}
Taeyoung Son, Juwon Kang, Namyup Kim, Sunghyun Cho, and Suha Kwak.
\newblock {URIE}: Universal image enhancement for visual recognition in the wild.
\newblock In \emph{European Conference on Computer Vision (ECCV)}, 2020.

\bibitem[Haris et~al.(2021)Haris, Shakhnarovich, and Ukita]{haris2021taskdriven}
Muhammad Haris, Greg Shakhnarovich, and Norimichi Ukita.
\newblock Task-driven super resolution: Object detection in low-resolution images.
\newblock In \emph{International Conference on Neural Information Processing (ICONIP)}, 2021.

\bibitem[Kim et~al.(2024)Kim, Oh, and Lee]{kim2024sr4ir}
Jaeha Kim, Junghun Oh, and Kyoung~Mu Lee.
\newblock Beyond image super-resolution for image recognition with task-driven perceptual loss.
\newblock In \emph{IEEE/CVF Conference on Computer Vision and Pattern Recognition (CVPR)}, 2024.

\bibitem[Zhang et~al.(2023)Zhang, Lei, Xie, Fang, Li, and Du]{zhang2023superyolo}
Jiaqing Zhang, Jie Lei, Weiying Xie, Zhenman Fang, Yunsong Li, and Qian Du.
\newblock {SuperYOLO}: Super resolution assisted object detection in multimodal remote sensing imagery.
\newblock \emph{IEEE Transactions on Geoscience and Remote Sensing}, 61:\penalty0 1--15, 2023.

\bibitem[Bai et~al.(2018)Bai, Zhang, Ding, and Ghanem]{bai2018sodmtgan}
Yancheng Bai, Yongqiang Zhang, Mingli Ding, and Bernard Ghanem.
\newblock {SOD-MTGAN}: Small object detection via multi-task generative adversarial network.
\newblock In \emph{European Conference on Computer Vision (ECCV)}, 2018.

\bibitem[Duan et~al.(2020)Duan, Liu, Yang, Huang, and Gao]{duan2020vcm}
Ling-Yu Duan, Jiaying Liu, Wenhan Yang, Tiejun Huang, and Wen Gao.
\newblock Video coding for machines: A paradigm of collaborative compression and intelligent analytics.
\newblock \emph{IEEE Transactions on Image Processing}, 29:\penalty0 8680--8695, 2020.

\bibitem[Tang et~al.(2022)Tang, Yuan, and Ma]{tang2022seafusion}
Linfeng Tang, Jiteng Yuan, and Jiayi Ma.
\newblock Image fusion in the loop of high-level vision tasks: A semantic-aware real-time infrared and visible image fusion network.
\newblock \emph{Information Fusion}, 82:\penalty0 28--42, 2022.

\bibitem[Liu et~al.(2022{\natexlab{a}})Liu, Fan, Huang, Wu, Liu, Zhong, and Luo]{liu2022tardal}
Jinyuan Liu, Xin Fan, Zhanbo Huang, Guanyao Wu, Risheng Liu, Wei Zhong, and Zhongxuan Luo.
\newblock Target-aware dual adversarial learning and a multi-scenario multi-modality benchmark to fuse infrared and visible for object detection.
\newblock In \emph{IEEE/CVF Conference on Computer Vision and Pattern Recognition (CVPR)}, 2022{\natexlab{a}}.

\bibitem[Liu et~al.(2023)Liu, Liu, Wu, Ma, Liu, Zhong, Luo, and Fan]{liu2023segmif}
Jinyuan Liu, Zhu Liu, Guanyao Wu, Long Ma, Risheng Liu, Wei Zhong, Zhongxuan Luo, and Xin Fan.
\newblock Multi-interactive feature learning and a full-time multi-modality benchmark for image fusion and segmentation.
\newblock In \emph{IEEE/CVF International Conference on Computer Vision (ICCV)}, 2023.

\bibitem[Dodge and Karam(2016)]{dodge2016quality}
Samuel Dodge and Lina Karam.
\newblock Understanding how image quality affects deep neural networks.
\newblock In \emph{International Conference on Quality of Multimedia Experience (QoMEX)}, 2016.

\bibitem[Johnson et~al.(2016)Johnson, Alahi, and Fei-Fei]{johnson2016perceptual}
Justin Johnson, Alexandre Alahi, and Li~Fei-Fei.
\newblock Perceptual losses for real-time style transfer and super-resolution.
\newblock In \emph{European Conference on Computer Vision (ECCV)}, 2016.

\bibitem[Wang et~al.(2018)Wang, Yu, Dong, and Loy]{wang2018sftgan}
Xintao Wang, Ke~Yu, Chao Dong, and Chen~Change Loy.
\newblock Recovering realistic texture in image super-resolution by deep spatial feature transform.
\newblock In \emph{IEEE/CVF Conference on Computer Vision and Pattern Recognition (CVPR)}, 2018.

\bibitem[Perez et~al.(2018)Perez, Strub, de~Vries, Dumoulin, and Courville]{perez2018film}
Ethan Perez, Florian Strub, Harm de~Vries, Vincent Dumoulin, and Aaron Courville.
\newblock {FiLM}: Visual reasoning with a general conditioning layer.
\newblock In \emph{AAAI Conference on Artificial Intelligence (AAAI)}, 2018.

\bibitem[Park et~al.(2019)Park, Liu, Wang, and Zhu]{park2019spade}
Taesung Park, Ming-Yu Liu, Ting-Chun Wang, and Jun-Yan Zhu.
\newblock Semantic image synthesis with spatially-adaptive normalization.
\newblock In \emph{IEEE/CVF Conference on Computer Vision and Pattern Recognition (CVPR)}, 2019.

\bibitem[Hoffman et~al.(2018)Hoffman, Tzeng, Park, Zhu, Isola, Saenko, Efros, and Darrell]{hoffman2018cycada}
Judy Hoffman, Eric Tzeng, Taesung Park, Jun-Yan Zhu, Phillip Isola, Kate Saenko, Alexei~A. Efros, and Trevor Darrell.
\newblock {CyCADA}: Cycle-consistent adversarial domain adaptation.
\newblock In \emph{International Conference on Machine Learning (ICML)}, 2018.

\bibitem[Clark et~al.(2024)Clark, Vicol, Swersky, and Fleet]{clark2024draft}
Kevin Clark, Paul Vicol, Kevin Swersky, and David~J. Fleet.
\newblock Directly fine-tuning diffusion models on differentiable rewards.
\newblock In \emph{International Conference on Learning Representations (ICLR)}, 2024.

\bibitem[Prabhudesai et~al.(2023)Prabhudesai, Goyal, Pathak, and Fragkiadaki]{prabhudesai2023alignprop}
Mihir Prabhudesai, Anirudh Goyal, Deepak Pathak, and Katerina Fragkiadaki.
\newblock Aligning text-to-image diffusion models with reward backpropagation.
\newblock \emph{arXiv preprint arXiv:2310.03739}, 2023.

\bibitem[Chefer et~al.(2023)Chefer, Alaluf, Vinker, Wolf, and Cohen-Or]{chefer2023attendexcite}
Hila Chefer, Yuval Alaluf, Yael Vinker, Lior Wolf, and Daniel Cohen-Or.
\newblock Attend-and-excite: Attention-based semantic guidance for text-to-image diffusion models.
\newblock \emph{ACM Transactions on Graphics}, 42\penalty0 (4), 2023.

\bibitem[Hu et~al.(2023)Hu, Liu, Kasai, Wang, Ostendorf, Krishna, and Smith]{hu2023tifa}
Yushi Hu, Benlin Liu, Jungo Kasai, Yizhong Wang, Mari Ostendorf, Ranjay Krishna, and Noah~A. Smith.
\newblock {TIFA}: Accurate and interpretable text-to-image faithfulness evaluation with question answering.
\newblock In \emph{IEEE/CVF International Conference on Computer Vision (ICCV)}, 2023.

\bibitem[Cho et~al.(2024)Cho, Hu, Baldridge, Garg, Anderson, Krishna, Bansal, Pont-Tuset, and Wang]{cho2024dsg}
Jaemin Cho, Yushi Hu, Jason~M. Baldridge, Roopal Garg, Peter Anderson, Ranjay Krishna, Mohit Bansal, Jordi Pont-Tuset, and Su~Wang.
\newblock Davidsonian scene graph: Improving reliability in fine-grained evaluation for text-to-image generation.
\newblock In \emph{International Conference on Learning Representations (ICLR)}, 2024.

\bibitem[Zhao et~al.(2023)Zhao, Pang, Du, Yang, Li, Cheung, and Lin]{zhao2023advvlm}
Yunqing Zhao, Tianyu Pang, Chao Du, Xiao Yang, Chongxuan Li, Ngai-Man Cheung, and Min Lin.
\newblock On evaluating adversarial robustness of large vision-language models.
\newblock In \emph{Advances in Neural Information Processing Systems (NeurIPS)}, 2023.

\bibitem[Milletari et~al.(2016)Milletari, Navab, and Ahmadi]{milletari2016vnet}
Fausto Milletari, Nassir Navab, and Seyed-Ahmad Ahmadi.
\newblock {V-Net}: Fully convolutional neural networks for volumetric medical image segmentation.
\newblock In \emph{International Conference on 3D Vision (3DV)}, 2016.

\bibitem[Sudre et~al.(2017)Sudre, Li, Vercauteren, Ourselin, and Cardoso]{sudre2017gdl}
Carole~H. Sudre, Wenqi Li, Tom Vercauteren, S{\'e}bastien Ourselin, and M.~Jorge Cardoso.
\newblock Generalised {Dice} overlap as a deep learning loss function for highly unbalanced segmentations.
\newblock In \emph{Deep Learning in Medical Image Analysis and Multimodal Learning for Clinical Decision Support (DLMIA)}, 2017.

\bibitem[Rahman and Wang(2016)]{rahman2016jaccard}
Md~Atiqur Rahman and Yang Wang.
\newblock Optimizing intersection-over-union in deep neural networks for image segmentation.
\newblock In \emph{International Symposium on Visual Computing (ISVC)}, 2016.

\bibitem[Berman et~al.(2018)Berman, Rannen~Triki, and Blaschko]{berman2018lovasz}
Maxim Berman, Amal Rannen~Triki, and Matthew~B. Blaschko.
\newblock The {Lov\'asz}-softmax loss: A tractable surrogate for the optimization of the intersection-over-union measure in neural networks.
\newblock In \emph{IEEE/CVF Conference on Computer Vision and Pattern Recognition (CVPR)}, 2018.

\bibitem[Wang et~al.(2023{\natexlab{b}})Wang, Ning, and Blaschko]{wang2023jaccard}
Zifu Wang, Xuefei Ning, and Matthew~B. Blaschko.
\newblock Jaccard metric losses: Optimizing the {J}accard index with soft labels.
\newblock In \emph{Advances in Neural Information Processing Systems (NeurIPS)}, 2023{\natexlab{b}}.

\bibitem[Aiazzi et~al.(2007)Aiazzi, Baronti, and Selva]{aiazzi2007gsa}
Bruno Aiazzi, Stefano Baronti, and Massimo Selva.
\newblock Improving component substitution pansharpening through multivariate regression of {MS}+{Pan} data.
\newblock \emph{IEEE Transactions on Geoscience and Remote Sensing}, 45\penalty0 (10):\penalty0 3230--3239, 2007.

\bibitem[Vivone(2019)]{vivone2019bdsdpc}
Gemine Vivone.
\newblock Robust band-dependent spatial-detail approaches for panchromatic sharpening.
\newblock \emph{IEEE Transactions on Geoscience and Remote Sensing}, 57\penalty0 (9):\penalty0 6421--6433, 2019.

\bibitem[Yang et~al.(2017)Yang, Fu, Hu, Huang, Ding, and Paisley]{yang2017pannet}
Junfeng Yang, Xueyang Fu, Yuwen Hu, Yue Huang, Xinghao Ding, and John Paisley.
\newblock {PanNet}: A deep network architecture for pan-sharpening.
\newblock In \emph{IEEE International Conference on Computer Vision (ICCV)}, 2017.

\bibitem[Deng et~al.(2021)Deng, Vivone, Jin, and Chanussot]{deng2021fusionnet}
Liang-Jian Deng, Gemine Vivone, Cheng Jin, and Jocelyn Chanussot.
\newblock Detail injection-based deep convolutional neural networks for pansharpening.
\newblock \emph{IEEE Transactions on Geoscience and Remote Sensing}, 59\penalty0 (8):\penalty0 6995--7010, 2021.

\bibitem[Jin et~al.(2022)Jin, Zhang, Jiang, Vivone, and Deng]{jin2022lagconv}
Zi-Rong Jin, Tian-Jing Zhang, Tai-Xiang Jiang, Gemine Vivone, and Liang-Jian Deng.
\newblock {LAGConv}: Local-context adaptive convolution kernels with global harmonic bias for pansharpening.
\newblock In \emph{AAAI Conference on Artificial Intelligence (AAAI)}, 2022.

\bibitem[Zhou et~al.(2022{\natexlab{b}})Zhou, Liu, and Wang]{zhou2022panformer}
Huanyu Zhou, Qingjie Liu, and Yunhong Wang.
\newblock {PanFormer}: A transformer based model for pan-sharpening.
\newblock In \emph{IEEE International Conference on Multimedia and Expo (ICME)}, 2022{\natexlab{b}}.

\bibitem[Cao et~al.(2026)Cao, He, Li, Zhu, Wang, Ma, Zhang, Zhou, Xie, and Zhang]{cao2026scaleformer}
Ke~Cao, Xuanhua He, Xueheng Li, Lingting Zhu, Yingying Wang, Ao~Ma, Zhanjie Zhang, Man Zhou, Chengjun Xie, and Jie Zhang.
\newblock Cross-scale pansharpening via {ScaleFormer} and the {PanScale} benchmark.
\newblock In \emph{IEEE/CVF Conference on Computer Vision and Pattern Recognition (CVPR)}, pages 13211--13221, 2026.

\bibitem[Xing et~al.(2024)Xing, Qu, Zhang, Zhang, Zhang, and Bruzzone]{xing2024crossdiff}
Yinghui Xing, Litao Qu, Shizhou Zhang, Kai Zhang, Yanning Zhang, and Lorenzo Bruzzone.
\newblock {CrossDiff}: Exploring self-supervised representation of pansharpening via cross-predictive diffusion model.
\newblock \emph{IEEE Transactions on Image Processing}, 33, 2024.

\bibitem[Meng et~al.(2023)Meng, Shi, Li, and Zhang]{meng2024pandiff}
Qingyan Meng, Wenxu Shi, Sijia Li, and Linlin Zhang.
\newblock {PanDiff}: A novel pansharpening method based on denoising diffusion probabilistic model.
\newblock \emph{IEEE Transactions on Geoscience and Remote Sensing}, 61, 2023.

\bibitem[Zhong et~al.(2024)Zhong, Wu, Deng, Cao, and Dou]{zhong2024ssdiff}
Yu~Zhong, Xiao Wu, Liang-Jian Deng, Zihan Cao, and Hong-Xia Dou.
\newblock {SSDiff}: Spatial-spectral integrated diffusion model for remote sensing pansharpening.
\newblock In \emph{Advances in Neural Information Processing Systems (NeurIPS)}, 2024.

\bibitem[He et~al.(2025)He, Cao, Zhang, Yan, Wang, Li, Xie, Hong, and Zhou]{he2025panmamba}
Xuanhua He, Ke~Cao, Jie Zhang, Keyu Yan, Yingying Wang, Rui Li, Chengjun Xie, Danfeng Hong, and Man Zhou.
\newblock {Pan-Mamba}: Effective pan-sharpening with state space model.
\newblock \emph{Information Fusion}, 115, 2025.

\bibitem[Wang et~al.(2025)Wang, Zheng, Shao, Duan, and Deng]{wang2025arconv}
Xueyang Wang, Zhixin Zheng, Jiandong Shao, Yule Duan, and Liang-Jian Deng.
\newblock Adaptive rectangular convolution for remote sensing pansharpening.
\newblock In \emph{IEEE/CVF Conference on Computer Vision and Pattern Recognition (CVPR)}, 2025.

\bibitem[Deng et~al.(2022)Deng, Vivone, Paoletti, Scarpa, He, Zhang, Chanussot, and Plaza]{deng2022pancollection}
Liang-Jian Deng, Gemine Vivone, Mercedes~E. Paoletti, Giuseppe Scarpa, Jiang He, Yongjun Zhang, Jocelyn Chanussot, and Antonio Plaza.
\newblock Machine learning in pansharpening: A benchmark, from shallow to deep networks.
\newblock \emph{IEEE Geoscience and Remote Sensing Magazine}, 10\penalty0 (3):\penalty0 279--315, 2022.

\bibitem[Wald et~al.(1997)Wald, Ranchin, and Mangolini]{wald1997fusion}
Lucien Wald, Thierry Ranchin, and Marc Mangolini.
\newblock Fusion of satellite images of different spatial resolutions: Assessing the quality of resulting images.
\newblock \emph{Photogrammetric Engineering and Remote Sensing}, 63\penalty0 (6):\penalty0 691--699, 1997.

\bibitem[Vivone et~al.(2021)Vivone, Dalla~Mura, Garzelli, Restaino, Scarpa, Ulfarsson, Alparone, and Chanussot]{vivone2021new}
Gemine Vivone, Mauro Dalla~Mura, Andrea Garzelli, Rocco Restaino, Giuseppe Scarpa, Magnus~O. Ulfarsson, Luciano Alparone, and Jocelyn Chanussot.
\newblock A new benchmark based on recent advances in multispectral pansharpening: Revisiting pansharpening with classical and emerging pansharpening methods.
\newblock \emph{IEEE Geoscience and Remote Sensing Magazine}, 9\penalty0 (1):\penalty0 53--81, 2021.

\bibitem[Liu et~al.(2024)Liu, Chen, Guan, Zhou, Zhu, Ye, Fu, and Zhou]{liu2024remoteclip}
Fan Liu, Delong Chen, Zhangqingyun Guan, Xiaocong Zhou, Jiale Zhu, Qiaolin Ye, Liyong Fu, and Jun Zhou.
\newblock {RemoteCLIP}: A vision language foundation model for remote sensing.
\newblock \emph{IEEE Transactions on Geoscience and Remote Sensing}, 62, 2024.

\bibitem[Zhang et~al.(2024)Zhang, Zhao, Guo, and Yin]{zhang2024georsclip}
Zilun Zhang, Tiancheng Zhao, Yulong Guo, and Jianwei Yin.
\newblock {RS5M} and {GeoRSCLIP}: A large-scale vision-language dataset and a large vision-language model for remote sensing.
\newblock \emph{IEEE Transactions on Geoscience and Remote Sensing}, 62, 2024.

\bibitem[Wang et~al.(2024)Wang, Prabha, Huang, Wu, and Rajagopal]{wang2024skyscript}
Zhecheng Wang, Rajanie Prabha, Tianyuan Huang, Jiajun Wu, and Ram Rajagopal.
\newblock {SkyScript}: A large and semantically diverse vision-language dataset for remote sensing.
\newblock In \emph{AAAI Conference on Artificial Intelligence}, 2024.

\bibitem[Yuan et~al.(2022)Yuan, Zhang, Fu, Li, Deng, Wang, and Sun]{yuan2022rsitmd}
Zhiqiang Yuan, Wenkai Zhang, Kun Fu, Xuan Li, Chubo Deng, Hongqi Wang, and Xian Sun.
\newblock Exploring a fine-grained multiscale method for cross-modal remote sensing image retrieval.
\newblock \emph{IEEE Transactions on Geoscience and Remote Sensing}, 60:\penalty0 1--19, 2022.

\bibitem[Lu et~al.(2018)Lu, Wang, Zheng, and Li]{lu2018rsicd}
Xiaoqiang Lu, Binqiang Wang, Xiangtao Zheng, and Xuelong Li.
\newblock Exploring models and data for remote sensing image caption generation.
\newblock \emph{IEEE Transactions on Geoscience and Remote Sensing}, 56\penalty0 (4):\penalty0 2183--2195, 2018.

\bibitem[Reed et~al.(2023)Reed, Gupta, Li, Brockman, Funk, Clipp, Keutzer, Candido, Uyttendaele, and Darrell]{reed2023scalemae}
Colorado~J. Reed, Ritwik Gupta, Shufan Li, Sarah Brockman, Christopher Funk, Brian Clipp, Kurt Keutzer, Salvatore Candido, Matt Uyttendaele, and Trevor Darrell.
\newblock {Scale-MAE}: A scale-aware masked autoencoder for multiscale geospatial representation learning.
\newblock In \emph{IEEE/CVF International Conference on Computer Vision (ICCV)}, 2023.

\bibitem[Cong et~al.(2022)Cong, Khanna, Meng, Liu, Rozi, He, Burke, Lobell, and Ermon]{cong2022satmae}
Yezhen Cong, Samar Khanna, Chenlin Meng, Patrick Liu, Erik Rozi, Yutong He, Marshall Burke, David~B. Lobell, and Stefano Ermon.
\newblock {SatMAE}: Pre-training transformers for temporal and multi-spectral satellite imagery.
\newblock In \emph{Advances in Neural Information Processing Systems (NeurIPS)}, 2022.

\bibitem[Oquab et~al.(2024)Oquab, Darcet, Moutakanni, Vo, Szafraniec, Khalidov, Fernandez, Haziza, Massa, El-Nouby, Assran, Ballas, Galuba, Howes, Huang, Li, Misra, Rabbat, Sharma, Synnaeve, Xu, J{\'e}gou, Mairal, Labatut, Joulin, and Bojanowski]{oquab2024dinov2}
Maxime Oquab, Timoth{\'e}e Darcet, Th{\'e}o Moutakanni, Huy Vo, Marc Szafraniec, Vasil Khalidov, Pierre Fernandez, Daniel Haziza, Francisco Massa, Alaaeldin El-Nouby, Mahmoud Assran, Nicolas Ballas, Wojciech Galuba, Russell Howes, Po-Yao Huang, Shang-Wen Li, Ishan Misra, Michael Rabbat, Vasu Sharma, Gabriel Synnaeve, Hu~Xu, Herv{\'e} J{\'e}gou, Julien Mairal, Patrick Labatut, Armand Joulin, and Piotr Bojanowski.
\newblock {DINOv2}: Learning robust visual features without supervision.
\newblock \emph{Transactions on Machine Learning Research}, 2024.

\bibitem[Ronneberger et~al.(2015)Ronneberger, Fischer, and Brox]{ronneberger2015unet}
Olaf Ronneberger, Philipp Fischer, and Thomas Brox.
\newblock U-net: Convolutional networks for biomedical image segmentation.
\newblock In \emph{Medical Image Computing and Computer-Assisted Intervention (MICCAI)}, 2015.

\bibitem[Wang et~al.(2004)Wang, Bovik, Sheikh, and Simoncelli]{wang2004ssim}
Zhou Wang, Alan~C. Bovik, Hamid~R. Sheikh, and Eero~P. Simoncelli.
\newblock Image quality assessment: From error visibility to structural similarity.
\newblock \emph{IEEE Transactions on Image Processing}, 13\penalty0 (4):\penalty0 600--612, 2004.

\bibitem[Garzelli and Nencini(2009)]{garzelli2009q2n}
Andrea Garzelli and Filippo Nencini.
\newblock Hypercomplex quality assessment of multi/hyperspectral images.
\newblock \emph{IEEE Geoscience and Remote Sensing Letters}, 6\penalty0 (4):\penalty0 662--665, 2009.

\bibitem[Xia et~al.(2017)Xia, Hu, Hu, Shi, Bai, Zhong, Zhang, and Lu]{xia2017aid}
Gui-Song Xia, Jingwen Hu, Fan Hu, Baoguang Shi, Xiang Bai, Yanfei Zhong, Liangpei Zhang, and Xiaoqiang Lu.
\newblock {AID}: A benchmark data set for performance evaluation of aerial scene classification.
\newblock \emph{IEEE Transactions on Geoscience and Remote Sensing}, 55\penalty0 (7):\penalty0 3965--3981, 2017.

\bibitem[Cheng et~al.(2017)Cheng, Han, and Lu]{cheng2017nwpu}
Gong Cheng, Junwei Han, and Xiaoqiang Lu.
\newblock Remote sensing image scene classification: Benchmark and state of the art.
\newblock \emph{Proceedings of the IEEE}, 105\penalty0 (10):\penalty0 1865--1883, 2017.

\bibitem[Christie et~al.(2018)Christie, Fendley, Wilson, and Mukherjee]{christie2018fmow}
Gordon Christie, Neil Fendley, James Wilson, and Ryan Mukherjee.
\newblock Functional map of the world.
\newblock In \emph{IEEE/CVF Conference on Computer Vision and Pattern Recognition (CVPR)}, 2018.

\bibitem[Etten(2020)]{vanetten2020roads}
Adam~Van Etten.
\newblock City-scale road extraction from satellite imagery v2: Road speeds and travel times.
\newblock In \emph{IEEE/CVF Winter Conference on Applications of Computer Vision (WACV)}, 2020.

\bibitem[Loshchilov and Hutter(2019)]{loshchilov2019adamw}
Ilya Loshchilov and Frank Hutter.
\newblock Decoupled weight decay regularization.
\newblock In \emph{International Conference on Learning Representations (ICLR)}, 2019.

\bibitem[Li et~al.(2023)Li, Li, Savarese, and Hoi]{li2023blip2}
Junnan Li, Dongxu Li, Silvio Savarese, and Steven Hoi.
\newblock {BLIP-2}: Bootstrapping language-image pre-training with frozen image encoders and large language models.
\newblock In \emph{International Conference on Machine Learning (ICML)}, 2023.

\bibitem[Liu et~al.(2022{\natexlab{b}})Liu, Mao, Wu, Feichtenhofer, Darrell, and Xie]{liu2022convnext}
Zhuang Liu, Hanzi Mao, Chao-Yuan Wu, Christoph Feichtenhofer, Trevor Darrell, and Saining Xie.
\newblock A {ConvNet} for the 2020s.
\newblock In \emph{IEEE/CVF Conference on Computer Vision and Pattern Recognition (CVPR)}, 2022{\natexlab{b}}.

\bibitem[Xie et~al.(2021)Xie, Cheng, Wang, Yao, and Han]{xie2021orcnn}
Xingxing Xie, Gong Cheng, Jiabao Wang, Xiwen Yao, and Junwei Han.
\newblock Oriented {R-CNN} for object detection.
\newblock In \emph{IEEE/CVF International Conference on Computer Vision (ICCV)}, 2021.

\bibitem[Rohrbach et~al.(2018)Rohrbach, Hendricks, Burns, Darrell, and Saenko]{rohrbach2018chair}
Anna Rohrbach, Lisa~Anne Hendricks, Kaylee Burns, Trevor Darrell, and Kate Saenko.
\newblock Object hallucination in image captioning.
\newblock In \emph{Conference on Empirical Methods in Natural Language Processing (EMNLP)}, 2018.

\bibitem[Hu et~al.(2022)Hu, Shen, Wallis, Allen-Zhu, Li, Wang, Wang, and Chen]{hu2022lora}
Edward~J. Hu, Yelong Shen, Phillip Wallis, Zeyuan Allen-Zhu, Yuanzhi Li, Shean Wang, Lu~Wang, and Weizhu Chen.
\newblock {LoRA}: Low-rank adaptation of large language models.
\newblock In \emph{International Conference on Learning Representations (ICLR)}, 2022.

\bibitem[Ehteshami~Bejnordi et~al.(2017)Ehteshami~Bejnordi, Veta, Johannes~van Diest, van Ginneken, Karssemeijer, Litjens, van~der Laak, and the CAMELYON16~Consortium]{bejnordi2017camelyon}
Babak Ehteshami~Bejnordi, Mitko Veta, Paul Johannes~van Diest, Bram van Ginneken, Nico Karssemeijer, Geert Litjens, Jeroen A. W.~M. van~der Laak, and the CAMELYON16~Consortium.
\newblock Diagnostic assessment of deep learning algorithms for detection of lymph node metastases in women with breast cancer.
\newblock \emph{JAMA}, 318\penalty0 (22):\penalty0 2199--2210, 2017.

\bibitem[Lin et~al.(2014)Lin, Maire, Belongie, Hays, Perona, Ramanan, Doll{\'a}r, and Zitnick]{lin2014coco}
Tsung-Yi Lin, Michael Maire, Serge Belongie, James Hays, Pietro Perona, Deva Ramanan, Piotr Doll{\'a}r, and C.~Lawrence Zitnick.
\newblock {Microsoft COCO}: Common objects in context.
\newblock In \emph{European Conference on Computer Vision (ECCV)}, 2014.

\bibitem[Bi{\'n}kowski et~al.(2018)Bi{\'n}kowski, Sutherland, Arbel, and Gretton]{binkowski2018kid}
Miko{\l}aj Bi{\'n}kowski, Danica~J. Sutherland, Michael Arbel, and Arthur Gretton.
\newblock Demystifying {MMD} {GANs}.
\newblock In \emph{International Conference on Learning Representations (ICLR)}, 2018.

\bibitem[Mittal et~al.(2013)Mittal, Soundararajan, and Bovik]{mittal2013niqe}
Anish Mittal, Rajiv Soundararajan, and Alan~C. Bovik.
\newblock Making a ``completely blind'' image quality analyzer.
\newblock \emph{IEEE Signal Processing Letters}, 20\penalty0 (3):\penalty0 209--212, 2013.

\bibitem[Yang et~al.(2022)Yang, Wu, Shi, Lao, Gong, Cao, Wang, and Yang]{yang2022maniqa}
Sidi Yang, Tianhe Wu, Shuwei Shi, Shanshan Lao, Yuan Gong, Mingdeng Cao, Jiahao Wang, and Yujiu Yang.
\newblock {MANIQA}: Multi-dimension attention network for no-reference image quality assessment.
\newblock In \emph{IEEE/CVF Conference on Computer Vision and Pattern Recognition Workshops (CVPRW)}, 2022.

\bibitem[Kuckreja et~al.(2024)Kuckreja, Danish, Naseer, Das, Khan, and Khan]{kuckreja2024geochat}
Kartik Kuckreja, Muhammad~Sohail Danish, Muzammal Naseer, Abhijit Das, Salman Khan, and Fahad~Shahbaz Khan.
\newblock {GeoChat}: Grounded large vision-language model for remote sensing.
\newblock In \emph{IEEE/CVF Conference on Computer Vision and Pattern Recognition (CVPR)}, 2024.

\bibitem[Lim et~al.(2017)Lim, Son, Kim, Nah, and Lee]{lim2017edsr}
Bee Lim, Sanghyun Son, Heewon Kim, Seungjun Nah, and Kyoung~Mu Lee.
\newblock Enhanced deep residual networks for single image super-resolution.
\newblock In \emph{IEEE/CVF Conference on Computer Vision and Pattern Recognition Workshops (CVPRW)}, 2017.

\bibitem[Liang et~al.(2021)Liang, Cao, Sun, Zhang, Van~Gool, and Timofte]{liang2021swinir}
Jingyun Liang, Jiezhang Cao, Guolei Sun, Kai Zhang, Luc Van~Gool, and Radu Timofte.
\newblock {SwinIR}: Image restoration using swin transformer.
\newblock In \emph{IEEE/CVF International Conference on Computer Vision Workshops (ICCVW)}, 2021.

\bibitem[Zamir et~al.(2022)Zamir, Arora, Khan, Hayat, Khan, and Yang]{zamir2022restormer}
Syed~Waqas Zamir, Aditya Arora, Salman Khan, Munawar Hayat, Fahad~Shahbaz Khan, and Ming-Hsuan Yang.
\newblock {Restormer}: Efficient transformer for high-resolution image restoration.
\newblock In \emph{IEEE/CVF Conference on Computer Vision and Pattern Recognition (CVPR)}, 2022.

\bibitem[Kumar et~al.(2017)Kumar, Verma, Sharma, Bhargava, Vahadane, and Sethi]{kumar2017monuseg}
Neeraj Kumar, Ruchika Verma, Sanuj Sharma, Surabhi Bhargava, Abhishek Vahadane, and Amit Sethi.
\newblock A dataset and a technique for generalized nuclear segmentation for computational pathology.
\newblock \emph{IEEE Transactions on Medical Imaging}, 36\penalty0 (7):\penalty0 1550--1560, 2017.

\bibitem[Huang et~al.(2023)Huang, Bianchi, Yuksekgonul, Montine, and Zou]{huang2023plip}
Zhi Huang, Federico Bianchi, Mert Yuksekgonul, Thomas~J. Montine, and James Zou.
\newblock A visual-language foundation model for pathology image analysis using medical {Twitter}.
\newblock \emph{Nature Medicine}, 29:\penalty0 2307--2316, 2023.

\bibitem[Gamper et~al.(2020)Gamper, Koohbanani, Benes, Graham, Jahanifar, Khurram, Azam, Hewitt, and Rajpoot]{gamper2020pannuke}
Jevgenij Gamper, Navid~Alemi Koohbanani, Ksenija Benes, Simon Graham, Mostafa Jahanifar, Syed~Ali Khurram, Ayesha Azam, Katherine Hewitt, and Nasir Rajpoot.
\newblock {PanNuke} dataset extension, insights and baselines.
\newblock \emph{arXiv preprint arXiv:2003.10778}, 2020.

\bibitem[Kather et~al.(2019)Kather, Krisam, Charoentong, Luedde, Herpel, Weis, Gaiser, Marx, Valous, Ferber, Jansen, Reyes-Aldasoro, Z{\"o}rnig, J{\"a}ger, Brenner, Chang-Claude, Hoffmeister, and Halama]{kather2019nctcrc}
Jakob~Nikolas Kather, Johannes Krisam, Pornpimol Charoentong, Tom Luedde, Esther Herpel, Cleo-Aron Weis, Timo Gaiser, Alexander Marx, Nektarios~A. Valous, Dyke Ferber, Lina Jansen, Constantino~Carlos Reyes-Aldasoro, Inka Z{\"o}rnig, Dirk J{\"a}ger, Hermann Brenner, Jenny Chang-Claude, Michael Hoffmeister, and Niels Halama.
\newblock Predicting survival from colorectal cancer histology slides using deep learning: A retrospective multicenter study.
\newblock \emph{PLOS Medicine}, 16\penalty0 (1):\penalty0 e1002730, 2019.

\bibitem[Cai et~al.(2019)Cai, Zeng, Yong, Cao, and Zhang]{cai2019realsr}
Jianrui Cai, Hui Zeng, Hongwei Yong, Zisheng Cao, and Lei Zhang.
\newblock Toward real-world single image super-resolution: A new benchmark and a new model.
\newblock In \emph{IEEE/CVF International Conference on Computer Vision (ICCV)}, 2019.

\bibitem[Kuznetsova et~al.(2020)Kuznetsova, Rom, Alldrin, Uijlings, Krasin, Pont-Tuset, Kamali, Popov, Malloci, Kolesnikov, Duerig, and Ferrari]{kuznetsova2020openimages}
Alina Kuznetsova, Hassan Rom, Neil Alldrin, Jasper Uijlings, Ivan Krasin, Jordi Pont-Tuset, Shahab Kamali, Stefan Popov, Matteo Malloci, Alexander Kolesnikov, Tom Duerig, and Vittorio Ferrari.
\newblock The {Open Images} dataset {V4}: Unified image classification, object detection, and visual relationship detection at scale.
\newblock \emph{International Journal of Computer Vision}, 128:\penalty0 1956--1981, 2020.

\end{thebibliography}
\endgroup
\end{document}